%% file: these.tex
\documentclass[french]{amu_these}

\usepackage{bm}
\usepackage{physics}
\usepackage{siunitx}
\usepackage{wrapfig}
\usepackage{booktabs}
\usepackage{stmaryrd}

\usepackage{hyperref}
\usepackage{cleveref}
\usepackage{tcolorbox}
\usepackage{amsmath}
\usepackage{amsfonts}
\usepackage{mathtools}
\usepackage{amssymb} %
\usepackage{fontawesome5} %
\tcbuselibrary{skins,breakable} %
\usepackage{epigraph}   %
\usepackage{pdfpages} %

\DeclareMathAlphabet{\mathcal}{OMS}{cmsy}{m}{n} %

\crefname{part}{Part}{Parts}      %
\Crefname{part}{Part}{Parts}
\crefname{section}{Section}{Sections}
\Crefname{section}{Section}{Sections}
\crefname{equation}{Eq.}{Eqs.}
\Crefname{equation}{Equation}{Equations}
\crefname{figure}{Fig.}{Figs.}
\Crefname{figure}{Figure}{Figures}
\crefname{chapter}{Chapter}{Chapters}
\Crefname{chapter}{Chapter}{Chapters}
\crefname{appendix}{appendix}{appendices}
\Crefname{appendix}{Appendix}{Appendices}

\newcommand{\apdx}[0]{Appendix}

\newcommand{\E}[1]{\mathbb{E} \left[{#1}\right]}

\newcommand{\avg}[1]{\left \langle #1 \right \rangle}
\DeclareMathOperator*{\argmax}{argmax}

\DeclareMathOperator{\erfc}{erfc}
\newcommand{\eulergamma}{\gamma} %

\newcommand{\Var}{\mathrm{Var}} %
\newcommand{\R}{\mathbb{R}}
\newcommand{\N}{\mathcal{N}}

\usepackage{graphicx}        %
\usepackage{tikzpagenodes}   %
\usepackage{lipsum}          %

\newcommand*{\framesdir}{frames/frame_} %
\newcommand*{\frameext}{.pdf}           %
\newcommand*{\nframes}{2}              %
\newcommand*{\cornerwidth}{2cm}         %
\newcommand*{\xshift}{-0.5cm}             %
\newcommand*{\yshift}{ 0.5cm}             %

\AddToHook{shipout/background}{%
  \edef\pageno{\number\value{page}}%
  \ifnum\pageno>\nframes
    \begin{tikzpicture}[remember picture,overlay]
      \node[anchor=south east,xshift=\xshift,yshift=\yshift] at (current page.south east)
        {\includegraphics[width=\cornerwidth]{\framesdir\pageno\frameext}};
    \end{tikzpicture}%
  \fi
}

\begin{document}
    \includepdf[pages=1, pagecommand=\thispagestyle{empty}]{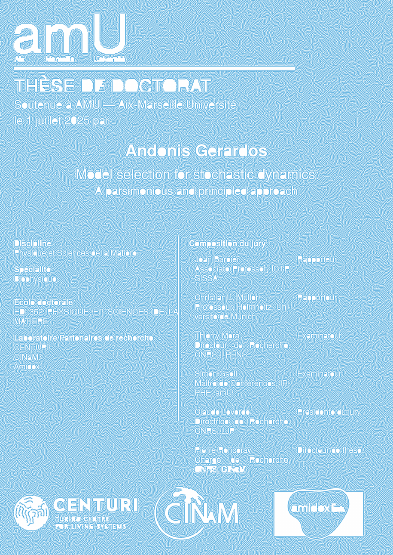}
    
    \newpage
    \thispagestyle{empty}
    \mbox{} %
    
	\input{tex_before/titre.tex}				%
										
	\input{tex_before/affidavit}				%

	\input{tex_before/publications}			%

	\input{tex_before/resume}					%

	\input{tex_before/abstract}				%

	\input{tex_before/remercie}				%

    \selectlanguage{english}
    
    \microtypesetup{protrusion=false}	%
	\tableofcontents					%
	\printglossary[						%
		type=\acronymtype,
		title={Liste des acronymes},
		toctitle={Liste des acronymes}
		]
	\printglossary[						%
		title={Glossaire},
		toctitle={Glossaire}
		]
	\printglossary[						%
		type=notation,
		title={Nomenclature},
		toctitle={Nomenclature}
		]
    \microtypesetup{protrusion=true}	%

	\ohead{\leftmark\Ifstr{\rightmark}{\leftmark}{}{ -- \rightmark}}	%

	  \input{tex_body/intro}

\input{tex_body/chap1_a}
    \input{tex_body/chap1_b}

    \input{tex_body/chap2_a}

\input{tex_body/chap2_b}
    \input{tex_body/chap2_c}
    \input{tex_body/chap2_d}

	\input{tex_body/chap3}
   \input{tex_body/chap4}
  \input{tex_body/chap5}
  \input{tex_body/chap6}

\input{tex_body/chap7}
  \input{tex_body/chap8}
  \input{tex_body/chap9}
	\input{tex_body/conc}

	\appendix

	\newpage
	\printbibliography[heading=bibintoc]%
	
	\newpage
	\printindex							%
	
	\newpage
	\printendnotes						%

\input{tex_behind/annexes}			%

\end{document}

%% file: tex_before/titre.tex
\chead{}
\pdfbookmark[0]{Page de titre}{titre}
\thispagestyle{empty}
\newgeometry{margin=2em}

\vspace{1em}

\begin{center}
	\begin{minipage}[c]{.5\linewidth}
		\raggedright\includegraphics[height=7em]{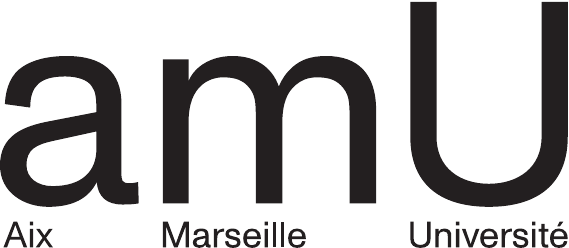}
	\end{minipage}\hfill
	\begin{minipage}[c]{.5\linewidth}
	\end{minipage}\hfill
\end{center}


\begin{center}
	\begin{minipage}[c]{.77\linewidth}
		\textcolor{yellowamu}{\noindent\rule{\textwidth}{4pt}}
	\end{minipage}\hfill
	\begin{minipage}[c]{.23\linewidth}
	\end{minipage}\hfill
\end{center}


\doublespacing
\begin{flushleft}
    \textsf{\HUGE\textcolor{blueamu}{THÈSE DE DOCTORAT}}\\
	\textsf{\Large Soutenue à AMU ― Aix-Marseille Université}\\
	\textsf{\Large le 1 juillet 2025 par}\\
\end{flushleft}
\vspace{1em}
\begin{center}
	\textsf{\textbf{\Huge Andonis Gerardos}}\\
    \vspace{1em}
	\textsf{\LARGE Model selection for stochastic dynamics:}\\ 
	\textsf{\Large A parsimonious and principled approach}\\
\end{center}
\singlespacing

\vspace{2em}

\begin{center}
	\begin{minipage}[t]{.45\linewidth}
    	    \vspace{.5em}
        	\textsf{\textbf{Discipline}}
        	
        	\textsf{Physique et Sciences de la Matière}
        	
    	    \vspace{1em}
        	\textsf{\textbf{Spécialité}}
        	
        	\textsf{Biophysique}
        	
    	    \vspace{2em}
        	\textsf{\textbf{École doctorale}}
        	
        	\textsf{ED 352 PHYSIQUE ET SCIENCES DE LA MATIERE}
        	
    	    \vspace{1em}
        	\textsf{\textbf{Laboratoire/Partenaires de recherche}}
        	
        	\textsf{CENTURI\\CINaM\\
        	Amidex
        	}

	\end{minipage}\hfill
	\begin{minipage}[t]{.03\linewidth}
	    \rule[-280pt]{1pt}{280pt}
	\end{minipage}\hfill
	\begin{minipage}[t]{.52\linewidth}
	    \vspace{.5em}
    	\textsf{\textbf{Composition du jury}}

	    \vspace{1em}
    	\textsf{
        \begin{tabular}{p{12em} p{10em}}
        	Jean Barbier & Rapporteur \\
        	  Associate Professor, ICTP, SISSA \\
        	\\
        	Christian L. Müller & Rapporteur \\
        	Professeur, Helmholtz, Université de Munich  \\
            \\
            Thierry Mora & Examinateur \\
        	Directeur de Recherche, CNRS, LPENS\\
            \\
            Simon Gsell & Examinateur \\
        	Maitre de Conférences, IRPHE, amU\\
            \\
        	Claude Loverdo & Présidente du jury \\
        	Directrice de Recherche, CNRS, LJP\\
            \\
        	Pierre Ronceray & Directeur de thèse \\
        	Chargé de Recherche, CNRS, CINaM\\
        \end{tabular}
        }
	\end{minipage}\hfill
\end{center}       

\vspace{0em}

\begin{center} 
	\begin{minipage}[c]{.33\linewidth}
		\centering\includegraphics[height=5em]{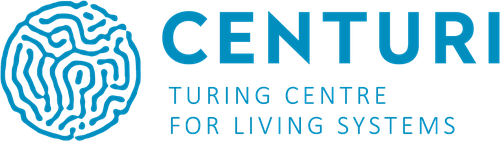} 
	\end{minipage}\hfill
    \begin{minipage}[c]{.33\linewidth}
		\centering\includegraphics[height=8em]{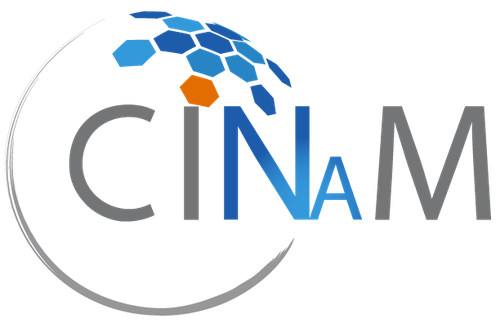} 
	\end{minipage}\hfill
	\begin{minipage}[c]{.33\linewidth}
		\centering\includegraphics[height=8em]{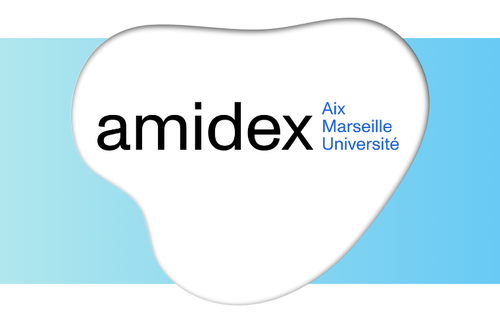}
	\end{minipage}\hfill
\end{center}

\restoregeometry

%% file: tex_before/affidavit.tex
\newpage
\addchap{Affidavit}
\label{chap:affidavit}
\thispagestyle{empty}

\iftrue 
    Je soussigné, Andonis Gerardos, 
    déclare par la présente que le travail présenté dans ce manuscrit est mon propre travail, réalisé sous la direction scientifique de Pierre Ronceray, 
    dans le respect des principes d’honnêteté, d'intégrité et de responsabilité inhérents à la mission de recherche. Les travaux de recherche et la rédaction de ce manuscrit ont été réalisés dans le respect à la fois de la charte nationale de déontologie des métiers de la recherche et de la charte AMU relative à la lutte contre le plagiat.
    
    Ce travail n'a pas été précédemment soumis en France ou à l'étranger dans une version identique ou similaire à un organisme examinateur.\\
    
    Fait à Marseille le 05/05/2025
    
    \begin{flushright}\includegraphics[height=80px]{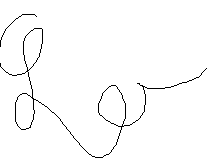}\end{flushright}

    ~\vfill
    \begin{center}
        \begin{minipage}[c]{0.25\linewidth}
            \includegraphics[height=35px]{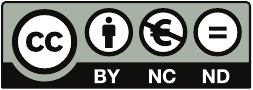}
        \end{minipage}\hfill
    \end{center}

    Cette \oe{}uvre est mise à disposition selon les termes de la \href{https://creativecommons.org/licenses/by-nc-nd/4.0/deed.fr}{Licence Creative Commons Attribution - Pas d’Utilisation Commerciale - Pas de Modification 4.0 International}. 
\fi

%% file: tex_before/publications.tex
\newpage

\addchap{Publications et Communications Scientifiques}
\label{chap:publications}

\addsec*{Publications issues des travaux de thèse}
\begin{enumerate}
    \item \textbf{[Statut : Publié]} \\
          S. Amiri, Y. Zhang, A. Gerardos, C. Sykes et P. Ronceray.
          \emph{Inferring geometrical dynamics of cell nucleus translocation}.
          Physical Review Research, 6, 043030 (2024). \\
          DOI: \href{https://doi.org/10.1103/PhysRevResearch.6.043030}{10.1103/PhysRevResearch.6.043030}

    \item \textbf{[Statut : Preprint / Soumis à Physical Review Letter - Jan. 2025]} \\ 
          A. Gerardos et P. Ronceray.
          \emph{Principled model selection for stochastic dynamics}. \\
        \href{https://arxiv.org/abs/2501.10339}{https://arxiv.org/abs/2501.10339}

\end{enumerate}

\addsec*{Communications et participation aux conférences et écoles}
\begin{enumerate}
    \item \textbf{iPoLS Annual Meeting}, Montpellier, France, Mai 2022.
    \item \textbf{École d'été de Physique de Beg Rohu}, Saint-Pierre-Quiberon, France, Juin 2022.
    \item \textbf{École d'été de Physique Théorique des Houches (Session Biophysique)}, Les Houches, France, Juillet 2023. (Poster)
    \item \textbf{CENTURI Conference}, Cargèse, France, Octobre 2023. (Poster)
    \item \textbf{DPG Spring Meeting (DPG Frühjahrstagung - Section Biological Physics)}, Berlin, Allemagne, Mars 2024. (Communication orale) \url{https://www.dpg-verhandlungen.de/year/2024/conference/berlin/part/bp/session/23/contribution/5}
    \item \textbf{BioInference 2024}, Warwick, Royaume-Uni, Juin 2024. (Communication orale) \url{https://bioinference.github.io/2024/}
    \item \textbf{École d'été Systèmes Complexes et Vitreux}, Cargèse, France, Juillet 2024. (Poster)
    \item \textbf{Physical Biology Circle Meeting}, Lisbonne, Portugal, Janvier 2025. (Communication orale)
    \item  \textbf{Stochastics processes: Inferences in complex systems}, Lausanne, Suisse, Mai 2025. (Poster) \url{https://www.cecam.org/workshop-details/stochastics-processes-inferences-in-complex-systems-1390}
    \item \textbf{Statphys 2025 (Statphys 29)}, Florence, Italie, Juillet 2025. (Communication orale)
\end{enumerate}

%% file: tex_before/resume.tex
\addchap{Résumé et mots clés}
\label{chap:resume}
\selectlanguage{french} 

Cette thèse porte sur la découverte d'équations différentielles stochastiques (EDS) et d'équations différentielles aux dérivées partielles stochastiques (EDPS) à partir de séries temporelles bruitées et discrètes. Un défi majeur est la sélection d'un modèle correct le plus simple possible parmi de vastes bibliothèques de modèles candidats, où les critères d'informations standards (AIC, BIC) sont souvent limités.

Nous introduisons PASTIS (Parsimonious Stochastic Inference), un nouveau critère d'information dérivé de la théorie des valeurs extrêmes. Sa pénalité, $n_\mathcal{B} \ln(n_0/p)$, intègre explicitement la taille de la bibliothèque initiale de paramètres candidats ($n_0$), le nombre de paramètres du modèle considéré ($n_\mathcal{B}$) et un seuil de significativité ($p$). Ce seuil de significativité représente la probabilité de sélectionner un modèle contenant plus de paramètres que nécessaire lors de la comparaison de nombreux modèles.

Des tests comparatifs (\textit{benchmarks}) sur divers systèmes (Lorenz, Ornstein-Uhlenbeck, Lotka-Volterra pour les EDS; Gray-Scott pour les EDPS) démontrent que PASTIS surpasse AIC, BIC, la validation croisée (CV) et SINDy (une méthode concurrente) en termes d'identification exacte du modèle et de capacité prédictive.

De plus, les données réelles peuvent être soumises à des intervalles d'échantillonnage ($\Delta t$) importants ou à du bruit de mesure ($\sigma$) qui peuvent détériorer les possibilités d'apprentissage et de sélection de modèle. Pour remédier à cela, nous avons développé des variantes robustes de PASTIS, PASTIS-$\Delta t$ et PASTIS-$\sigma$, étendant ainsi l'applicabilité de l'approche aux données expérimentales imparfaites.

PASTIS fournit ainsi un cadre méthodologique statistiquement fondé, validé et pratique pour la découverte de modèles simples pour les processus à dynamique stochastique.

\vspace{0.5cm} 
Mots clés: Équations Différentielles Stochastiques, Équations aux Dérivées Partielles Stochastiques, Sélection de Modèles, Inférence Parcimonieuse, Critères d'Information, Théorie des Valeurs Extrêmes, PASTIS, Dynamique de Langevin, Analyse de Séries Temporelles, Découverte Guidée par les Données

%% file: tex_before/abstract.tex

\addchap{Abstract and keywords}
\label{chap:abstract}
\selectlanguage{english} 
This thesis focuses on the discovery of stochastic differential equations (SDEs) and stochastic partial differential equations (SPDEs) from noisy and discrete time series. A major challenge is selecting the simplest possible correct model from vast libraries of candidate models, where standard information criteria (AIC, BIC) are often limited.

We introduce PASTIS (Parsimonious Stochastic Inference), a new information criterion derived from extreme value theory. Its penalty term, $n_\mathcal{B} \ln(n_0/p)$, explicitly incorporates the size of the initial library of candidate parameters ($n_0$), the number of parameters in the considered model ($n_\mathcal{B}$), and a significance threshold ($p$). This significance threshold represents the probability of selecting a model containing more parameters than necessary when comparing many models.

Benchmarks on various systems (Lorenz, Ornstein-Uhlenbeck, Lotka-Volterra for SDEs; Gray-Scott for SPDEs) demonstrate that PASTIS outperforms AIC, BIC, cross-validation (CV), and SINDy (a competing method) in terms of exact model identification and predictive capability.

Furthermore, real-world data can be subject to large sampling intervals ($\Delta t$) or measurement noise ($\sigma$), which can impair model learning and selection capabilities. To address this, we have developed robust variants of PASTIS, PASTIS-$\Delta t$ and PASTIS-$\sigma$, thus extending the applicability of the approach to imperfect experimental data.

PASTIS thus provides a statistically grounded, validated, and practical methodological framework for discovering simple models for processes with stochastic dynamics.

\vspace{0.5cm} 


%% file: tex_before/remercie.tex
\addchap{Remerciements}
Le 1er juillet 2025 marque la fin des montagnes russes que fut ma thèse. Ce manège, avec ses vertiges et ses accélérations, aurait sans doute déraillé sans les qualités humaines et scientifiques exceptionnelles de son incroyable co-pilote : mon directeur de thèse, Pierre Ronceray. Un immense merci à toi, Pierre, pour m'avoir maintenu sur les rails tout en me laissant les tracer.

Mes remerciements vont également à l'ensemble des membres de mon jury pour l'intérêt et le temps précieux qu'ils ont accordé à mon travail. Merci à Jean Barbier, Christian L. Müller, Thierry Mora, Simon Gsell et Claude Loverdo. Je suis particulièrement reconnaissant envers les membres de mon comité de suivi, Thierry Mora et Anna Frishman, pour leurs conseils avisés tout au long de ce parcours. Les discussions scientifiques stimulantes ont été un moteur, et je remercie chaleureusement Nicolas Levernier, Florence Bansept, Simon Gsell et Philippe Roudot. Une pensée spéciale pour João Pedro Valeriano Miranda, avec qui j'ai eu des échanges scientifiques profonds.

L'ambiance d'un laboratoire est essentielle, et je voudrais remercier tous ses membres pour leur accueil. Plus particulièrement, merci à Kheya Sengupta et Pierre Müller pour leur constante bienveillance dans mes péripéties. Merci également à Didier Tonneau, Étienne Loiseau, Carole Fauquet, Emmanuèl Helfer, Dominique Chatain et Annie Viallat. La science moderne ne serait rien sans une infrastructure robuste ; merci donc à Andres Saul et Baptiste Desmoulin pour l'entretien méticuleux du cluster de calcul du CINaM.

Cette thèse n'aurait pu voir le jour sans le soutien financier d'institutions françaises et Marseillaise. Je remercie vivement Centuri (Turing Centre for Living Systems) pour le financement de mes trois années de thèse, ainsi que le projet «Anita» d'AMIDEX pour les neuf derniers mois. L'aventure Centuri, d'ailleurs, ne serait pas aussi riche sans son incroyable équipe administrative qui orchestre des événements mémorables. Un grand merci à Marlène Salom, Melina De Oliveira, Pauline Becherel, Rémi Wojciechowski, Simon Legendre et Alizée Guarino. Dans ce quotidien, la gentillesse et le sourire des personnels du Crous ont été de vrais rayons de soleil.

Ces dernières années n'auraient pas eu la même saveur sans les discussions animées, souvent à propos d'urbanisme et du fameux Boulevard Urbain Sud, autour d'une andouillette grillée par le Crous. Merci à Sham Tlili, Marc-Eric Perrin, Martin Lardy, Jules Vanaret et Alice Gros pour ces moments précieux. Après ces festins, j'ai eu la chance de retrouver une ambiance tout aussi chaleureuse au laboratoire grâce à Alice Briole, Pauline Lahure, Marion Marchand, Sacha Soulerin et Zakaria Marmri.

Mon parcours n'aurait sans doute jamais commencé sans le goût pour la science que mes professeurs de classes préparatoires ont su m'insuffler. Merci à Monsieur Minetti, Monsieur Turner, Monsieur Petrov et Monsieur Boulesteix. Ce chemin a ensuite été jalonné de rencontres précieuses durant mes études et expériences en recherche ; merci à Yohann Geli, Ayoub Hihi, Luca Castello, Jérôme Garnier-Brun, Nicolas Dietler et Loïc Marrec pour leur soutien indéfectible. Enfin, il y a les rencontres qui changent une trajectoire. C'est pendant mon Master que j'ai eu la chance de croiser la route de ma très chère Joséphine Baillet. Son soutien a été si déterminant que, sans elle, je n'écrirais pas ces lignes aujourd'hui : j'aurais tout simplement raté la date limite de candidature à mon Master 2. Pour cela et pour tout le reste, merci ma cocotte.

Enfin, et surtout, cette thèse n'aurait tout simplement pas existé sans ma mère, qui m'a accordé une confiance absolue et a sans cesse cultivé ma curiosité par ses questions. Elle n'aurait pas existé non plus sans mon défunt grand-père, qui me répétait au potager qu'il valait mieux travailler avec son cerveau qu'avec ses mains. Je remercie aussi ma grand-mère, pour sa douceur et son incroyable talent à nous unir autour d'un bon repas.

\begin{flushright}
\textit{À ma maman, mon papou et ma mamie.}
\end{flushright}

\newpage

\section*{Code Availability}

The source code developed for the research presented in this manuscript is openly and publicly available.

The primary development repository is hosted on GitHub and can be accessed at:
\url{https://github.com/odannis/PASTIS_PhD}

For long-term preservation and to ensure the specific version of the code used to generate the results in this thesis is permanently accessible, the repository has been archived on Zenodo. The archived version, which corresponds to the analyses and simulations described herein (Version [X.X.X] or commit hash \texttt{[specific-commit-hash]}), is available via the following Digital Object Identifier (DOI):
\url{https://doi.org/10.5281/zenodo.15814187}

\section*{Use of AI tools}
Generative artificial intelligence (AI) tools were employed as an aid during the research and writing process for this thesis. Among the specific uses were checking the written explanation for spelling and grammar mistakes, making it clear and to the point, creating partial code for plotting figures, debugging software implementations, finding relevant literature, and explaining math demonstrations. The core research ideas, methodologies (including the development of the PASTIS criterion), results, and interpretations presented are the original work of the author.

%% file: tex_body/intro.tex
\chapter*{General Introduction: Discovering Governing Equations from Stochastic Dynamics}
\label{chap:general_intro}

The mathematical modeling of complex systems across diverse scientific fields, from physics and biology to finance and climate science, increasingly relies on the framework of stochastic differential equations (SDEs) and stochastic partial differential equations (SPDEs) \cite{gardinerStochasticMethodsHandbook2010, riskenFokkerPlanckEquationMethods1996a}. These tools provide a powerful language for describing systems whose dynamics are influenced not only by deterministic rules but also by inherent randomness or fluctuating environments \cite{amiriInferringGeometricalDynamics2024,andersenModellingHeatDynamics2000,baoStochasticPopulationDynamics2012,bechingerActiveParticlesComplex2016,blackPricingOptionsCorporate1973,bolleyMeanfieldLimitStochastic2012,bricardEmergenceMacroscopicDirected2013,bruckneretal.LearningDynamicsCell,chepizhkoOptimalNoiseMaximizes2013,degondContinuumLimitSelfdriven2008,doi:10.1098/rsos.221177,ferrerTransitionPathTime2024,goldwynStochasticDifferentialEquation2011,gulerStochasticHodgkinHuxleyEquations2013,hasselmannStochasticClimateModels1976,helbingSocialForceModel1995,hozeHeterogeneityAMPAReceptor2012,huangDirectObservationFull2011,hullOptionsFuturesOther2021a,huynhInferringDensitydependentPopulation2023,korbmacherTimeContinuousMicroscopicPedestrian2023,liMeasurementInstantaneousVelocity2010,majdaMathematicalFrameworkStochastic2001,marchettiHydrodynamicsSoftActive2013,mccannThermallyActivatedTransitions1999,palacciLivingCrystalsLightActivated2013,penlandPredictionNino31993,pouwDatadrivenPhysicsbasedModeling2024,preislerModelingAnimalMovements2004,romanczukActiveBrownianParticles2012,singhStochasticDynamicsPredatorprey2021,spagnoloNOISEINDUCEDPHENOMENA2003,teradaLandscapeBasedViewStepping2022,tordeuxStopandgoWavesInduced2020,turkcanBayesianInferenceScheme2012,unamiStochasticDifferentialEquation2010,vandenbergEcologicalModellingApproaches2022,xuProportionalStochasticGeneralized2023,xuStochasticGeneralizedLotkaVolterra2020,yaoStochasticModelKinesin2014,zapien-camposInferringInteractionsMicrobiome2023,zarate-minanoContinuousWindSpeed2013,zottlModelingActiveColloids2023}.
While the foundational theory of SDEs, originating from the study of Brownian motion \cite{alberteinsteinUberMolekularkinetischenTheorie1905, mariansmoluchowskiSMOLUCHOWSKICheminMoyen1906, paullangevinPaulLangevins19081908}, provides a basis for understanding these processes, the \textit{discovery} of governing equations directly from time-series data remains a critical challenge. 

Often, while we might hypothesize a range of potential mechanisms or interactions that could govern a system's evolution, the true underlying dynamics are expected to be \textit{parsimonious} – involving only a subset of these possibilities. This leads to a fundamental problem in data-driven modeling: how can we reliably identify the correct, minimal set of terms describing the system's drift and diffusion from noisy, finite data, especially when faced with a large library of candidate functions representing potential physical laws or interactions? This task falls within the rapidly growing field of data-driven discovery of dynamical systems \cite{gaoDatadrivenInferenceComplex2023, northReviewDataDrivenDiscovery2023}.

Standard statistical approaches to model selection, such as those based on the Akaike Information Criterion (AIC) \cite{akaikeNewLookStatistical1974} or Bayesian Information Criterion (BIC) \cite{schwarzEstimatingDimensionModel1978}, often struggle in this sparse discovery context. While valuable for comparing a small number of pre-defined models, these criteria can be prone to overfitting, selecting overly complex models laden with superfluous terms when searching through large model spaces. This "multiple comparisons" problem fundamentally limits their ability to reliably uncover the true sparse structure underlying stochastic dynamics.

This thesis addresses this critical gap by developing and validating a novel, statistically principled framework for Parsimonious Stochastic Inference, termed PASTIS. Rooted in likelihood-based inference \cite{frishmanLearningForceFields2020} and leveraging insights from extreme value theory \cite{leadbetterExtremesRelatedProperties1983}, PASTIS introduces an information criterion specifically designed to penalize model complexity in a way that explicitly accounts for the size of the candidate function library being searched. By establishing a direct link between the penalty term and a user-defined statistical significance level, PASTIS aims to control the rate of false discoveries, enabling more robust identification of the truly relevant terms governing the system's dynamics.

The work presented herein details the theoretical derivation of the PASTIS criterion, explores its properties, and demonstrates its efficacy through comprehensive benchmarking against established methods (including AIC, BIC, and sparse regression techniques like SINDy \cite{bruntonDiscoveringGoverningEquations2016, kaptanogluPySINDyComprehensivePython2022}) across a variety of SDE systems. Furthermore, the framework is extended to the significantly more challenging domain of identifying sparse SPDEs from spatio-temporal data. Finally, recognizing that real-world data are rarely ideal, we investigate and propose adaptations to the PASTIS methodology to enhance its robustness against common data imperfections, namely finite sampling intervals and the presence of measurement noise.

This thesis is structured as follows:
\begin{itemize}
    \item Part \ref{part:Introdution_langevin} 
    lays the necessary groundwork. Chapters \ref{chap:langevin_intro} and \ref{chap:sde_toolbox} introduce the Langevin equation, SDEs, and the essential tools of stochastic calculus.
    \item Part \ref{part:Learning_Langevin_equation} establishes the techniques for inferring parameters for a \textit{given} SDE or SPDE model, discussing likelihood-based approaches and estimators robust to data imperfections.
    \item Part \ref{part:Model_selection} focuses on the core contribution: sparse model selection. Chapter \ref{chap:likelihood_aic} introduces likelihood-based inference in the context of model selection and highlights the limitations of standard criteria like AIC and BIC. 
    Chapter \ref{chap:pastis} derives the PASTIS information criterion using extreme value statistics. 
    Chapter \ref{chap:validation} presents extensive validation of PASTIS on benchmark SDE systems. Chapter \ref{chap:pastis_spde} extends the PASTIS framework to SPDE discovery. 
    Chapter \ref{chap:robustness_sde} details the adaptation of PASTIS for robustness against non-ideal data. Finally, Chapter \ref{chap:part2_discussion_conclusion} provides a concluding discussion for this part, summarizing the advantages and limitations of the PASTIS framework.
\end{itemize}
Through this work, we aim to provide a robust and practical tool for uncovering the fundamental equations governing complex stochastic phenomena directly from data.

\begin{figure}[h!]
    \centering
    \includegraphics[width=\textwidth]{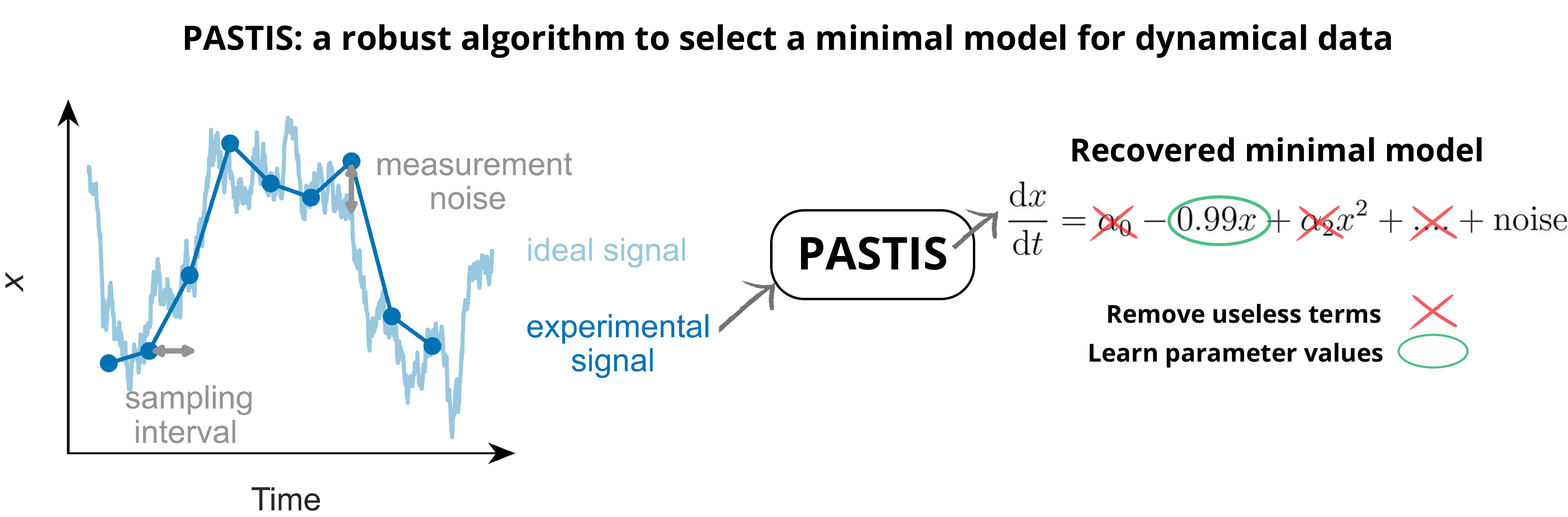}
    \caption{Visual Abstract of the PASTIS method.}
    \label{fig:visual_abstract}
\end{figure}

%% file: tex_body/chap1_a.tex
\part{Introduction to the Langevin equation}
\label{part:Introdution_langevin}

\chapter{An Introduction to the Langevin Equation and Stochastic Calculus}

\epigraph{%
“Life is far too important a thing ever to talk seriously about.”
}{Oscar Wilde, Lady Windermere's Fan}

\chaptertoc{} 
\label{chap:langevin_intro}

\section{The Birth of the Langevin Equation}
\label{sec:birth_langevin} 

\subsection{Emergence of a New Mathematical Object from Observations of Brownian Motion}
\label{subsec:observations_brownian} 

\begin{figure}[htbp] 
    \centering
    \subfloat[Putty grain trajectories]{\includegraphics[width=.5\textwidth, max height=2in]{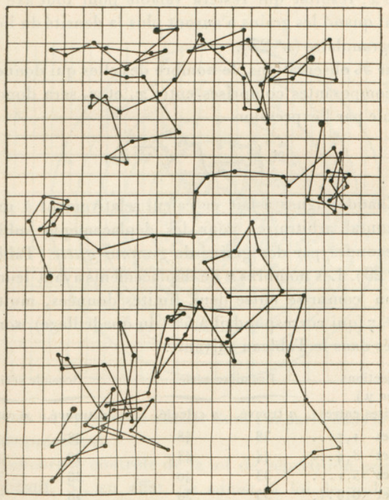}\label{fig:Brownian_motion_perrin}} 
    \qquad 
    \subfloat[Ink diffusion in water]{\includegraphics[width=.5\textwidth, max height=2in]{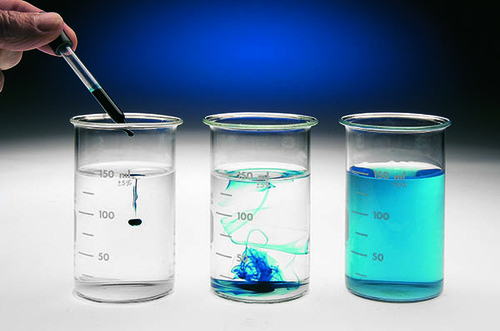}\label{fig:Brownian_motion_ink}} 
    \caption{(a) From Jean Perrin's 1909 work \cite[p. 81]{jeanperrinMouvementBrownienRealite1909}: Three tracings show the motion of a putty grain (radius \SI{0.53}{\micro\metre}) observed under a microscope. Successive positions recorded every 30 seconds are connected by straight lines (grid mesh size is \SI{3.2}{\micro\metre}). (b) Diffusion of an ink drop in a beaker of water, a macroscopic consequence of Brownian motion and convection (Credit: McGraw-Hill/Charles D. Winters).}
    \label{fig:Brownian_motion_exemple}
\end{figure}

The study of microscopic dynamics often begins with Brownian motion: the incessant, seemingly random jiggling of microscopic particles suspended in a fluid (liquid or gas). While the particles themselves are typically too small to see with the naked eye, their erratic dance becomes strikingly apparent under a microscope. \Cref{fig:Brownian_motion_exemple}(a) captures this, showing trajectories recorded by Jean Perrin in 1909. This microscopic chaos has macroscopic consequences; the gradual spreading of a drop of ink in water, depicted in \cref{fig:Brownian_motion_exemple}(b), is a visible manifestation driven by the underlying Brownian motion of countless ink particles (alongside convection).

The phenomenon owes its name to the Scottish botanist Robert Brown, who, in 1827, provided the first detailed scientific account while observing pollen grains suspended in water. He meticulously documented the motion but could not determine its origin.

Explaining the physical mechanism behind this ceaseless motion became a major scientific puzzle. A pivotal breakthrough arrived in 1905 when Albert Einstein, building upon the kinetic theory of heat, developed a purely theoretical description \cite{alberteinsteinUberMolekularkinetischenTheorie1905}. He derived a quantitative relationship linking the particle's \textbf{mean squared displacement (MSD)} to observable macroscopic quantities:
\begin{equation}
    \text{MSD} = \E{(x(t) - x(0))^2} = 2 D t
    \label{eq:chap_1_einstein_relation}
\end{equation}
Here, $\E{\cdot}$ denotes the statistical average, $x(t)$ is the particle's position at time $t$, and $D$ is the \textbf{diffusion coefficient}. Einstein further related $D$ to the thermal energy and fluid properties via the Stokes-Einstein relation: $D = k_B T \mu$, where $k_B$ is Boltzmann's constant, $T$ is the temperature, and $\mu$ is the particle's \textbf{mobility} (the inverse of the friction coefficient, $\gamma = \mu^{-1}$). For a spherical particle of radius $a$ in a fluid with viscosity $\eta$, Stokes' law gives the friction coefficient as $\gamma = 6 \pi \eta a$, so $\mu = (6 \pi \eta a)^{-1}$. Einstein's theory provided a powerful connection between the microscopic random walk and macroscopic thermodynamics, but precise experimental verification was initially lacking.

Independently, in 1906, Marian Smoluchowski approached the problem from a different statistical perspective, also deriving results consistent with a diffusive process, although some initial calculations yielded slightly different numerical factors compared to Einstein's specific predictions \cite{mariansmoluchowskiSMOLUCHOWSKICheminMoyen1906}. These theoretical advances spurred the need for careful experimental tests.

The decisive experimental confirmation came from the meticulous work of Jean Perrin around 1909 \cite{jeanperrinMouvementBrownienRealite1909}. His experiments validated Einstein's MSD relation \cref{eq:chap_1_einstein_relation} with remarkable accuracy. Furthermore, by measuring the MSD and knowing the other parameters, Perrin was able to determine Avogadro's number $N_A$ ($k_B = R/N_A$, where $R$ is the gas constant), providing compelling quantitative evidence for the existence of atoms and molecules. This groundbreaking work earned Jean Perrin the Nobel Prize in Physics in 1926.

Amidst these developments, in 1908, Paul Langevin proposed a conceptually different, strikingly direct approach to derive Einstein's result \cite{paullangevinPaulLangevins19081908}. Instead of focusing solely on statistical descriptions, Langevin applied Newton's second law ($F=ma$) directly to the Brownian particle. His crucial insight was to model the total force on the particle as the sum of two components: a systematic frictional drag force opposing the particle's velocity, and a rapidly fluctuating \textbf{random force} representing the incessant, unbalanced impacts from the surrounding fluid molecules:
\begin{equation}
    m\dv[2]{x}{t} = \underbrace{- \gamma \dv{x}{t}}_{\text{Frictional force}} + \underbrace{ F_{\text{random}}(t) }_{\text{Random force}}
    \label{eq:langevin_equation_conceptual}
\end{equation}
Here, $m$ is the particle's mass, and $\gamma = \mu^{-1}$ is the friction coefficient. Langevin characterized $F_{\text{random}}(t)$ as being "sometimes positive, sometimes negative" and having a magnitude sufficient to counteract the damping effect of friction, thus maintaining the particle's thermal agitation \cite{paullangevinPaulLangevins19081908}. He argued that because the molecular impacts are extremely rapid and random compared to the particle's velocity changes, this random force should be statistically uncorrelated with the particle's position $x(t)$. Hence, his method relied on the assumption that the expectation of the product of the particle's velocity and this random force is zero. By multiplying \cref{eq:langevin_equation_conceptual} by $x$, using integration by parts, averaging, and invoking the equipartition theorem of statistical mechanics ($\E{\frac{1}{2} m v^2} = \frac{1}{2} k_B T$ for thermal equilibrium in one dimension), Langevin elegantly recovered Einstein's relation \cref{eq:chap_1_einstein_relation}.

To achieve this consistency, the statistical properties of $F_{\text{random}}(t)$ must be precisely linked to the friction $\gamma$ and temperature $T$. This requirement dictates the magnitude of the random force. It is common today to express the random force in terms of a standardized noise term $\xi_t$, leading to the modern form of the \textbf{Langevin equation}:
\begin{equation}
    m\dv[2]{x}{t} = - \gamma \dv{x}{t} +  \gamma \sqrt{2 D} \xi_t 
    \label{eq:langevin_equation}
\end{equation}
Langevin's work was pioneering in its explicit inclusion of a stochastic forcing term within a dynamical equation based on physical laws. Consequently, differential equations featuring such random terms and derived from physical principles are often referred to as \textbf{Langevin equations} in physics. From a mathematical perspective, they fall under the broader category of \textbf{Stochastic Differential Equations (SDEs)}.

While Langevin's physical intuition about $F_{\text{random}}(t)$ was powerful, its precise mathematical nature remained undefined. What mathematical properties must this "random force" possess to be consistent with observations? This question motivates the need for a more formal mathematical description. Thus, we will first look at a mathematical definition motivated by physical arguments.

\subsection{Mathematical Formalism for the Physicists' Random Term $\xi_t$}
\label{subsec:physicist_interpretation} 

We now delve deeper into the properties required of the normalized random force $\xi_t$, guided by physical considerations and the known behavior of Brownian motion, particularly: (1) its prominence for microscopic particles where fluid friction is significant, and (2) the linear-in-time growth of the MSD \cref{eq:chap_1_einstein_relation}.

From (1), a crucial simplification arises, which is called \textbf{the overdamped limit}, relevant for many systems like colloidal particles in liquids. Here, viscous damping forces ($\propto \gamma \propto a$) dominate inertial forces ($m \propto a^3$) because the particle's mass is very small. On the characteristic timescale of velocity relaxation ($\tau_v = m/\gamma$), the acceleration term $m\dv[2]{x}{t}$ becomes negligible compared to the frictional term $-\gamma \dv{x}{t}$.

Setting the inertial term to zero in \cref{eq:langevin_equation} yields the \textbf{overdamped Langevin equation}:
\begin{equation}
    \gamma \dv{x}{t} = \sqrt{2 \gamma k_B T} \xi_t \implies \dv{x}{t} = \sqrt{2 (\gamma^{-1}) k_B T} \xi_t = \sqrt{2 \mu k_B T} \xi_t = \sqrt{2D} \xi_t
    \label{eq:chap_1_langevin_overdamped_equation_deriv}
\end{equation}
This can be written in differential form as:
\begin{equation}
    \dd{x} = \sqrt{2D} \xi_t \dd{t}
    \label{eq:chap_1_langevin_overdamped_equation}
\end{equation}
In this picture, $\xi_t$ represents the effective, normalized random impulse delivered by the countless fluid molecule collisions over an infinitesimal time interval $\dd{t}$. Because this net impulse results from a vast number of quasi-independent molecular impacts, the Central Limit Theorem suggests that its integral over any finite time interval should be Gaussian distributed. Furthermore, by symmetry (no preferred direction for the net impacts), the average force must be zero:
\begin{equation}
    \E{\xi_t} = 0
\end{equation}
Now consider the temporal correlations. The molecular collisions at time $t$ are essentially independent of those at a different time $t'$, provided $|t-t'|$ is larger than the extremely short duration of molecular collision events. This strongly suggests that the random force has a very short memory, ideally zero for $t \neq t'$. To quantify this, we examine what correlation structure is needed to recover Einstein's MSD relation from the overdamped equation \cref{eq:chap_1_langevin_overdamped_equation}. Integrating \cref{eq:chap_1_langevin_overdamped_equation} gives $x(t) - x(0) = \int_0^t \sqrt{2D} \xi_s \dd{s}$. Squaring and taking the expectation yields:
\begin{equation}
    \E{(x(t) - x(0))^2} = 2D \int_0^t \int_0^t \E{\xi_s \xi_{s'}} \dd{s} \dd{s'}
    \label{eq:chap_1_msd_from_overdamped}
\end{equation}
Comparing this with the target result $\E{(x(t) - x(0))^2} = 2 D t$, we require the double integral to equal $t$:
\begin{equation}
   \int_0^t \int_0^t  \E{\xi_s \xi_{s'}} \dd{s} \dd{s'} =  t
   \label{eq:chap_1_int_noise}
\end{equation}
The simplest mathematical object satisfying both the requirement of zero correlation for $s \neq s'$ and this integral identity (\cref{eq:chap_1_int_noise}) is the \textbf{Dirac delta function}:
\begin{equation}
    \E{\xi_s \xi_{s'}} = \delta(s-s')
    \label{eq:chap_1_delta_correlation}
\end{equation}
This delta correlation precisely yields $\int_0^t \int_0^t \delta(s-s') \dd{s} \dd{s'} = \int_0^t \dd{s} = t$.

Thus, the physicist's operational picture of $\xi_t$ is that of \textbf{Gaussian white noise}: a stationary stochastic process characterized by a Gaussian distribution at any time $t$, zero mean ($\E{\xi_t} = 0$), and a delta-function autocorrelation ($\E{\xi_s \xi_{s'}} = \delta(s-s')$).

While intuitive and useful, this description presents profound mathematical difficulties. A process with delta-correlated noise intrinsically has infinite variance ($\E{\xi_t^2} \rightarrow \infty$ as suggested by $\delta(0)$), which is physically unrealistic and mathematically problematic. Furthermore, the trajectory $x(t)$ resulting from \cref{eq:chap_1_langevin_overdamped_equation}, while continuous, turns out to be nowhere differentiable. This lack of a well-defined derivative $\dd{x}/\dd{t}$ echoes Jean Perrin's own remarks on the seemingly "pathological" nature of the experimentally observed Brownian paths\footnote{Perrin noted the jaggedness of the paths, remarking on how mathematical constructs like curves without tangents at every point seemed closer to reality than idealized smooth curves \cite{jeanperrinMouvementBrownienRealite1909}.}. Interpreting equations like \cref{eq:chap_1_langevin_overdamped_equation} requires a more sophisticated mathematical framework: the Wiener process.

\subsection{A Rigorous Mathematical Formulation: The Wiener Process}
\label{subsec:wiener_process}

While the physical description of Langevin captured the essence of Brownian motion, and Perrin provided compelling experimental evidence, the mathematical nature of the "random force" or "white noise" remained problematic from a rigorous standpoint. The concept of a function that is continuous everywhere but differentiable nowhere challenged classical calculus. Furthermore, defining integrals involving the highly irregular white noise required new mathematical tools. Norbert Wiener, working in the early 1920s, sought to address these challenges. His goal was to construct a rigorous mathematical model for such paths. This endeavor was pivotal not only for understanding Brownian motion itself but also for laying the foundations of the modern theory of stochastic processes and stochastic calculus.

Central to Wiener's framework is the \textbf{Wiener process} (or Brownian motion), denoted here as $W_t$, which is defined by the following properties:
\begin{enumerate}
    \item $W_0 = 0$ (starts at the origin almost surely).
    \item Independent increments: For any time sequence $0 \le t_0 < t_1 < \dots < t_n$, the random variables (increments) $W_{t_1} - W_{t_0}, W_{t_2} - W_{t_1}, \dots, W_{t_{n}} - W_{t_{n-1}}$ are mutually independent.
    \item Gaussian increments: For any $0 \le t_0 < t_1$, the increment $W_{t_1} - W_{t_0}$ follows a normal (Gaussian) distribution with mean 0 and variance $t_1 - t_0$, denoted as $\mathcal{N}(0, t_1 - t_0)$.
    \item Continuous paths: The function $t \mapsto W_t$ is continuous in $t$ (almost surely).
\end{enumerate}
We use $W_t$ to represent the mathematical construct, distinguishing it from the physical phenomenon of Brownian motion itself.

Using the Wiener process, the overdamped Langevin equation \cref{eq:chap_1_langevin_overdamped_equation} is rigorously rewritten in differential form as:
\begin{equation}
    \dd{x_t} = \sqrt{2 D} \, \dd{W_t}
    \label{eq:chap_1_math_brownian_motion}
\end{equation}
Here, $\dd{x_t}$ is shorthand for $x_{t+\dd{t}} - x_t$, and $\dd{W_t}$ represents the infinitesimal increment $W_{t+\dd{t}} - W_t$, which has mean 0 and variance $\dd{t}$. Crucially, \cref{eq:chap_1_math_brownian_motion} is understood as a symbolic representation of the corresponding \textbf{stochastic integral equation}:
\begin{equation}
    x(t) - x(0) = \int_0^t \sqrt{2D} \, \dd{W_s}
    \label{eq:chap_1_ito_integral_simple}
\end{equation}
This integral, known as the \textbf{Itô stochastic integral}, is defined formally as a limit of sums over partitions of the interval $[0, t]$. For a partition $0 = s_0 < s_1 < \dots < s_n = t$:
\begin{equation}
    \int_0^t \sqrt{2D} \, \dd{W_s} = \lim_{n \to \infty} \sum_{i=0}^{n-1} \sqrt{2D} \left( W_{s_{i+1}} - W_{s_i} \right)
    \label{eq:chap_1_ito_sum_definition}
\end{equation}
where the limit is taken such that the mesh size $\max(s_{i+1}-s_i) \to 0$. Using the properties of the Wiener process (independence and variance of increments), we can rigorously calculate the MSD:
\begin{align}
    \E{[x(t) - x(0)]^2} &= \E{\left( \int_0^t \sqrt{2D} \, \dd{W_s} \right)^2} \nonumber \\
    &= 2 D \, \E{\left( \lim_{n \to \infty} \sum_{i=0}^{n-1} (W_{s_{i+1}} - W_{s_i}) \right)^2} \nonumber \\
    &= 2 D \, \lim_{n \to \infty} \E{\left( \sum_{i=0}^{n-1} (W_{s_{i+1}} - W_{s_i}) \right)^2} \quad \text{(Interchange limit and expectation)} \nonumber \\
    &= 2 D \, \lim_{n \to \infty} \sum_{i=0}^{n-1} \E{(W_{s_{i+1}} - W_{s_i})^2} \quad (\E{\Delta W_i \Delta W_j} = 0 \text{ for } i \neq j) \nonumber \\
    &= 2 D \, \lim_{n \to \infty} \sum_{i=0}^{n-1} (s_{i+1} - s_i) \quad (\text{Since } \E{(\Delta W_i)^2} = \Delta s_i = s_{i+1}-s_i) \nonumber \\
    &= 2 D \, \lim_{n \to \infty} (s_n - s_0) = 2 D (t - 0) = 2 D t
\end{align}
This derivation formally reproduces Einstein's result \cref{eq:chap_1_einstein_relation} within a mathematically sound framework.

Both the physicist's formulation \cref{eq:chap_1_langevin_overdamped_equation} using delta-correlated white noise $\xi_t$ and the mathematician's formulation \cref{eq:chap_1_math_brownian_motion} using the Wiener process differential $\dd{W_t}$ effectively describe the same physical process in the overdamped limit. They yield consistent results for quantities like the MSD. The persistence of both notations often reflects the conventions and historical backgrounds of different scientific communities. Formally, the relationship is often expressed heuristically as:
\begin{equation}
    \dd{W_t} \coloneqq \xi_t \dd{t}
    \label{eq:dw_xi_heuristic}
\end{equation}
This heuristic relationship emphasizes that the Wiener increment $\dd{W_t}$ captures the integrated effect of the idealized white noise $\xi_t$ over an infinitesimal time interval $\dd{t}$. It highlights that $\dd{W_t}$ scales as $\sqrt{\dd{t}}$ in magnitude (due to $\E{(\dd{W_t})^2} = \dd{t}$), unlike ordinary differentials which scale as $\dd{t}$. In the remainder of this manuscript, we will primarily adopt the $\dd{W_t}$ notation, grounding our discussion in the rigorous framework of stochastic calculus based on the Wiener process. Next, we will show how this knowledge is applied to real-world systems.

\section{The General Overdamped Langevin Equation (SDE)}
\label{sec:general_overdamped_sde}

\begin{figure}[htbp] 
    \centering
    \includegraphics[width=0.9\linewidth]{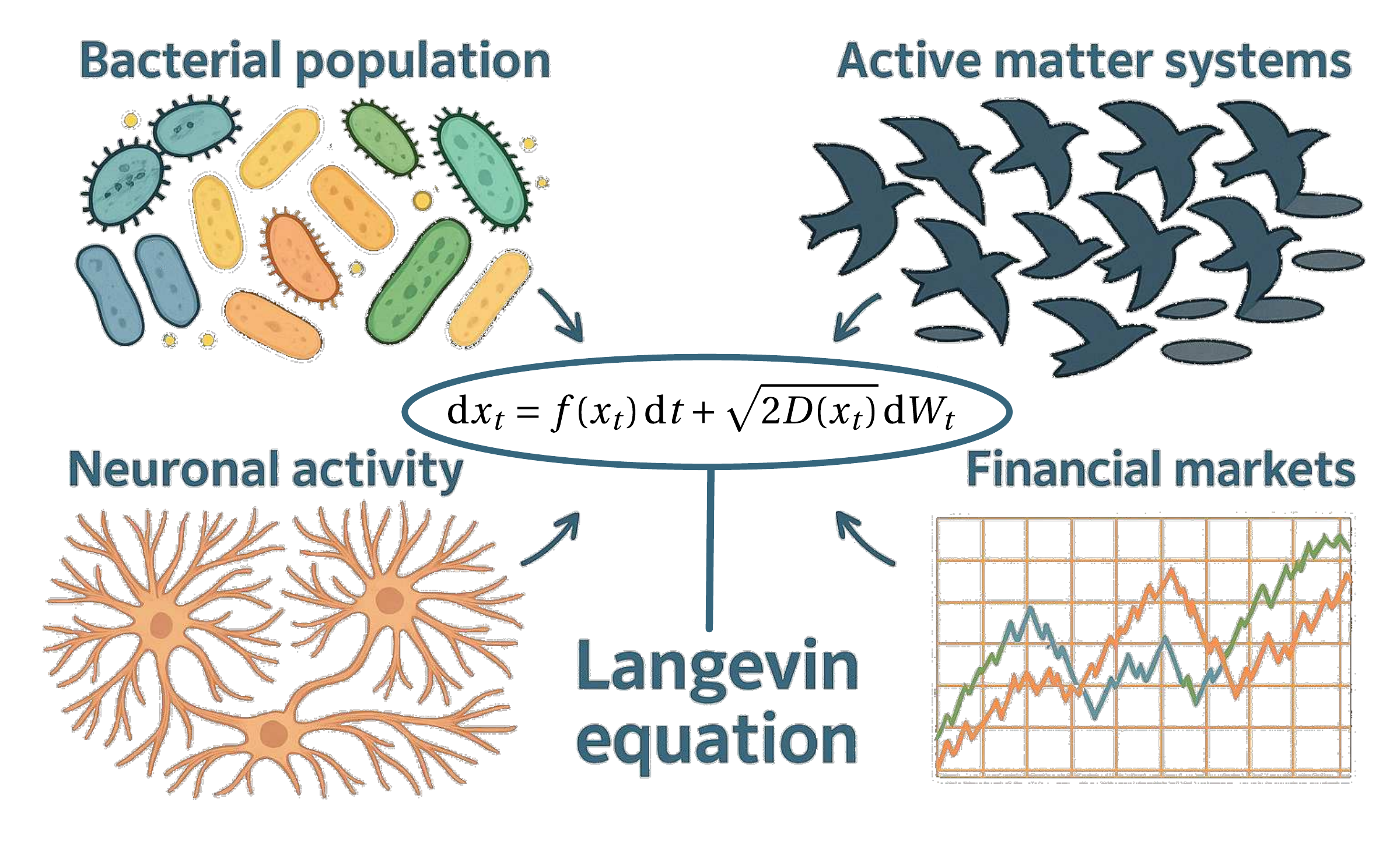}
    \caption{The general overdamped Langevin equation, $\dd{\bm{x}_{t}}=\bm{f}(\bm{x}_{t}, t)\dd{t}+\sqrt{2\bm{D}(\bm{x}_{t}, t)}\dd{\bm{W}_{t}}$, provides a powerful framework for modeling diverse systems exhibiting stochastic dynamics. Examples shown span fields such as bacterial population dynamics, active matter (flocking/swarming), neuronal modeling, and financial market analysis. Created with the help of Sora.} 
    \label{fig:Langevin_applications} 
\end{figure}

Real-world systems rarely involve only random fluctuations; they are typically subject to deterministic forces as well. The mathematical framework developed for Brownian motion can be readily generalized to model such dynamics through the \textbf{general (overdamped) Langevin equation}, more formally known as a \textbf{Stochastic Differential Equation (SDE)} of the Itô type:
\begin{equation}
    \dd{\bm{x}_t} = \underbrace{\bm{f}(\bm{x}_t)}_{\text{Drift}}\dd{t} + \underbrace{\bm{g}(\bm{x}_t) \, \dd{\bm{W}_t}}_{\text{Noise}}
    \label{eq:chap_1_the_overdamped_langevin_general}
\end{equation}
Here:
\begin{itemize}
    \item $\bm{x}_t$ is a vector representing the state of the system (e.g., position, concentration, price) at time $t$. It can reside in $\mathbb{R}^d$.
    \item $\bm{f}(\bm{x}_t)$ is a vector-valued function called the \textbf{drift term} or \textbf{drift vector field}. It represents the deterministic part of the dynamics (e.g., forces divided by friction, mean reaction rates, expected trends). It maps from the state space (and possibly time) to $\mathbb{R}^d$.
    \item $\bm{W_t}$ is a vector of $d$ independent standard Wiener processes, $\bm{W_t} = (W_t^1, \dots, W_t^d)^T$.
    \item $\bm{g}(\bm{x}_t)$ is a $d \times d$ matrix-valued function, often referred to as the \textbf{diffusion term} or \textbf{diffusion matrix} (though strictly speaking, the diffusion matrix is $\bm{D} = \frac{1}{2} \bm{g} \bm{g}^T$). It determines how the different noise components influence the state variables.
    \item The term $\bm{g}(\bm{x}_t) \dd{\bm{W}_t}$ represents the \textbf{stochastic} or \textbf{noise term}. If $\bm{g}$ depends explicitly on $\bm{x}_t$, the noise is termed \textbf{multiplicative}; if $\bm{g}$ is constant (or only depends on $t$), the noise is \textbf{additive}.
\end{itemize}

The notation $\sqrt{2\bm{D}}$ used previously (e.g., in the caption of \cref{fig:Langevin_applications} and \cref{eq:langevin_equation}) corresponds to choosing $\bm{g}$ such that $\bm{g}\bm{g}^T = 2\bm{D}$. This requires $\bm{D}$ to be positive semi-definite, which is physically necessary as it relates to the covariance matrix of the noise increments.

\Cref{eq:chap_1_the_overdamped_langevin_general} is, like its simpler counterpart for Brownian motion \cref{eq:chap_1_math_brownian_motion}, shorthand for the integral equation:
\begin{equation}
    \bm{x}_\tau - \bm{x}_0 = \int_0^\tau \bm{f}(\bm{x}_t) \dd{t} + \int_0^\tau \bm{g}(\bm{x}_t) \dd{\bm{W_t}}
    \label{eq:sde_integral_form}
\end{equation}
The stochastic integral $\int \dots \dd{\bm{W_t}}$ is interpreted in the \textbf{Itô sense}. This means the stochastic part is defined as the limit of sums where the integrand ($\bm{g}$) is evaluated at the beginning of each small time interval $[t_i, t_{i+1}]$:
\begin{equation}
\int_0^\tau \bm{g}(\bm{x}_t) \dd{\bm{W_t}} \approx \sum_{i} \bm{g}(\bm{x_{t_i}}) \left(\bm{W_{t_{i+1}}} - \bm{W_{t_{i}}}\right)
\end{equation}
This non-anticipating definition is often preferred in physical modeling as it reflects causality (the system's response at time $t$ depends on the noise integrated up to time $t$, not future noise).

While originating from the physics of Brownian motion (where $\bm{x}_t$ is the position, $\bm{f}$ relates to the deterministic forces, and $\bm{g}$ relates to the temperature and friction), the SDE framework \cref{eq:chap_1_the_overdamped_langevin_general} has proven remarkably versatile. It is now widely employed as a powerful empirical and theoretical modeling tool across diverse scientific domains. In many contemporary applications, the specific forms of the drift $\bm{f}(\bm{x}_t)$ and diffusion $\bm{g}(\bm{x}_t)$ terms are determined based on phenomenological understanding, data analysis, or specific modeling assumptions, rather than being strictly derived from first physical principles.

Examples of this broad applicability are illustrated in \cref{fig:Langevin_applications}. More precisely, these include:

\begin{itemize}
    \item \textbf{Active Matter Physics:} SDEs model the collective behavior of self-propelled agents (e.g., bird flocks, bacterial colonies, synthetic swimmers), exemplified by models like the Vicsek model \cite{vicsekNovelTypePhase1995}, which was originally formulated as a discrete-time stochastic model. In continuous-time frameworks related to such models, $\bm{x}_t$ typically describes positions and orientations. The drift $\bm{f}$ captures self-propulsion and alignment interactions, while the noise term $\bm{g}\dd{\bm{W}_t}$ represents randomness in individual decisions or environmental factors \cite{preislerModelingAnimalMovements2004,romanczukActiveBrownianParticles2012,degondContinuumLimitSelfdriven2008,bolleyMeanfieldLimitStochastic2012,marchettiHydrodynamicsSoftActive2013,chepizhkoOptimalNoiseMaximizes2013}.

    \item \textbf{Quantitative Finance:} Asset prices (like stock prices $S_t$) are frequently modeled using geometric Brownian motion, a specific type of SDE \cite{hullOptionsFuturesOther2021a}. In this case, the drift $\bm{f}$ reflects the expected growth rate (return), and the multiplicative noise $\bm{g}$ represents market volatility. This SDE underpins the famous \textbf{Black-Scholes-Merton} option pricing formula \cite{blackPricingOptionsCorporate1973}, a cornerstone of financial engineering recognized with the Nobel Prize in Economic Sciences in 1997.

    \item \textbf{Population Dynamics:} SDEs provide a powerful framework for modeling the fluctuating evolution of population sizes—ranging from bacteria and animals to plants, cell populations, and entire ecological communities. The drift term ($\bm{f}$) typically describes the average or deterministic population dynamics. This often draws inspiration from classical Ordinary Differential Equation (ODE) models, such as logistic growth for density-dependent regulation or the Lotka-Volterra equations governing predator-prey and competitive interactions. The crucial addition in the SDE framework is the random term ($\bm{g}\dd{\bm{W}_t}$), which explicitly incorporates stochasticity. This accounts for random fluctuations arising from demographic noise (inherent randomness in individual birth and death events) and/or environmental noise (unpredictable changes in external factors affecting the population) \cite{spagnoloNOISEINDUCEDPHENOMENA2003,baoStochasticPopulationDynamics2012,xuStochasticGeneralizedLotkaVolterra2020,singhStochasticDynamicsPredatorprey2021,vandenbergEcologicalModellingApproaches2022,huynhInferringDensitydependentPopulation2023,zapien-camposInferringInteractionsMicrobiome2023,xuProportionalStochasticGeneralized2023}.

    \item \textbf{Climate Science:} SDEs can model the variability of climate indices (e.g., El Niño Southern Oscillation) or physical variables. The drift $\bm{f}$ may capture seasonal cycles or long-term trends forced by external factors, while the random term $\bm{g}\dd{\bm{W}_t}$ represents unpredictable short-term weather fluctuations and internal variability within the climate system \cite{hasselmannStochasticClimateModels1976,penlandPredictionNino31993,andersenModellingHeatDynamics2000,majdaMathematicalFrameworkStochastic2001,unamiStochasticDifferentialEquation2010,zarate-minanoContinuousWindSpeed2013}.

    \item \textbf{Colloidal Systems:} SDEs are essential tools for modeling the dynamics of colloidal particles, both passive and active.
         \begin{itemize}
             \item For \textbf{passive colloids}, whose motion is driven by thermal energy and external influences, SDEs describe their Brownian motion within potential landscapes \cite{paullangevinPaulLangevins19081908,liMeasurementInstantaneousVelocity2010,huangDirectObservationFull2011}. The drift term ($\bm{f}$) captures deterministic forces arising from external fields (e.g., optical traps, gravity, flow gradients) or inter-particle interactions (e.g., electrostatic, depletion). The random term ($\bm{g}\dd{\bm{W}_t}$) represents the random displacements due to thermal fluctuations, typically linked to the solvent viscosity and particle friction via the fluctuation-dissipation theorem \cite{liMeasurementInstantaneousVelocity2010,mccannThermallyActivatedTransitions1999}.
             \item For \textbf{active colloids} (e.g., self-propelled Janus particles, active Brownian particles), which convert stored or ambient energy into directed motion, SDEs model their persistent random walks and complex collective behaviors \cite{bechingerActiveParticlesComplex2016,zottlModelingActiveColloids2023,ferrerTransitionPathTime2024,bricardEmergenceMacroscopicDirected2013,palacciLivingCrystalsLightActivated2013}. In this case, the SDE framework often includes the self-propulsion velocity within the drift term $\bm{f}$ (alongside other deterministic forces), while the noise term incorporates both thermal Brownian motion and additional 'active noise' sources, such as the stochastic reorientation dynamics.
         \end{itemize}

    \item \textbf{Cell Biology and Biophysics:} SDEs model numerous intrinsically stochastic processes at the molecular and cellular level:
         \begin{itemize}
             \item \textbf{Diffusion and interactions of molecules} (e.g., receptors, lipids) within cell membranes. While influenced by thermal Brownian motion, this process is often more complex due to membrane compartmentalization, binding kinetics, and interactions with the cytoskeleton \cite{hozeHeterogeneityAMPAReceptor2012,turkcanBayesianInferenceScheme2012,doi:10.1098/rsos.221177}.
             \item \textbf{Motion of molecular motors} (like kinesin or myosin) along cytoskeletal filaments, driven by energy consumption (e.g., ATP hydrolysis) and subject to thermal noise and stochastic detachment/attachment events \cite{yaoStochasticModelKinesin2014,teradaLandscapeBasedViewStepping2022}.
             \item \textbf{Stochastic trajectories and dynamics of migrating cells}. These are typically modeled using overdamped SDEs or persistent random walk models that incorporate internal cell polarity dynamics and responses to chemotactic cues, reflecting the highly damped nature of cell movement in tissues \cite{bruckneretal.LearningDynamicsCell,amiriInferringGeometricalDynamics2024}.
             \item The behavior of \textbf{semi-flexible polymers} (like DNA or actin filaments), where SDEs describe the fluctuating dynamics of the chain's conformation or tangent vector under tension and thermal forces \cite{winklerActivePolymersEmergent2017}.
         \end{itemize}
         In these biological contexts, the drift $\bm{f}$ represents systematic influences (potential gradients, active propulsion forces, directed motility cues, biochemical reaction rates), while $\bm{g}\dd{\bm{W}_t}$ captures not only thermal fluctuations but also inherent biological stochasticity originating from sources like gene expression noise, random channel openings, or the probabilistic nature of molecular interactions and motor stepping.

    \item \textbf{Neuroscience:} SDEs capture neuronal membrane-potential dynamics. The drift $\bm{f}$ encodes deterministic ion currents and synaptic inputs, while the random term $\bm{g}\,\mathrm d\bm{W}_t$ represents channel noise and random synaptic inputs \cite{goldwynStochasticDifferentialEquation2011,gulerStochasticHodgkinHuxleyEquations2013}.

    \item \textbf{Pedestrian Dynamics:} SDEs can describe individual movement in crowds. The drift $\bm{f}$ represents desired velocity and collision avoidance maneuvers, while the noise term $\bm{g}\dd{\bm{W}_t}$ accounts for fluctuations in walking speed, deviations from the intended path, and unpredictable decision-making \cite{pouwDatadrivenPhysicsbasedModeling2024,helbingSocialForceModel1995,korbmacherTimeContinuousMicroscopicPedestrian2023,tordeuxStopandgoWavesInduced2020}.
\end{itemize}

This extensive array of examples underscores the remarkable adaptability and power of the general overdamped Langevin equation. By providing a flexible framework to incorporate both deterministic trends and stochastic fluctuations, SDEs have become an indispensable tool for quantitative modeling across an impressive spectrum of scientific and engineering disciplines. From the microscopic colloids to the macroscopic dynamics of populations, financial markets, and even climate patterns, the SDE framework allows researchers to capture and analyze complex, noise-influenced behaviors. However, the very nature of these equations, driven by the erratic Wiener process, means that their analysis and manipulation demand a distinct set of mathematical techniques beyond those of classical calculus.

Working with SDEs requires specialized mathematical tools, distinct from standard calculus, due to the non-differentiable nature of the Wiener process. The next section, titled 'SDE Toolbox', introduces fundamental tools for working with SDEs.

\label{chap:end_langevin_intro} 

%% file: tex_body/chap1_b.tex
\chapter{The SDE Toolbox}
\chaptertoc{} 
\label{chap:sde_toolbox}

\section{Introduction}
\label{sec:sde_toolbox_intro}

This chapter delves into the essential mathematical framework required for analyzing Stochastic Differential Equations (SDEs), such as those introduced previously. We will explore key concepts from Itô calculus, starting with the Itô isometry for calculating expectations of stochastic integrals. We then introduce the cornerstone of this calculus, Itô's formula (or Itô's Lemma), which provides the correct chain rule for functions of stochastic processes. Finally, we discuss the different conventions for defining stochastic integrals – Itô and Stratonovich – highlighting their relationship and respective advantages, while establishing the \textbf{Itô convention as the standard for this manuscript}.

This chapter introduces fundamental results and techniques from stochastic calculus essential for analyzing SDEs such as \cref{eq:chap_1_the_overdamped_langevin_general}. For simplicity, we will focus primarily on the one-dimensional Itô SDE:
\begin{equation}
    \dd{x_t} = f(x_t) \dd{t} + g(x_t) \dd{W_t}
    \label{eq:chap_1_overdamped_1d} 
\end{equation}
where $f(x_t)$ is the drift term, $g(x_t)$ is the diffusion function (potentially state-dependent, often related to noise strength), and $W_t$ is a standard Wiener process. We assume $f$ and $g$ are sufficiently smooth functions for the operations below to be valid.

\section{Itô Isometry}
\label{sec:ito_isometry}

The Itô isometry is a crucial property relating the expectation of the product of two Itô integrals to the expectation of an ordinary integral involving the product of the integrands. Let $h_1(x_t)$ and $h_2(x_t)$ be two suitable functions of the stochastic process $x_t$. The Itô isometry states:
\begin{equation}
    \E{\left(\int_0^\tau h_1(x_t) \dd{W_t}\right) \left(\int_0^\tau h_2(x_t) \dd{W_t}\right)} = \E{\int_0^\tau h_1(x_t) h_2(x_t) \dd{t}}
    \label{eq:ito_isometry}
\end{equation}
A special case, often used for calculating variances (second moments), arises when $h_1=h_2=h$:
\begin{equation}
    \E{\left(\int_0^\tau h(x_t) \dd{W_t}\right)^2} = \E{\int_0^\tau h(x_t)^2 \dd{t}}
    \label{eq:ito_isometry_variance}
\end{equation}

\subsection*{Proof Sketch:}
We start with the limit definition of the Itô integrals:
\begin{equation}
    \E{\left(\lim_{n\to \infty} \sum_{i=0}^{n - 1} h_1(x_{t_i})\Delta W_i \right) \left(\lim_{n\to \infty} \sum_{j=0}^{n - 1} h_2(x_{t_j})\Delta W_j \right)}
\end{equation}
where $t_k = k\tau/n$, $\Delta t = \tau/n$, and $\Delta W_k = W_{t_{k+1}} - W_{t_k}$. Interchanging expectation and limit (assuming conditions allow), we examine the expectation of the product of sums:
\begin{equation}
     \lim_{n\to \infty} \E{ \sum_{i=0}^{n - 1} \sum_{j=0}^{n - 1} h_1(x_{t_i}) h_2(x_{t_j}) \Delta W_i \Delta W_j }
\end{equation}
We use the properties of Wiener increments: $\E{\Delta W_k} = 0$, $\E{\Delta W_k \Delta W_l} = \delta_{kl} \Delta t$, and importantly, $\Delta W_k$ is independent of $h_1(x_{t_i})$ and $h_2(x_{t_j})$ for $i, j \le k$.
\begin{itemize}
    \item If $i < j$: $\E{h_1(x_{t_i}) h_2(x_{t_j}) \Delta W_i \Delta W_j} = \E{h_1(x_{t_i}) h_2(x_{t_j}) \Delta W_i} \E{\Delta W_j} = 0$.
    \item If $j < i$: Similar logic gives 0.
    \item If $i = j$: $\E{h_1(x_{t_i}) h_2(x_{t_i}) (\Delta W_i)^2} = \E{h_1(x_{t_i}) h_2(x_{t_i})} \E{(\Delta W_i)^2} = \E{h_1(x_{t_i}) h_2(x_{t_i})} \Delta t$.
\end{itemize}
Only the terms with $i=j$ survive:
\begin{equation}
     \lim_{n\to \infty} \sum_{i=0}^{n - 1} \E{h_1(x_{t_i}) h_2(x_{t_i})} \Delta t = \E{ \int_0^\tau h_1(x_t) h_2(x_t) \dd{t} }
\end{equation}
This completes the sketch, recovering \cref{eq:ito_isometry}.

\section{Itô's Formula (Itô's Lemma)}
\label{sec:ito_formula}

Itô's formula is the cornerstone of stochastic calculus, providing the chain rule for functions of processes driven by Wiener noise. It differs from the standard chain rule of ordinary calculus due to the non-zero quadratic variation of the Wiener process (heuristically, $(\dd{W_t})^2 \sim \dd{t}$).

Let $Y(x)$ be a twice continuously differentiable function (denoted $Y \in C^2$). If $x_t$ follows the SDE \cref{eq:chap_1_overdamped_1d}, then the process $Y_t = Y(x_t)$ follows the SDE given by:
\begin{equation}
    \dd{Y(x_t)} = \left( f(x_t) \dv{Y}{x}(x_t) + \frac{1}{2} g(x_t)^2 \dv[2]{Y}{x}(x_t) \right)\dd{t} + g(x_t) \dv{Y}{x}(x_t) \dd{W_t}
    \label{eq:chap_1_Itô_formula} 
\end{equation}
Note the crucial additional term involving the second derivative, $\frac{1}{2} g(x_t)^2 \dv[2]{Y}{x}(x_t)$, compared to the ordinary chain rule.

\subsection{Derivation Sketch}
\label{subsec:ito_derivation}

We begin with the Taylor expansion for the function $Y \in C^2$ over a small increment $\Delta x_t = x_{t+\Delta t} - x_t$:
\begin{equation}
    Y(x_{t+\Delta t}) - Y(x_{t}) = \dv{Y}{x}(x_t) \Delta x_t + \frac{1}{2} \dv[2]{Y}{x}(x_t) (\Delta x_t)^2 + \mathcal{O}(|\Delta x_t|^3)
    \label{eq:taylor_increment}
\end{equation}
Using the approximation for the SDE increment $\Delta x_t \approx f(x_t)\Delta t + g(x_t) \Delta W_t$, we expand $(\Delta x_t)^2$:
\begin{align*}
(\Delta x_t)^2 &\approx (f\Delta t + g \Delta W_t)^2 = f^2(\Delta t)^2 + 2fg\Delta t \Delta W_t + g^2 (\Delta W_t)^2
\end{align*}
Plugging this into the Taylor expansion:
\begin{align*}
\Delta Y(x_t) &\approx \dv{Y}{x}(f\Delta t + g \Delta W_t) + \frac{1}{2} \dv[2]{Y}{x}(f^2(\Delta t)^2 + 2fg\Delta t \Delta W_t + g^2(\Delta W_t)^2) \\ 
&= \left(\dv{Y}{x} f \right)\Delta t + \left(\dv{Y}{x} g \right) \Delta W_t + \left(\frac{1}{2} \dv[2]{Y}{x} g^2 \right)(\Delta W_t)^2 + \text{Higher Order Terms (H.O.T.)}
\end{align*}
Isolating the key Itô term using $(\Delta W_t)^2 \approx \Delta t$:
\begin{align*}
\Delta Y(x_t) &\approx \left(\dv{Y}{x} f + \frac{1}{2} \dv[2]{Y}{x} g^2 \right)\Delta t + \dv{Y}{x} g \Delta W_t \\
&\quad + \frac{1}{2} \dv[2]{Y}{x} g^2 ((\Delta W_t)^2 - \Delta t) + \text{H.O.T.}
\end{align*}

Now, consider the total change $Y(x_{\tau}) - Y(x_{0})$ by summing these increments over a partition $0 = t_0 < t_1 < \dots < t_n = \tau$. As $n \to \infty$:
\begin{align}
    Y(x_{\tau}) - Y(x_{0}) &= \sum_{i=0}^{n-1} \Delta Y(x_{t_i}) \notag \\
    &\approx \sum_{i=0}^{n-1} \left[ \left(\dv{Y}{x} f + \frac{1}{2} \dv[2]{Y}{x} g^2 \right)_{t_i}\Delta t + \left(\dv{Y}{x} g \right)_{t_i} \Delta W_i \right] \notag \\
    &\quad + \sum_{i=0}^{n-1} \left(\frac{1}{2} \dv[2]{Y}{x} g^2 \right)_{t_i} ((\Delta W_i)^2 - \Delta t) + \sum \text{H.O.T.} \notag \\
    &\xrightarrow{n\to\infty} \int_0^\tau \left(\dv{Y}{x} f + \frac{1}{2} \dv[2]{Y}{x} g^2 \right) \dd{t} + \int_0^\tau \dv{Y}{x} g \dd{W_t} + J
    \label{eq:chap_1_Itô_formula_derivation_revised}
\end{align}
where $J$ is the limit of the sum of discrepancy terms:
\begin{equation}
    J = \lim_{n \to \infty} \sum_{i=0}^{n-1} \underbrace{\frac{1}{2} \dv[2]{Y}{x}(x_{t_i}) g(x_{t_i})^2 ((\Delta W_i)^2 - \Delta t)}_{Z_i}
\end{equation}
The Itô formula \cref{eq:chap_1_Itô_formula} holds if $J$ converges to 0 in mean square. We calculate $\E{J_n^2}$:
\begin{align}
    \E{J_n^2} &= \E{\left(\sum_{i=0}^{n-1} Z_i \right)^2} = \sum_{i=0}^{n-1} \E{Z_i^2} + \sum_{i \neq j} \E{Z_i Z_j}
\end{align}
As before, the cross terms vanish ($\E{Z_i Z_j} = 0$ for $i \neq j$ because $Z_i$ and $Z_j$ involve non-overlapping and independent increments $(\Delta W_i)^2 - \Delta t_i$ and $(\Delta W_j)^2 - \Delta t_j$). We are left with:
\begin{align}
    \E{J_n^2} &= \sum_{i=0}^{n-1} \E{Z_i^2} = \sum_{i=0}^{n-1} \E{\left( \frac{1}{2} \dv[2]{Y}{x}(x_{t_i}) g(x_{t_i})^2 \right)^2 \left( (\Delta W_i)^2 - \Delta t \right)^2} \\
    &= \sum_{i=0}^{n-1} \E{\left( \frac{1}{2} \dv[2]{Y}{x}(x_{t_i}) g(x_{t_i})^2 \right)^2} \E{\left( (\Delta W_i)^2 - \Delta t \right)^2} \label{eq:meansquare_step}
\end{align}
Using $\E{\left( (\Delta W_i)^2 - \Delta t \right)^2} = 2(\Delta t)^2$:
\begin{equation}
    \E{J_n^2} = \sum_{i=0}^{n-1} \E{\left( \frac{1}{2} \dv[2]{Y}{x}(x_{t_i}) g(x_{t_i})^2 \right)^2} 2(\Delta t)^2
\end{equation}
Assume the term $H(x) = \left(\frac{1}{2} \dv[2]{Y}{x}(x) g(x)^2\right)^2$ is bounded by $M$. Then:
\begin{align}
    \E{J_n^2} &\le \sum_{i=0}^{n-1} M \cdot 2(\Delta t)^2 = 2 M \sum_{i=0}^{n-1} (\Delta t)^2 \\
    &= 2 M n \left(\frac{\tau}{n}\right)^2 = 2 M \frac{\tau^2}{n}
\end{align}
Taking the limit as $n \to \infty$:
\begin{equation}
    \lim_{n \to \infty} \E{J_n^2} \le \lim_{n \to \infty} 2 M \frac{\tau^2}{n} = 0
\end{equation}
Thus, $J$ converges to zero in mean square, justifying Itô's formula \cref{eq:chap_1_Itô_formula}. (For a full proof, see e.g., \cite[p.~90]{kloedenStochasticTaylorExpansions1992}).

From the previous calculation, we can understand the heuristic notation commonly used:
\begin{equation}
    (\dd{W_t})^2 = \dd{t}
    \label{eq:heuristic_nototation_dw2_dt}
\end{equation}
which intuitively captures the reason for the Itô correction term and corresponds to neglecting $J$ from the start.

\subsection{A Matter of Convention: Itô vs. Stratonovich Integration}
\label{subsec:ito_vs_strato}
Having established Itô's formula (\cref{eq:chap_1_Itô_formula}), which provides the rule for differentiating functions of a process defined by an Itô SDE (\cref{eq:chap_1_overdamped_1d}), it is crucial to recognize that the very definition of the stochastic integral underpinning these concepts is not unique. Specifically, the meaning of the term $\int_0^t g(x_s) \dd{W_s}$ depends on a choice of convention.

In standard Riemann integration, $\int_0^t h(s) \dd{s}$ is defined as a limit of sums, $\lim \sum h(s_i^*) \Delta s_i$. For well-behaved functions $h$, the limit is independent of where the evaluation point $s_i^*$ is chosen within each interval $[s_i, s_{i+1}]$. However, stochastic integration with respect to the Wiener process $W_t$ is fundamentally different. The paths of $W_t$, while continuous, are highly irregular. This high degree of irregularity means that the limit of the sum $\lim \sum h(x_{s_i^*}) \Delta W_i$ \textit{does} depend on how the evaluation points $s_i^*$ are chosen relative to the interval $[s_i, s_{i+1}]$ (e.g., start, midpoint, end).

This ambiguity necessitates adopting a specific mathematical convention for defining the stochastic integral. The two most prevalent conventions in physics, mathematics, and related fields are the Itô integral and the Stratonovich integral. Understanding their definitions and the relationship between them is essential for correctly interpreting SDEs, applying calculus rules, and comparing results from different sources or modeling approaches. This subsection details these two conventions.

The definition of a stochastic integral, such as $\int h(x_t) \dd{W_t}$, is therefore more subtle than its deterministic counterpart. When approximating the integral as a sum over small intervals $[t_i, t_{i+1}]$, a choice must be made about \textit{where} within each interval the function $h(x_t)$ is evaluated.

Let us consider a partition $0 = t_0 < t_1 < \dots < t_n = \tau$ with $\Delta t_i = t_{i+1} - t_i$. A generalized definition for the stochastic integral can be formulated using a parameter $\lambda \in [0, 1]$ specifying the evaluation point:
\begin{equation}
    I_\lambda(h) = \lim_{\max \Delta t_i \to 0} \sum_{i=0}^{n-1}\left((1-\lambda) h(x_{t_i}) + \lambda h\left(x_{t_{i+1}} \right) \right)\left(W_{t_{i+1}} -  W_{t_i}\right)
    \label{eq:chap_1_generalized_integral_sum}
\end{equation}
Different choices of $\lambda$ lead to different integration conventions:

\begin{itemize}
    \item \textbf{Itô Integral ($\lambda = 0$):} Evaluate $h$ at $t_i$. Non-anticipating.
    \begin{equation}
        \int_0^\tau h(x_t) \dd{W_t} := \lim_{\max \Delta t_i \to 0} \sum_{i=0}^{n-1} h(x_{t_i}) (W_{t_{i+1}} - W_{t_i})
        \label{eq:chap_1_ito_definition}
    \end{equation}
    \item \textbf{Stratonovich Integral ($\lambda = 1/2$):} Midpoint evaluation, often using an average. Obeys classical chain rule.
    \begin{equation}
        \int_0^\tau h(x_t) \circ \dd{W_t} := \lim_{\max \Delta t_i \to 0} \sum_{i=0}^{n-1} \frac{h(x_{t_i}) + h(x_{t_{i+1}})}{2} (W_{t_{i+1}} - W_{t_i})
        \label{eq:chap_1_stratonovich_definition}
    \end{equation}
\end{itemize}

To relate them, assume $x_t$ follows the general Itô SDE:
\begin{equation}
    \dd{x_t} = f(x_t) \dd{t} + g(x_t) \dd{W_t}
    \label{eq:chap_1_generic_sde_for_conversion} 
\end{equation}
Consider the Stratonovich sum \cref{eq:chap_1_stratonovich_definition}. We approximate $h(x_{t_{i+1}}) \approx h(x_{t_i}) + h'(x_{t_i}) \Delta x_i$, where $\Delta x_i \approx f(x_{t_i}) \Delta t_i + g(x_{t_i}) \Delta W_i$. Substituting this into the sum (using shorthand $h_i = h(x_{t_i})$, $f_i=f(x_{t_i})$, $g_i=g(x_{t_i})$ and so on):
\begin{align*}
    \sum_{i=0}^{n-1} \frac{h_i + h_{i+1}}{2} \Delta W_i &\approx \sum_{i=0}^{n-1} \frac{h_i + (h_i + h'_i [f_i \Delta t_i + g_i \Delta W_i])}{2} \Delta W_i \\
    &\approx \sum_{i=0}^{n-1} \left( h_i + \frac{1}{2} h'_i f_i \Delta t_i + \frac{1}{2} h'_i g_i \Delta W_i \right) \Delta W_i \\
    &\approx \sum_{i=0}^{n-1} \left( h_i \Delta W_i + \frac{1}{2} h'_i f_i \Delta t_i \Delta W_i + \frac{1}{2} h'_i g_i (\Delta W_i)^2 \right)
\end{align*}
In the limit $\max \Delta t_i \to 0$, the term $\Delta t_i \Delta W_i$ vanishes, while the term containing $(\Delta W_i)^2$ contributes $\frac{1}{2} \int_0^\tau h'(x_t) g(x_t) \dd{t}$ (where $g(x_t)$ is the diffusion coefficient of $x_t$). The sum converges to:
\begin{equation*}
    \int_0^\tau h(x_t) \dd{W_t} + \frac{1}{2} \int_0^\tau h'(x_t) g(x_t) \dd{t}
\end{equation*}
This establishes the general conversion formula between Stratonovich and Itô integrals when $x_t$ follows the Itô SDE \cref{eq:chap_1_generic_sde_for_conversion} and we are integrating $h(x_t)$ with respect to $W_t$:
\begin{equation}
    \int_0^\tau h(x_t) \circ \dd{W_t} = \int_0^\tau h(x_t) \dd{W_t} + \frac{1}{2} \int_0^\tau \dv{h}{x}(x_t) g(x_t) \dd{t}
    \label{eq:chap_1_Strato_to_Ito_general} 
\end{equation}
For standard deterministic integrals (Riemann integral), where the integrator is time, e.g., the evaluation point within the interval does not affect the limit; thus:
\begin{equation}
    \int_0^\tau h(x_t) \circ \dd{t} = \int_0^\tau h(x_t) \dd{t}
    \label{eq:chap_1_equality_standard_integral}
\end{equation}

So far, we have highlighted the distinction between Itô and Stratonovich conventions when integrating a function $h(x_t)$ with respect to the Wiener process, $\dd{W_t}$. A related question arises when considering integration with respect to the process $x_t$ itself. Since $\dd{x_t} = f(x_t) \dd{t} + g(x_t) \dd{W_t}$, an integral such as $\int Y'(x_t) \dd{x_t}$ (where $Y'$ is the derivative of some function $Y$) must be interpreted carefully. Formally, it expands as:
\begin{equation*}
    \int Y'(x_t) \dd{x_t} = \int Y'(x_t) f(x_t) \dd{t} + \int Y'(x_t) g(x_t) \dd{W_t}
\end{equation*}
The first term is a standard Riemann integral with respect to time. However, the second term is a stochastic integral with respect to $\dd{W_t}$. As we have just seen, the definition and value of this stochastic part depend on whether we use the Itô or Stratonovich convention (\cref{eq:chap_1_ito_definition,eq:chap_1_stratonovich_definition,eq:chap_1_Strato_to_Ito_general}). Consequently, the interpretation of $\int Y'(x_t) \dd{x_t}$ also depends on the chosen convention. This difference is fundamental and manifests most clearly when examining the rules for differentiation (the chain rule) within each framework, as discussed next.

\subsection{Advantages and Disadvantages of the Stratonovich Convention}
\label{subsec:strato_adv_disadv}


The choice between the Itô and Stratonovich integration conventions involves a trade-off between mathematical properties and adherence to familiar calculus rules.

\textbf{Advantage of Stratonovich: Obeys Classical Calculus Rules}

The most significant advantage of the Stratonovich convention is that it preserves the structure of the standard chain rule from ordinary differential calculus. As we will demonstrate, if $x_t$ follows a Stratonovich SDE (i.e., an SDE written using $\circ \dd{W_t}$ or $\circ \dd{x_t}$), then for a sufficiently smooth function $Y(x_t)$, its differential is given by:
\begin{equation}
    \dd{Y(x_t)} = \dv{Y}{x}(x_t) \circ \dd{x_t} 
    \label{eq:chap_1_strato_chain_rule}
\end{equation}
This means that variable transformations and coordinate changes within Stratonovich SDEs follow the same rules as in deterministic calculus, which can be more intuitive.

\textbf{Disadvantage of Stratonovich: Complicates Expectation Calculations}

Despite the elegance of its chain rule, the Stratonovich convention introduces complexities in other crucial areas, primarily related to calculating expectations. The fundamental building block of the Itô SDE is the Itô integral $\int_0^t h(x_s) \dd{W_s}$. A key property of the Itô integral is that:
\begin{equation*}
    \E{\int_0^t h(x_s) \dd{W_s}} = 0
\end{equation*}
This property is immensely useful in theoretical analysis. When taking the expectation of an Itô SDE or a function derived via Itô's lemma (\cref{eq:chap_1_Itô_formula}), the entire stochastic integral term vanishes on average, simplifying the analysis considerably (e.g., when deriving equations for moments or analyzing stability).

In contrast, the expectation of the Stratonovich integral $ \int_0^t h(x_s) \circ \dd{W_s}$ is typically non-zero. To calculate it, one must first convert the Stratonovich integral back to an Itô integral using the conversion formula (\cref{eq:chap_1_Strato_to_Ito_general}):
\begin{equation*}
    \E{\int_0^t h(x_s) \circ \dd{W_s}} = \E{\int_0^t h(x_s) \dd{W_s}} + \frac{1}{2} \E{\int_0^t \dv{h}{x}(x_s) g(x_s) \dd{s}} = \frac{1}{2} \E{\int_0^t \dv{h}{x}(x_s) g(x_s) \dd{s}}
\end{equation*}
This introduces an additional term (the "spurious drift" or Itô correction term) that needs to be evaluated, often making the calculation of expectations more cumbersome than in the Itô framework. Similarly, the Itô isometry (\cref{eq:ito_isometry_variance}), which provides a straightforward way to calculate the second moment $\E{[ \int h \dd{W} ]^2}$, does not directly apply to Stratonovich integrals.

\textbf{Manuscript Convention}

Because many theoretical derivations and analytical results in stochastic processes heavily rely on expectation calculations, the \textbf{Itô convention is the standard choice throughout this manuscript}. Its non-anticipating nature also often aligns better with the notion of causality in physical and biological systems where future noise cannot influence the present state or dynamics. While calculations might sometimes seem less intuitive initially due to the modified chain rule (Itô's formula), the benefits for analysis involving expectations often outweigh this.

\subsubsection*{Proof for $\dd{Y(x_t)} = \dv{Y}{x}(x_t) \circ \dd{x_t}$}
Let $x_t$ follow the Itô SDE \cref{eq:chap_1_overdamped_1d}: $\dd{x_t} = f(x_t) \dd{t} + g(x_t) \dd{W_t}$.
Let $Y(x)$ be a smooth function. Itô's formula \cref{eq:chap_1_Itô_formula} gives:
\begin{equation*}
    dY(x_t) = \left( Y' f + \frac{1}{2} Y'' g^2 \right) \dd{t} + Y' g \dd{W_t}
\end{equation*}
Our goal is to show this equals $Y' \circ \dd{x_t}$.

First, find the Stratonovich form of \cref{eq:chap_1_overdamped_1d}. Apply \cref{eq:chap_1_Strato_to_Ito_general} (with $h(x_t)$ in that formula being $g(x_t)$ here, and $g(x_t)$ from the SDE for $x_t$ also being $g(x_t)$ here), then rearrange:
\begin{equation}
    \int g(x_t) \dd{W_t} = \int g(x_t) \circ \dd{W_t} - \frac{1}{2} \int g'(x_t) g(x_t) \dd{t}
    \label{eq:chap_1_convert_g_term}
\end{equation}
Substitute into \cref{eq:chap_1_overdamped_1d}:
\begin{align}
    \dd{x_t} &= f(x_t) \dd{t} + \left( g(x_t) \circ \dd{W_t} - \frac{1}{2} g'(x_t) g(x_t) \dd{t} \right) \nonumber \\
    &= \underbrace{\left( f(x_t) - \frac{1}{2} g'(x_t) g(x_t) \right)}_{\tilde{f}(x_t) \text{: Stratonovich Drift}} \dd{t} + g(x_t) \circ \dd{W_t}
    \label{eq:chap_1_strato_sde} 
\end{align}
This is the SDE in Stratonovich form.

Now, return to Itô's formula for $dY(x_t)$. Apply the conversion \cref{eq:chap_1_Strato_to_Ito_general} to the Itô term $Y' g \dd{W_t}$. Let $h(x_t) = Y'(x_t)g(x_t)$ in the conversion formula, and the $g(x_t)$ in that formula's correction term is the diffusion coefficient of $x_t$ (which is $g(x_t)$ from \cref{eq:chap_1_overdamped_1d}):
\begin{equation*}
    \int Y'(x_t) g(x_t) \dd{W_t} = \int Y'(x_t) g(x_t) \circ \dd{W_t} - \frac{1}{2} \int \dv{(Y' g)}{x}(x_t) g(x_t) \dd{t}
\end{equation*}
Substitute this into Itô's formula for $dY(x_t)$:
\begin{align*}
    dY(x_t) &= \left( Y' f + \frac{1}{2} Y'' g^2 \right) \dd{t} + \left( Y' g \circ \dd{W_t} - \frac{1}{2} (Y' g)' g \dd{t} \right) \\
    &= \left( Y' f + \frac{1}{2} Y'' g^2 - \frac{1}{2} (Y'' g + Y' g') g \right) \dd{t} + Y' g \circ \dd{W_t} \\
    &= \left( Y' f + \frac{1}{2} Y'' g^2 - \frac{1}{2} Y'' g^2 - \frac{1}{2} Y' g' g \right) \dd{t} + Y' g \circ \dd{W_t} \\
    &= Y'(x_t) \left( f(x_t) - \frac{1}{2} g'(x_t) g(x_t) \right) \dd{t} + Y'(x_t) g(x_t) \circ \dd{W_t} \\
    &= Y'(x_t) \left( \tilde{f}(x_t) \dd{t} + g(x_t) \circ \dd{W_t} \right) \\
    &= \dv{Y}{x}(x_t) \circ \dd{x_t} \qquad \text{(using \cref{eq:chap_1_strato_sde})}
\end{align*}
This confirms that under the Stratonovich convention, the chain rule \cref{eq:chap_1_strato_chain_rule} holds.

\section{Conclusion}
\label{sec:sde_toolbox_conclusion}

This chapter has laid the mathematical groundwork for describing complex systems where deterministic evolution is intertwined with inherent randomness. Starting with the physical insights of Brownian motion and Langevin's equation, we formalized the description using Stochastic Differential Equations (SDEs). The essential tools and pitfalls for working effectively with SDEs were introduced: the Wiener process ($W_t$) to represent the underlying noise, the Itô stochastic integral to rigorously quantify its cumulative effect, and Itô's formula (\cref{eq:chap_1_Itô_formula}) to understand how functions of the stochastic state change in response to both deterministic drift and random diffusion. The choice of the Itô convention was motivated by its analytical advantages, particularly concerning expectations, though the relationship with the Stratonovich interpretation, which preserves the classical chain rule (\cref{eq:chap_1_strato_chain_rule}), was explored.

This "SDE toolbox" is not merely a collection of mathematical abstractions; it provides the essential equipment for constructing realistic models, deriving analytical insights, and performing quantitative predictions for the stochastic phenomena investigated throughout this thesis. The following chapters will demonstrate the practical application of these powerful tools.

\newpage

\begin{tcolorbox}[
    enhanced, sharp corners, boxrule=0.5pt, colframe=black!75!white, colback=white, coltitle=black, fonttitle=\bfseries, title=Chapter Takeaways,
    attach boxed title to top left={yshift=-0.1in, xshift=0.15in},
    boxed title style={ colback=white, sharp corners, boxrule=0pt, frame code={\draw[black!75!white, line width=0.5pt] ([yshift=-1pt]frame.south west) -- ([yshift=-1pt]frame.south east);} },
    boxsep=5pt, left=5pt, right=5pt, top=12pt, bottom=5pt
    ]
    \textbf{Context:} This chapter introduced the mathematical framework for Itô Stochastic Differential Equations (SDEs), essential for describing systems influenced by continuous random noise. The general 1D form considered is:
    \begin{equation*}
        \dd{x_t} = f(x_t) \dd{t} + g(x_t)\, \dd{W_t}
    \end{equation*}
    We explored the core tools needed to rigorously define, interpret, and manipulate such equations using the Itô convention.
    \medskip

    \begin{itemize}
        \item \textbf{Wiener Process ($W_t$):} Model for Brownian motion; continuous paths, independent Gaussian increments $\Delta W_t \sim \mathcal{N}(0, \Delta t)$.
        \item \textbf{Itô Integral ($\int h(t) \dd{W_t}$):} Defined by evaluating the integrand $h$ at the start of the interval ($t_i$). Non-anticipating. \textbf{Default convention in this manuscript.}
        \item \textbf{Stratonovich Integral ($\int h(t) \circ \dd{W_t}$):} Midpoint evaluation. Obeys classical chain rule but lacks simple expectation property. Conversion formula relates it to Itô integral (\cref{eq:chap_1_Strato_to_Ito_general}).
        \item \textbf{Itô's Lemma (\cref{eq:chap_1_Itô_formula}):} The "chain rule" for Itô SDEs. For a function $Y(x_t)$ where $x_t$ follows the 1D SDE above:
        \begin{equation*}
        \dd{Y(x_t)} = \left( f(x_t) \dv{Y}{x}(x_t) + \frac{1}{2} g(x_t)^2 \dv[2]{Y}{x}(x_t) \right)\dd{t} + g(x_t) \dv{Y}{x}(x_t) \dd{W_t}
        \end{equation*}
        (Generalizes to functions $Y(x,t)$ and multi-dimensional processes).
    \end{itemize}
\end{tcolorbox}

%% file: tex_body/chap2_a.tex
\chapter{Likelihood-Based Parameter Estimation} 
\label{chap:inference_methods}
\chaptertoc{}

\section*{Introduction} 
\addcontentsline{toc}{section}{Introduction} 

The previous chapter equipped us with the "SDE Toolbox" – the essential mathematical framework for describing and analyzing systems governed by Stochastic Differential Equations. We learned how to model dynamics subject to random fluctuations using the Wiener process, how to interpret and manipulate these equations via stochastic calculus (particularly the Itô convention), and how to describe the evolution of functions of these stochastic processes using Itô's lemma.

However, formulating an SDE model, such as $\dd{\bm{x}_t} = \bm{f}(\bm{x}_t\mid \bm{\alpha}) \dd{t} + \sqrt{2 \bm{D}(\bm{x}_t\mid \bm{\alpha})} \dd{\bm{W_t}}$, often involves unknown parameters, denoted collectively by $\bm{\alpha}$. These parameters might represent physical constants, reaction rates, interaction strengths, or noise intensities that define the specific behavior of the system. A critical task in applying these models to real-world phenomena is therefore \emph{parameter estimation}: determining the values of $\bm{\alpha}$ that best explain observed data.

This chapter delves into parameter estimation, focusing on one of the most fundamental and widely used approaches: the \textbf{likelihood principle}. We will define the likelihood function and its maximization to obtain the Maximum Likelihood Estimator (MLE). Subsequently, we will discuss methods for evaluating the quality of estimators using the Mean Squared Error (MSE) and its decomposition into bias and variance. We will also explore theoretical limits on estimator precision via the Cramér-Rao Lower Bound and examine the key asymptotic properties of the MLE, namely its consistency and normality under certain conditions. Finally, we will briefly touch upon the connection between maximum likelihood and Bayesian estimation paradigms. This chapter lays the statistical foundation for inferring model parameters from data, a crucial step in bridging SDE models with empirical observations.

\section{Likelihood and Maximum Likelihood Estimation}
\label{sec:likelihood_mle} 

Consider a set of $N$ observed data points, denoted $\bm{X} = \{ \bm{x}_1, \bm{x}_2, \dots, \bm{x}_N\}$. We assume these data are generated from a parametric statistical model characterized by a probability density function (or probability mass function for discrete data) $p(\bm{x}\mid \bm{\alpha})$, where $\bm{\alpha} = (\alpha_1, \dots, \alpha_{m})$ is the vector of parameters we wish to estimate, belonging to a parameter space $\mathcal{A} \subseteq \mathbb{R}^{m}$.

If we assume the data points $\{\bm{x}_i\}$ are independent realizations from this model, the joint probability density of observing the entire dataset $\bm{X}$ is given by the product of the individual densities:
\begin{equation}
    p(\bm{X}\mid \bm{\alpha}) = \prod_{i=1}^{N} p(\bm{x}_i\mid \bm{\alpha}).
    \label{eq:joint_density_iid} 
\end{equation}
When this joint density is viewed as a function of the parameters $\bm{\alpha}$ for a fixed, observed dataset $\bm{X}$, it is called the \textbf{likelihood function}:
\begin{equation}
    \mathcal{L}(\bm{\alpha}\mid \bm{X}) = p(\bm{X}\mid \bm{\alpha}).
    \label{eq:likelihood_definition} 
\end{equation}

Intuitively, the likelihood $\mathcal{L}(\bm{\alpha}\mid \bm{X})$ measures the plausibility of different parameter values $\bm{\alpha}$ having generated the observed data $\bm{X}$.

The principle of \textbf{Maximum Likelihood Estimation} (MLE) postulates that the best estimate for $\bm{\alpha}$ is the value that makes the observed data most probable, i.e., the value that maximizes the likelihood function. The MLE, denoted by $\hat{\bm{\alpha}}_{\mathrm{MLE}}$, is thus defined as:
\begin{equation}
    \hat{\bm{\alpha}}_{\mathrm{MLE}} = \argmax_{\bm{\alpha} \in \mathcal{A}} \mathcal{L}(\bm{\alpha}\mid \bm{X}).
    \label{eq:mle_definition} 
\end{equation}

In practice, maximizing the likelihood function $\mathcal{L}$ is often computationally challenging due to the product form. Since the natural logarithm $\ln(\cdot)$ is a monotonically increasing function, maximizing $\mathcal{L}$ is equivalent to maximizing its logarithm. Therefore, we typically work with the \textbf{log-likelihood function}, $\ell(\bm{\alpha}\mid \bm{X})$:
\begin{equation}
    \ell(\bm{\alpha}\mid \bm{X}) = \ln \mathcal{L}(\bm{\alpha}\mid \bm{X}) = \sum_{i=1}^{N} \ln p(\bm{x}_{i}\mid \bm{\alpha}).
    \label{eq:loglikelihood_definition}
\end{equation}
Maximizing the log-likelihood turns products into sums, which is often numerically more stable and analytically more tractable.

Note, for dependent data like SDE time series, $p(\bm{X}\mid \bm{\alpha})$ is the joint density $p(\bm{x_1}, ..., \bm{x_N}\mid \bm{\alpha})$ and its calculation is more complex, often involving transition densities. This section introduces the independent measurement concept first. In the next chapter, we will deal with SDE time series.

\section{Evaluating Estimator Performance: Bias, Variance, and MSE}
\label{sec:estimator_performance}

Having defined the Maximum Likelihood Estimator (MLE) $\hat{\bm{\alpha}}$ as a principled method for parameter estimation based on observed data $\bm{X}$, a crucial next step is to evaluate its quality. How close can we expect our estimate $\hat{\bm{\alpha}}$ to be to the true, unknown parameter vector $\bm{\alpha}^*$ that generated the data? Since the data $\bm{X}$ are random variables (drawn from $p(\bm{x}\mid \bm{\alpha}^*)$), any estimator $\hat{\bm{\alpha}} = \hat{\bm{\alpha}}(\bm{X})$ derived from the data is also a random variable. Therefore, we need statistical measures to quantify its performance on average. We typically analyze estimators component by component; let $\hat{\alpha}_i$ be the estimator for the $i$-th parameter $\alpha_i^*$.

\subsection{Bias}
\label{subsec:bias}

The \textbf{bias} of an estimator $\hat{\alpha}_i$ measures the systematic difference between the expected value of the estimator and the true parameter value:
\begin{equation}
    \mathrm{Bias}(\hat{\alpha}_i) = \E{\hat{\alpha}_i} - \alpha_i^*.
    \label{eq:bias_definition}
\end{equation}
Here, the expectation $\E{\cdot}$ is taken with respect to the probability distribution of the data $\bm{X}$, which is governed by the true parameter $\bm{\alpha}^*$.
\begin{itemize}
    \item If $\mathrm{Bias}(\hat{\alpha}_i) = 0$, the estimator is called \textbf{unbiased}. On average, an unbiased estimator yields the correct parameter value.
    \item If $\mathrm{Bias}(\hat{\alpha}_i) \neq 0$, the estimator systematically overestimates ($\mathrm{Bias} > 0$) or underestimates ($\mathrm{Bias} < 0$) the true parameter value.
\end{itemize}
While unbiasedness is often a desirable property, it is not the only criterion for a good estimator. An unbiased estimator might still exhibit large fluctuations around the true value.

\subsection{Variance}
\label{subsec:variance}

The \textbf{variance} of an estimator $\hat{\alpha}_i$ measures the spread or variability of the estimates around their expected value $\E{\hat{\alpha}_i}$:
\begin{equation}
    \mathrm{Var}(\hat{\alpha}_i) = \E{ \left( \hat{\alpha}_i - \E{\hat{\alpha}_i} \right)^2 }.
    \label{eq:variance_definition}
\end{equation}
The variance quantifies the estimator's precision. A smaller variance indicates that the estimates obtained from different datasets (generated under the same true parameter $\bm{\alpha}^*$) tend to be closer to each other (and closer to the average estimate $\E{\hat{\alpha}_i}$). Low variance implies high reliability or reproducibility of the estimate.

\subsection{Mean Squared Error (MSE)}
\label{subsec:mse}

Neither bias nor variance alone provides a complete picture of estimator quality. We might prefer a slightly biased estimator with very low variance over an unbiased one with high variance. The \textbf{Mean Squared Error} (MSE) combines both aspects by measuring the average squared difference between the estimator and the true parameter value:
\begin{equation}
    \mathrm{MSE}(\hat{\alpha}_i) = \E{ (\hat{\alpha}_i - \alpha_i^*)^2 }.
    \label{eq:mse_definition_expanded}
\end{equation}
The MSE provides a single measure of the overall accuracy of the estimator. A smaller MSE indicates a better estimator on average, in the sense that its estimates are expected to be closer to the true value.

A fundamental result connects MSE to bias and variance. We can decompose the MSE as follows:
\begin{align}
    \mathrm{MSE}(\hat{\alpha}_i) &= \E{ (\hat{\alpha}_i - \alpha_i^*)^2 } \nonumber \\
    &= \E{ (\hat{\alpha}_i - \E{\hat{\alpha}_i} + \E{\hat{\alpha}_i} - \alpha_i^*)^2 } \nonumber \\
    &= \E{ [(\hat{\alpha}_i - \E{\hat{\alpha}_i}) + (\E{\hat{\alpha}_i} - \alpha_i^*)]^2 } \nonumber \\
    &= \E{ (\hat{\alpha}_i - \E{\hat{\alpha}_i})^2 + 2(\hat{\alpha}_i - \E{\hat{\alpha}_i})(\E{\hat{\alpha}_i} - \alpha_i^*) + (\E{\hat{\alpha}_i} - \alpha_i^*)^2 } \nonumber \\
    &= \E{ (\hat{\alpha}_i - \E{\hat{\alpha}_i})^2 } + \E{ 2(\hat{\alpha}_i - \E{\hat{\alpha}_i})(\E{\hat{\alpha}_i} - \alpha_i^*) } + \E{ (\E{\hat{\alpha}_i} - \alpha_i^*)^2 } \nonumber \\
    &= \mathrm{Var}(\hat{\alpha}_i) + 2(\E{\hat{\alpha}_i} - \alpha_i^*) \E{ (\hat{\alpha}_i - \E{\hat{\alpha}_i}) } + (\E{\hat{\alpha}_i} - \alpha_i^*)^2 \nonumber \\
    &= \mathrm{Var}(\hat{\alpha}_i) + 2(\mathrm{Bias}(\hat{\alpha}_i)) (0) + (\mathrm{Bias}(\hat{\alpha}_i))^2 \nonumber \\
    &= \mathrm{Var}(\hat{\alpha}_i) + (\mathrm{Bias}(\hat{\alpha}_i))^2.
    \label{eq:mse_bias_variance_decomp}
\end{align}
This is the \textbf{bias-variance decomposition} of the MSE. It shows explicitly that the MSE incorporates both the variance (random error, related to precision) and the squared bias (systematic error, related to accuracy).

\begin{figure}[htbp]
    \centering
    \includegraphics[width=0.6\textwidth]{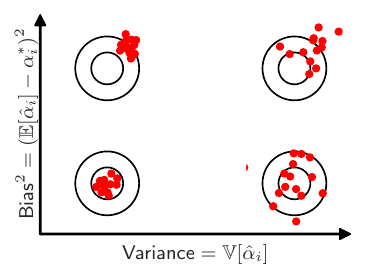} 
    \caption{Illustration of the bias–variance tradeoff using a target analogy. Each red dot represents an estimate $\hat{\alpha}_i$ from a different hypothetical dataset; the bullseye (center) represents the true parameter value $\alpha_i^*$. Low bias means estimates are centered on the target; low variance means estimates are tightly clustered.}
    \label{fig:bias_variance_decomposition}
\end{figure}

\paragraph{The Bias-Variance Trade-off}
The decomposition \cref{eq:mse_bias_variance_decomp} highlights a crucial concept in statistical modeling and estimation: the \textbf{bias-variance trade-off}, illustrated conceptually in \cref{fig:bias_variance_decomposition}. Often, attempts to decrease the bias of an estimator lead to an increase in its variance, and vice versa. For example, complex models might achieve lower bias by fitting the training data very closely, but suffer from high variance when applied to new data (overfitting). Conversely, simpler models might have higher bias (failing to capture the true underlying structure) but often exhibit lower variance. The goal is often to find an estimator or model that achieves a good balance, minimizing the overall MSE, rather than minimizing bias or variance in isolation. Understanding this trade-off is essential for choosing appropriate estimation methods and controlling model complexity.

In the following sections, we will explore theoretical limits on the variance of estimators (specifically, unbiased estimators) and examine the properties of the MLE in terms of bias and variance, particularly in the limit of large datasets.

\section{Cramér--Rao Lower Bound: The Limit of Precision}
\label{sec:crlb}

While the Mean Squared Error (MSE) helps us evaluate the overall performance of an estimator, combining bias and variance, the \textbf{Cramér--Rao Lower Bound} (CRLB) addresses a more specific question: What is the \textit{best possible precision} (i.e., minimum variance) that any \textit{unbiased} estimator can achieve for a given statistical model and dataset size? It provides a fundamental benchmark against which we can compare the performance of specific unbiased estimators like the Maximum Likelihood Estimator (MLE).

\subsection{Fisher Information: Measuring Data's Information Content}
\label{subsec:fisher_info}

The key quantity underpinning the CRLB is the \textbf{Fisher Information Matrix} (FIM), denoted $\bm{I}_N(\bm{\alpha}^*)$, defined for the $N$ data points observed. This matrix quantifies the amount of information that the observable data $\bm{X}$ carries about the true parameter vector $\bm{\alpha}^*$. Intuitively, high Fisher information implies that the data are very sensitive to changes in the parameters, allowing for more precise estimation. Conversely, low Fisher information suggests that the data provide little constraint on the parameter values.

For a model with true parameter vector $\bm{\alpha}^*$ and log-likelihood function $\ell(\bm{\alpha}^*\mid \bm{X}) = \ln p(\bm{X}\mid \bm{\alpha}^*)$, the elements of the FIM $\bm{I}_N(\bm{\alpha}^*)$ are defined as:
\begin{equation}
    [\bm{I}_N(\bm{\alpha}^*)]_{ij} = \E{ \left(\pdv{\ell(\bm{\alpha}^*\mid \bm{X})}{\alpha_i^*}\right) \left(\pdv{\ell(\bm{\alpha}^*\mid \bm{X})}{\alpha_j^*}\right)  } = - \E{  \pdv[2]{\ell(\bm{\alpha}^*\mid \bm{X})}{\alpha_i^*}{\alpha_j^*} }.
    \label{eq:fisher_info_definition_revised}
\end{equation}
The expectation is taken with respect to the data distribution $p(\bm{X}\mid \bm{\alpha}^*)$. The second form shows that the Fisher information is related to the expected curvature (negative Hessian) of the log-likelihood function. A sharply peaked log-likelihood function (high curvature) corresponds to high Fisher information, indicating that the maximum is well-localized and the parameter can be estimated precisely.

\subsection{The Bound for Unbiased Estimators}
\label{subsec:crlb_unbiased}

Let $\hat{\bm{\alpha}} = \hat{\bm{\alpha}}(\bm{X})$ be any \textit{unbiased} estimator of the true parameter vector $\bm{\alpha}^*$, meaning $\E{\hat{\bm{\alpha}}} = \bm{\alpha}^*$. The CRLB states that the covariance matrix of $\hat{\bm{\alpha}}$ is bounded below by the inverse of the Fisher Information Matrix evaluated at the \textit{true} parameter value $\bm{\alpha}^*$:
\begin{equation}
    \mathrm{Cov}(\hat{\bm{\alpha}}) = \E{ [ (\hat{\bm{\alpha}} - \bm{\alpha}^*)(\hat{\bm{\alpha}} - \bm{\alpha}^*)^T ] }_{p(\bm{X}\mid \bm{\alpha}^*)} \geq \bm{I}_N(\bm{\alpha}^*)^{-1}.
    \label{eq:crlb_vector_revised}
\end{equation}
Here, the matrix inequality $\bm{A} \geq \bm{B}$ means that the matrix $\bm{A} - \bm{B}$ is positive semi-definite.

For a single scalar parameter $\alpha$, the bound simplifies to a lower bound on the variance:
\begin{equation}
    \mathrm{Var}(\hat{\alpha}) \geq \frac{1}{I_N(\alpha^*)}.
    \label{eq:crlb_scalar}
\end{equation}
where $I_N(\alpha^*)$ is the scalar Fisher information evaluated at the true value $\alpha^*$.

\subsection{Proof Sketch for a Scalar Unbiased Estimator}
\label{subsec:crlb_proof}

We demonstrate the CRLB \cref{eq:crlb_scalar} for an unbiased scalar estimator $\hat{\alpha}$ of $\alpha^*$. Let $\ell(\alpha\mid \bm{X}) = \ln p(\bm{X}\mid \alpha)$ be the log-likelihood. The score function is $U(\alpha) = \frac{\partial \ell}{\partial \alpha}$. Assume regularity conditions hold, allowing differentiation under the integral sign when taking expectations with respect to $p(\bm{X}\mid \alpha)$.

\begin{enumerate}
    \item \textbf{Zero Expectation of Score:} The expectation of the score function, evaluated at the true parameter $\alpha^*$, is zero:
    \begin{equation*}
        \E{U(\alpha^*)} = \E{\frac{\partial}{\partial \alpha^*} \ln p(\bm{X}\mid \alpha^*)} = 0.
        \label{eq:score_expectation_zero}
    \end{equation*}
    (See footnote\footnote{%
    $\E{\frac{\partial}{\partial \alpha^*}\ln p(\bm{X}\mid\alpha^*)} = \int p(\bm{X}\mid\alpha^*) \frac{1}{p(\bm{X}\mid\alpha^*)} \frac{\partial p(\bm{X}\mid\alpha^*)}{\partial \alpha^*} \dd{\bm{X}} = \int \frac{\partial p(\bm{X}\mid\alpha^*)}{\partial \alpha^*} \dd{\bm{X}} = \frac{d}{d\alpha^*} \int p(\bm{X}\mid\alpha^*) \dd{\bm{X}} = \frac{d}{d\alpha^*}(1) = 0$.} for derivation.) 

    \item \textbf{Derivative of the Unbiasedness Condition:} Since we have that $\E{\hat{\alpha}} = \alpha^*$, we can differentiate with respect to $\alpha^*$:
    \begin{align*}
        \frac{d}{d\alpha^*} \E{\hat{\alpha}} &= \frac{d}{d\alpha^*} \int \hat{\alpha}(\bm{X}) p(\bm{X}\mid\alpha^*) \dd{\bm{X}} \\ 
        &= \int \hat{\alpha}(\bm{X}) \frac{\partial p(\bm{X}\mid\alpha^*)}{\partial \alpha^*} \dd{\bm{X}} \\ 
        &= \int \hat{\alpha}(\bm{X}) \left( p(\bm{X}\mid\alpha^*) \frac{\partial \ln p(\bm{X}\mid\alpha^*)}{\partial \alpha^*} \right) \dd{\bm{X}} \\ 
        &= \int \hat{\alpha}(\bm{X}) U(\alpha^*) p(\bm{X}\mid\alpha^*) \dd{\bm{X}} \\
        &= \E{\hat{\alpha} U(\alpha^*)}.
    \end{align*}
    Since $\frac{d}{d\alpha^*} (\alpha^*) = 1$, the unbiasedness condition implies 
    $$\E{\hat{\alpha} U(\alpha^*)} = 1$$.

    \item \textbf{Covariance of Estimator and Score:} Using $\E{U(\alpha^*)} = 0$, we have $\E{\alpha^* U(\alpha^*)} = \alpha^* \E{U(\alpha^*)} = 0$. Subtracting this from the previous result gives the covariance between the estimator and the score function:
    \begin{equation*}
        \mathrm{Cov}(\hat{\alpha}, U(\alpha^*)) = \E{(\hat{\alpha} - \E{\hat{\alpha}}) (U(\alpha^*) - \E{U(\alpha^*)})} = \E{(\hat{\alpha} - \alpha^*) U(\alpha^*)} = 1.
    \end{equation*}

    \item \textbf{Cauchy--Schwarz Inequality:} Apply the Cauchy--Schwarz inequality $|\mathrm{Cov}(X,Y)|^2 \le \mathrm{Var}(X) \mathrm{Var}(Y)$ with $X = \hat{\alpha}$ and $Y = U(\alpha^*)$:
    \begin{equation*}
        1^2 = (\mathrm{Cov}(\hat{\alpha}, U(\alpha^*)))^2 \le \mathrm{Var}(\hat{\alpha}) \cdot \mathrm{Var}(U(\alpha^*)).
    \end{equation*}

    \item \textbf{Identify Terms:} We know $\mathrm{Var}(\hat{\alpha})$ is the variance of the estimator. The variance of the score is:
    \begin{equation*}
        \mathrm{Var}(U(\alpha^*)) = \E{(U(\alpha^*) - \E{U(\alpha^*)})^2} = \E{U(\alpha^*)^2} = I_N(\alpha^*),
    \end{equation*}
    which is exactly the Fisher information $I_N(\alpha^*)$ for a scalar parameter. 

    \item \textbf{Result:} Substituting these into the inequality gives:
    \begin{equation*}
        1 \le \mathrm{Var}(\hat{\alpha}) \cdot I_N(\alpha^*).
    \end{equation*}
    Rearranging ($I_N(\alpha^*) > 0$) yields the scalar Cramér-Rao lower bound:
    \begin{equation*}
        \mathrm{Var}(\hat{\alpha}) \ge \frac{1}{I_N(\alpha^*)}.
    \end{equation*}
\end{enumerate}
The proof for the vector case \cref{eq:crlb_vector_revised} follows similar logic using vector/matrix extensions of the concepts and the multivariate Cauchy-Schwarz inequality. A detailed derivation can be found in standard statistical inference texts, such as \cite[p.~70]{kayFundamentalsStatisticalSignal1993}. 

\section{Asymptotic Properties of Maximum Likelihood Estimators}
\label{sec:mle_asymptotics}

Maximum Likelihood Estimators are widely used partly due to their excellent properties in the limit of large sample sizes ($N \to \infty$). Assuming the data consists of $N$ independent and identically distributed (i.i.d.) observations, the MLE $\hat{\bm{\alpha}}_{N}$ (explicitly showing dependence on $N$) exhibits the following crucial properties:

\begin{enumerate}
    \item \textbf{Consistency:} The MLE converges in probability to the true parameter value as the sample size increases:
        \begin{equation}
            \hat{\bm{\alpha}}_{N} \xrightarrow[N \to \infty]{p} \bm{\alpha}^*.
            \label{eq:mle_consistency}
        \end{equation}
    \item \textbf{Asymptotic Normality:} The distribution of the MLE, when properly scaled and centered, approaches a multivariate normal distribution:
        \begin{equation}
            \sqrt{N} (\hat{\bm{\alpha}}_{N} - \bm{\alpha}^*) \xrightarrow[N \to \infty]{d} \mathcal{N}\left(\bm{0},\, \bm{I}_1(\bm{\alpha}^*)^{-1}\right).
            \label{eq:mle_asymptotic_normality_revised}
        \end{equation}
        Here, $\xrightarrow{d}$ denotes convergence in distribution, $\bm{0}$ is the zero vector, and $\bm{I}_1(\bm{\alpha}^*)$ is the Fisher information matrix for a \textit{single} observation $p(\bm{x}\mid \bm{\alpha}^*)$. For i.i.d. data, the Fisher information for $N$ observations is $\bm{I}_N(\bm{\alpha}^*) = N \bm{I}_1(\bm{\alpha}^*)$.

    \item \textbf{Asymptotic Efficiency:} The MLE achieves the Cramér-Rao Lower Bound (\cref{eq:crlb_vector_revised}) asymptotically. From \cref{eq:mle_asymptotic_normality_revised}, the asymptotic covariance matrix of the MLE is $ \frac{1}{N} \bm{I}_1(\bm{\alpha}^*)^{-1} = \bm{I}_N(\bm{\alpha}^*)^{-1}$. Since $\bm{I}_N(\bm{\alpha}^*)^{-1}$ is the CRLB for an unbiased estimator based on $N$ samples, the MLE is asymptotically the most precise unbiased estimator possible.
\end{enumerate}

These properties make the MLE a cornerstone of parametric inference. Consistency ensures accuracy with large data, asymptotic normality provides a basis for constructing confidence intervals and hypothesis tests, and asymptotic efficiency guarantees optimal precision in the large-sample limit. The ability to approximate the distribution of $\hat{\bm{\alpha}}_{N}$ using the normal distribution is particularly powerful for practical applications. For instance, an approximate $100(1-\beta)\%$ confidence interval for a scalar parameter $\alpha_i^*$ can be constructed as $\hat{\alpha}_{i,N} \pm z_{\beta/2} \sqrt{[\bm{I}_N(\hat{\bm{\alpha}}_N)^{-1}]_{ii}}$, where $z_{\beta/2}$ is the appropriate quantile of the standard normal distribution and the FIM is estimated by plugging in the MLE $\hat{\bm{\alpha}}_N$.

\subsection{Proof Sketch for Asymptotic Normality (Scalar Parameter)}
\label{subsec:mle_normality_proof}

We outline the proof for the asymptotic normality \cref{eq:mle_asymptotic_normality_revised} in the case of a scalar parameter $\alpha$. Let $X = \{x_1, \dots, x_N\}$ be $N$ i.i.d.\ observations from $p(x\mid \alpha)$, with true parameter $\alpha^*$ and log-likelihood $\ell(\alpha\mid X) = \sum_{i=1}^N \ln p(x_i\mid \alpha)$. The score function is $U_N(\alpha) = \frac{\partial \ell}{\partial \alpha} = \sum_{i=1}^N \frac{\partial}{\partial \alpha} \ln p(x_i\mid \alpha)$. The MLE $\hat{\alpha}_N$ is defined by the condition $U_N(\hat{\alpha}_N) = 0$.

\subsubsection*{Step 1: Taylor Expansion of the Score Function}
We expand the score function $U_N(\alpha)$ around the true value $\alpha^*$ using Taylor's theorem. Evaluating the expansion at $\alpha = \hat{\alpha}_N$:
\begin{equation}
    U_N(\hat{\alpha}_N) \approx U_N(\alpha^*) + (\hat{\alpha}_N - \alpha^*) U_N'(\alpha^*) 
    \label{eq:taylor_score_revised}
\end{equation}
where $U_N'(\alpha) = \frac{\partial^2 \ell}{\partial \alpha^2}$ is the Hessian (scalar here) of the log-likelihood. Since $U_N(\hat{\alpha}_N) = 0$ by definition of the MLE, we rearrange:
\begin{equation}
    -(\hat{\alpha}_N - \alpha^*) U_N'(\alpha^*) \approx U_N(\alpha^*) 
    \label{eq:taylor_rearranged}
\end{equation}
This allows the approximation:
\begin{equation}
    \hat{\alpha}_N - \alpha^* \approx -\frac{U_N(\alpha^*)}{U_N'(\alpha^*)}.
    \label{eq:alpha_diff_approx}
\end{equation}
Multiplying by $\sqrt{N}$ gives the key relationship:
\begin{equation}
    \sqrt{N}(\hat{\alpha}_N - \alpha^*) \approx \frac{U_N(\alpha^*) / \sqrt{N}}{-U_N'(\alpha^*) / N}.
    \label{eq:scaled_diff_approx_revised}
\end{equation}

\subsubsection*{Step 2: Limiting Behavior of Scaled Score and Hessian}

\textbf{Numerator:} $U_N(\alpha^*) = \sum_{i=1}^N \left. \frac{\partial}{\partial \alpha} \ln p(x_i\mid \alpha) \right|_{\alpha=\alpha^*}$. This is a sum of $N$ i.i.d.\ random variables (the individual scores evaluated at $\alpha^*$). Each term has mean $\E{\left. \frac{\partial}{\partial \alpha} \ln p(x\mid \alpha) \right|_{\alpha=\alpha^*}} = 0$ and variance $\E{\left( \left. \frac{\partial}{\partial \alpha} \ln p(x\mid \alpha) \right|_{\alpha=\alpha^*} \right)^2} = I_1(\alpha^*)$ (the Fisher information for one observation). By the Central Limit Theorem (CLT):
\begin{equation}
    \frac{1}{\sqrt{N}} U_N(\alpha^*) = \frac{\sum_{i=1}^N \left. \frac{\partial}{\partial \alpha} \ln p(x_i\mid \alpha) \right|_{\alpha=\alpha^*}}{\sqrt{N}} \xrightarrow[N \to \infty]{d} \mathcal{N}\bigl(0, I_1(\alpha^*)\bigr).
    \label{eq:clt_score_revised}
\end{equation}

\textbf{Denominator:} $-U_N'(\alpha^*) / N = -\frac{1}{N} \sum_{i=1}^N \left. \frac{\partial^2}{\partial \alpha^2} \ln p(x_i\mid \alpha) \right|_{\alpha=\alpha^*}$. This is the sample average of $N$ i.i.d.\ random variables. By the Law of Large Numbers (LLN):
\begin{equation}
    -\frac{1}{N} U_N'(\alpha^*) \xrightarrow[N \to \infty]{p} -\E{\left. \frac{\partial^2}{\partial \alpha^2} \ln p(x\mid \alpha) \right|_{\alpha=\alpha^*}} =  I_1(\alpha^*).
    \label{eq:lln_hessian_raw}
\end{equation}
($\xrightarrow{p}$ denotes convergence in probability).

\subsubsection*{Step 3: Applying Slutsky's Theorem}
We now combine the limits for the numerator \cref{eq:clt_score_revised} and the denominator \cref{eq:lln_hessian_raw} using the approximation \cref{eq:scaled_diff_approx_revised}. Slutsky's Theorem states that if $X_N \xrightarrow{d} X$ and $Y_N \xrightarrow{p} c$ (where $c$ is a constant), then $X_N / Y_N \xrightarrow{d} X / c$. Applying this here:
\begin{equation}
    \sqrt{N}(\hat{\alpha}_N - \alpha^*) \xrightarrow[N \to \infty]{d} \frac{\mathcal{N}\bigl(0, I_1(\alpha^*)\bigr)}{I_1(\alpha^*)}.
\end{equation}
Since dividing a $\mathcal{N}(0, \sigma^2)$ variable by a constant $c$ results in a $\mathcal{N}(0, \sigma^2/c^2)$ variable, we get:
\begin{equation}
    \sqrt{N}(\hat{\alpha}_N - \alpha^*) \xrightarrow[N \to \infty]{d} \mathcal{N}\left(0, \frac{I_1(\alpha^*)}{[I_1(\alpha^*)]^2}\right) = \mathcal{N}\left(0, I_1(\alpha^*)^{-1}\right).
    \label{eq:mle_asymptotic_normality_final}
\end{equation}
This completes the sketch of the proof for the asymptotic normality of the scalar MLE. The proof for the vector case $\hat{\bm{\alpha}}_N$ follows analogous steps using multivariate versions of the Taylor expansion, CLT, LLN, and Slutsky's theorem, involving the score vector and the Fisher Information matrix \cite{kayFundamentalsStatisticalSignal1993}.

\section{Connection Between Maximum Likelihood and Bayesian Estimation}
\label{sec:mle_bayes_connection}

While Maximum Likelihood Estimation (MLE) provides a powerful framework based solely on the data, it is instructive to understand its close relationship to Bayesian estimation methods. In many circumstances, the MLE can be interpreted as a specific case of a Bayesian estimator, highlighting the conceptual links between frequentist and Bayesian perspectives.

In the Bayesian framework, inference revolves around the \textbf{posterior distribution} $p(\bm{\alpha} | \bm{X})$. This distribution represents our updated beliefs about the parameters $\bm{\alpha}$ after observing the data $\bm{X}$. It is obtained by combining our \textit{prior beliefs} $p(\bm{\alpha})$ (what we believe about $\bm{\alpha}$ before seeing the data) with the \textit{likelihood} $\mathcal{L}(\bm{\alpha}\mid \bm{X})$ (information from the data) using Bayes' theorem:
\begin{equation}
    p(\bm{\alpha}|\bm{X}) = \frac{p(\bm{X} | \bm{\alpha})\, p(\bm{\alpha})}{p(\bm{X})} = \frac{\mathcal{L}(\bm{\alpha}\mid \bm{X})\, p(\bm{\alpha})}{\int_{\mathcal{A}} \mathcal{L}(\bm{\alpha}'\mid \bm{X})\, p(\bm{\alpha}')\, \dd{\bm{\alpha}'}}.
    \label{eq:bayes_theorem}
\end{equation}
Here, $p(\bm{X} | \bm{\alpha})$ is simply the probability of the data given the parameters, which is identical to the likelihood function $\mathcal{L}(\bm{\alpha}\mid \bm{X})$. The denominator $p(\bm{X}) = \int_{\mathcal{A}} \mathcal{L}(\bm{\alpha}'\mid \bm{X})\, p(\bm{\alpha}')\, \dd{\bm{\alpha}}'$ is the marginal likelihood or evidence, which serves as a normalization constant ensuring the posterior integrates to one. While the full posterior distribution is the primary target of Bayesian inference, point estimates can also be derived.

A common Bayesian point estimate is the \textbf{Maximum A Posteriori} (MAP) estimator, $\hat{\bm{\alpha}}_{\mathrm{MAP}}$. This is the value of $\bm{\alpha}$ that maximizes the posterior probability density:
\begin{equation}
    \hat{\bm{\alpha}}_{\mathrm{MAP}} = \argmax_{\bm{\alpha} \in \mathcal{A}} p(\bm{\alpha}|\bm{X}).
    \label{eq:map_definition_argmax_posterior}
\end{equation}
where $\mathcal{A}$ is our parameter space for $\bm{\alpha}$. Since the evidence $p(\bm{X})$ does not depend on $\bm{\alpha}$, maximizing the posterior is equivalent to maximizing the product of the likelihood and the prior, or equivalently, their logarithms:
\begin{equation}
    \hat{\bm{\alpha}}_{\mathrm{MAP}} = \argmax_{\bm{\alpha} \in \mathcal{A}} \left[ \mathcal{L}(\bm{\alpha}\mid\bm{X}) \, p(\bm{\alpha}) \right] = \argmax_{\bm{\alpha} \in \mathcal{A}} \left[ \ell(\bm{\alpha}\mid\bm{X}) + \ln p(\bm{\alpha}) \right].
    \label{eq:map_definition}
\end{equation}
Comparing the MAP objective function in \cref{eq:map_definition} with the objective function for the MLE (which is simply maximizing $\ell(\bm{\alpha}\mid\bm{X})$, see \cref{eq:loglikelihood_definition}), we clearly see the influence of the prior term $\ln p(\bm{\alpha})$ in the MAP estimation. The prior effectively regularizes the likelihood, pulling the estimate towards regions favored by the prior beliefs.

The connection to MLE becomes explicit when we consider a specific choice for the prior. If the prior distribution $p(\bm{\alpha})$ is chosen to be \textbf{uniform} over the allowed parameter space $\mathcal{A}$, meaning $p(\bm{\alpha}) \propto \text{constant}$, then its logarithm $\ln p(\bm{\alpha})$ is also a constant. In this case, the $\ln p(\bm{\alpha})$ term in the MAP objective \eqref{eq:map_definition} does not affect the location of the maximum. Therefore, maximizing the posterior becomes equivalent to maximizing the likelihood:
\begin{equation}
    \text{Uniform prior } p(\bm{\alpha}) \implies \hat{\bm{\alpha}}_{\mathrm{MAP}} = \argmax_{\bm{\alpha} \in \mathcal{A}} \ell(\bm{\alpha}\mid\bm{X}) = \hat{\bm{\alpha}}_{\mathrm{MLE}}.
\end{equation}
Thus, the MLE can be rigorously interpreted as a MAP estimate under a uniform prior assumption. This corresponds to a scenario where no specific prior knowledge is incorporated, and the parameter estimation relies entirely on the information provided by the data through the likelihood function.

While Bayesian methods offer a powerful way to incorporate prior knowledge and obtain full distributional estimates, specifying a meaningful and justifiable prior $p(\bm{\alpha})$ can be challenging, especially for complex models like the SDEs considered later in this work. When prior information is weak, unavailable, or deliberately excluded to focus purely on data-driven inference, the MLE provides a well-established and asymptotically optimal approach, as discussed in \cref{sec:mle_asymptotics}. 

\section{Conclusion and Outlook} 
\label{sec:inference_conclusion_outlook} 

This chapter has laid out the cornerstones of parameter estimation using the likelihood approach. We defined the likelihood function $\mathcal{L}(\bm{\alpha}\mid \bm{X})$ and the principle of Maximum Likelihood Estimation (MLE) to find the parameters $\hat{\bm{\alpha}}_{\mathrm{MLE}}$ that best explain observed data (\cref{eq:mle_definition}). We explored how to evaluate estimator performance through the Mean Squared Error (\cref{eq:mse_definition_expanded}) and its fundamental decomposition into bias and variance (\cref{eq:mse_bias_variance_decomp}). The Cramér-Rao Lower Bound (\cref{eq:crlb_vector_revised}) established theoretical limits on the precision of unbiased estimators via the Fisher Information matrix (\cref{eq:fisher_info_definition_revised}). Furthermore, we discussed the desirable asymptotic properties of the MLE under regularity conditions, including consistency, asymptotic normality (\cref{eq:mle_asymptotic_normality_revised}), and efficiency, which justify its widespread use. Finally, the connection to Bayesian inference via the MAP estimator (\cref{eq:map_definition}) was highlighted, showing MLE corresponds to MAP with a uniform prior.

These concepts provide a powerful and general statistical foundation for inference. However, applying them effectively requires tailoring them to the specific structure of the models generating the data. Having established the mathematical framework for Stochastic Differential Equations, particularly the Langevin equation (\cref{eq:chap_1_the_overdamped_langevin_general}), in the "\textit{SDE Toolbox}" chapter, we are now positioned to bridge the general estimation theory developed here with the specific context of continuous-time stochastic processes.

The challenge now shifts to the practical inference of parameters for SDE models. The next \cref{part:Learning_Langevin_equation} tackles this problem directly.

%% file: tex_body/chap2_b.tex
\part{Learning Langevin Equations}
\label{part:Learning_Langevin_equation}

\chapter{Inference Methods for Langevin Equations}
\label{chap:inference_langevin}
\epigraph{Be you, be proud of you because you can be do what we want to do}{Francois Hollande, French president 2012-2017}
\chaptertoc{}

\section{Introduction}
\label{sec:introduction_inference_langevin}
In the previous \cref{part:Introdution_langevin}, 
we explored the foundational concepts for parameter estimation of independent and identically distributed (i.i.d.) random variables. We now apply this knowledge to our central stochastic process: the Langevin equation. This chapter focuses on methods to infer the drift \( \bm{f}(\bm{x}) \) and diffusion matrix \( \bm{D}(\bm{x}) \) of a \( d \)-dimensional Langevin equation from a discretely observed trajectory.

We assume that a continuous process \( \bm{x}_t \) is measured at discrete time intervals \( \Delta t \). Consequently, we obtain a trajectory
\[
\bm{X}_\tau = \{ \bm{x}_{t_0}, \bm{x}_{t_1}, \dots, \bm{x}_{t_N} \},
\]
with \( t_N = \tau \). For example, \( \bm{x}_t \) may represent the position of a self-propelled particle.

Furthermore, we assume that \( \bm{x}_t \) obeys a \( d \)-dimensional Langevin equation:
\begin{equation}
    \dd{\bm{x}_t} = \bm{f}(\bm{x}_t) \dd{t} + \sqrt{2\, \bm{D}(\bm{x}_t)}\, \dd{\bm{W}_t},
    \label{eq:chap_2a_langevin}
\end{equation}
which belongs to a widely used class of continuous stochastic dynamical models. Here, the drift \( \bm{f}(\bm{x}) \) represents the deterministic component, the positive-definite symmetric diffusion matrix \( \bm{D}(\bm{x}) \) accounts for the stochastic fluctuations, and \( \bm{W}_t \) is a \( d \)-dimensional Wiener process. While defined generally with a state-dependent diffusion matrix \( \bm{D}(\bm{x}_t) \), to obtain tractable analytical estimators, we will often focus on the case of additive noise where \( \bm{D} \) is constant.

Our objective is to reconstruct an inferred drift \( \hat{\bm{f}}(\bm{x}) \) from the observed trajectory \( \bm{X}_\tau \) that closely approximates the true drift \( \bm{f}(\bm{x}) \). To this end, we express the drift as a linear combination of basis functions:
\begin{equation}
    \bm{f}(\bm{x}) = \sum_{i=1}^{n_{\mathcal{B}}} \alpha_i^* \, \bm{b}_i(\bm{x}),
\end{equation}
where \( \mathcal{B} = \{\bm{b}_1, \dots, \bm{b}_{n_{\mathcal{B}}}\} \) is a chosen set of basis functions and the true coefficients \( \alpha_i^* \) determine the contribution of each basis function. Approximating the drift using a linear combination of basis functions offers several advantages:

\begin{enumerate}
    \item \textbf{Dimensionality Reduction:} It reduces the infinite-dimensional problem of function estimation to a finite-dimensional parameter estimation problem (finding the coefficients \( \alpha_i^* \)).
    \item \textbf{Flexibility:} By selecting a sufficiently rich set of basis functions (e.g., polynomials, Fourier modes, radial basis functions, wavelets), a wide variety of drift functional forms can be accurately approximated.
    \item \textbf{Incorporation of Prior Knowledge:} The choice of basis can incorporate prior physical knowledge about the system. For example, if the system is known to be confined near an equilibrium point, polynomial basis functions might be appropriate. If periodicity is expected, Fourier modes could be a natural choice.
    \item \textbf{Regularization and Model Selection:} Limiting the number of basis functions \( n_{\mathcal{B}} \) provides intrinsic regularization, helping to prevent overfitting the noise in the data, especially when the trajectory length \( \tau \) is limited. Furthermore, this framework naturally lends itself to model selection: setting a coefficient \( \alpha_i \) to zero effectively removes the corresponding basis function \( \bm{b}_i \) from the model. This allows for the development of principled methods to simplify the model by eliminating unnecessary basis functions (i.e., setting their coefficients to zero), which we will discuss further in \cref{part:Model_selection}.
\end{enumerate}

For these reasons, approximating drift or force fields using basis functions is a well-established approach in the study of stochastic differential equations, as evidenced by numerous studies in the literature \cite{boninsegnaSparseLearningStochastic2018,brucknerLearningDynamicsCell2021,callahamNonlinearStochasticModelling2021,gaoAutonomousInferenceComplex2022,huangSparseInferenceActive2022,wannerHigherOrderDrift2024,courseStateEstimationPhysical2023,amiriInferringGeometricalDynamics2024,nabeelDiscoveringStochasticDynamical2025,ferrettiBuildingGeneralLangevin2020}.

Beyond the strategy of approximating the drift via basis functions, alternative inference techniques have been explored in the literature. One notable direction employs Gaussian processes (GPs) to perform non-parametric inference, directly learning the drift function's form from data without assuming a fixed parametric structure \cite{batzApproximateBayesLearning2018}. More recently, the emergence of Physics-Informed Neural Networks (PINNs) \cite{raissiPhysicsinformedNeuralNetworks2019,karniadakisPhysicsinformedMachineLearning2021,wangExpertsGuideTraining2023} has provided another powerful toolset. PINNs leverage deep learning architectures constrained by physical laws; in this context, they are used to infer the drift and diffusion coefficients directly \cite{olearyStochasticPhysicsinformedNeural2022a} or by incorporating knowledge of the corresponding Fokker-Planck equation that governs the probability density evolution of the SDE \cite{chenSolvingInverseStochastic2020,huScoreBasedPhysicsInformedNeural2024}.

While these advanced techniques like Gaussian Processes and Physics-Informed Neural Networks offer powerful alternative perspectives, this chapter will return to the foundational approach of parameter estimation within a defined model structure. Specifically, we will focus on developing the maximum likelihood estimation (MLE) framework, often employed in conjunction with the basis function approximations discussed earlier. This work will be the foundation for our next topic: the selection of a minimal model, as detailed in \cref{part:Model_selection}. 

\section{Maximum Likelihood Estimation Framework}
\label{sec:mle_framework}

\subsection{Likelihood for Data Modeled by a Langevin Equation}
\label{subsection:likelihood_derivation}
To derive a tractable likelihood function, we employ the Euler-Maruyama discretization of \cref{eq:chap_2a_langevin}. Furthermore, as introduced in \cref{sec:introduction_inference_langevin}, we will focus primarily on the case of \textbf{additive noise}, meaning the diffusion matrix \( \bm{D} \) is assumed constant and independent of the state \( \bm{x}_t \). While the general framework can be extended to state-dependent diffusion, the constant \( \bm{D} \) case allows for a clearer derivation of the core estimation principles.

To construct the likelihood, we employ the Euler-Maruyama discretization of \cref{eq:chap_2a_langevin} over a small time step \( \Delta t \). This scheme assumes \( \bm{f}(\bm{x}_t) \) and \( \bm{D}(\bm{x}_t) \) are approximately constant within each interval \( [t_i, t_i+\Delta t] \), evaluating them at \( t_i \). For constant \( \bm{D} \), rearranging the discretized equation gives:

\begin{equation}
        \frac{\Delta {\bm{W}_{t_i}}}{\sqrt{\Delta{t}}} = \frac{\bm{x}_{t_i+\Delta{t}} - \bm{x}_{t_i} - \bm{f}(\bm{x}_{t_i}) \Delta{t}}{ \sqrt{2 \Delta{t} \bm{D}}}
        \label{eq:chap_2a_discretized_noise}
\end{equation}

Recall that \( \Delta \bm{W}_{t_i} = \bm{W}_{t_i+\Delta t} - \bm{W}_{t_i} \) is a normally distributed random vector with mean \( \bm{0} \) and covariance matrix \( \Delta t\,\bm{I} \). Consequently, \( \bm{x}_{t_i+\Delta t} \) conditioned on \( \bm{x}_{t_i} \) follows a Gaussian distribution. The conditional probability density function (transition probability) is

\begin{equation}
    P(\bm{x}_{t_i+\Delta t} \mid \bm{x}_{t_i}; \bm{f}, \bm{D}) = \frac{1}{\sqrt{\det\left(4\pi\,\bm{D}\Delta t\right)}}
    \exp\!\left\{-\frac{1}{4\Delta t}\left[\Delta \bm{x}_{t_i} - \bm{f}(\bm{x}_{t_i})\Delta t\right]^\top \!\bm{D}^{-1}\!\left[\Delta \bm{x}_{t_i} - \bm{f}(\bm{x}_{t_i})\Delta t\right]\right\},
    \label{eq:chap_2a_transition_prob}
\end{equation}
with the increment
\[
\Delta \bm{x}_{t_i} = \bm{x}_{t_i+\Delta t} - \bm{x}_{t_i}.
\]
This expression can be rewritten as:
\begin{equation}
    P(\bm{x}_{t_i+\Delta t} \mid \bm{x}_{t_i}; \bm{f}, \bm{D}) = \frac{1}{\sqrt{\det\left(4\pi\,\bm{D}\Delta t\right)}}
    \exp\!\left\{-\left[\frac{\Delta \bm{x}_{t_i}}{\Delta t} - \bm{f}(\bm{x}_{t_i})\right]^\top \!\left(4\,\bm{D}\right)^{-1}\!\left[\frac{\Delta \bm{x}_{t_i}}{\Delta t} - \bm{f}(\bm{x}_{t_i})\right]\,\Delta t\right\}.
\end{equation}

By applying the chain rule of probability and using the Markov property inherent in the Langevin equation (i.e., \( P(\bm{x}_{t_{i+1}} \mid \bm{x}_{t_i}, \dots, \bm{x}_{t_1}) = P(\bm{x}_{t_{i+1}} \mid \bm{x}_{t_i}) \)), the probability density function, or likelihood, for the entire trajectory becomes
\begin{align*}
    P(\bm{X}_\tau \mid \bm{f}, \bm{D})
    &= P(\bm{x}_{t_N} \mid \bm{x}_{t_{N-1}}) \, P(\bm{x}_{t_{N-1}} \mid \bm{x}_{t_{N-2}}) \cdots P(\bm{x}_{t_1} \mid \bm{x}_{t_0}) \, P(\bm{x}_{t_0})\\[1mm]
    &= \left[\prod_{i=0}^{N-1} P(\bm{x}_{t_{i+1}} \mid \bm{x}_{t_i})\right] P(\bm{x}_{t_0})\\[1mm]
    &= \frac{P(\bm{x}_{t_0})}{Z_{\bm{D}}} \exp\!\left\{-\sum_{i=0}^{N-1}\left[\frac{\Delta \bm{x}_{t_i}}{\Delta t} - \bm{f}(\bm{x}_{t_i})\right]^\top \!\left(4\,\bm{D}\right)^{-1}\!\left[\frac{\Delta \bm{x}_{t_i}}{\Delta t} - \bm{f}(\bm{x}_{t_i})\right]\Delta t \right\},
\end{align*}
where the normalization constant depends on \( \bm{D} \) but not the drift parameters:
\[
Z_{\bm{D}} = \prod_{i=0}^{N-1} \sqrt{\det\left(4\pi\,\bm{D}\Delta t\right)} = \det\left(4\pi\,\bm{D}\Delta t\right)^{N/2}.
\]
For a more detailed derivation, see \cite{riskenFokkerPlanckEquationMethods1996}, p.73.

The full likelihood expression derived above includes the initial state probability density \( P(\bm{x}_{t_0}) \). For the purpose of estimating the parameters \( \bm{f} \) and \( \bm{D} \) that govern the system's dynamics, this term is commonly omitted. This is justified either by considering the likelihood \textit{conditional} on the observed starting point \( \bm{x}_{t_0} \) (in which case \( P(\bm{x}_{t_0}) \) is treated as a constant factor independent of the parameters \( \bm{f}, \bm{D} \) to be estimated) or by noting that for sufficiently long trajectories (large \( N \)), the product of the \( N \) transition terms \( P(\bm{x}_{t_{i+1}} \mid \bm{x}_{t_i}) \) dominates the contribution of the single initial state term, making its influence asymptotically negligible. Consequently, we will proceed by maximizing the product of the transition probabilities, which is equivalent to maximizing the conditional likelihood \( P(\bm{x}_{t_1}, \dots, \bm{x}_{t_N} \mid \bm{x}_{t_0}, \bm{f}, \bm{D}) \).

The log-likelihood \( \ell \) of observing the trajectory \( \bm{X}_\tau \) given the model parameters \( \bm{f} \) and \( \bm{D} \) is then defined as \( \ell( \bm{f}, \bm{D} \mid \bm{X}_\tau) = \ln P(\bm{X}_\tau \mid \bm{x}_{t_0}, \bm{f}, \bm{D}) \):
\begin{equation}
    \ell( \bm{f}, \bm{D} \mid \bm{X}_\tau)
    = -\sum_{i=0}^{N-1}\left[\frac{\Delta \bm{x}_{t_i}}{\Delta t} - \bm{f}(\bm{x}_{t_i})\right]^\top \!\left(4\,\bm{D}\right)^{-1}\!\left[\frac{\Delta \bm{x}_{t_i}}{\Delta t} - \bm{f}(\bm{x}_{t_i})\right]\Delta t - \ln Z_{\bm{D}}.
    \label{eq:chap_2a_loglikelihood_sum}
\end{equation}
It is often convenient to express this using a trajectory average, denoted by \( \langle \cdot \rangle \):
\begin{equation}
    \ell( \bm{f}, \bm{D} \mid \bm{X}_\tau)
    = -\tau\,\Biggl\langle \left[\frac{\Delta \bm{x}_t}{\Delta t} - \bm{f}(\bm{x}_t)\right]^\top \!\left(4\,\bm{D}\right)^{-1}\!\left[\frac{\Delta \bm{x}_t}{\Delta t} - \bm{f}(\bm{x}_t)\right] \Biggr\rangle - \ln Z_{\bm{D}},
    \label{eq:chap_2a_loglikelihood_average}
\end{equation}
where the time average over the trajectory \( \bm{X}_\tau \) for a test function $h$ is defined by
\[
\langle h(\bm{x}_t) \rangle =  \frac{1}{N}\sum_{i=0}^{N-1} h(\bm{x}_{t_i}) = \frac{1}{\tau}\sum_{i=0}^{N-1} h(\bm{x}_{t_i})\,\Delta t,
\]
with \( \tau = N \Delta t \) being the total observation time considered in the average. 

\subsection{Maximizing the Likelihood}
\label{subsec:mle_maximization}

To obtain an analytical solution for the maximum likelihood estimators (MLEs), we parametrize the unknown drift $\bar{\bm{f}}$ as a linear combination of basis functions:
\begin{equation}
    \bar{\bm{f}}(\bm{x}) = \sum_{i=1}^{n_{\mathcal{B}}} \alpha_i\,\bm{b}_i(\bm{x}).
\end{equation}
Here, \( \bm{\alpha} = (\alpha_1, \dots, \alpha_{n_{\mathcal{B}}})^\top \) is the vector of parameters to be estimated. Additionally, we treat the elements of the \( d \times d \) diffusion matrix \( \bm{D} \) as scalar parameters to be estimated. We discussed these approximations in the introduction (\cref{sec:introduction_inference_langevin}). 

Within this framework, the log-likelihood \cref{eq:chap_2a_loglikelihood_average} becomes a function of \( \bm{\alpha} \) and \( \bm{D} \):
\begin{equation}
    \ell( \bm{\alpha}, \bm{D} \mid \bm{X}_\tau)
    = -\tau\,\Biggl\langle \left[\frac{\Delta \bm{x}_t}{\Delta t} - \sum_{i=1}^{n_{\mathcal{B}}} \alpha_i\,\bm{b}_i(\bm{x}_t)\right]^\top \!\left(4\,\bm{D}\right)^{-1}\!\left[\frac{\Delta \bm{x}_t}{\Delta t} - \sum_{j=1}^{n_{\mathcal{B}}} \alpha_j\,\bm{b}_j(\bm{x}_t)\right] \Biggr\rangle - \ln Z_{\bm{D}}.
\end{equation}
Note the change of summation index in the second bracket for clarity in differentiation.

The maximum likelihood estimators \( \hat{\bm{\alpha}}_{\text{MLE}} \) and \( \hat{\bm{D}}_{\text{MLE}} \) are found by setting the partial derivatives of the log-likelihood with respect to each parameter to zero:
\begin{equation}
    \frac{\partial \ell( \hat{\bm{\alpha}}_{\text{MLE}}, \hat{\bm{D}}_{\text{MLE}} \mid \bm{X}_\tau )}{\partial \alpha_j} = 0
    \quad \text{for } j=1, \dots, n_{\mathcal{B}},
    \label{eq:chap_2b_likelihood_conditions_alpha}
\end{equation}
\begin{equation}
    \frac{\partial \ell( \hat{\bm{\alpha}}_{\text{MLE}}, \hat{\bm{D}}_{\text{MLE}} \mid \bm{X}_\tau )}{\partial D_{\beta\gamma}} = 0
    \quad \text{for } 1 \le \beta \le \gamma \le d.
    \label{eq:chap_2b_likelihood_conditions_D}
\end{equation}

In the remainder of this chapter, we adopt the convention that Latin indices (e.g., \( i, j \)) denote the components of the parameter vector \( \bm{\alpha} \) and the basis function index, while Greek indices (e.g., \( \beta, \gamma \)) denote spatial components (dimensions \(1, \dots, d\)).

We note that applying the maximum likelihood principle (\cref{eq:chap_2a_loglikelihood_average}) to estimate parameters in Langevin-type equations is a well-established methodology. It has been explored for various model structures and parameterizations in prior work, including contributions by \cite{kleinhansIterativeProcedureEstimation2005,kleinhansMaximumLikelihoodEstimation2007,fricksTimeDomainMethodsDiffusive2009}.

\subsubsection{Drift Estimator}
\label{subsubsec:conditional_drift_mle}

To estimate the drift parameters \( \bm{\alpha} \), we enforce the condition given in \cref{eq:chap_2b_likelihood_conditions_alpha}. Differentiating the log-likelihood with respect to \(\alpha_j\) yields (using the symmetry of \( \bm{D}^{-1} \)):
\begin{align}
    \pdv{\ell}{\alpha_j} &= -\tau \left\langle \left[-\bm{b}_j(\bm{x}_t)\right]^\top \!\left(4\,\bm{D}\right)^{-1}\!\left[\frac{\Delta \bm{x}_t}{\Delta t} - \sum_{i=1}^{n_{\mathcal{B}}} \alpha_i\,\bm{b}_i(\bm{x}_t)\right] \right. \nonumber \\
    &\quad \left. - \left[\frac{\Delta \bm{x}_t}{\Delta t} - \sum_{i=1}^{n_{\mathcal{B}}} \alpha_i\,\bm{b}_i(\bm{x}_t)\right]^\top \!\left(4\,\bm{D}\right)^{-1}\!\left[-\bm{b}_j(\bm{x}_t)\right] \right\rangle \nonumber\\
    &= 2\tau \left\langle \bm{b}_j(\bm{x}_t)^\top (4\bm{D})^{-1}\left(\frac{\Delta \bm{x}_t}{\Delta t} - \sum_{i=1}^{n_{\mathcal{B}}} \alpha_i\,\bm{b}_i(\bm{x}_t)\right) \right\rangle = 0.
\end{align}
This condition implies that for the MLE \( \hat{\bm{\alpha}}_{\text{MLE}} \):
\begin{equation}
    \sum_{i=1}^{n_{\mathcal{B}}} \hat{\alpha}_{\text{MLE},i} \left\langle \bm{b}_j^\top (4\bm{D})^{-1}\,\bm{b}_i \right\rangle = \left\langle \bm{b}_j^\top (4\bm{D})^{-1}\frac{\Delta \bm{x}_t}{\Delta t} \right\rangle.
\end{equation}
This is a system of linear equations for \( \hat{\bm{\alpha}}_{\text{MLE}} \). Defining the matrix \( \bm{G} \) and vector \( \bm{Y} \) (both dependent on \( \bm{D} \)) as
\begin{align}
    G_{ji} &= \left\langle \bm{b}_j^\top (4\bm{D})^{-1}\,\bm{b}_i \right\rangle, \label{eq:chap2a_G_def_mle} \\
    Y_j &= \left\langle \bm{b}_j^\top (4\bm{D})^{-1}\frac{\Delta \bm{x}_t}{\Delta t} \right\rangle, \label{eq:chap2a_Y_def_mle}
\end{align}
the system becomes \( \sum_{i=1}^{n_{\mathcal{B}}} G_{ji} \hat{\alpha}_{\text{MLE},i} = Y_j \), or in matrix form \( \bm{G} \hat{\bm{\alpha}}_{\text{MLE}} = \bm{Y} \). The solution is:
\begin{equation}
    \hat{\bm{\alpha}}_{\text{MLE}} = \bm{G}^{-1} \bm{Y},
\end{equation}
or element-wise:
\begin{equation}
    \hat{\alpha}_{\text{MLE},i} = \sum_{j=1}^{n_{\mathcal{B}}} \left(G^{-1}\right)_{ij} \left\langle \bm{b}_j^\top (4\bm{D})^{-1}\frac{\Delta \bm{x}_t}{\Delta t} \right\rangle.
    \label{eq:chap2a_alpha_likelihood}
\end{equation}
This provides an explicit estimator for the drift coefficients \( \hat{\bm{\alpha}}_{\text{MLE}} \), \textbf{provided the diffusion matrix \( \bm{D} \) is known}.

\subsubsection{Diffusion Matrix Estimator}
\label{sec:chap_2b_Diffusion_maximum_likelihood_estimator}
To estimate the diffusion matrix, we enforce the corresponding condition from \cref{eq:chap_2b_likelihood_conditions_D}. In doing so, we will need the following standard identities:
\begin{equation}
    \dv{\bm{D}^{-1}}{D_{\beta\gamma}} = -\bm{D}^{-1}\bm{E}_{\beta\gamma}\bm{D}^{-1}, \qquad
    \dv{\det\bm{D}}{D_{ij}} = \det\bm{D}\, (\bm{D}^{-1})_{ji}, \qquad
    \ln Z_{\bm{D}} = \frac{\tau}{2\Delta t}\ln\det\bm{D} + C,
\end{equation}
where \(\bm{E}_{\beta\gamma}\) is a matrix whose entries are zero except for a one in the \((\beta,\gamma)\) position, and \(C\) is a constant independent of \(\bm{D}\). (A proof of the first two identities can be found in standard texts; see also Footnote\footnote{From $\bm{D}\bm{D}^{-1} = \bm{I}$ where $\bm{I}$ is the identity matrix, we obtain that $\bm{D}\dv{\bm{D}^{-1}}{D_{\beta \gamma}} + \dv{\bm{D}}{D_{\beta \gamma}}\bm{D}^{-1} = 0$. Hence, $\dv{\bm{D}^{-1}}{D_{\beta \gamma}} = - \bm{D}^{-1} \bm{E}_{\beta \gamma} \bm{D}^{-1}$}.)

Defining the residual vector
\[
\bm{r}_t = \frac{\Delta \bm{x}_t}{\Delta t} - \sum_{i=1}^{n_{\mathcal{B}}} \alpha_i\,\bm{b}_i(\bm{x}_t),
\]
we seek an extremum of the log-likelihood with respect to \(D_{\beta\gamma}\):
\begin{equation*}
    \pdv{\ell}{D_{(\beta\gamma)}} = \frac{\tau}{4}\left\langle \bm{r}_t^\top\,\bm{D}^{-1}\bm{E}_{\beta\gamma}\bm{D}^{-1}\,\bm{r}_t \right\rangle - \frac{\tau}{2\Delta t} (\hat{D}^{-1})_{\gamma\beta} = 0.
\end{equation*}
Since \(\bm{D}\) is symmetric, we have
\[
\sum_{\zeta,\eta} (r_t)_\zeta\,D^{-1}_{\zeta\beta}\,D^{-1}_{\gamma\eta}\,(r_t)_\eta = \sum_{\zeta,\eta} D^{-1}_{\beta\zeta}\,(r_t)_\zeta\,(r_t)_\eta\,D^{-1}_{\eta\gamma},
\]
where \(\zeta,\eta,\beta,\gamma\) denote spatial components. In matrix form, this condition can be written as
\begin{equation}
    \frac{\Delta t}{2}\,\bm{D}^{-1}\left\langle \bm{r}_t\,\bm{r}_t^\top \right\rangle\bm{D}^{-1} = \bm{D}^{-1}.
\end{equation}
Thus, the maximum likelihood estimator for the diffusion matrix is given by
\begin{equation}
    \hat{\bm{D}}_{\text{MLE}} = \frac{\Delta t}{2}\left\langle \left(\frac{\Delta \bm{x}_t}{\Delta t} - \sum_{i=1}^{n_{\mathcal{B}}} \hat{\alpha}_i\,\bm{b}_i(\bm{x}_t)\right) \left(\frac{\Delta \bm{x}_t}{\Delta t} - \sum_{i=1}^{n_{\mathcal{B}}} \hat{\alpha}_i\,\bm{b}_i(\bm{x}_t)\right)^\top \right\rangle.
    \label{eq:chap2a_eq_likelihood_maximization}
\end{equation}

Equations \eqref{eq:chap2a_alpha_likelihood} and \eqref{eq:chap2a_eq_likelihood_maximization} constitute a coupled system for the maximum likelihood estimators \( \hat{\bm{\alpha}}_{\text{MLE}} \) and \( \hat{\bm{D}}_{\text{MLE}} \). Specifically, the estimator for \( \hat{\bm{\alpha}}_{\text{MLE}} \) depends on \( \bm{D} \) through the matrix \( \bm{G} \) and vector \( \bm{Y} \) (\cref{eq:chap2a_G_def_mle,eq:chap2a_Y_def_mle}), while the estimator for \( \hat{\bm{D}}_{\text{MLE}} \) explicitly depends on the estimated coefficients \( \hat{\bm{\alpha}}_{\text{MLE}} \) used to compute the residuals \( \bm{r}_t \). This interdependence precludes obtaining a simultaneous, closed-form analytical solution for both \( \hat{\bm{\alpha}}_{\text{MLE}} \) and \( \hat{\bm{D}}_{\text{MLE}} \). Consequently, these parameters must typically be determined numerically. A common approach is an iterative procedure: one alternates between updating \( \hat{\bm{\alpha}} \) using the current estimate of \( \hat{\bm{D}} \) via \cref{eq:chap2a_alpha_likelihood}, and then updating \( \hat{\bm{D}} \) using the latest estimate of \( \hat{\bm{\alpha}} \) via \cref{eq:chap2a_eq_likelihood_maximization}, repeating this cycle until the estimates converge. While this iterative approach or direct numerical maximization can find the true MLEs, the computational cost and the complexity arising from the coupling motivate the exploration of alternative, potentially simpler, estimation strategies, such as those based on linear regression principles, which we discuss next.

\section{Estimators via Linear Regression}

While maximum likelihood estimation provides a rigorous framework, alternative methods based on linear regression principles can also be employed to estimate the parameters of the Langevin equation. This approach often yields estimators that are computationally simpler or have more straightforward analytical forms. This section presents such alternative estimators for the drift and diffusion terms. It should be noted that related derivations, often focusing on moment-based estimation of drift and diffusion coefficients (often called Kramers–Moyal coefficients in the literature), have been previously established for general Langevin processes in works such as \cite{siegertAnalysisDataSets1998,suraNoteEstimatingDrift2002,friedrichHowQuantifyDeterministic2000,lamourouxKernelbasedRegressionDrift2009,gottschallDefinitionHandlingDifferent2008}.

\subsection{Diffusion Matrix Estimator via Linear Regression}
We derive a simple estimator for the diffusion matrix, particularly suited for the case of additive noise (constant \( \bm{D} \)). From the discretized Langevin equation:
\[
\Delta \bm{x}_t \approx \bm{f}(\bm{x}_t)\Delta t + \sqrt{2\bm{D}}\,\Delta \bm{W}_t.
\]
Then,
\begin{align}
    \Delta \bm{x}_t\,\Delta \bm{x}_t^\top &\approx {\bm{f}(\bm{x}_t)\bm{f}(\bm{x}_t)^\top \Delta t^2} + \bm{f}(\bm{x}_t)(\sqrt{2\bm{D}}\,\Delta \bm{W}_t)^\top \Delta t + \sqrt{2\bm{D}}\,\Delta \bm{W}_t \bm{f}(\bm{x}_t)^\top \Delta t \nonumber \\
    &\quad + (\sqrt{2\bm{D}}\,\Delta \bm{W}_t)(\sqrt{2\bm{D}}\,\Delta \bm{W}_t)^\top.
\end{align}
Taking the expectation (conditional on \( \bm{x}_t \)) yields, using \( \mathbb{E}[\Delta \bm{W}_t] = 0 \) and \( \mathbb{E}[\Delta \bm{W}_t \Delta \bm{W}_t^\top] = \bm{I} \Delta t \):
\begin{equation}
    \mathbb{E}[\Delta \bm{x}_t\,\Delta \bm{x}_t^\top \mid \bm{x}_t] = \underbrace{\bm{f}(\bm{x}_t)\bm{f}(\bm{x}_t)^\top \Delta t^2}_{\mathcal{O}(\Delta t^2)} + \underbrace{2\bm{D}\,\Delta t}_{\mathcal{O}(\Delta t)}.
\end{equation}
For sufficiently small \( \Delta t \), the \( \mathcal{O}(\Delta t^2) \) term is negligible compared to the \( \mathcal{O}(\Delta t) \) term. Replacing the expectation with the sample average \( \langle \Delta \bm{x}_t\,\Delta \bm{x}_t^\top \rangle \), we obtain an estimator for \( \bm{D} \):
\begin{equation}
    \bar{\bm{D}} = \frac{1}{2\Delta t}\left\langle \Delta \bm{x}_t\,\Delta \bm{x}_t^\top \right\rangle.
    \label{eq:chap2a_approximation_D}
\end{equation}
This simple estimator is valid for additive noise and small \( \Delta t \). It corresponds to the maximum likelihood estimator \( \hat{\bm{D}}_{\text{MLE}} \) from \cref{eq:chap2a_eq_likelihood_maximization} if the estimated drift contribution \( \hat{\bm{f}}(\bm{x}_t)\Delta t \) is neglected compared to the displacement \( \Delta \bm{x}_t \).

\subsection{Drift Estimation via Linear Regression}

We start from the discretized Langevin equation (ignoring bias terms of \( \mathcal{O}(\Delta t) \) and fluctuating terms of $\mathcal{O}(1)$ and higher): 
\[
\frac{\Delta \bm{x}_t}{\Delta t} \approx \bm{f}(\bm{x}_t) + \frac{\sqrt{2\bm{D}(\bm{x}_t)}\,\Delta \bm{W}_t}{\Delta t}.
\]
Substituting the basis function expansion for the true drift, \( \bm{f}(\bm{x}) = \sum_{j=1}^{n_{\mathcal{B}}}\alpha_j^*\,\bm{b}_j(\bm{x}) \), we have:
\[
\frac{\Delta \bm{x}_t}{\Delta t} \approx \sum_{j=1}^{n_{\mathcal{B}}}\alpha_j^*\,\bm{b}_j(\bm{x}_t) + \text{noise term}.
\]
To estimate the coefficients \( \alpha_j^* \), we can perform a weighted linear regression. We multiply both sides by \( \bm{b}_i(\bm{x}_t)^\top (4\bm{A})^{-1} \) (where \( \bm{A} \) is some positive-definite weighting matrix, potentially related to \( \bm{D} \)) and take the trajectory average:
\begin{equation}
    \left\langle \bm{b}_i^\top (4\bm{A})^{-1} \frac{\Delta \bm{x}_t}{\Delta t} \right\rangle \approx \sum_{j=1}^{n_{\mathcal{B}}} \alpha_j^*\,\left\langle \bm{b}_i^\top (4\bm{A})^{-1} \bm{b}_j \right\rangle + \left\langle \bm{b}_i^\top (4\bm{A})^{-1} \frac{\sqrt{2\bm{D}}\,\Delta \bm{W}_t}{\Delta t} \right\rangle.
    \label{eq:chap2a_linear_regression_step}
\end{equation}
If the weighting matrix \( \bm{A} \)  can be considered independent of the noise increment \( \Delta \bm{W}_t \), the expectation of the noise term vanishes. This motivates defining the linear regression estimator \( \hat{\bm{\alpha}}_{\text{LR}} \) by ignoring the noise term and solving the resulting linear system:
\begin{equation}
    \hat{\alpha}_{\text{LR}, i} = \sum_{j=1}^{n_{\mathcal{B}}} \left( G_A^{-1} \right)_{ij} \left\langle \bm{b}_j^\top (4\bm{A})^{-1} \frac{\Delta \bm{x}_t}{\Delta t} \right\rangle,
    \label{eq:chap2a_alpha_linear_regression}
\end{equation}
where the matrix \( \bm{G}_A \) is defined by
\[
(G_A)_{ji} = \left\langle \bm{b}_j^\top (4\bm{A})^{-1}\,\bm{b}_i \right\rangle.
\]

\paragraph{Connection to Frishman \& Ronceray (2020)}
It is worth explicitly noting the connection to the related work by Frishman \& Ronceray \cite{frishmanLearningForceFields2020}. The estimator they developed for learning force fields aligns with a specific configuration of our linear regression estimator \( \hat{\bm{\alpha}}_{\text{LR}} \) (\cref{eq:chap2a_alpha_linear_regression}). Specifically, their approach corresponds to setting the weighting matrix to the identity, \( \bm{A} = \bm{I} \). This choice simplifies the regression, effectively making it an unweighted ordinary least-squares fit between the empirical velocities \( \Delta \bm{x}_t / \Delta t \) and the basis functions, minimizing \( \langle \lVert \Delta \bm{x}_t / \Delta t - \sum_i \alpha_i \bm{b}_i \rVert^2 \rangle \). Furthermore, their formulation typically employs scalar basis functions \( b_k(\bm{x}) \) applied component-wise to the drift vector (i.e., modeling each \( f_\beta(\bm{x}) = \sum_k \alpha_{\beta k} b_k(\bm{x}) \)), differing from the direct vector basis function formulation \( \bm{f}(\bm{x}) = \sum_i \alpha_i \bm{b}_i(\bm{x}) \) used here. The framework presented in this chapter thus offers greater generality through the arbitrary positive-definite weighting matrix \( \bm{A} \) (allowing for connections to MLE via \( \bm{A}=\bm{D} \)) and the direct use of vector basis functions.

\section{A Practical Approximate Maximum Likelihood (AML) Drift Estimator}
\label{sec:approx_mle_drift}

Building upon the insights from maximum likelihood and linear regression, we now introduce a computationally efficient method to estimate the drift parameters. Recall that the exact MLE suffers from coupled equations for \( \hat{\bm{\alpha}}_{\text{MLE}} \) and \( \hat{\bm{D}}_{\text{MLE}} \) (\cref{subsec:mle_maximization}), while the optimal linear regression requires knowledge of \( \bm{D} \). We can circumvent these issues by using the simple, drift-independent diffusion estimator \( \bar{\bm{D}} \) from \cref{eq:chap2a_approximation_D} as a reasonable proxy for the true diffusion matrix within the drift estimation step.

This leads to a practical two-step approach:
\begin{enumerate}
    \item Estimate the diffusion matrix using the simple estimator: \( \bar{\bm{D}} = \frac{1}{2\Delta t}\left\langle \Delta \bm{x}_t\,\Delta \bm{x}_t^\top \right\rangle \).
    \item Substitute this fixed \( \bar{\bm{D}} \) into the formal MLE equation for the drift parameters (\cref{eq:chap2a_alpha_likelihood}).
\end{enumerate}
This procedure yields the \textbf{approximate maximum likelihood drift estimator} \( \hat{\bm{\alpha}} \):
\begin{equation}
    \hat{\alpha}_i = \sum_{j=1}^{n_{\mathcal{B}}} \left(G^{-1}\right)_{ij} \left\langle \bm{b}_j^\top \left(4\bar{\bm{D}}\right)^{-1} \frac{\Delta \bm{x}_t}{\Delta t} \right\rangle,
    \label{eq:chap_2b_alpha_approx_mle}
\end{equation}
where the Gram matrix \( G \) is now computed using the estimated \( \bar{\bm{D}} \):
\[
G_{ij} = \left\langle \bm{b}_i^\top \left(4\bar{\bm{D}}\right)^{-1} \bm{b}_j \right\rangle.
\]

This estimator \( \hat{\bm{\alpha}} \) directly corresponds to the drift estimator derived via linear regression (\cref{eq:chap2a_alpha_linear_regression}) if we set the weighting matrix \( \bm{A} = \bar{\bm{D}} \). Conceptually, \( \hat{\bm{\alpha}} \) maximizes an approximate log-likelihood where \( \bar{\bm{D}} \) is treated as a fixed, known parameter:
\begin{equation}
    \ell_{\text{approx}}(\bm{X}_\tau \mid \bm{\alpha}) = -\tau \left\langle \left(\frac{\Delta \bm{x}_t}{\Delta t} - \sum_{i=1}^{n_{\mathcal{B}}} \alpha_i \bm{b}_i(\bm{x}_t)\right)^\top \left(4\bar{\bm{D}}\right)^{-1} \left(\frac{\Delta \bm{x}_t}{\Delta t} - \sum_{k=1}^{n_{\mathcal{B}}} \alpha_k \bm{b}_k(\bm{x}_t)\right) \right\rangle - \ln Z_{\bar{\bm{D}}}.
\end{equation}
This approach decouples the estimation problem, providing a computationally tractable estimator (\( \bar{\bm{D}} \) is calculated first, then \( \bm{G} \) and the required averages, followed by a single matrix inversion for \( \hat{\bm{\alpha}} \)) that retains a close connection to the true MLE structure. This efficiency becomes particularly valuable when performing model selection involving multiple candidate drift models, as discussed in \cref{part:Model_selection}.

\subsection{Bias Analysis of the Approximate Estimator}

We now analyze the bias of our approximate estimator \( \hat{\bm{\alpha}} \) defined in \cref{eq:chap_2b_alpha_approx_mle}. We aim to show that it is asymptotically unbiased as the observation time \( \tau \to \infty \).

To compute the bias, we substitute the discretized Langevin dynamics (assuming the true drift is \( \bm{f}(\bm{x}) = \sum_k \alpha_k^* \bm{b}_k(\bm{x}) \)):
\[
\frac{\Delta \bm{x}_t}{\Delta t} = \sum_k \alpha_k^* \bm{b}_k(\bm{x}_t) + \frac{\sqrt{2\bm{D}}\,\Delta \bm{W}_t}{\Delta t} + \mathcal{O}(\Delta t)
\]
into \cref{eq:chap_2b_alpha_approx_mle} (where \( \bm{D} \) is the true diffusion matrix). We get:
\begin{align}
    \hat{\alpha}_i &= \sum_{j=1}^{n_{\mathcal{B}}} \left(G^{-1}\right)_{ij} \left\langle \bm{b}_j^\top \left(4\bar{\bm{D}}\right)^{-1} \left( \sum_k \alpha_k^* \bm{b}_k + \frac{\sqrt{2\bm{D}}\,\Delta \bm{W}_t}{\Delta t} \right) \right\rangle + \mathcal{O}(\Delta t) \nonumber \\
    &= \sum_{j,k=1}^{n_{\mathcal{B}}} \left(G^{-1}\right)_{ij} \alpha_k^* \underbrace{\left\langle \bm{b}_j^\top \left(4\bar{\bm{D}}\right)^{-1} \bm{b}_k \right\rangle}_{G_{kj}} + \underbrace{\sum_{j=1}^{n_{\mathcal{B}}} \left(G^{-1}\right)_{ij} \left\langle \bm{b}_j^\top \left(4\bar{\bm{D}}\right)^{-1} \frac{\sqrt{2\bm{D}}\,\Delta \bm{W}_t}{\Delta t} \right\rangle}_{Z_i} + \mathcal{O}(\Delta t) \nonumber\\[1mm]
    \hat{\alpha}_i &= \alpha_i^* + Z_i + \mathcal{O}(\Delta t),
    \label{eq:chap_2b_Z_I_alpha}
\end{align}
where the error term is
\begin{equation}
    Z_i = \sum_{j=1}^{n_{\mathcal{B}}} \left(G^{-1}\right)_{ij} \left\langle \bm{b}_j^\top \left(4\bar{\bm{D}}\right)^{-1} \frac{\sqrt{2\bm{D}}\,\Delta \bm{W}_t}{\Delta t} \right\rangle.
\end{equation}
The bias of the estimator is
\begin{equation}
    \mathbb{E}[\hat{\alpha}_i - \alpha_i^*] = \mathbb{E}[Z_i] + \mathcal{O}(\Delta t).
\end{equation}

If \( \bm{G} \) and \( \bar{\bm{D}} \) were fixed quantities independent of the specific noise realization \( \{ \Delta \bm{W}_t \} \), then \( \mathbb{E}[Z_i] \) would be zero because \( \mathbb{E}[\Delta \bm{W}_t] = 0 \). However, both \( \bm{G} \) and \( \bar{\bm{D}} \) are computed from the same trajectory \( \bm{X}_\tau \), which itself depends on \( \{ \Delta \bm{W}_t \} \), inducing correlations.

To assess the bias, we assume the process has sufficient mixing properties such that trajectory averages converge to expectations for large \( \tau \). The central limit theorem suggests that quantities computed as trajectory averages fluctuate around their means. We can decompose:
\begin{align}
    \bm{G} &= \mathbb{E}[\bm{G}] + \bm{R}_{G}, \\
    \bar{\bm{D}} &= \mathbb{E}[\bar{\bm{D}}] + \bm{R}_{\bar{D}},
\end{align}
where \( \mathbb{E}[\bm{G}] \) and \( \mathbb{E}[\bar{\bm{D}}] \) are the expectations (long-time averages), and \( \bm{R}_{G}, \bm{R}_{\bar{D}} \) are random fluctuation terms. For large \( \tau = N \Delta t \), we expect \( \mathbb{E}[\bar{\bm{D}}] \approx \bm{D} \). The fluctuations typically scale as:
\begin{align}
    (\bm{R}_{G})_{ij} &\sim \mathcal{O}(1/\sqrt{\tau}), \\
    (\bm{R}_{\bar{D}})_{\beta\gamma} &\sim \mathcal{O}(1/\sqrt{N}) = \mathcal{O}(\sqrt{\Delta t / \tau}).
\end{align}
This implies that
\begin{align}
\bm{G}^{-1} &= (\mathbb{E}[\bm{G}])^{-1} + \bm{R}_{G^{-1}}\\
(4\bar{\bm{D}})^{-1} &= (4\mathbb{E}[\bar{\bm{D}}])^{-1} + \bm{R}_{D^{-1}}
\end{align}
 with \( \bm{R}_{G^{-1}} \sim \mathcal{O}(1/\sqrt{\tau}) \), and \( \bm{R}_{D^{-1}} \sim \mathcal{O}(\sqrt{\Delta t / \tau}) \).

We then decompose the fluctuation term \(Z_i\) into four parts:
\begin{multline}
    Z_i = \sum_{j=1}^{n_{\mathcal{B}}} \underbrace{ (\mathbb{E}[G]^{-1})_{ij} \left\langle \bm{b}_j^\top (4\mathbb{E}[\bar{D}])^{-1} \frac{\sqrt{2\bm{D}}\,\Delta \bm{W}_t}{\Delta t} \right\rangle }_{Z_{ij,1}} \\
    + \sum_{j=1}^{n_{\mathcal{B}}} \underbrace{ (R_{G^{-1}})_{ij} \left\langle \bm{b}_j^\top (4\mathbb{E}[\bar{D}])^{-1} \frac{\sqrt{2\bm{D}}\,\Delta \bm{W}_t}{\Delta t} \right\rangle }_{Z_{ij,2}} \\
    + \sum_{j=1}^{n_{\mathcal{B}}} \underbrace{ (\mathbb{E}[G]^{-1})_{ij} \left\langle \bm{b}_j^\top R_{D^{-1}} \frac{\sqrt{2\bm{D}}\,\Delta \bm{W}_t}{\Delta t} \right\rangle }_{Z_{ij,3}} \\
    + \sum_{j=1}^{n_{\mathcal{B}}} \underbrace{ (R_{G^{-1}})_{ij} \left\langle \bm{b}_j^\top R_{D^{-1}} \frac{\sqrt{2\bm{D}}\,\Delta \bm{W}_t}{\Delta t} \right\rangle }_{Z_{ij,4}}.
    \label{eq:chap2b_decomposition_Z_revised}
\end{multline}
We estimate the expectation of each term for large \( \tau \):

\begin{itemize}
    \item For \(Z_{ij,1}\): Since \( \mathbb{E}[\Delta \bm{W}_t] = 0 \) and the prefactors are (asymptotically) constants, we have \( \mathbb{E}[Z_{ij,1}] = 0 \).
    \item For \(Z_{ij,2}\): This involves the correlation between the fluctuation in \( \bm{G}^{-1} \) and the noise average term. Applying Cauchy–Schwarz and using Itô isometry arguments, one can argue that \( |\mathbb{E}[Z_{ij,2}]| \lesssim \mathcal{O}(1/\sqrt{\tau}) \times \mathcal{O}(1/\sqrt{\tau}) = \mathcal{O}(1/\tau) \).
    \item For \(Z_{ij,3}\): This involves the correlation between the fluctuation in \( \bar{\bm{D}}^{-1} \) and the noise average. Careful analysis is needed due to \( \bar{\bm{D}} \) depending on \( \Delta \bm{x}_t \), which contains \( \Delta \bm{W}_t \). Under reasonable assumptions about correlations decaying, it can be argued that \( |\mathbb{E}[Z_{ij,3}]| \lesssim \mathcal{O}(\sqrt{\Delta t / \tau}) \times \mathcal{O}(1/\sqrt{\tau}) = \mathcal{O}(\sqrt{\Delta t}/\tau) \).
    \item For \(Z_{ij,4}\): This involves correlations between fluctuations of both \( \bm{G}^{-1} \) and \( \bar{\bm{D}}^{-1} \) with the noise term. It is expected to be of higher order, potentially \( |\mathbb{E}[Z_{ij,4}]| \lesssim \mathcal{O}(1/\sqrt{\tau}) \times \mathcal{O}(\sqrt{\Delta t / \tau}) \times \mathcal{O}(1/\sqrt{\tau}) = \mathcal{O}(\sqrt{\Delta t} / \tau^{3/2}) \).
\end{itemize}
A detailed analysis of these expectations requires careful treatment of correlations which is very difficult to achieve for the general case. We provide this analysis for the Ornstein-Uhlenbeck case in \apdx~\ref{appendix:OU_neglecting_correlation}. 
Nevertheless, the scaling analysis suggests that the dominant bias terms vanish at least as \( \mathcal{O}(1/\tau) \) or \( \mathcal{O}(\sqrt{\Delta t}/\tau) \).

Thus, we conclude that the bias vanishes asymptotically:
\begin{equation}
    \mathbb{E}[\hat{\alpha}_i] = \alpha_i^* + O\left(\frac{1}{\tau}\right) + \mathcal{O}(\Delta t),
\end{equation}
so that the estimator \( \hat{\bm{\alpha}} \) is asymptotically unbiased as \( \tau \to \infty \) and $\Delta t \to 0$.

\subsection{Mean Square Error of the Drift Estimator}
\label{sec:MSE_drift_estimator}

In addition to the bias, another key quantity for assessing the quality of an estimator is its variance. The mean square error (MSE) combines both bias and variance, and serves as a standard measure of estimator accuracy (\cref{subsec:mse}). In the case of our approximate maximum likelihood estimator $\hat{\alpha}_i$, we now derive an explicit expression for its MSE in the large-$\tau$ limit. The mean square error (MSE) is defined by
\begin{equation}
    \text{MSE}(\hat{\alpha}_i) = \mathbb{E}\!\left[\left(\hat{\alpha}_i - \alpha_i^*\right)^2\right].
\end{equation}
To derive the MSE, we use that
\[
\hat{\alpha}_i = \alpha_i^* + Z_i + \mathcal{O}(\Delta t) \quad \Longrightarrow \quad \text{MSE}(\hat{\alpha}_i) = \mathbb{E}[Z_i^2] +  O\left(\frac{1}{\tau^2}\right) + \mathcal{O}(\Delta t^2).
\]
Using the decomposition of \(Z_i\) from \cref{eq:chap2b_decomposition_Z_revised}, we write
\begin{equation}
    \mathbb{E}[Z_i^2] = \sum_{j,k=1}^{n_{\mathcal{B}}} \mathbb{E}[Z_{ij,1} Z_{ik,1}] + 2\mathbb{E}[Z_{ij,1} Z_{ik,2}] + \cdots.
    \label{eq:chap2b_MSE_decomposition_Z}
\end{equation}
The leading term is given by
\begin{align}
    \sum_{j,k=1}^{n_{\mathcal{B}}} \mathbb{E}[Z_{ij,1} Z_{ik,1}] &=  \sum_{j,k=1}^{n_{\mathcal{B}}}  \left(\E{G}^{-1}\right)_{ij}\left(\E{G}^{-1}\right)_{ik}\E{\left\langle \frac{\sqrt{2\bm{D}}\,\Delta \bm{W}_t}{\Delta t} \cdot \left(4\bm{D}\right)^{-1} \bm{b}_j \right\rangle \left\langle \frac{\sqrt{2\bm{D}}\,\Delta \bm{W}_t}{\Delta t} \cdot \left(4\bm{D}\right)^{-1} \bm{b}_k \right\rangle} \nonumber\\
    &= \sum_{j,k=1}^{n_{\mathcal{B}}}\left(\E{G}^{-1}\right)_{ij}\left(\E{G}^{-1}\right)_{ik} \frac{\E{G}_{jk}}{2\tau},\quad \text{by Ito-Isometry} \nonumber\\[1mm]
    &= \frac{1}{2 \tau}\, \left(\E{G}^{-1}\right)_{ii}.
\end{align}
The remaining terms (e.g. \(\mathbb{E}[Z_{ij,1} Z_{ik,2}]\), \(\mathbb{E}[Z_{ij,1} Z_{ik,3}]\), etc.) can be shown, via successive applications of the Cauchy–Schwarz inequality and the Itô isometry, to be of lower order—typically \(O\left(\tau^{-3/2}\right)\). Hence, the leading order MSE is:
\begin{equation}
    \text{MSE}(\hat{\alpha}_i) = \frac{1}{2\tau}\, \left(\mathbb{E}[G]^{-1}\right)_{ii} + O\!\left(\frac{1}{\tau^{3/2}}\right) + \mathcal{O}(\Delta t^2).
\end{equation}
This crucial result shows that the variance of the estimator decreases inversely with the observation time \( \tau \). Furthermore, a similar calculation for the covariance gives:
\begin{equation}
    \mathbb{E}\!\left[(\hat{\alpha}_i - \alpha_i^*) (\hat{\alpha}_j - \alpha_j^*)\right] =  \frac{1}{2 \tau}\, \left(\mathbb{E}[G]^{-1}\right)_{ij} + O\!\left(\frac{1}{\tau^{3/2}}\right) + \mathcal{O}(\Delta t^2).
    \label{eq:chap2b_MSE_covariance}
\end{equation}

This asymptotic variance matches the Cramér-Rao lower bound (CRLB) for this estimation problem under the approximation that \( \bar{\bm{D}} \) is fixed\footnote{The CRLB states that for an unbiased estimator, $\text{Var}(\hat{\alpha}_i) \geq [\bm{I}^{-1}]_{ii}$ (or $\text{MSE}(\hat{\alpha}_i) \geq [\bm{I}^{-1}]_{ii}$), where $\bm{I}$ is the total Fisher Information matrix. For the model with fixed $\bar{\bm{D}}$, the Fisher Information \( \bm{I} \) scales with \( \tau \). Specifically, \( I_{ij} = -\mathbb{E}[\partial^2 \ell_{\text{approx}} / \partial \alpha_i \partial \alpha_j] = 2\tau \mathbb{E}[G_{ij}] \), where $\bm{G}$ is computed using $\bar{\bm{D}}$. The CRLB for the variance of \( \hat{\alpha}_i \) within this model is thus \( [\bm{I}^{-1}]_{ii} = [(2\tau \mathbb{E}[\bm{G}])^{-1}]_{ii} = \frac{1}{2\tau} [(\mathbb{E}[\bm{G}])^{-1}]_{ii} \), matching asymptotically our derived MSE for small $\Delta t$.}, suggesting the approximate estimator is asymptotically efficient under these conditions.

\subsection{Error in Reconstructed Drift}
While the parameter error \( \text{MSE}(\hat{\alpha}_i) \) is fundamental, a more intuitive measure of performance is the error in the reconstructed drift function itself. We quantify how well the inferred drift \( \hat{\bm{f}}(\bm{x}) = \sum_i \hat{\alpha}_i \bm{b}_i(\bm{x}) \) approximates the true drift \( \bm{f}(\bm{x}) = \sum_i \alpha^*_i \bm{b}_i(\bm{x}) \).

We define the mean squared drift error along the trajectory, weighted by the inverse diffusion matrix (which naturally appears in the likelihood), as:
\begin{equation}
    \mathcal{E}\left(\hat{\bm{f}}\right) = \left\langle \left(\hat{\bm{f}}(\bm{x}_t) - \bm{f}(\bm{x}_t)\right)^\top \left(4\bar{\bm{D}}\right)^{-1} \left(\hat{\bm{f}}(\bm{x}_t) - \bm{f}(\bm{x}_t)\right) \right\rangle.
\end{equation}
Expressing the drifts in terms of the basis functions and parameters:
\[
\hat{\bm{f}}(\bm{x}_t) - \bm{f}(\bm{x}_t) = \sum_{i=1}^{n_{\mathcal{B}}} (\hat{\alpha}_i - \alpha^*_i)\,\bm{b}_i(\bm{x}_t).
\]
Substituting this into the error definition yields:
\begin{align}
    \mathcal{E}\left(\hat{\bm{f}}\right) &= \left\langle \left( \sum_{i} (\hat{\alpha}_i - \alpha^*_i)\,\bm{b}_i \right)^\top \left(4\bar{\bm{D}}\right)^{-1} \left( \sum_{j} (\hat{\alpha}_j - \alpha^*_j)\,\bm{b}_j \right) \right\rangle \nonumber \\
    &= \sum_{i,j=1}^{n_{\mathcal{B}}} (\hat{\alpha}_i - \alpha^*_i) \left\langle \bm{b}_i^\top \left(4\bar{\bm{D}}\right)^{-1} \bm{b}_j \right\rangle (\hat{\alpha}_j - \alpha^*_j) \nonumber \\
    &= \sum_{i,j=1}^{n_{\mathcal{B}}} \left(\hat{\alpha}_i - \alpha^*_i\right) \,G_{ij} \,\left(\hat{\alpha}_j - \alpha^*_j\right).
\end{align}
Now, we take the expectation of this error over many realizations of the trajectory. Assuming \( \tau \) is large enough that fluctuations in \( G_{ij} \) around its mean \( \mathbb{E}[G_{ij}] \) are small and can be neglected in this calculation (i.e., \( G_{ij} \approx \mathbb{E}[G_{ij}] \)), we use the result for the parameter covariance matrix from \cref{eq:chap2b_MSE_covariance}:
\begin{align}
    \E{\mathcal{E}\left(\hat{\bm{f}}\right)} &= \sum_{i,j=1}^{n_{\mathcal{B}}} \E{(\hat{\alpha}_i - \alpha^*_i) (\hat{\alpha}_j - \alpha^*_j)}\left(\E{G_{ij}} +  O\left(\frac{1}{\tau}\right)\right)\nonumber \\
    &=\sum_{i,j=1}^{n_{\mathcal{B}}} \E{\frac{1}{2\tau}\, (\E{G}^{-1})_{ij}  + O\left(\frac{1}{\tau^{3/2}}\right)}\left(\E{G_{ij}} +  O\left(\frac{1}{\tau}\right)\right)\nonumber \\
    &= \frac{1}{2\tau} \sum_{i,j=1}^{n_{\mathcal{B}}} (\mathbb{E}[G]^{-1})_{ij} (\mathbb{E}[G])_{ji} +  O\left(\frac{1}{\tau^{3/2}}\right) \quad \text{(using symmetry } \mathbb{E}[G]_{ij} = \mathbb{E}[G]_{ji}) \nonumber \\
    &=  \frac{1}{2\tau} \mathrm{Tr}(\bm{I}_{n_{\mathcal{B}} \times n_{\mathcal{B}}}) + O\left(\frac{1}{\tau^{3/2}}\right) \nonumber \\
    &= \frac{n_{\mathcal{B}}}{2\tau} + O\left(\frac{1}{\tau^{3/2}}\right).
    \label{eq:chap2b_norm_mse_drift}
\end{align}
This important result indicates that the expected squared error in the reconstructed drift, weighted appropriately and averaged over the trajectory, scales inversely with the observation time \( \tau \) and linearly with the number of basis functions \( n_{\mathcal{B}} \). This result has been obtained previously by Frishman \& Ronceray in \cite{frishmanLearningForceFields2020}.

\subsection{Benchmark for additive noise}

We will confirm \cref{eq:chap2b_norm_mse_drift} with numerical simulations based on the model detailed below. 
They are simulated assuming isotropic additive noise following
\begin{equation}
    \dd{\bm{x}_t} = \bm{f}(\bm{x}_t) \dd{t} + \sqrt{2 D_0} \dd{\bm{W}_t},
    \label{eq:benchmark_additive_noise_iso}
\end{equation}
where $D_0$ is a scalar diffusion constant. For each system, we define both a minimal basis set $\mathcal{B}^*$, containing only the functions essential to represent the true drift, and a larger, over-complete basis set $\mathcal{B}_0$, typically constructed from standard polynomial terms, which includes $\mathcal{B}^*$.

\subsubsection{The Lorenz System}
\label{sec:lorenz_system_description}
The Lorenz system provides a classic example of deterministic chaos \cite{lorenzDeterministicNonperiodicFlow1963} widely used as a benchmark in the ODE case \cite{bruntonDiscoveringGoverningEquations2016,EnsembleSINDyRobustSparse}. 
We analyze its stochastic counterpart governed by the Langevin equation \eqref{eq:benchmark_additive_noise_iso}  in $d=3$ dimensions, with the state vector $\bm{x} = (x, y, z)^\top$. The drift term $\bm{f}(\bm{x})$ retains the form of the Lorenz equations:
\begin{equation}
    \bm{f}(\bm{x}) =
    \begin{pmatrix}
        \sigma (z - x) \\
        x z - \beta y\\
        x (\rho - y ) - z
    \end{pmatrix},
    \label{eq:lorenz_drift}
\end{equation}
where $\sigma$, $\rho$, and $\beta$ are the parameters controlling the dynamics. We use the standard parameter values $\sigma = 10$, $\rho = 28$, and $\beta = 8/3$, known to produce a certain butterfly-like shape. 

To express this drift precisely within our framework, $\bm{f}(\bm{x}) = \sum_{i=1}^{n_{\mathcal{B}}} \alpha_i^* \bm{b}_i(\bm{x})$, we identify the necessary polynomial terms. The minimal set of basis functions required to exactly span this drift, denoted $\mathcal{B}^*_{\text{Lorenz}}$, consists of $n_{\mathcal{B}^*} = 7$ specific vector basis functions:
\begin{align}
    \mathcal{B}^*_{\text{Lorenz}} = \{
    &(x, 0, 0)^\top, (z, 0, 0)^\top,  (0, xz, 0)^\top, (0, y, 0)^\top,\nonumber \\
    &(0, 0, x)^\top, (0, 0, xy)^\top, (0, 0, z)^\top
    \}. \label{eq:lorenz_basis_minimal}
\end{align}
The corresponding true coefficients $\alpha_i^*$ for these basis functions (in the order listed) are $(-\sigma, \sigma,1 ,-\beta, \rho, -1, -1)$.

For benchmarking purposes, particularly when testing model selection methods, we also define an over-complete basis set, $\mathcal{B}_{0, \text{Lorenz}}$. This set includes all vector basis functions formed from scalar polynomial monomials in $x, y, z$ up to degree 2. Specifically, it contains functions of the form $p(\bm{x})\bm{e}_\beta$, where $\beta \in \{1, 2, 3\}$, $\bm{e}_\beta$ is the standard basis vector, and $p(\bm{x})$ is any monomial from the set $\{1, x, y, z, x^2, y^2, z^2, xy, xz, yz\}$ (10 monomials). This results in a total of $n_{\mathcal{B}_0} = 10 \times 3 = 30$ basis functions in $\mathcal{B}_{0, \text{Lorenz}}$. This basis naturally includes the minimal basis $\mathcal{B}^*_{\text{Lorenz}}$.

Trajectories for analysis were generated using the Euler-Maruyama numerical integration scheme with a fine integration time step $\dd{t}_{\text{sim}} = 0.001$. To simulate typical experimental observation conditions, the resulting high-resolution trajectory was then subsampled every 10 steps, yielding the final trajectory data with an effective time step of $\Delta t = 10 \times dt_{\text{sim}} = 0.01$. Furthermore, the diffusion scalar is $D_0=100$.

\subsubsection{The Ornstein-Uhlenbeck Process}
\label{sec:OU_process_description}
The Ornstein-Uhlenbeck (OU) process serves as a fundamental model for stochastic dynamics \cite{uhlenbeckTheoryBrownianMotion1930}. We consider a $d=10$ dimensional OU process governed by \eqref{eq:benchmark_additive_noise_iso} with state vector $\bm{x} = (x_1, \dots, x_{10})^\top$. We define the drift term as:
\begin{equation}
    \bm{f}(\bm{x}) = -\bm{A} \bm{x},
\end{equation}
where $\bm{A}$ is the $10 \times 10$ damping matrix. For our benchmark simulations (e.g., Figure~\ref{fig:drift_error_vs_tau}(b)), we use the specific sparse matrix:
\begin{equation}
\bm{A} =
\begin{bmatrix}
 1. &  0. &  0. &  0. &  0. &  0. &  0. &  0. &  0. &  0. \\
 0. &  1. &  0. &  0. &  0. &  0. &  0. &  0. &  1. &  0. \\
 0. &  0. &  1. & -1. &  0. &  0. &  0. &  0. &  0. &  0. \\
 0. &  0. &  0. &  1. &  0. &  0. &  0. &  0. &  0. &  0. \\
 0. &  0. &  0. &  0. &  1. &  0. &  0. &  0. &  0. &  0. \\
 0. &  0. &  0. &  0. &  0. &  1. &  0. &  0. &  0. &  0. \\
-1. &  0. &  0. &  0. &  0. &  0. &  1. &  0. &  0. &  0. \\
 0. &  0. &  0. &  0. &  0. &  0. &  0. &  1. &  0. &  0. \\
 0. & -1. &  0. &  0. &  0. & -1. &  1. &  0. &  1. &  0. \\
 0. &  0. &  1. &  0. &  0. &  1. &  0. &  1. &  0. &  1.
\end{bmatrix}.
\label{eq:OU_A_matrix}
\end{equation}
The drift components are $f_\beta(\bm{x}) = -\sum_{\delta=1}^{10} A_{\beta\delta} x_\delta$. An exact representation $\bm{f}(\bm{x}) = \sum_i \alpha_i^* \bm{b}_i(\bm{x})$ only requires basis functions corresponding to the non-zero entries of $\bm{A}$. The minimal exact basis, $\mathcal{B}^*_{\text{OU}}$, thus consists of vector basis functions $\bm{b}_{\beta\delta}(\bm{x}) = x_\delta \bm{e}_\beta$ only for those pairs $(\beta, \delta)$ where $A_{\beta\delta} \neq 0$. By counting the non-zero elements in the matrix $\bm{A}$ \eqref{eq:OU_A_matrix}, we find that this minimal basis contains $n_{\mathcal{B}^*} = 19$ functions. The true coefficient associated with $\bm{b}_{\beta\delta}(\bm{x})$ in this basis is $\alpha_{\beta\delta}^* = -A_{\beta\delta}$.

For comparison and testing, we also define an over-complete basis set $\mathcal{B}_{0, \text{OU}}$, which includes all vector basis functions constructed from scalar polynomial monomials up to degree 1. This comprises constant terms and linear terms. Specifically, it contains the $d=10$ constant basis functions $\{\bm{e}_\beta \mid \beta=1,\dots,10\}$ and the $d^2=100$ linear basis functions $\{x_\delta \bm{e}_\beta \mid \beta, \delta=1,\dots,10\}$. The total number of functions in this standard affine basis is $n_{\mathcal{B}_0} = d + d^2 = 10 + 100 = 110$.

Trajectories for analysis were generated using the Euler-Maruyama numerical integration scheme with a fine integration time step $\dd{t}_{\text{sim}} = 0.001$. To simulate typical experimental observation conditions, the resulting high-resolution trajectory was then subsampled every 10 steps, yielding the final trajectory data with an effective time step of $\Delta t = 10 \times dt_{\text{sim}} = 0.01$. Furthermore, the diffusion scalar is $D_0=100$.

\subsubsection{Numerical Validation: Additive Noise Benchmarks}
\begin{figure}[htbp]
    \centering
    \includegraphics[width=\linewidth]{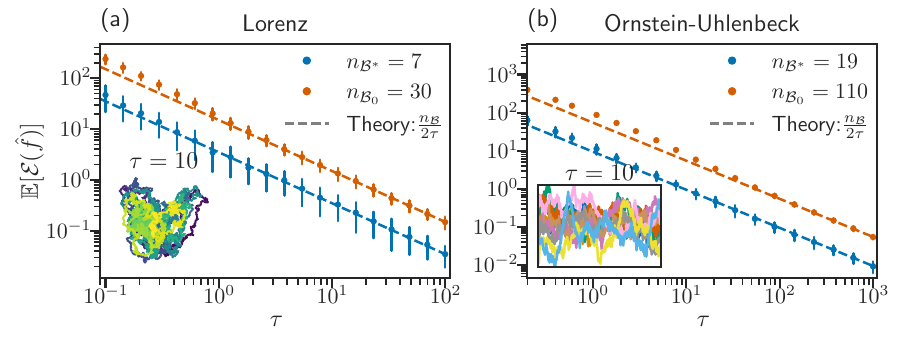} 
   \caption{Log--log plot illustrating the decay of the expected normalized mean square drift error, $\mathbb{E}[\mathcal{E}(\hat{\bm{f}})]$, as a function of the trajectory length $\tau$. Points with error bars represent the mean and standard deviation obtained from multiple simulation runs, while dashed lines show the theoretical prediction $n_{\mathcal{B}}/(2\tau)$ derived in \cref{eq:chap2b_norm_mse_drift}. (a) Results for the $d=3$ Lorenz system (\cref{eq:lorenz_drift}), comparing the minimal exact basis ($n_{\mathcal{B}^*}=7$) with the over-complete polynomial basis up to degree 2 ($n_{\mathcal{B}_0}=30$). Inset shows a projection of an example trajectory segment of length $\tau=10$. (b) Results for the $d=10$ Ornstein-Uhlenbeck process (\cref{eq:OU_A_matrix}), comparing its minimal exact basis ($n_{\mathcal{B}^*}=19$) with the over-complete affine basis ($n_{\mathcal{B}_0}=110$).}
    \label{fig:drift_error_vs_tau}
\end{figure}

As seen in \cref{fig:drift_error_vs_tau}, the theoretical prediction for the error, $\mathbb{E}[\mathcal{E}(\hat{\bm{f}})] \approx \frac{n_{\mathcal{B}}}{2\tau}$, holds well for sufficiently long trajectories, validating our theoretical analysis derived in \cref{eq:chap2b_norm_mse_drift}. 
The error also clearly increases with the number of basis functions \( n_{\mathcal{B}} \) used in the fit (comparing results for $\mathcal{B}^*$ vs $\mathcal{B}_0$ for each system), quantifying the statistical 'cost' of adding each basis function as \( 1/(2\tau) \). This provides a key insight for model selection, which aims to balance model complexity ($n_{\mathcal{B}}$) against goodness-of-fit. It is crucial to remember that this result relies on the assumption that the true drift \( \bm{f}(\bm{x}) \) lies within the span of the chosen basis functions \( \{\bm{b}_i\} \). If the basis is inadequate (model misspecification), a systematic bias error will dominate the statistical error for large $\tau$.

\section{Inference under Multiplicative Noise}
\label{sec:multiplicative_noise_impact_analysis}

The preceding sections largely focused on inference assuming either explicitly that the diffusion matrix \( \bm{D} \) was constant (additive noise). However, in many physical, biological, or financial systems, the noise intensity itself depends on the system's state \( \bm{x}_t \). This leads to the more general Langevin equation with multiplicative noise:
\begin{equation}
 \dd{\bm{x}_t} = \bm{f}(\bm{x}_t) \dd{t} + \sqrt{2\, \bm{D}(\bm{x}_t)}\, \dd{\bm{W}_t},
 \label{eq:chap_2c_langevin_multiplicative}
\end{equation}
where \( \bm{D}(\bm{x}_t) \) is now a state-dependent, positive-definite symmetric matrix function.

The presence of state-dependent diffusion introduces significant challenges. In this section, rather than developing new estimators specifically for multiplicative noise, we analyze the consequences of applying the approximate estimator derived for additive noise (\cref{eq:chap_2b_alpha_approx_mle}) directly to data generated by \cref{eq:chap_2c_langevin_multiplicative}. This corresponds to a scenario of model misspecification, where the inference procedure assumes an incorrect noise structure.

\subsection{Analysis of the Estimator under Misspecified Noise}
\label{subsec:multiplicative_impact_on_additive_estimator}

We now investigate the performance of the specific Approximate Maximum Likelihood (AML) estimator derived for additive noise when the data \( \bm{X}_\tau \) is generated by a process with true multiplicative noise \( \bm{D}(\bm{x}_t) \) and true drift \( \bm{f}(\bm{x}) = \sum_k \alpha_k^* \bm{b}_k(\bm{x}) \). Recall the estimator structure:
\begin{enumerate}
    \item Compute an estimate of a constant diffusion matrix:
    \[ \bar{\bm{D}} = \frac{1}{2\Delta t}\left\langle \Delta \bm{x}_t\,\Delta \bm{x}_t^\top \right\rangle. \]
    \item Use this \( \bar{\bm{D}} \) to compute the drift parameters:
    \[ \hat{\alpha}_i = \sum_{j=1}^{n_{\mathcal{B}}} \left(G^{-1}\right)_{ij} \left\langle \bm{b}_j^\top \left(4\bar{\bm{D}}\right)^{-1} \frac{\Delta \bm{x}_t}{\Delta t} \right\rangle, \quad \text{with } G_{ij} = \left\langle \bm{b}_i^\top \left(4\bar{\bm{D}}\right)^{-1} \bm{b}_j \right\rangle. \]
\end{enumerate}

\subsubsection{Effect on the Estimated Diffusion \( \bar{\bm{D}} \)}
First, what does \( \bar{\bm{D}} \) estimate in this case? Using \( \mathbb{E}[\Delta \bm{x}_t\,\Delta \bm{x}_t^\top \mid \bm{x}_t] \approx 2\bm{D}(\bm{x}_t)\,\Delta t + \mathcal{O}(\Delta t^2) \), the trajectory average computed by the estimator \( \bar{\bm{D}} \) will converge to:
\begin{equation}
 \bar{\bm{D}} = \frac{1}{2\Delta t}\left\langle \Delta \bm{x}_t\,\Delta \bm{x}_t^\top \right\rangle \xrightarrow{\tau \to \infty} \frac{1}{2\Delta t} \left\langle \mathbb{E}[\Delta \bm{x}_t\,\Delta \bm{x}_t^\top \mid \bm{x}_t] \right\rangle \approx \left\langle \bm{D}(\bm{x}_t) \right\rangle.
    \label{eq:chap2d_Dbar_limit}
\end{equation}
Thus, when applied to multiplicative noise data, the simple estimator \( \bar{\bm{D}} \) approximates the \emph{trajectory average} of the true state-dependent diffusion matrix. Obviously, this estimate will fail to capture the state-specific variations of \( \bm{D}(\bm{x}_t) \).

\subsubsection{Effect on Drift Parameter for Bias and MSE}
The drift estimator \( \hat{\alpha}_i \) uses the incorrect, constant weighting \( \bar{\bm{D}} \approx \langle \bm{D} \rangle \). Let's analyze the impact on the MSE. Following the same steps as before (\cref{sec:MSE_drift_estimator}), we write: 
\[
\hat{\alpha}_i = \alpha_i^* + Z_i' + \mathcal{O}(\Delta t) \quad \Longrightarrow \quad \text{MSE}(\hat{\alpha}_i) = \mathbb{E}[(Z_i')^2] +  O\left(\frac{1}{\tau^2}\right) + \mathcal{O}(\Delta t^2).
\]
with \( G \) being the Gram matrix using the average diffusion: \( (G)_{ij} = \bm{b}_i^\top (4 \bar{\bm{D}})^{-1} \bm{b}_j\) and
\[
 Z'_i = \sum_{j=1}^{n_{\mathcal{B}}} \left(G^{-1}\right)_{ij} \left\langle \bm{b}_j^\top \left(4\bar{\bm{D}}\right)^{-1} \frac{\sqrt{2\bm{D}(\bm{x}_t)}}{\Delta t} \Delta \bm{W}_t \right\rangle.
\]
Thus, we still have that the bias vanishes asymptotically:
\begin{equation}
    \mathbb{E}[\hat{\alpha}_i] = \alpha_i^* + O\left(\frac{1}{\tau}\right) + \mathcal{O}(\Delta t),
\end{equation}

Let \( G^*_{\text{multi}} \) be the matrix arising from the noise term calculation, involving the true diffusion matrix:
\begin{equation}
 (G^*_{\text{multi}})_{jk} =  \avg{\bm{b}_j^\top (4 \bar{\bm{D}})^{-1} (4\bm{D}(\bm{x}_t)) (4 \bar{\bm{D}})^{-1} \bm{b}_k}.
 \label{eq:chap2d_G_star_multi}
\end{equation}
The MSE for the $i$-th parameter then becomes:
\begin{equation}
 \text{MSE}(\hat{\alpha}_i) \approx  \frac{1}{2 \tau} \left[ (\mathbb{E}[\bm{G}])^{-1} (\mathbb{E}[\bm{G}^*_{\text{multi}}]) (\mathbb{E}[\bm{G}])^{-1} \right]_{ii}.
 \label{eq:chap2d_MSE_multiplicative}
\end{equation}
This differs from the additive noise result (\( \text{MSE}(\hat{\alpha}_i) \approx \frac{1}{2\tau} (\mathbb{E}[\bm{G}]^{-1})_{ii} \)). The presence of \( \bm{D}(\bm{x}_t) \) inside the expectation for \( \mathbb{E}[G^*]_{\text{multi}} \) prevents the earlier simplification.

\subsubsection{Effect on Reconstructed Drift Error and Its Estimation}
We now examine the expected normalized mean squared drift error, \( \mathbb{E}[\mathcal{E}(\hat{\bm{f}})] = \mathbb{E}[\langle (\hat{\bm{f}} - \bm{f})^\top (4\bar{\bm{D}})^{-1} (\hat{\bm{f}} - \bm{f}) \rangle] \). Using the parameter covariance derived from \cref{eq:chap2d_MSE_multiplicative}:
\begin{align}
 \mathbb{E}\!\left[\mathcal{E}\left(\hat{\bm{f}}\right)\right] &\approx \sum_{i,j=1}^{n_{\mathcal{B}}} \mathbb{E}\!\left[(\hat{\alpha}_i - \alpha^*_i) (\hat{\alpha}_j - \alpha^*_j)\right] \mathbb{E}[\bm{G}]_{ij}\nonumber \\
 &\approx \sum_{i,j=1}^{n_{\mathcal{B}}} \left( \frac{1}{2 \tau} \left[ \mathbb{E}[\bm{G}]^{-1} \mathbb{E}[\bm{G}^*_{\text{multi}}] \mathbb{E}[\bm{G}]^{-1} \right]_{ij} \right) \mathbb{E}[\bm{G}_{ij}] \nonumber \\
 &= \frac{1}{2 \tau} \mathrm{Tr} \left( (\mathbb{E}[\bm{G}])^{-1} (\mathbb{E}[\bm{G}^*_{\text{multi}}]) \right).
 \label{eq:chap2d_drift_error_multiplicative_revised}
\end{align}
The result \( \frac{1}{\tau} \mathrm{Tr}(...) \) no longer simplifies to \( n_{\mathcal{B}}/(2\tau) \). The simple scaling law breaks down due to the mismatch between the assumed constant noise \( \bar{\bm{D}} \) and the true state-dependent noise \( \bm{D}(\bm{x}_t) \).

While we cannot compute the true error \( \mathcal{E}(\hat{\bm{f}}) \) as it depends on \( \bm{f} \), we can attempt to \textbf{estimate} the expected error \( \mathbb{E}[\mathcal{E}(\hat{\bm{f}})] \) given by \cref{eq:chap2d_drift_error_multiplicative_revised}. This requires estimating the matrix \( \bm{G}^*_{\text{multi}} \). A possible approach involves using an instantaneous estimate of the local diffusion:
\begin{equation}
 \hat{\bm{D}}_{\text{inst.}}(\bm{x}_t) = \frac{1}{2\Delta t}\Delta \bm{x}_t\,\Delta \bm{x}_t^\top.
\end{equation}
We know that \( \mathbb{E}[\hat{\bm{D}}_{\text{inst.}}(\bm{x}_t) | \bm{x}_t] \approx \bm{D}(\bm{x}_t) \) (neglecting drift terms of \(\mathcal{O}(\Delta t)\)). Therefore, we can construct an estimator \( \bm{G}_{\text{multi}} \) for \( \bm{G}^*_{\text{multi}} \) by replacing \( \bm{D}(\bm{x}_t) \) with \( \hat{\bm{D}}_{\text{inst.}}(\bm{x}_t) \). We define:
\begin{equation}
 (\bm{G}_{\text{multi}})_{jk} = \left\langle \bm{b}_j^\top (4 \bar{\bm{D}})^{-1} (4\hat{\bm{D}}_{\text{inst.}}(\bm{x}_t)) (4 \bar{\bm{D}})^{-1} \bm{b}_k \right\rangle.
 \label{eq:G_multi_estimator}
\end{equation}

Using the trajectory-computed Gram matrix \( \bm{G} \) (based on \( \bar{\bm{D}} \)) and this estimate \( \bm{G}_{\text{multi}} \), we can formulate an estimator \( \hat{\mathcal{E}}_{\text{est}} \) for the expected normalized mean squared drift error:
\begin{equation}
 \hat{\mathcal{E}}_{\text{est}} = \frac{1}{2 \tau} \mathrm{Tr} \left( \bm{G}^{-1} \bm{G}_{\text{multi}} \right).
    \label{eq:drift_error_estimator}
\end{equation}
This provides a computable quantity from a single trajectory that aims to approximate the expected error \( \mathbb{E}[\mathcal{E}(\hat{\bm{f}})] \) (\cref{eq:chap2d_drift_error_multiplicative_revised}) incurred when applying the additive-noise estimator to multiplicative-noise data.

While the estimator $\hat{\mathcal{E}}_{\text{est}}$ (\cref{eq:drift_error_estimator}) allows us to approximate the increased error resulting from applying an additive-noise framework to multiplicative noise, it does not resolve the underlying model misspecification. Addressing this directly requires methods capable of accurately estimating the state-dependent diffusion matrix $\bm{D}(\bm{x})$ itself. Several approaches exist in the literature that tackle this challenge, often drawing from the framework of Approximate Maximum Likelihood (AML) coefficient estimation. 
These methods typically involve estimating both drift and diffusion terms from trajectory data, potentially using basis function expansions or non-parametric methods for $\bm{D}(\bm{x})$ \cite{frishmanLearningForceFields2020,siegertAnalysisDataSets1998, friedrichHowQuantifyDeterministic2000, suraNoteEstimatingDrift2002}.

\subsection{Benchmark for Multiplicative Noise}
\subsubsection{Benchmark System: Lotka-Volterra}
\label{sec:lotka_volterra_benchmark}

As a benchmark system exhibiting non-linear dynamics and multiplicative noise, we employ a generalized Lotka-Volterra model for competing species. The system is defined in $d=7$ dimensions with the state vector $\bm{x} = (x_1, \dots, x_7)^\top$, representing the populations of seven species. We consider the following dynamical equation:
\begin{equation}
 \dd{x_i}(t) = x_i(t) \left( r_i - \sum_{j=1}^{d} A_{ij} x_j(t) \right) \dd{t} + \sqrt{2\, D_0 x_i^2}\, \dd{{W}_i(t)},
    \label{eq:lv_drift}
\end{equation}
Here, $r_i$ represents the intrinsic growth rate of species $i$, and $A_{ij}$ represents the interaction coefficient between species $i$ and species $j$. For our simulations, we set the growth rates to unity, $r_i = 1$ for all $i$, and the interaction matrix $\bm{A}$ is constructed as a $7 \times 7$ matrix based on nearest-neighbor interactions, plus a few additional specific interactions:
\begin{equation}
\bm{A} =
\begin{bmatrix}
 1 &  1 &  1 &  0 &  0 &  0 & -1 \\
-1 &  1 &  1 &  0 &  0 &  0 &  0 \\
 1 & -1 &  1 &  1 &  1 &  0 &  0 \\
 0 &  0 & -1 &  1 &  1 &  0 &  0 \\
 0 &  0 &  1 & -1 &  1 &  1 &  0 \\
 0 &  0 &  0 &  0 & -1 &  1 &  1 \\
 1 &  0 &  0 &  0 &  0 & -1 &  1
\end{bmatrix}.
\label{eq:lv_A_matrix}
\end{equation}

The drift \eqref{eq:lv_drift} contains linear terms ($r_i x_i$) and quadratic terms ($-A_{ij} x_i x_j$). To represent this drift exactly in the form $\bm{f}(\bm{x}) = \sum \alpha_k^* \bm{b}_k(\bm{x})$, the minimal basis set $\mathcal{B}^*_{\text{LV}}$ must include vector basis functions corresponding to these terms. Specifically, it requires:
\begin{itemize}
    \item 7 linear basis functions: $\bm{b}_{i, \text{linear}}(\bm{x}) = x_i \bm{e}_i$ (associated coefficient $r_i=1$).
    \item Basis functions for the quadratic terms: $\bm{b}_{ij, \text{quadratic}}(\bm{x}) = x_i x_j \bm{e}_i$ for each pair $(i, j)$ where $A_{ij} \neq 0$ (associated coefficient $-A_{ij}$).
\end{itemize}
The 25 non-zero entries in $\bm{A}$ \eqref{eq:lv_A_matrix} leads to a minimal basis $\mathcal{B}^*_{\text{LV}}$ containing $n_{\mathcal{B}^*} = 7 (\text{linear}) + 25 (\text{quadratic}) = 32$ functions, corresponding to the value used in Figure~\ref{fig:drift_error_lv_multiplicative}.

In addition to the deterministic drift \eqref{eq:lv_drift}, the system includes multiplicative noise. The noise term $\sqrt{2 \bm{D}(\bm{x})} \dd{\bm{W}_t}$ is defined via $\sqrt{2 D_{ii}(\bm{x})} = \sqrt{D_0} x_i$ for the diagonal elements (and zero off-diagonal), corresponding to a diffusion matrix with $D_{ii}(\bm{x}) = \frac{D_0}{2} x_i^2$. For the simulations presented, the noise strength parameter is set to $D_0=0.05$. This form of noise, where fluctuation magnitude scales with the square of the population ($x_i^2$), is often characteristic of environmental stochasticity affecting population dynamics proportionally.

Trajectories for analysis were generated using the Euler-Maruyama numerical integration scheme with a fine integration time step $\dd{t}_{\text{sim}} = 0.001$. To simulate typical experimental observation conditions, the resulting high-resolution trajectory was then subsampled every 10 steps, yielding the final trajectory data with an effective time step of $\Delta t = 10 \times dt_{\text{sim}} = 0.01$.

For benchmarking purposes, particularly when testing model selection, we also define an over-complete basis set, $\mathcal{B}_{0, \text{LV}}$. This set is constructed to include all linear and quadratic terms potentially relevant to the Lotka-Volterra dynamics within our vector basis framework. Specifically, it contains the $d=7$ linear basis functions $\{\bm{b}_{i, \text{linear}}(\bm{x}) = x_i \bm{e}_i \mid i=1,\dots,7\}$ plus all $d^2=49$ possible quadratic basis functions of the form $\{\bm{b}_{ij, \text{quadratic}}(\bm{x}) = x_i x_j \bm{e}_i \mid i=1,\dots,7, j=1,\dots,7\}$. This construction results in a total of $n_{\mathcal{B}_0} = 7 + 49 = 56$ basis functions in $\mathcal{B}_{0, \text{LV}}$, encompassing the minimal basis $\mathcal{B}^*_{\text{LV}}$.

\subsubsection{Numerical Validation: Multiplicative Noise Benchmark}

\begin{figure}[htbp]
 \centering
 \includegraphics[width=1\linewidth]{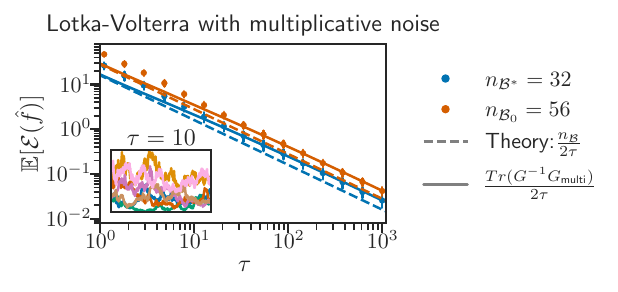} 
    \caption{Log–log plot showing the evolution of the inferred drift error, $\mathbb{E}[\mathcal{E}(\hat{\bm{f}})]$, as a function of the trajectory length $\tau$ for the Lotka–Volterra system under multiplicative noise, when analyzed using the estimator designed for additive noise (\cref{eq:chap_2b_alpha_approx_mle}). Results are presented for the minimal exact basis $\mathcal{B}^*$ ($n_{\mathcal{B}^*}=32$) and the with the over-complete basis ($n_{\mathcal{B}_0}=56$). The error curves (blue points with error bars showing the standard deviation) are compared against the theoretical prediction for additive noise ($n_{\mathcal{B}}/(2\tau)$, lower dashed line) and the corrected prediction accounting for multiplicative noise misspecification ($\frac{1}{2\tau}\mathrm{Tr}(\bm{G}^{-1}\bm{G}_{\text{multi}})$ derived from \cref{eq:drift_error_estimator}, upper dashed line). Inset: an example trajectory of length $\tau=10$.}
 \label{fig:drift_error_lv_multiplicative}
\end{figure}

As demonstrated in Figure~\ref{fig:drift_error_lv_multiplicative}, when applying the estimator designed for additive noise to the Lotka-Volterra system with true multiplicative noise, the actual mean squared drift error $\mathbb{E}[\mathcal{E}(\hat{\bm{f}})]$ is slightly higher than the prediction $n_{\mathcal{B}}/(2\tau)$ (lower dashed line) that would be expected under the idealized additive noise assumption. This discrepancy arises from the model misspecification inherent in using the averaged diffusion estimate $\bar{\bm{D}}$, which neglects the state-dependence of the true diffusion matrix $\bm{D}(\bm{x}_t)$. 
Despite this, the error clearly retains the characteristic $1/\tau$ decay rate for large trajectory lengths $\tau$. The magnitude of the error aligns well with the modified theoretical prediction given by $\frac{1}{2\tau}\mathrm{Tr}((\mathbb{E}[\bm{G}])^{-1}(\mathbb{E}[\bm{G}^*_{\text{multi}}]))$ (\cref{eq:chap2d_drift_error_multiplicative_revised}), as indicated by the upper dashed line in Figure~\ref{fig:drift_error_lv_multiplicative}.

Furthermore, the estimator $\hat{\mathcal{E}}_{\text{est}}$ defined in \cref{eq:drift_error_estimator} offers a practical means to approximate the drift error directly from observed data. While the increased error magnitude due to misspecification should be acknowledged, the fact that the estimator's error still converges to zero at the optimal $1/\tau$ rate suggests that using the simpler additive-noise estimator might remain a pragmatic, albeit potentially suboptimal, approach for drift inference, especially in scenarios where implementing estimators specifically designed for multiplicative noise is significantly more complex.

\section{Conclusion}
\label{sec:conclusion}

In this chapter, we established a framework for inferring the drift \( \bm{f}(\bm{x}) \) and diffusion matrix \( \bm{D}(\bm{x}) \) of a Langevin equation from discretely sampled trajectory data. 
Our goal was to develop and analyze practical methods for parameter estimation based on likelihood principles.

For the common case of additive noise (constant \( \bm{D} \)), we derived the formal Maximum Likelihood Estimators (MLEs) for drift parameters \( \bm{\alpha} \) (assuming a basis function expansion \( \bm{f}(\bm{x}) = \sum \alpha_i \bm{b}_i(\bm{x}) \)) and the diffusion matrix \( \bm{D} \). We identified the inherent coupling between these estimators, which complicates direct analytical solutions. To overcome this, we introduced and analyzed a practical Approximate Maximum Likelihood (AML) for the drift parameters, \( \hat{\bm{\alpha}} \), which utilizes a preliminary simple estimate of the constant diffusion matrix, \( \bar{\bm{D}} \). Under the assumption of additive noise, this approximate estimator was shown to be asymptotically unbiased, with a Mean Squared Error (MSE) scaling favourably as \( 1/\tau \) and achieving the Cramér-Rao lower bound. Furthermore, we derived a simple and insightful scaling law for the expected normalized drift reconstruction error, \( \mathbb{E}[\mathcal{E}(\hat{\bm{f}})] \approx n_{\mathcal{B}}/(2\tau) \), quantifying the error contribution per basis function.

Recognizing that many real-world systems exhibit state-dependent noise, we extended our analysis to investigate the consequences of applying the additive-noise estimator $\hat{\bm{\alpha}}$ to data generated from systems with true multiplicative noise (\cref{sec:multiplicative_noise_impact_analysis}). We found that the simple diffusion estimator $\bar{\bm{D}}$ converges to the trajectory average of the true diffusion matrix $\langle \bm{D}(\bm{x}_t) \rangle$, and derived a modified expression for the expected drift error (\cref{eq:chap2d_drift_error_multiplicative_revised}). While the simple $n_{\mathcal{B}}/(2\tau)$ scaling breaks down due to the model misspecification, the analysis revealed that the $1/\tau$ convergence rate is preserved. Furthermore, we developed an estimator, $\hat{\mathcal{E}}_{\text{est}}$ (\cref{eq:drift_error_estimator}), capable of approximating this modified error directly from data, providing a valuable tool for assessing estimator performance even under misspecified noise conditions, as demonstrated with the Lotka-Volterra example (Figure~\ref{fig:drift_error_lv_multiplicative}).

In summary, this chapter has established a practical framework for inferring Langevin drift dynamics from trajectory data, focusing on a computationally tractable Approximate Maximum Likelihood (AML) approach. We have provided a thorough analysis of its statistical errors under both additive and multiplicative noise conditions, characterizing the estimator's performance and its dependence on the number of basis functions $n_{\mathcal{B}}$ and trajectory length $\tau$. The derived error scaling laws are crucial not only for understanding estimator behavior but also serve as an essential foundation for model building. However, the methods developed herein rely on the direct observation of the system's state $\bm{x}_t$ and assume a sufficiently small discretization step $\Delta t$. The subsequent chapter will address these practical challenges, exploring inference techniques designed to handle the complexities introduced by measurement noise (where observations are imperfect) and the difficulties associated with larger sampling intervals. Building upon the foundations laid in both this chapter and the next, the techniques for model selection, aimed at identifying the most predictive yet parsimonious drift model by selecting the optimal subset of basis functions, will then be the central topic of \cref{part:Model_selection}.
\newpage

\begin{tcolorbox}[
    enhanced, 
    sharp corners, 
    boxrule=0.5pt, 
    colframe=black!75!white, 
    colback=white, 
    coltitle=black, 
    fonttitle=\bfseries, 
    title=Chapter Takeaways, 
    attach boxed title to top left={yshift=-0.1in, xshift=0.15in}, 
    boxed title style={ 
        colback=white, 
        sharp corners, 
        boxrule=0pt, 
        frame code={
            \draw[black!75!white, line width=0.5pt]
                ([yshift=-1pt]frame.south west) -- ([yshift=-1pt]frame.south east);
        }
    },
    boxsep=5pt, 
    left=5pt,
    right=5pt,
    top=12pt, 
    bottom=5pt
    ]
    \textbf{Context:} We aimed to infer parameters for the Langevin equation
    \begin{equation*}
        \dd{\bm{x}_t} = \bm{f}(\bm{x}_t) \dd{t} + \sqrt{2\, \bm{D}(\bm{x}_t)}\, \dd{\bm{W}_t}
    \end{equation*}
     from discrete data, modeling the drift as
     \begin{equation*}
     \bm{f}(\bm{x}) = \sum_{i=1}^{n_{\mathcal{B}}} \alpha_i \bm{b}_i(\bm{x}).
     \end{equation*}
    We developed a practical approximation for the MLE for drift coefficients \( \alpha_i \):
                \[ \hat{\alpha}_i = \sum_{j} (\bm{G}^{-1})_{ij} \left\langle \bm{b}_j^\top (4\bar{\bm{D}})^{-1} \frac{\Delta \bm{x}_t}{\Delta t} \right\rangle \]
                using \( \bar{\bm{D}} = \frac{1}{2\Delta t}\langle \Delta \bm{x}_t\,\Delta \bm{x}_t^\top \rangle \) and \( G_{ij} = \langle \bm{b}_i^\top (4\bar{\bm{D}})^{-1} \bm{b}_j \rangle \).
    \medskip 

    \begin{itemize}
        \item \textbf{Additive Noise (\(\bm{D}(\bm{x}_t) = \bm{D}\) constant):}
            \begin{itemize}
                \item This estimator is asymptotically unbiased and efficient.
                \item Simple drift error: \( {\mathbb{E}[\mathcal{E}(\hat{\bm{f}})] \approx \frac{n_{\mathcal{B}}}{2\tau}} \). 
            \end{itemize}
        \item \textbf{Multiplicative Noise (Impact of Misspecification):}
            \begin{itemize}
                \item Actual expected error: \( \mathbb{E}[\mathcal{E}(\hat{\bm{f}})] \approx \frac{1}{2\tau} \mathrm{Tr}( G^{-1} G_{\text{multi}} ) \neq \frac{n_{\mathcal{B}}}{2\tau}\) (cf. \cref{eq:chap2d_drift_error_multiplicative_revised}). 
            \end{itemize}
    \end{itemize}
\end{tcolorbox}

%% file: tex_body/chap2_c.tex
\chapter{Techniques to Improve SDE Inference}
\label{chap:advanced_inference}

\epigraph{"C'est par la logique qu'on démontre, c'est par l'intuition qu'on invente" \\[0.6em] \textit{It is by logic that we prove, but by intuition that we discover.}}{Henri Poincaré}

\chaptertoc{}

\section{Presentation of Practical Challenges}
\label{sec:chap3_challenges}

Applying Stochastic Differential Equation (SDE) inference methods, such as those introduced in the previous chapter, to real-world experimental data presents several practical challenges. This chapter focuses on two primary challenges: the sampling interval and measurement noise.

First, the \textbf{sampling interval} \( \Delta t \), defined as the time between consecutive measurements, is often constrained. While high temporal resolution (small \( \Delta t \)) is desirable for accurately capturing fast system dynamics, practical limitations can impose larger intervals. For instance, in fluorescence microscopy experiments, fluorophores photobleach during prolonged acquisition. To extend the total observation duration \( \tau \) before excessive photobleaching degrades the signal, it might be necessary to increase the sampling interval \( \Delta t \), thereby reducing the sampling frequency and accepting lower temporal resolution.

Second, all physical measurements are subject to \textbf{measurement noise}, representing the discrepancy between a measured value and the true underlying quantity. This noise can originate from various sources, including inherent detector limitations and environmental fluctuations. Consequently, robustness to measurement noise is a critical requirement for any practical inference method intended for use with experimental data.

\begin{figure}[htbp] 
    \centering
    \includegraphics[width=0.7\linewidth]{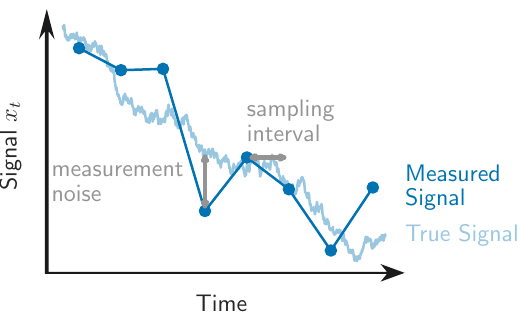} 
    \caption{Schematic illustration comparing an underlying 'True Signal' (solid curve) with the corresponding 'Measured Signal' (dots). The measurements deviate from the true signal due to: (1) \textbf{discrete sampling} at intervals \( \Delta t \), which misses information between points, and (2) \textbf{measurement noise}, which causes vertical scatter around the true value at each sampling time.}
    \label{fig:sampling_noise_illustration} 
\end{figure}

A schematic representation illustrating the combined impact of discrete sampling and measurement noise is shown in Figure~\ref{fig:sampling_noise_illustration}. While this chapter concentrates on these two issues, analyzing real-world experimental data often involves other significant challenges. For example, \textbf{partial observability} can arise when experimental limitations prevent the measurement of all relevant degrees of freedom. Furthermore, the underlying system dynamics may not be constant, exhibiting \textbf{non-stationarity} where the governing equations change over time (e.g., $\bm{f}(t, \bm{x_t})$). 
Addressing the issue of non-stationarity was beyond the scope of this thesis.

\section{Robustness with Large Sampling Intervals}
\label{sec:chap3_large_dt}

\subsection{The Trapeze Derivation} 

The approximate MLE drift estimator developed in the previous chapter relied on the Euler-Maruyama approximation \( \Delta \bm{x}_t \approx \bm{f}(\bm{x}_t)\Delta t + \sqrt{2 \bm{D}} \Delta \bm{W}_t\), an approximation that is formally accurate only in the limit \( \Delta t \to 0 \). 
How does this estimator perform when the sampling interval \( \Delta t \) increases?

We now explore this dependence on \( \Delta t \) informally, starting from the exact relation obtained by integrating the SDE (assuming the Itô interpretation, see e.g., \cref{eq:chap_2a_langevin} from the previous chapter for the SDE form). (For a rigorous treatment based on the stochastic Taylor expansion, readers are referred to \cref{appendix:mse_dt_calculation}.)
\begin{equation}
\frac{\Delta\bm{x}_t}{\Delta t} = \frac{1}{\Delta t}\int_t^{t+\Delta t} \bm{f}(\bm{x}_{t'}) \dd{t'} + \frac{1}{\Delta t}\int_t^{t+\Delta t} \sqrt{2\bm{D}(\bm{x}_{t'})} \dd{\bm{W}_{t'}}
\label{eq:delta_x_integrated}
\end{equation}
Recall that the Approximate Maximum Likelihood (AML) based estimator approximates the first term on the right-hand side (the time-averaged drift) simply by its value at the start of the interval, \( \bm{f}(\bm{x}_t) \). 
This approximation is expected to deteriorate as the sampling interval \( \Delta t \) increases.

To improve the approximation of the drift integral term for larger \( \Delta t \), we can employ higher-order integration rules. The trapeze rule, for instance, approximates the integral as: 
\[ \frac{1}{\Delta t}\int_t^{t+\Delta t} \bm{f}(\bm{x}_{t'}) \dd{t'} \approx \frac{\bm{f}(\bm{x}_t) + \bm{f}(\bm{x}_{t+\Delta t})}{2}. \]
This rule is known to have a local truncation error of \( \mathcal{O}(\Delta t^2) \) for deterministic integrals and serves as a basis for higher-order Ordinary Differential Equation (ODE) integration schemes like Heun's method. It suggests that averaging the drift at the start and end points might provide a more accurate estimate of the mean drift over the interval compared to the $\mathcal{O}(\Delta t)$ Euler-Maruyama approximation, especially for larger $\Delta t$. Substituting this approximation into \cref{eq:delta_x_integrated} yields:
\begin{equation}
  \frac{\Delta\bm{x}_t}{\Delta t} = \frac{\bm{f}(\bm{x}_{t}) + \bm{f}(\bm{x}_{t+\Delta t})}{2} + \frac{1}{\Delta t}\int_t^{t+\Delta t} \sqrt{2\bm{D}(\bm{x}_{t'})} \dd{\bm{W}_{t'}} + \mathcal{O}(\Delta t^2) + \mathcal{O}_{\text{fluc}}(1)
  \label{eq:delta_x_trapeze_approx} 
\end{equation}
where $\mathcal{O}_{\text{fluc}}$ represents the order for the fluctuating terms (cf. \cref{eq:appendix_final_taylor} for the exact derivation). 
This relation suggests that for larger \( \Delta t \), the trapeze average \( (\bm{f}(\bm{x}_t) + \bm{f}(\bm{x}_{t+\Delta t}))/2 \) provides a better approximation for the drift's contribution to the observed increment \( \Delta \bm{x}_t / \Delta t \) than the simple Euler-Maruyama term \( \bm{f}(\bm{x}_t) \).

We can leverage this potentially improved approximation to define a new estimator that might be more robust to larger \( \Delta t \). Assuming the true drift lies in the span of a known basis \( \mathcal{B} = \{\bm{b}_i\}_{i=1,n_{\mathcal{B}}} \), such that \( \bm{f}(\bm{x}) = \sum_j \alpha_j^* \bm{b}_j(\bm{x}) \), we substitute this functional form into the trapeze approximation \cref{eq:delta_x_trapeze_approx}: 
\begin{equation}
  \frac{\Delta\bm{x}_t}{\Delta t} \approx \sum_j \alpha_j^* \frac{\bm{b_j}(\bm{x}_{t}) + \bm{b_j}(\bm{x}_{t+\Delta t})}{2} + \frac{1}{\Delta t}\int_t^{t+\Delta t} \sqrt{2\bm{D}(\bm{x}_{t'})} \dd{\bm{W}_{t'}}
  \label{eq:delta_x_trapeze_approx_alpha} 
\end{equation}
We then follow a similar regression procedure as in the previous chapter: multiply by a suitable weighting term \( \bm{b}_i(\bm{x}_t)^\top (4\bar{\bm{D}})^{-1} \) and average over the observed trajectory to obtain the linear system:
\begin{equation}
  \avg{\bm{b}_i(\bm{x}_t)^\top (4\bar{\bm{D}})^{-1} \frac{\Delta\bm{x}_t}{\Delta t}} \approx \sum_j \alpha_j^* \avg{ \bm{b}_i(\bm{x}_t)^\top (4\bar{\bm{D}})^{-1} \frac{\bm{b_j}(\bm{x}_{t}) + \bm{b_j}(\bm{x}_{t+\Delta t})}{2}} + \text{Fluctuating terms}
  \label{eq:delta_x_trapeze_approx_alpha_system} 
\end{equation}
where \( \bar{\bm{D}} \) is still computed via \cref{eq:chap2a_approximation_D} for consistency, though its interpretation as a simple average diffusion might be less accurate for large \( \Delta t \). 
Defining the 'trapeze' Gram matrix \( \bm{G}^{Tr} \) and the input vector \( \bm{Y} \) (which is identical to the one used in the Approximate Maximum Likelihood (AML) estimator) as: 
\begin{align}
(\bm{G}^{Tr})_{ij} &= \avg{ \bm{b}_i(\bm{x}_t)^\top (4\bar{\bm{D}})^{-1} \frac{\bm{b_j}(\bm{x}_{t}) + \bm{b_j}(\bm{x}_{t+\Delta t})}{2}}, \label{eq:G_Tr_def}\\
Y_i &= \avg{\bm{b}_i(\bm{x}_t)^\top (4\bar{\bm{D}})^{-1} \frac{\Delta\bm{x}_t}{\Delta t}} , \label{eq:Y_def_chap3}
\end{align}
the system becomes \( \bm{Y} \approx \bm{G}^{Tr} \bm{\alpha} \). Inverting this system yields the trapeze estimator \( \hat{\bm{\alpha}}^{Tr} \): 
\begin{equation}
  \hat{\bm{\alpha}}^{Tr} = (\bm{G}^{Tr})^{-1} \bm{Y}, \quad \text{or element-wise: } \hat{\alpha}_i^{Tr} = \sum_j ((\bm{G}^{Tr})^{-1})_{ij} Y_j.
  \label{eq:alpha_Tr_estimator}
\end{equation}
We recall the previous Approximate Maximum Likelihood (AML) based estimator for comparison: 
\begin{equation}
    \hat{\bm{\alpha}} = (\bm{G})^{-1} \bm{Y}
    \label{eq:euler_estimator_recall}
\end{equation}
where $\bm{G}$ is the standard Gram matrix from the previous chapter, derived from a basis $\mathcal{B}$. We assume the basis functions are chosen such that $\bm{G}$ and the corresponding trapeze matrix $\bm{G}^{Tr}$ are invertible given the data; otherwise, indicating an ill-suited basis, a pseudo-inverse would be necessary. 

The trapeze estimator \( \hat{\bm{\alpha}}^{Tr} \) is derived from a linear regression approach based on the trapeze approximation \cref{eq:delta_x_trapeze_approx}, not directly from maximizing a likelihood function corresponding to this approximation. 
However, it retains a desirable property: in the limit \( \Delta t \to 0 \), we have \( \bm{b}_j(\bm{x}_{t+\Delta t}) \to \bm{b}_j(\bm{x}_t) \), causing the trapeze Gram matrix \( \bm{G}^{Tr} \to \bm{G} \). 
Consequently, \( \hat{\bm{\alpha}}^{Tr} \to \hat{\bm{\alpha}} \), recovering the Approximate Maximum Likelihood (AML) derived from the "Euler-Maruyama" approximation.

\subsection{Normalized Mean Square Error for the Reconstructed Drift $\mathcal{E}(\hat{\bm{f}})$}
The primary motivation for introducing \( \hat{\bm{\alpha}}^{Tr} \) is that the trapeze rule approximates the drift integral with a local truncation error of \( \mathcal{O}(\Delta t^2) \), whereas the "Euler-Maruyama" scheme used for \( \hat{\bm{\alpha}} \) has an error of \( \mathcal{O}(\Delta t) \). 
Therefore, we hypothesize that the systematic \textit{bias} induced by a finite \( \Delta t \) in the drift estimate \( \hat{\bm{f}}^{Tr}(\bm{x}) = \sum_i \hat{\alpha}_i^{Tr} \bm{b}_i(\bm{x}) \) scales as \( \mathcal{O}(\Delta t^2) \), compared to a \( \mathcal{O}(\Delta t) \) bias scaling for the Approximate Maximum Likelihood (AML) based estimate \( \hat{\bm{f}}(\bm{x}) = \sum_i \hat{\alpha}_i \bm{b}_i(\bm{x}) \).

Given the bias-variance decomposition (MSE = $\text{Bias}^2$ + Variance), we expect the overall normalized mean squared error (NMSE)
$$
\mathcal{E}(\hat{\bm{f}}) = \langle (\hat{\bm{f}} - \bm{f})^\top (4\bar{\bm{D}})^{-1} (\hat{\bm{f}} - \bm{f}) \rangle
$$
of the drift estimate to be dominated by the squared bias for large sampling intervals \( \Delta t \). This implies an expected NMSE $\mathcal{E}(\hat{\bm{f}})$ scaling of approximately \( \mathcal{O}(\Delta t^2) \) for the Approximate Maximum Likelihood (AML) estimator and \( \mathcal{O}(\Delta t^4) \) for the trapeze estimator in this regime. 
A precise derivation is given in \cref{appendix:mse_dt_calculation}, which \textbf{confirms} this scaling.

\begin{figure}[htbp]
    \begin{center}
    \includegraphics[width=1\linewidth]{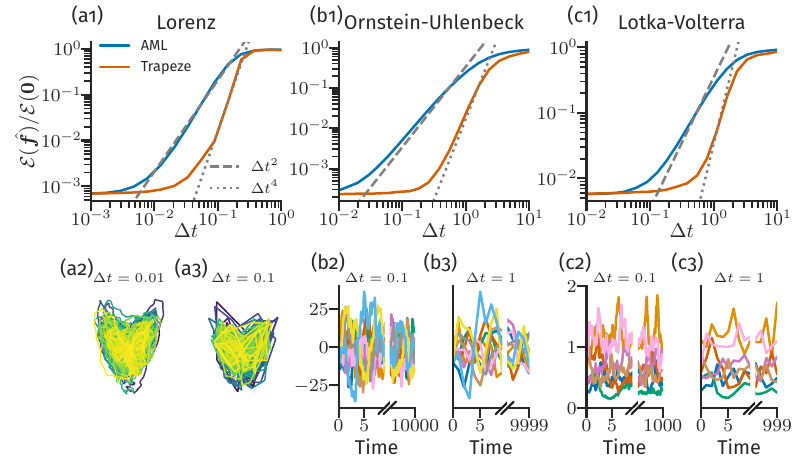} 
    \end{center}
    \caption{Numerical simulation results comparing the Approximate Maximum Likelihood (AML) \cref{eq:euler_estimator_recall} and Trapeze estimators \cref{eq:alpha_Tr_estimator} for different stochastic differential equation systems inferred with $\mathcal{B}^*$ model. (a1, b1, c1) Relative error $\mathcal{E}(\bm{\hat{f}})/\mathcal{E}(\bm{0})$ versus sampling interval $\Delta t$ for the Lorenz system with $\tau=100$ (a1), Ornstein-Uhlenbeck process with $\tau=10000$ (b1), and Lotka-Volterra system with $\tau=1000$ (c1). Dashed lines indicate $\Delta t^2$ and $\Delta t^4$ convergence slopes for reference. (a2, a3, b2, b3, c2, c3) Lorenz, Ornstein-Uhlenbeck, and Lotka-Volterra trajectories sampled with two different sampling intervals $\Delta t$.} 
    \label{fig:delta_t_benchmark}
\end{figure}

Furthermore, this theoretical scaling is approximately numerically validated by the benchmark results presented in \cref{fig:delta_t_benchmark}. 
Panels (a1, b1, c1) show the relative error \( \mathcal{E}(\bm{\hat{f}})/\mathcal{E}(\bm{0}) \) versus the sampling interval \( \Delta t \) for the Lorenz, Ornstein-Uhlenbeck, and Lotka-Volterra systems, respectively. As hypothesized, the relative error for the Approximate Maximum Likelihood (AML) method follows the reference slope proportional to \( \Delta t^2 \) for larger \( \Delta t \), while the error for the trapeze method aligns well with the steeper \( \Delta t^4 \) reference slope. 

Notably, this improved scaling for the trapeze method holds both for systems modeled with additive noise, such as Lorenz (a1) and Ornstein-Uhlenbeck (b1), and for the Lotka-Volterra system (c1), which inherently features multiplicative noise. 
This convergence behavior confirms that for sufficiently large \( \Delta t \), the estimator bias becomes the dominant contribution to the overall error for both methods in these systems.

Note that the lower plateau observed in Panels (a1, b1, c1) of \cref{fig:delta_t_benchmark} is controlled by the length of the trajectory $\tau$ and the number of basis functions $n_\mathcal{B}$, which was computed in the previous chapter for the $\Delta t \to 0$ limit as $\mathbb{E}[\mathcal{E}(\hat{f})] \approx \frac{n_{\mathcal{B}}}{2\tau}$. 

Beyond this numerical confirmation, we provide in \cref{appendix:mse_dt_calculation} a detailed calculation of the bias and MSE for both the Approximate Maximum Likelihood (AML) estimator (\cref{appendix:section_euler_maruyama_mse}) and the trapeze estimator (\cref{appendix:section_trapeze_mse}). 

The analysis framework presented here contributed to a publication investigating the dynamics of cell nucleus translocation through tight constrictions using a data-driven approach~\cite{amiriInferringGeometricalDynamics2024}. In that work (Amiri et al., 2024), Stochastic Force Inference (SFI) \cite{frishmanLearningForceFields2020} was applied to experimental trajectories of nuclei, including geometric shape descriptors, to derive predictive stochastic differential equations describing the process~\cite{amiriInferringGeometricalDynamics2024}. A key methodological challenge addressed in the paper was the large sampling interval (\( \Delta t \) = 10 min) inherent in the experiments due to phototoxicity~\cite{amiriInferringGeometricalDynamics2024}. To mitigate the resulting discretization bias, an improved SFI algorithm incorporating a \textbf{trapeze estimator} was introduced. 
This modification reduced the estimator bias (to \( \mathcal{O}(\Delta t^2) \)) compared to the previous methods in SFI, enabling more accurate inference from the available time-lapse data.

It is noteworthy that similar higher-order estimators addressing large sampling intervals have been independently developed, for instance, by Wanner et al.~\cite{wannerHigherOrderDrift2024}, employing alternative derivations.

A light note on the development of the trapeze estimator: the initial motivation arose somewhat serendipitously. While "working" remotely during the summer after my first PhD year, I faced practical limitations inferring Stochastic Partial Differential Equations (SPDEs) from simulated video data due to memory constraints. Increasing the sampling interval \( \Delta t \) to manage data size unfortunately compromised the inference accuracy using standard methods. Facing an upcoming meeting with my advisor, Pierre Ronceray, and needing a solution to this discretization error problem, I found the trapeze trick to improve the inference for large  \( \Delta t \). I remember Pierre looking skeptical during the online meeting that this simple trick was indeed working.

\subsection{Alternative Methods for Large $\Delta t$}

Beyond the regression-based approaches using higher-order integration rules like the trapeze estimator discussed here and in Wanner et al.~\cite{wannerHigherOrderDrift2024}, another significant line of research addresses the large $\Delta t$ challenge through direct approximation of the likelihood function. 
A prominent example is the work by Aït-Sahalia~\cite{ait-sahaliaClosedFormLikelihoodExpansions2002, ait-sahaliaClosedformLikelihoodExpansions2008}, who developed closed-form asymptotic expansions based on Hermite polynomials for the likelihood function. 
This work was discovered by the author relatively recently, and thus a thorough comparative analysis was not conducted.
Moreover, a Python package implementing the method developed by Aït-Sahalia \cite{kirkbyPymlePythonPackage2024} was recently developed. Thus, benchmarking against these methods represents a possible direction for future work.

\section{Robustness Against Measurement Noise}
\label{sec:chap3_measurement_noise}

We now shift our focus to the second challenge: the impact of random measurement errors on the inference process. We model the observed position \( \bm{y}_t \) at time \( t \) as the true underlying system state \( \bm{x}_t \) corrupted by additive, independent Gaussian noise:
\begin{equation}
\bm{y}_t = \bm{x}_t + \bm{\eta}_t, \quad \text{with } \bm{\eta}_t \sim \mathcal{N}(\bm{0}, \bm{\sigma}^2),
\label{eq:measurement_noise_model}
\end{equation}
where $\bm{\sigma}^2$ is the covariance matrix of the measurement noise. The noise \( \bm{\eta}_t \) is assumed to be independent at different time points, i.e., \( \mathbb{E}[\bm{\eta}_t \bm{\eta}_{t'}^\top] = \bm{\sigma}^2 \delta_{tt'} \), and also independent of the system's dynamical noise \( \dd{\bm{W}_t} \).

How does this measurement noise affect our drift estimators, which fundamentally rely on calculating increments from the observed data? The observed increment is now calculated from the noisy positions: \( \Delta \bm{y}_t = \bm{y}_{t+\Delta t} - \bm{y}_t = (\bm{x}_{t+\Delta t} - \bm{x}_t) + (\bm{\eta}_{t+\Delta t} - \bm{\eta}_t) = \Delta \bm{x}_t + \Delta \bm{\eta}_t \), where \( \Delta \bm{\eta}_t = \bm{\eta}_{t+\Delta t} - \bm{\eta}_t \). The term \( \frac{\Delta \bm{y}_t}{\Delta t} \), which approximates the velocity and is used in the estimators (e.g., in the vector \( \bm{Y} \) in \cref{eq:Y_def_chap3}), becomes:
\[ \frac{\Delta \bm{y}_t}{\Delta t} = \frac{\Delta \bm{x}_t}{\Delta t} + \frac{\Delta \bm{\eta}_t}{\Delta t}. \]
The measurement noise contribution \( \frac{\Delta \bm{\eta}_t}{\Delta t} \) has zero mean (\( \mathbb{E}[\Delta \bm{\eta}_t] = \bm{0} \)), but its variance is \( \Var\left(\frac{\Delta \bm{\eta}_t}{\Delta t}\right) = \frac{\Var(\bm{\eta}_{t+\Delta t}) + \Var(\bm{\eta}_t)}{(\Delta t)^2} = \frac{2 \bm{\sigma}^2}{(\Delta t)^2} \). This variance dramatically increases, or "blows up," as \( \Delta t \to 0 \). This indicates that for high sampling frequencies (small \( \Delta t \)), the noise term \( \Delta \bm{\eta}_t / \Delta t \) can overwhelm the true physical increment \( \Delta \bm{x}_t / \Delta t \), posing a significant challenge for inference.

Furthermore, a more subtle issue arises because the basis functions in the estimators are now evaluated at the noisy positions \( \bm{y}_t \) instead of the true positions \( \bm{x}_t \). Consider the term \( Y_i \) used in both the Approximate Maximum Likelihood (AML) estimator \( \hat{\bm{\alpha}} \) (\cref{eq:euler_estimator_recall}) and the trapeze estimator \( \hat{\bm{\alpha}}^{Tr} \) (\cref{eq:alpha_Tr_estimator}). 
When substituting the observed quantities \( \Delta\bm{y}_t \) and \( \bm{y}_t \), it becomes:
\begin{align}
  Y_i &= \avg{\frac{\Delta \bm{y}_t}{\Delta t}\cdot (4\bar{\bm{D}})^{-1}\bm{b}_i(\bm{y}_t)} \nonumber \\
  &= \avg{\left(\frac{\Delta \bm{x}_t}{\Delta t} + \frac{\Delta \bm{\eta}_t}{\Delta t}\right) \cdot (4\bar{\bm{D}})^{-1}\bm{b}_i(\bm{x}_t + \bm{\eta}_t)} \nonumber \\
  &= \avg{\frac{\Delta \bm{x}_t}{\Delta t}\cdot (4\bar{\bm{D}})^{-1}\bm{b}_i(\bm{y}_t)} + \underbrace{\avg{\frac{\Delta \bm{\eta}_t}{\Delta t}\cdot (4\bar{\bm{D}})^{-1}\bm{b}_i(\bm{y}_t)}}_{\text{Principal bias term}}.
  \label{eq:noise_bias_term_def}
\end{align}
The second term in \cref{eq:noise_bias_term_def} introduces a systematic bias. This bias arises because the basis function evaluation \( \bm{b}_i(\bm{y}_t) = \bm{b}_i(\bm{x}_t + \bm{\eta}_t) \) depends on the noise \( \bm{\eta}_t \) present at time \( t \). Since the noise increment \( \Delta \bm{\eta}_t = \bm{\eta}_{t+\Delta t} - \bm{\eta}_t \) also involves \( \bm{\eta}_t \), these two terms are correlated. Specifically, assuming \( \bm{b}_i \) is sufficiently smooth and the noise magnitude \( \sigma \) is small, we can approximate \( \bm{b}_i(\bm{y}_t) \approx \bm{b}_i(\bm{x}_t) + (\bm{\eta}_t \cdot \nabla)\bm{b}_i(\bm{x}_t) \). The first-order problematic correlation within the bias term expectation is then approximately: 
\[ \mathbb{E}\left[ \frac{\Delta \bm{\eta}_t}{\Delta t} \cdot (4\bar{\bm{D}})^{-1} (\bm{\eta}_t \cdot \nabla)\bm{b}_i(\bm{x}_t) \right] = \mathbb{E}\left[ \frac{(\bm{\eta}_{t+\Delta t} - \bm{\eta}_t)}{\Delta t} \cdot (4\bar{\bm{D}})^{-1} (\bm{\eta}_t \cdot \nabla)\bm{b}_i(\bm{x}_t) \right]. \]
Using the independence of noise at different times (\( \mathbb{E}[\bm{\eta}_{t+\Delta t} \bm{\eta}_t^\top] = \bm{0} \)) and \( \mathbb{E}[\bm{\eta}_t (\bm{\eta}_t \cdot \nabla)\bm{b}_i] = (\bm{\sigma}^2 \nabla) \cdot \bm{b}_i \), this expectation becomes \( - \frac{1}{\Delta t} (4\bar{\bm{D}})^{-1} ((\bm{\sigma}^2 \nabla) \cdot \bm{b}_i) \). This leads to a bias term in \( Y_i \) that is approximately proportional to \( \sigma^2 / \Delta t \), which again diverges as \( \Delta t \to 0 \). Mitigation strategies are therefore required, especially when dealing with high-frequency sampling or significant noise levels.

The noise impact starts to dominate the underlying dynamics when this bias term is comparable in magnitude to the drift term itself, which occurs roughly when $\sigma^2 / \Delta t \approx \langle \bm{f}^2 \rangle$ (with $\bm{\sigma} = \sigma \bm{I}$). Consequently, mitigation strategies are crucial, particularly for high-frequency sampling (small $\Delta t$) or significant noise levels ($\sigma^2$).

\subsection{Shift Estimator}
\label{subsec:shifted_estimator}

To mitigate the bias identified in \cref{eq:noise_bias_term_def}, we need to break the correlation between the noise increment \( \Delta \bm{\eta}_t \) and the noise component within the evaluated basis function \( \bm{b}_i(\bm{y}_t) \). One strategy is to evaluate the basis function at a time point \emph{not} involved in the increment calculation \( \Delta \bm{\eta}_t = \bm{\eta}_{t+\Delta t} - \bm{\eta}_t \). A simple choice is to use the basis function evaluated at the \emph{previous} time step, \( \bm{b}_i(\bm{y}_{t-\Delta t}) \).

If we modify the problematic term in \cref{eq:noise_bias_term_def} using this 'time-shifted' or 'lagged' evaluation point, it becomes:
\[ \avg{\frac{\Delta \bm{\eta}_t}{\Delta t}\cdot (4\bar{\bm{D}})^{-1}\bm{b}_i(\bm{y}_{t-\Delta t})}. \]
Now, the noise increment involves \( \bm{\eta}_t \) and \( \bm{\eta}_{t+\Delta t} \), while the basis function evaluation \( \bm{y}_{t-\Delta t} = \bm{x}_{t-\Delta t} + \bm{\eta}_{t-\Delta t} \) involves noise \( \bm{\eta}_{t-\Delta t} \). Since the measurement noise is assumed independent at different time steps (\(t-\Delta t\), \(t\), \(t+\Delta t\)), the noise terms in the increment are independent of the noise term in the basis function evaluation. Therefore, the expectation of this modified term vanishes:
\[ \mathbb{E}\left[\frac{\Delta \bm{\eta}_t}{\Delta t}\cdot (4\bar{\bm{D}})^{-1}\bm{b}_i(\bm{y}_{t-\Delta t})\right] = \mathbb{E}\left[\frac{\Delta \bm{\eta}_t}{\Delta t}\right] \cdot \mathbb{E}\left[(4\bar{\bm{D}})^{-1}\bm{b}_i(\bm{y}_{t-\Delta t})\right] = 0, \]
because \( \mathbb{E}[\Delta \bm{\eta}_t] = 0 \). This suggests defining a 'shift' estimator by modifying the calculation of the input vector \( Y_i \) and potentially the Gram matrix \( G_{ij} \). Let the modified input vector be:
\begin{equation}
  Y_i^{\text{shift}} = \avg{\frac{\Delta \bm{y}_t}{\Delta t}\cdot (4\bar{\bm{D}})^{-1}\bm{b}_i(\bm{y}_{t-\Delta t})}.
  \label{eq:Y_lag_def}
\end{equation}
Similarly modifying the Gram matrix used in the Approximate Maximum Likelihood (AML) scheme (\cref{eq:euler_estimator_recall}) leads to:
\begin{equation}
(\bm{G}^{\text{shift}})_{ij} = \avg{\bm{b}_j(\bm{y}_t) \cdot (4\bar{\bm{D}})^{-1}\bm{b}_i(\bm{y}_{t-\Delta t})}.
\label{eq:G_lag_def}
\end{equation}
A bias-reduced estimator, which we term the 'Shift estimator', can then be constructed as:
\begin{equation}
    \hat{\bm{\alpha}}^{\text{shift}} = (\bm{G}^{\text{shift}})^{-1} \bm{Y}^{\text{shift}}.
    \label{eq:alpha_shift_estimator}
\end{equation}
Although this approach removes the primary bias term proportional to \( \sigma^2 / \Delta t \), its use with large \( \Delta t \) presents a new difficulty. The issue is that for large time steps, the observed increment \( \frac{\Delta \bm{y}_t}{\Delta t} \) can decorrelate from the basis function \( \bm{b}_i \) evaluated at the preceding time \( t-\Delta t \). Consequently, the expected correlation term used in the estimation may approach zero, \( \mathbb{E}\left[\left\langle \frac{\Delta \bm{y}_t}{\Delta t}\cdot (4\bar{\bm{D}})^{-1}\bm{b}_i(\bm{y}_{t-\Delta t}) \right\rangle\right] \approx 0 \), significantly hindering the inference process.

The development of this shift estimator occurred during my studies. To mine knowledge, this specific shift method has not been extensively reported in the literature for SDE inference, despite its simplicity.

\subsection{Stratonovich Estimator}
\label{subsubsec:averaging}

Instead of shifting the evaluation point, an alternative strategy, inspired by the evaluation point used in Stratonovich calculus, involves averaging the basis function evaluated at the start and end points of the interval over which the increment \( \Delta \bm{y}_t \) is calculated. Let's examine the critical bias term (from \cref{eq:noise_bias_term_def}) when this averaged evaluation is used in the correlation:
\[ \text{Bias term} = \avg{\frac{\Delta \bm{\eta}_t}{\Delta t}\cdot (4\bar{\bm{D}})^{-1} \left( \frac{\bm{b}_i(\bm{y}_t) + \bm{b}_i(\bm{y}_{t+\Delta t})}{2} \right)}. \]
We analyze the expectation of the core correlation term, expanding \( \bm{b}_i \) around the true states \( \bm{x}_t, \bm{x}_{t+\Delta t} \) and keeping leading-order noise terms:
\begin{align*}
  &\quad \mathbb{E}\left[ \Delta \bm{\eta}_t \cdot \frac{\bm{b}_i(\bm{y}_t) + \bm{b}_i(\bm{y}_{t+\Delta t})}{2} \right] \\
  &\approx \frac{1}{2} \mathbb{E}\left[ (\bm{\eta}_{t+\Delta t} - \bm{\eta}_t) \cdot \left( \bm{b}_i(\bm{x}_t) + (\bm{\eta}_t \cdot \nabla)\bm{b}_i(\bm{x}_t) + \bm{b}_i(\bm{x}_{t+\Delta t}) + (\bm{\eta}_{t+\Delta t} \cdot \nabla)\bm{b}_i(\bm{x}_{t+\Delta t}) \right) \right].
\end{align*}
Assuming \( \nabla \bm{b}_i \) is approximately constant between \( \bm{x}_t \) and \( \bm{x}_{t+\Delta t} \) for simplicity, the expectation becomes:
\begin{align*}
  &\approx \frac{1}{2} \mathbb{E}\left[ (\bm{\eta}_{t+\Delta t} - \bm{\eta}_t) \cdot \left( (\bm{\eta}_t + \bm{\eta}_{t+\Delta t}) \cdot \nabla \right) \bm{b}_i \right] \\
  &= \frac{1}{2} \left( \mathbb{E}[\bm{\eta}_{t+\Delta t} (\bm{\eta}_t \cdot \nabla)\bm{b}_i] + \mathbb{E}[\bm{\eta}_{t+\Delta t} (\bm{\eta}_{t+\Delta t} \cdot \nabla)\bm{b}_i] - \mathbb{E}[\bm{\eta}_t (\bm{\eta}_t \cdot \nabla)\bm{b}_i] - \mathbb{E}[\bm{\eta}_t (\bm{\eta}_{t+\Delta t} \cdot \nabla)\bm{b}_i] \right).
\end{align*}
Due to the independence of \( \bm{\eta}_t \) and \( \bm{\eta}_{t+\Delta t} \), the cross-term expectations (\( \mathbb{E}[\bm{\eta}_{t+\Delta t} \bm{\eta}_t] \)-related terms) are zero. Using the identity \( \mathbb{E}[\bm{\eta} (\bm{\eta} \cdot \nabla)\bm{b}_i] = (\bm{\sigma}^2 \nabla) \cdot \bm{b}_i \), the expectation simplifies to:
\[ \mathbb{E}[\dots] \approx \frac{1}{2} \left( 0 + (\bm{\sigma}^2 \nabla) \cdot \bm{b}_i - (\bm{\sigma}^2 \nabla) \cdot \bm{b}_i - 0 \right) = \bm{0}. \]
Thus, using the average of the basis function evaluated at the endpoints \( \bm{y}_t \) and \( \bm{y}_{t+\Delta t} \) also causes the leading-order measurement noise bias term (proportional to \( \sigma^2 / \Delta t \)) to vanish due to cancellation.

This motivates defining estimators using this averaged basis function evaluation. Let:
\begin{align}
  Y_i^{\text{Strato}} &= \avg{\frac{\Delta \bm{y}_t}{\Delta t}\cdot (4\bar{\bm{D}})^{-1}\left( \frac{\bm{b}_i(\bm{y}_t) + \bm{b}_i(\bm{y}_{t+\Delta t})}{2} \right)}, \label{eq:Y_avg_def} \\
  (\bm{G}^{\text{Strato}})_{ij} &= \avg{\bm{b}_j(\bm{y}_t)  \cdot (4\bar{\bm{D}})^{-1}\left( \frac{\bm{b}_i(\bm{y}_t) + \bm{b}_i(\bm{y}_{t+\Delta t})}{2} \right)}. \label{eq:G_avg_def}
\end{align}
However, simply using \(  (\bm{G}^{\text{Strato}})^{-1} \bm{Y}^{\text{Strato}} \) does not lead to an unbiased estimate of the coefficients \( \bm{\alpha} \) corresponding to the original Itô drift term \( \bm{f}(\bm{x}) \) in the SDE \( \dd{\bm{x}} = \bm{f}(\bm{x}) \dd{t} + \sqrt{2\bm{D}} \dd{\bm{W}}_t \). This is because the Stratonovich-like averaging introduces its own difference compared to the Itô definition. To obtain an estimator that is unbiased with respect to the Itô drift, we must relate \(Y_i^{\text{Strato}}\) back to the original Itô-based term \(Y_i\) (\cref{eq:Y_def_chap3}) and account for the difference.

We can approximate this difference (often called the Itô-Stratonovich correction term in this context):
\begin{align}
  Y_i^{\text{Strato}} &= \avg{\frac{\Delta \bm{y}_t}{\Delta t}\cdot (4\bar{\bm{D}})^{-1}\left(\frac{\bm{b}_i(\bm{y}_{t+\Delta t}) + \bm{b}_i(\bm{y}_t)}{2} \right)} \nonumber \\
  &\approx \avg{\frac{\Delta \bm{y}_t}{\Delta t}\cdot (4\bar{\bm{D}})^{-1}\left(\frac{\bm{b}_i(\bm{x}_{t+\Delta t}) + \bm{b}_i(\bm{x}_t)}{2} \right)} \nonumber \quad \text{(based on the previous derivation)} \\
  &\approx Y_i + \avg{\frac{\Delta \bm{y}_t}{\Delta t}\cdot (4\bar{\bm{D}})^{-1}\frac{\Delta \bm{x}_t \cdot \nabla\bm{b}_i(\bm{x}_{t})}{2}}\\
    &\approx Y_i + \avg{\frac{ \sqrt{2\bm{D}}\Delta\bm{W}_t}{\Delta t}\cdot (4\bar{\bm{D}})^{-1}\frac{ \sqrt{2\bm{D}} \Delta \bm{W}_t \cdot \nabla\bm{b}_i}{2}}\\
    &\approx Y_i + \avg{\Tr{(4\bar{\bm{D}})^{-1} \bm{D} \cdot \nabla \bm{b}_i}} \quad \text{(using heuristic } \Delta W^2 \sim \Delta t \text{)}
    \label{eq:ito_strato_correction_approx}
\end{align}
where in the last step (\cref{eq:ito_strato_correction_approx}), we used the approximation involving \( \mathbb{E}[ \Delta \bm{W} \Delta \bm{W}^\top ] \approx \Delta t \bm{I} \) discussed heuristically in the introduction (e.g., \cref{eq:heuristic_nototation_dw2_dt} if this label refers to a previous chapter's equation).

From this calculation, we observe that \(Y_i^{\text{Strato}}\) is not equal to \(Y_i\), containing an additional term related to the diffusion \( \bm{D} \) and the gradient of the basis function. Thus, simply inverting \(\bm{G}^{\text{Strato}}\) and multiplying it by \(Y^{\text{Strato}}\) will not yield an unbiased estimate of \( \bm{\alpha} \). 
Fortunately, since we have an estimate for this bias term (\cref{eq:ito_strato_correction_approx}), we can subtract it.

Correcting this bias requires an estimate of the diffusion tensor \( \bm{D} \) that is itself robust to measurement noise and works for additive or multiplicative dynamical noise. To achieve this, we use the instantaneous diffusion tensor estimator inspired by the work of Vestergaard et al.~\cite{vestergaardOptimalEstimationDiffusion2014}:
\begin{equation}
    \hat{\bm{D}}_{\sigma}(t) = \frac{1}{2\Delta t} \Delta \bm{y}_t \Delta \bm{y}_t^\top + \frac{1}{\Delta t}\avg{\Delta \bm{y}_{t+\Delta t} \Delta \bm{y}_t}^\top
    \label{eq:D_vestergaard}
\end{equation}
Then, we define an approximately unbiased estimator for the Itô drift coefficients, which we call the 'Stratonovich' estimator:
\begin{equation}
    \hat{\bm{\alpha}}^{\text{Strato}} =  (\bm{G}^{\text{Strato}})^{-1} \left( \bm{Y}^{\text{Strato}} - \avg{\Tr{(4\bar{\bm{D}})^{-1} \hat{\bm{D}}_{\sigma} \nabla \bm{b}}} \right)
    \label{eq:alpha_strato_corrected}
\end{equation}
where the average \( \avg{\dots} \) in the correction term is taken over the trajectory, evaluating \( \nabla \bm{b}_i \) at the observed positions \( \bm{y}_t \). This "Stratonovich form" combined with an Itô correction for the drift estimator coefficients was first derived and used in a similar form by Frishman and Ronceray~\cite{frishmanLearningForceFields2020}.

\subsection{Comparative Discussion of Techniques}
\label{sec:chap3_comparison}

\begin{figure}[htbp]
    \centering
    \includegraphics[width=1.1\linewidth]{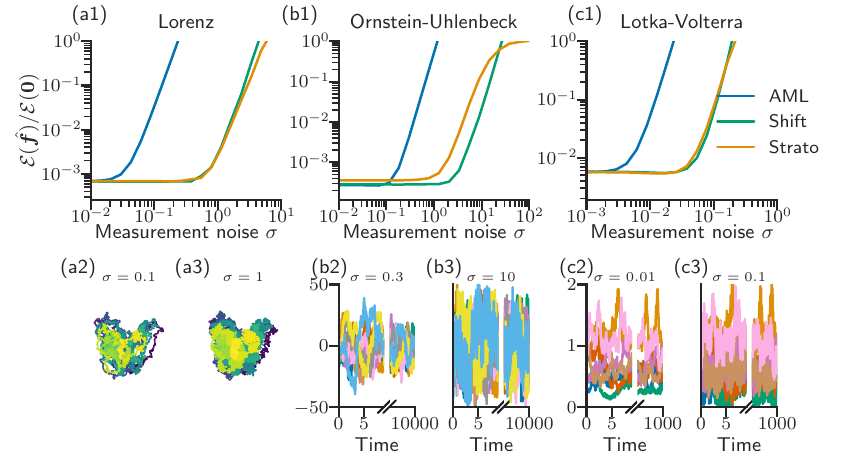}
    \caption{Comparison of numerical drift estimation methods under measurement noise. The top row (a1, b1, c1) shows the relative drift error (e.g., $\mathcal{E}(\hat{\bm{f}})/\mathcal{E}(\bm{0})$) as a function of the measurement noise standard deviation $\bm{\sigma} = \sigma \bm{I}$ for the Approximate Maximum Likelihood (AML) (\cref{eq:euler_estimator_recall}), Shift (\cref{eq:alpha_shift_estimator}), and Stratonovich (\cref{eq:alpha_strato_corrected}) methods. Results are shown for the Lorenz system (a1), Ornstein-Uhlenbeck process (b1), and Lotka-Volterra system (c1) with a fixed sampling interval $\Delta t$. The bottom rows show sample trajectories where the measurement noise $\sigma$ was selected to match 1\% of the relative drift error for the Approximate Maximum Likelihood (AML) and Shift methods.} 
    \label{fig:measurement_noise_benchmark}
\end{figure}

Having introduced two distinct strategies to mitigate the adverse effects of measurement noise – the Shift estimator (\cref{subsec:shifted_estimator}) and the Stratonovich estimator (\cref{subsubsec:averaging}) – we now numerically compare their performance against the standard Approximate Maximum Likelihood (AML) based estimator (\cref{eq:euler_estimator_recall}). The goal is to assess their robustness when applied to data corrupted by varying levels of additive measurement noise, as modeled by \cref{eq:measurement_noise_model}. 

\Cref{fig:measurement_noise_benchmark} presents the results of this comparison for three different dynamical systems: the Lorenz system, the Ornstein-Uhlenbeck process, and the Lotka-Volterra system. 
The top panels (a1, b1, c1) display the relative error in the estimated drift field, \( \mathcal{E}(\hat{\bm{f}})/\mathcal{E}(\bm{0}) \), as a function of the measurement noise standard deviation \( \sigma \) ($\bm{\sigma} = \sigma \bm{I}$). The simulation details are available in the previous \cref{sec:OU_process_description,sec:lorenz_system_description,sec:lotka_volterra_benchmark}.

The results clearly demonstrate the susceptibility of the standard Approximate Maximum Likelihood (AML) estimator to measurement noise. Across all three systems, its relative error increases substantially as \( \sigma \) grows.

In stark contrast, both the Shift estimator (\cref{eq:alpha_shift_estimator}) and the Stratonovich estimator (\cref{eq:alpha_strato_corrected}) exhibit remarkable robustness to measurement noise. Their relative errors remain significantly lower than the standard (AML) estimator's error and stay relatively constant over a wide range of \( \sigma \) values depicted in the plots. This observation validates the effectiveness of the bias mitigation strategies employed in their derivation – namely, breaking the correlation between the noise increment and the basis function evaluation (Shift method) or cancelling the leading-order bias through averaging and applying an Itô correction (Stratonovich method).

In this benchmark, the performance of the Shift and Stratonovich methods appears quantitatively very similar, with both offering a dramatic improvement over the AML approach. However, for the Ornstein-Uhlenbeck case, the Shift method shows an improvement over the Stratonovich method. 
The sample trajectories shown in the lower panels of \cref{fig:measurement_noise_benchmark} (a2, a3, b2, b3, c2, c3) illustrate how increasing measurement noise degrades the observed signal, highlighting the challenging conditions under which these robust estimators maintain good performance. 

In conclusion, the numerical results strongly support the theoretical findings. When dealing with experimental data potentially affected by non-negligible measurement noise ($\sigma^2 / \Delta t > \langle \bm{f}^2 \rangle$), employing estimators specifically designed for noise robustness, such as the Shift or Stratonovich methods presented here, is highly recommended over standard approaches based directly on the Approximate Maximum Likelihood (AML). Given its relative simplicity compared to the Stratonovich method, the shift estimator often presents a more practical and advantageous choice. 

\section{Conclusion}
\label{sec:chap3_conclusion}

This chapter addressed two significant practical challenges encountered when inferring Stochastic Differential Equation models from experimental time-series data: the presence of potentially large sampling intervals \( \Delta t \) and the unavoidable contamination by measurement noise \( \sigma \). We demonstrated that inference techniques based on the Approximate Maximum Likelihood (AML), while effective in the limit of high sampling frequency and low noise, suffer from systematic biases that degrade their accuracy under more realistic conditions. Specifically, we identified biases scaling as \( \mathcal{O}(\Delta t) \) due to finite sampling intervals and as \( \mathcal{O}(\sigma^2 / \Delta t) \) due to measurement noise.

To overcome these limitations, we developed and analyzed several enhanced estimation techniques targeting each specific challenge. First, addressing the large sampling interval challenge, we introduced the 'Trapeze' estimator (\cref{sec:chap3_large_dt}). By incorporating a higher-order approximation of the integrated drift term, this method exhibits significantly improved performance, with its dominant error scaling as \( \mathcal{O}(\Delta t^4) \) compared to \( \mathcal{O}(\Delta t^2) \) for the Approximate Maximum Likelihood (AML) approach, as confirmed by numerical benchmarks (\cref{fig:delta_t_benchmark}).

Second, to combat the detrimental effects of measurement noise, we proposed two distinct strategies (\cref{sec:chap3_measurement_noise}). The 'Shift' estimator (\cref{subsec:shifted_estimator}) mitigates the noise-induced bias by evaluating basis functions at a time point uncorrelated with the noise increment. The 'Stratonovich' estimator (\cref{subsubsec:averaging}) achieves robustness by averaging basis function evaluations across the interval and applying a specific correction term derived from the Itô-Stratonovich relationship, requiring a noise-robust estimate of the diffusion tensor. Numerical simulations (\cref{fig:measurement_noise_benchmark}) confirmed that both methods dramatically outperform the AML estimator in the presence of substantial measurement noise, effectively removing the leading-order bias. 

It is important to note that the solutions presented here were developed largely independently to address either the large \( \Delta t \) problem or the measurement noise problem. Creating estimators that are simultaneously robust to both challenges presents additional difficulties. Specifically, many techniques for correcting measurement noise implicitly assume that \( \Delta t \) is sufficiently small for certain approximations to hold. Developing practical and theoretically well-founded estimators that effectively handle both large sampling intervals and significant measurement noise concurrently remains an active area of research, potentially requiring different approaches or more complex correction schemes.

In summary, the estimators developed in this chapter provide valuable tools for robust SDE inference when facing common experimental constraints in isolation. By specifically targeting the biases introduced by discrete sampling and measurement noise, these methods extend the applicability and reliability of data-driven SDE modeling to a wider range of real-world dynamical systems under those specific conditions. 

\newpage

\begin{tcolorbox}[
    enhanced, 
    sharp corners, 
    boxrule=0.5pt, 
    colframe=black!75!white, 
    colback=white, 
    coltitle=black, 
    fonttitle=\bfseries, 
    title=Chapter Takeaways, 
    attach boxed title to top left={yshift=-0.1in, xshift=0.15in}, 
    boxed title style={ 
        colback=white, 
        sharp corners, 
        boxrule=0pt, 
        frame code={
            \draw[black!75!white, line width=0.5pt]
                ([yshift=-1pt]frame.south west) -- ([yshift=-1pt]frame.south east);
        }
    },
    boxsep=5pt, 
    left=5pt,
    right=5pt,
    top=12pt, 
    bottom=5pt
    ]
    \textbf{Context:} This chapter focused on improving SDE drift inference from discrete data when facing practical challenges: large sampling intervals ($\Delta t$) and measurement noise ($\sigma$). Approximate Maximum Likelihood (AML) estimators (\cref{eq:euler_estimator_recall}) exhibit significant biases under these conditions. We developed and tested methods to mitigate these issues.

    \medskip 

    \begin{itemize}
        \item \textbf{Challenge: Large Sampling Intervals ($\Delta t$):}
            \begin{itemize}
                \item AML estimator bias scales as $\mathcal{O}(\Delta t)$ (see \cref{fig:delta_t_benchmark} or \cref{appendix:mse_dt_calculation}).
                \item \textbf{Solution: Trapeze Estimator} (\cref{eq:alpha_Tr_estimator}) uses a higher-order drift approximation. 
                    \[ \hat{\bm{\alpha}}^{Tr} = (\bm{G}^{Tr})^{-1} \bm{Y} \]
                    \begin{equation*} 
                        (\bm{G}^{Tr})_{ij} = \avg{ \bm{b}_i(\bm{x}_t)^\top (4\bar{\bm{D}})^{-1} \frac{\bm{b_j}(\bm{x}_{t}) + \bm{b_j}(\bm{x}_{t+\Delta t})}{2}}; \quad Y_i = \avg{\bm{b}_i(\bm{x}_t)^\top (4\bar{\bm{D}})^{-1} \frac{\Delta\bm{x}_t}{\Delta t}}
                    \end{equation*}
                \item \textit{Benefit:} Reduces bias, leading to improved error scaling (MSE $\propto \Delta t^4$ compared to $\Delta t^2$ for standard method). Validated in \cref{fig:delta_t_benchmark}. 
            \end{itemize}
        \item \textbf{Challenge: Measurement Noise ($\sigma$):}
            \begin{itemize}
                \item Standard estimator bias scales as $\mathcal{O}(\sigma^2 / \Delta t)$ due to noise correlations.
                \item \textbf{Solution 1: Shift Estimator} (\cref{eq:alpha_shift_estimator}).
                    \[ \hat{\bm{\alpha}}^{\text{shift}} = (\bm{G}^{\text{shift}})^{-1} \bm{Y}^{\text{shift}} \]
                    \begin{equation*} 
                        (\bm{G}^{\text{shift}})_{ij} = \avg{\bm{b}_j(\bm{y}_t) \cdot (4\bar{\bm{D}})^{-1}\bm{b}_i(\bm{y}_{t-\Delta t})}; \quad  Y_i^{\text{shift}} = \avg{\frac{\Delta \bm{y}_t}{\Delta t}\cdot (4\bar{\bm{D}})^{-1}\bm{b}_i(\bm{y}_{t-\Delta t})}
                    \end{equation*}
                \item \textbf{Solution 2: Stratonovich Estimator} (\cref{eq:alpha_strato_corrected}) uses $\frac{1}{2}(\bm{b}_i(\bm{y}_t) + \bm{b}_i(\bm{y}_{t+\Delta t}))$ instead of $\bm{b}_i(\bm{y}_t)$ in $\bm{Y}$ plus an Itô correction term (requires noise-robust $\hat{\bm{D}}_{\sigma}$, \cref{eq:D_vestergaard}).
                    \[ \hat{\bm{\alpha}}^{\text{Strato}} =  (\bm{G}^{\text{Strato}})^{-1} \left( \bm{Y}^{\text{Strato}} - \text{Correction}(\hat{\bm{D}}_{\sigma}) \right) \]
                \item Both noise-robust methods showed significantly improved performance over the standard method in benchmarks (\cref{fig:measurement_noise_benchmark}). 
            \end{itemize}
    \end{itemize}
\end{tcolorbox}

%% file: tex_body/chap2_d.tex
\chapter{Inferring Stochastic Partial Differential Equations}
\label{chap:spde_inference}
\chaptertoc{}

The field of Stochastic Partial Differential Equations (SPDEs) is relatively young, with significant developments occurring over the last few decades \cite{pardouxStochasticPartialDifferential2021}. As a notable example, Martin Hairer was awarded a Fields Medal in 2014 for "his outstanding contributions to the theory of stochastic partial differential equations" \cite{FieldsMedallists2014}. His lectures on the presentation of the basic theory of stochastic partial differential equations are freely available on arXiv \cite{hairerIntroductionStochasticPDEs2023}. In the following, we adopt a more heuristic approach, focusing on the inference of SPDEs from data.

\section{Introduction: From SDEs to Spatially Extended Systems}
\label{sec:chap4_intro}

The preceding chapters focused on inferring the dynamics of systems described by Stochastic Differential Equations (SDEs), which model the evolution of variables over time under the influence of noise. However, many complex phenomena encountered in physics, biology, finance, and engineering are characterized by fields that vary continuously in both space and time. Modeling such spatially extended systems necessitates the framework of Stochastic Partial Differential Equations (SPDEs).

An SPDE describes the evolution of a field, denoted as $u(\bm{x}, t)$, where $\bm{x} \in \mathcal{D} \subset \mathbb{R}^d$ represents the spatial coordinates within a domain $\mathcal{D}$ (with $d$ being the spatial dimension) and $t$ denotes time. A general form common to many SPDEs encountered in applications can be written as:
\begin{equation}
 \pdv{u(\bm{x}, t)}{t} = \mathcal{F}[u](\bm{x}, t) + \sqrt{2 D} \, \xi(\bm{x}, t)
 \label{eq:spde_general_vector}
\end{equation}
Here, $\mathcal{F}[u]$ is a (potentially nonlinear) differential operator acting on the field $u$. This operator encapsulates the deterministic part of the dynamics, which might include processes like diffusion (e.g., involving $\nabla^2 u$), advection (e.g., $\bm{v} \cdot \grad u$), and local reactions (e.g., $f(u)$). The term $\xi(\bm{x}, t)$ represents a spatio-temporal stochastic noise field, modeling unresolved degrees of freedom or external random influences. The parameter $D \ge 0$ scales the intensity of this noise.

The noise $\xi(\bm{x}, t)$ is frequently idealized as Gaussian white noise, uncorrelated in both space and time. This idealization is characterized by its covariance structure:
\begin{equation}
 \E{ \xi(\bm{x}', t') \xi(\bm{x}, t) } = \delta(\bm{x} - \bm{x}') \delta(t - t')
 \label{eq:noise_correlation_physicist}
\end{equation}
where $\E{\cdot}$ denotes the expectation value, and $\delta(\cdot)$ is the Dirac delta function. While mathematically abstract, this representation is convenient in many physical models.

The general form \cref{eq:spde_general_vector} encompasses many fundamental equations, including:
\begin{itemize}
   \item The Kardar-Parisi-Zhang (KPZ) equation, often used for modeling interface growth (1D field: $u(x,t)$) \cite{kardarDynamicScalingGrowing1986}.
   \item The stochastic Allen-Cahn equation for phase separation dynamics (typically 2D field: $u(x,y,t)$) \cite{funakiScalingLimitStochastic1995}.
   \item The stochastic Burgers equation \cite{dapratoStochasticBurgersEquation1994} and the stochastic Navier-Stokes equations \cite{flandoliDissipativityInvariantMeasures1994}, which are models for turbulence in fluid flow (often 2D or 3D fields). 
\end{itemize}

The central challenge addressed in this chapter is \textbf{SPDE inference}: determining the structure of the deterministic operator $\mathcal{F}$ from observations of the field $u(\bm{x}, t)$. Typically, these observations are discrete in both space (measured at locations $\bm{x}_i$) and time (sampled at times $t_k$), and may themselves be corrupted by measurement noise. This chapter extends the inference methodologies previously developed for SDEs to the more complex domain of SPDEs, primarily by leveraging spatial discretization techniques to transform the infinite-dimensional problem into a high-dimensional SDE system.

\section{Challenges Specific to SPDE Inference}
\label{sec:chap4_spde_challenges_vector}

Inferring SPDEs from data presents several practical challenges:

\begin{itemize}
    \item \textbf{High Dimensionality and Computational Cost:} The dimension of the discretized state vector \( \bm{u} \), denoted \( N \), corresponds to the number of spatial grid points and can be very large (e.g., millions for 3D simulations). This significantly increases the computational costs associated with storing data and calculating functions of $u$ (such as the vector basis functions \( \bm{b}_j(\bm{u}_t) \) that will be introduced later). 

    \item \textbf{Spatial Structure of Noise:} The basic framework often assumes spatially uncorrelated and uniform dynamical noise (i.e., driven by \( \xi(\bm{x},t) \) with constant \(D\)). Real-world systems, however, may exhibit complex spatial correlations or non-uniform intensity. Inferring such complex noise structures is undoubtedly relevant, though this question was not pursued in the present work. 

    \item \textbf{Boundary Conditions:} Learning the boundary conditions is crucial for accurately modeling the system and predicting its evolution. In the following analyses, we circumvent this difficulty by assuming known periodic boundary conditions. When dealing with real data where boundary conditions are unknown or complex, one might exclude data points near the boundary to mitigate edge effects, but this approach prevents learning the boundary conditions themselves.

    \item \textbf{Spatial Resolution ($\Delta x$):} Real-world data is typically available on a discrete spatial grid with spacing $\Delta x$. This raises the question of whether this resolution is adequate. This consideration is important because many processes modeled by SPDEs are, in fact, effective descriptions arising from coarse-graining underlying microscopic interactions. 
    Thus, with high-resolution data, it might be possible to coarse-grain further to determine if a larger effective grid size would be more appropriate. However, in most practical cases, the question is often the opposite: Is the available grid spacing $\Delta x$ too large (i.e., is the resolution insufficient)? This is frequently the scenario because while coarse-graining is possible, reducing the grid size (increasing resolution) post-acquisition is generally impossible.
    Determining the optimal $\Delta x$ or assessing the adequacy of a given $\Delta x$ remains a complex issue dependent on the specific system and available data.

    \item \textbf{Combined Challenges:} As for SDEs, addressing large time steps \( \Delta t \) and significant measurement noise \( \sigma \) is important since real data will typically be subjected to these two issues. 
    We will extend the previous corrections developed for SDEs to SPDEs. 
    Developing inference methods robust to both simultaneously is an ongoing research area for both SDEs and SPDEs.
\end{itemize}

Despite these challenges, several groups have addressed the question of SPDE inference from data. 
A comprehensive overview of methodologies for SPDE inference was provided by \cite{cialencoStatisticalInferenceSPDEs2018}. A significant portion of the existing literature, as highlighted in that review, focuses on the \textbf{spectral approach} \cite{hubnerTwoExamplesParameter1993, lototskyStatisticalInferenceStochastic2009, lototskyStochasticPartialDifferential2017}. This method is typically applied to SPDEs where the differential operators $\mathcal{F}[u]$ have a simple formulation in Fourier space. 
Within this framework, inference relies on observing the dynamics of a finite number $N$ of the solution's Fourier modes, $u_k(t)$, continuously over a time interval $[0, T]$. Statistical procedures, such as constructing Maximum Likelihood Estimators (MLEs) or using linear regression, are then used to learn coefficients by comparing data and model in Fourier space \cite{cialencoTrajectoryFittingEstimators2018}. 

More recently, research has shifted towards data-driven \textit{discovery} of SPDEs, aiming to learn both the structure and parameters from data, often focusing on parsimony. 
For instance, Mathpati et al. \cite{mathpatiDiscoveringStochasticPartial2024} introduce a framework combining an extended Euler-Maruyama expansion with sparse Bayesian learning. Their method to learn the coefficients is identical to the one developed here in \cref{sec:chap4_adapting_methods_vector}. This practical method relies on linear regression using $\Delta_t u$, a technique also employed in the context of PDE learning \cite{rudyDatadrivenDiscoveryPartial2017}.

\subsection{Noise: Mathematical Definition}
\label{sec:noise_math_numeric}

Rigorously, the noise term $\xi(\bm{x}, t) \dd{t}$ in \cref{eq:spde_general_vector} corresponds to the increment of a cylindrical Wiener process, which we denote as $\xi(\bm{x}, t) \dd{t}=\dd B(t, \bm{x})$. Formally, this process admits an expansion:
\begin{equation}
 B(t, \bm{x}) = \sum_{k=1}^{\infty} W_k(t) e_k(\bm{x})
 \label{eq:cylindrical_brownian_motion_expansion_repeat}
\end{equation}
where $\{e_k(\bm{x})\}$ form a complete orthonormal basis, and $\{W_k(t)\}$ are independent 1D Wiener processes (see \cref{sec:sde_toolbox_intro} for background).

When simulating or analyzing SPDEs numerically, spatial discretization is performed. The continuous field $u(\bm{x}, t)$ is approximated by a finite-dimensional vector $\bm{u}(t) = (u_1(t), u_2(t), ..., u_N(t))^{\top}$, where each component $u_i(t)$ represents the field's value at or associated with a specific grid point (or region) $\bm{x}_i$. The index $i$ typically runs over points on a 1, 2, or $d$-dimensional grid. This discretization transforms the infinite-dimensional SPDE \cref{eq:spde_general_vector} into a large system of coupled SDEs for the vector $\bm{u}(t)$:
\begin{equation}
 \dd{\bm{u}}(t) = \bm{F}(\bm{u}(t)) \dd{t} + \sqrt{2D} \, \bm{G} \, \dd{\bm{W}}(t)
 \label{eq:sde_system_general_discretized}
\end{equation}
Here, $\bm{F}$ is the discretized version of the deterministic operator $\mathcal{F}$. The noise term involves $\dd{\bm{W}}(t)$, a vector containing the increments of the underlying independent Wiener processes $W_k(t)$ from the expansion \cref{eq:cylindrical_brownian_motion_expansion_repeat}. The matrix $\bm{G}$ encodes how the spatial basis functions $\{e_k\}$ project onto the discrete grid representation.

Handling the infinite sum in \cref{eq:cylindrical_brownian_motion_expansion_repeat} within a finite-dimensional simulation involves two key aspects:

\begin{enumerate}
    \item \textbf{Truncation:} The infinite sum must be truncated to a finite number of terms, $M$. The truncated process is $B_M(t, \bm{x}) = \sum_{k=1}^{M} W_k(t) e_k(\bm{x})$. The choice of $M$ is critical and is often linked to the spatial discretization level $N$. For simulating standard spatio-temporal white noise ($\delta$-correlated in space), it is common practice to choose $M=N$, effectively associating one independent Wiener process $W_k(t)$ with each degree of freedom $u_k(t)$ (or grid point/cell) in the discretized system \cite[Ch. 13]{lordIntroductionComputationalStochastic2014}.

    \item \textbf{Projection onto the Discrete Grid / Basis:} The effect of the truncated noise $B_M(t, \bm{x})$ (or rather, its increments $\dd B_M$) on the discrete variables $u_i(t)$ depends on the specific discretization method and the choice of basis $\{e_k\}$. A common choice is to use the noise basis $e_i(x) =  \mathbb{1}_{\text{cell } i}(x)$ (where $\mathbb{1}$ is the indicator function) on the grid cell with volume $\Delta V$ (e.g., $\Delta V = \Delta x$ in 1D, $\Delta V = (\Delta x)^d$ in $d$ dimensions) \cite[Ch. 10.5]{lordIntroductionComputationalStochastic2014}.
    The corresponding noise term associated with $\dd{B_M}$ is $\frac{1}{\sqrt{\Delta V}} \dd{W_i}(t)$, where $W_i(t)$ is an independent Wiener process associated with grid point $i$. This scaling ensures the variance matches the $\delta$-correlation property of $\xi$ in the limit ($\Delta V \to 0$) \cite{hairerIntroductionStochasticPDEs2023}. In this case, the matrix $\bm{G}$ in \cref{eq:sde_system_general_discretized} is diagonal with entries $1/\sqrt{\Delta V}$.
\end{enumerate}

Thus, for a $d$-dimensional field discretized into $N$ grid points with uniform cell volume $\Delta V = (\Delta x)^d$, this leads to the SDE system:
\begin{equation}
 \dd{\bm{u}}(t) = \bm{F}(\bm{u}(t)) \dd{t} + \sqrt{\frac{2D}{\Delta V}} \, \dd{\bm{W}}(t)
 \label{eq:sde_system_discretized_fd}
\end{equation}
where $\dd{\bm{u}}(t)$ is the $N$-dimensional vector of state variable increments, $\dd{\bm{W}}(t)$ is an $N$-dimensional vector of independent Wiener increments, $\bm{F}$ represents the discretized deterministic operator (incorporating finite difference approximations of derivatives), and the scaling factor $\sqrt{1/\Delta V}$ arises from the normalization required for delta-correlated noise on a discrete grid.

\subsection{Spatial Discretization and the Resulting SDE System}
\label{sec:chap4_discretization_vector}

Real-world data are inherently discrete in both space and time. As discussed, spatial discretization transforms the SPDE into a high-dimensional SDE system, forming the basis for our inference approach. We will employ the Finite Difference discretization introduced previously.
Consider a spatial grid with points \( \bm{x}_i \), where \( i = 1, \dots, N \), and let \( u_i(t) \equiv u(\bm{x}_i, t) \) denote the field value at these points. The index $i$ enumerates the $N$ grid points, which can be arranged in a 1, 2, or $d$-dimensional spatial layout. This discretization effectively converts the single SPDE \cref{eq:spde_general_vector} into the high-dimensional system of coupled SDEs previously derived, \cref{eq:sde_system_discretized_fd}.

The spatial derivatives within the operator \( \mathcal{F} \) are then approximated using methods like finite differences. For example, using finite differences on a 1D grid with spacing \( \Delta x \), common approximations include:
\begin{align*}
 (\nabla u)_i &\approx \frac{u_{i+1}(t) - u_{i}(t)}{ \Delta x} \quad (\text{Forward difference for } \partial u / \partial x) \\
 (\nabla^2 u)_i &\approx \frac{u_{i+1}(t) - 2u_i(t) + u_{i-1}(t)}{(\Delta x)^2} \quad (\text{Central difference for } \partial^2 u / \partial x^2)
\end{align*}
Appropriate modifications are needed near boundaries depending on the boundary conditions; in the following, we will assume periodic boundary conditions. 

The inference problem thus becomes: given time series data \( \{\bm{u}_{t_k}\}_{k=1,\dots,N_t} \), representing snapshots of the field on the discrete grid, potentially subjected to measurement noise and sampled at intervals \( \Delta t = t_{k+1} - t_k \), can we estimate the functional form of the drift vector \( \bm{F}(\bm{u}) \) in the discretized system \cref{eq:sde_system_discretized_fd}?

\section{A Stochastic Gray-Scott Model}
\label{sec:gray_scott_spde}

The Gray-Scott model is one example of a reaction-diffusion system capable of generating complex spatial patterns, similar to other models like the Turing model. It involves two chemical species represented by concentration fields $u(\bm{x}, t)$ and $v(\bm{x}, t)$. Here, we consider a stochastic version by adding noise terms:
\begin{align} \partial_t u &= D_{u}\nabla^{2}u - u v^{2} + F(1 - u) + \sqrt{2 D} \xi_u(\bm{x}, t) \\ \partial_t v &= D_{v}\nabla^{2}v + u v^{2} - (F + k)v + \sqrt{2 D} \xi_v(\bm{x}, t)
\label{eq:gray_scott_spde}
\end{align}
where $\xi_u$ and $\xi_v$ are independent spatio-temporal Gaussian white noise fields with properties as defined in \cref{eq:noise_correlation_physicist}, $D_u, D_v$ are diffusion coefficients, $F, k$ are reaction parameters, and $D$ represents the noise intensity (assumed equal for both species here).
Depending on the parameters $D_u, D_v, F, k$, this model gives rise to a wide variety of patterns. An interactive simulation demonstrating the various patterns generated by non-stochastic Gray-Scott equations is available at \url{https://pmneila.github.io/jsexp/grayscott/} \cite{pmneilaPmneilaJsexp2025}. Readers are invited to explore the beautiful interactive simulations on this website. 

In the simulations presented later in this chapter, we use a spatial discretization with $\Delta x = 1$, a time step $\Delta t = 0.01$, on a $100 \times 100$ grid (total $N=10000$ points) with periodic boundary conditions. The parameters are set to $D_u=0.2097, D_v=0.105, F=0.029, k=0.057$, and noise intensity $D=0.001$.
When not specified otherwise, the total trajectory duration used for inference is $\tau=10$.

For the inference task demonstrated in \cref{fig:spde_euler_gray_scott}, two different basis sets are employed to represent the drift term $\bm{F}(\bm{u})$ where $\bm{u} = (u, v)^\top$. The first is the "exact" basis, denoted $\mathcal{B}^*$, which comprises the $n_{\mathcal{B}^*}=7$ vector functions corresponding precisely to the terms present in the true Gray-Scott drift equations (\cref{eq:gray_scott_spde}). The second is an "over-complete" basis, denoted $\mathcal{B}_0$, containing $n_{\mathcal{B}_0}=78$ candidate terms. This larger library includes the exact terms plus additional functions, such as higher-order polynomials in $u$ and $v$, and various spatial derivatives.

\section{Adapting Inference Techniques to the Discretized System}
\label{sec:chap4_adapting_methods_vector}

Since the discretized system \cref{eq:sde_system_discretized_fd} is a high-dimensional system of SDEs, we can adapt the inference methods developed in previous chapters for SDEs. 

\subsection{Baseline Inference: Approximate Maximum Likelihood (AML)}
\label{subsec:chap4_euler_vector}

We assume the true drift vector \( \bm{F}(\bm{u}) \) can be represented as a linear combination of known vector basis functions \( \bm{b}_{j}(\bm{u}) \):
\begin{equation}
    \bm{F}(\bm{u}) = \sum_{j=1}^{n_{\mathcal{B}}} \alpha_{j}^* \bm{b}_{j}(\bm{u}).
    \label{eq:chap4_basis_expansion_vector_only}
\end{equation}
Each vector basis function \( \bm{b}_j(\bm{u}) \) typically represents a specific operation (e.g., identity, square, a discretized spatial derivative like the finite-difference Laplacian) applied component-wise or locally across the spatial grid. The goal is to estimate the scalar coefficients \( \bm{\alpha}^* = [\alpha_1^*, \dots, \alpha_{n_{\mathcal{B}}}^*]^\top \).

Assuming the sampling interval $\Delta t$ is sufficiently small, the Euler-Maruyama approximation of \cref{eq:sde_system_discretized_fd} is:
\begin{equation}
\frac{\Delta \bm{u}_t}{\Delta t} \approx \sum_{j=1}^{n_{\mathcal{B}}} \alpha_{j} \bm{b}_{j}(\bm{u}_t) + \sqrt{\frac{2D}{\Delta V}} \frac{\Delta \bm{W}_t}{\Delta t}
\label{eq:spde_discretized_euler}
\end{equation}
where $\Delta \bm{u}_t = \bm{u}_{t+\Delta t} - \bm{u}_t$, $\Delta V = (\Delta x)^d$ is the volume of a grid cell (assuming a uniform grid in $d$ dimensions), and $\Delta \bm{W}_t = \sqrt{\Delta t} \bm{\mathcal{N}}(\bm{0}, \bm{I})$ represents the increment of the $N$-dimensional Wiener process over $\Delta t$. To find the coefficients \( \bm{\alpha} \), we perform a least-squares regression. We define the input vector \( \bm{Y} \) and the Gram matrix \( \bm{G} \) using vector dot products averaged over time \( \langle\dots\rangle_t \):
\begin{align}
 (G_{\mathcal{B}})_{kj} &= \left\langle \bm{b}_k\left(\bm{u}_t\right)^{\top} \bm{b}_j(\bm{u}_t) \right\rangle_t, \label{eq:G_spde_def_vector}\\
 Y_k &= \left\langle \bm{b}_k\left(\bm{u}_t\right)^{\top} \left(\frac{\Delta\bm{u}_t}{\Delta t}\right) \right\rangle_t. \label{eq:Y_spde_def_vector}
\end{align}
The standard vector dot product \( \bm{a}^{\top} \bm{b} = \sum_{i=1}^N a_i b_i \) sums contributions over all $N$ spatial grid points. The time average \( \langle\dots\rangle_t = \frac{1}{N_t} \sum_{k=1}^{N_t} \) averages these scalar dot products over the $N_t$ available time steps in the observed trajectory. The estimator for the coefficients is then found by solving the linear system \( \bm{Y} = \bm{G} \bm{\alpha} \):
\begin{equation}
    \hat{\bm{\alpha}} = (\bm{G})^{-1} \bm{Y}.
    \label{eq:alpha_euler_spde_vector}
\end{equation}
This approach is identical to the derivation of the Approximate Maximum Likelihood. Consequently, this estimator is referred to as the AML estimator.

As in the previous chapter, we define a normalized mean squared error (NMSE) for the drift estimate:
\begin{equation}
\mathcal{E}(\hat{\bm{F}}) = \left\langle (\hat{\bm{F}}_t - \bm{F}(\bm{u}_t))^\top (4\bar{D}_{\text{eff}})^{-1} (\hat{\bm{F}}_t - \bm{F}(\bm{u}_t)) \right\rangle_t
\end{equation}
where $\hat{\bm{F}}_t = \sum_{j=1}^{n_{\mathcal{B}}} \hat{\alpha}_j \bm{b}_j(\bm{u}_t)$ is the reconstructed drift vector at time $t$, and $\bar{D}_{\text{eff}} = \frac{1}{2\Delta t N} \langle (\Delta \bm{u}_t)^\top (\Delta \bm{u}_t) \rangle_t$ is the effective noise intensity on the grid.
Extending the calculation from \cref{chap:inference_langevin}, with a uniform noise intensity $D$ (scalar) and spatially uncorrelated noise increments leads to: 
\begin{equation}
    \E{\mathcal{E}(\hat{\bm{F}})} \approx \frac{n_\mathcal{B}}{2 \tau}
    \label{eq:expected_mse_spde}
\end{equation}

In this case, the ordinary least squares approach using the standard dot product in \cref{eq:G_spde_def_vector,eq:Y_spde_def_vector} is equivalent to generalized least squares with diffusion weighting, since the diffusion is a scalar. The normalization by $(4\bar{D}_{\text{eff}})^{-1}$ in the MSE definition serves to make the error measure dimensionless.

\begin{figure}[htbp]
    \centering
    \includegraphics[width=0.8\linewidth]{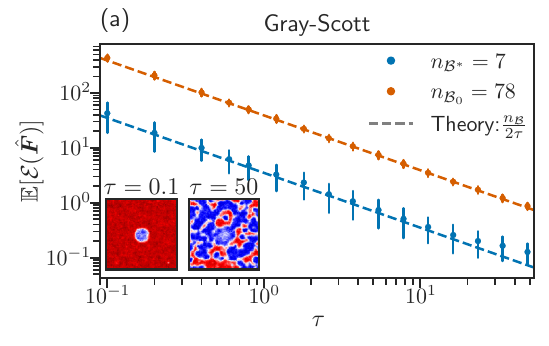}
    \caption{Log--log plot illustrating the decay of the expected normalized mean square drift error, $\mathbb{E}[\mathcal{E}(\hat{\bm{F}})]$, as a function of the trajectory length $\tau$ for the stochastic Gray-Scott system. Points with error bars represent the mean and standard deviation obtained from 100 simulation runs using the Approximate Maximum Likelihood (AML) inference method (\cref{eq:alpha_euler_spde_vector}), while dashed lines show the theoretical prediction $n_{\mathcal{B}}/(2\tau)$ (cf. \cref{eq:expected_mse_spde}). Results for two different sizes of the basis set are compared: $n_{\mathcal{B}^*}=7$ (blue) and $n_{\mathcal{B}_0}=78$ (orange).}
    \label{fig:spde_euler_gray_scott} 
\end{figure}

Figure \ref{fig:spde_euler_gray_scott} presents numerical results for inferring the drift of the Gray-Scott reaction-diffusion system using the Approximate Maximum Likelihood (AML) approach (\cref{eq:alpha_euler_spde_vector}). The plot shows the expected normalized mean squared error of the drift estimate, \( \mathbb{E}[\mathcal{E}(\hat{\bm{F}})] \), as a function of the total trajectory time \( \tau \). As expected from \cref{eq:expected_mse_spde}, the error decreases as \( \tau \) increases. The results align reasonably well with the theoretical prediction, indicated by the dashed line \( n_{\mathcal{B}}/(2\tau) \), confirming the convergence properties of the estimator.

\subsection{Robustness to Large Sampling Intervals ($\Delta t$)}
\label{subsec:chap4_large_dt_vector}

\begin{figure}[htbp]
    \centering
    \includegraphics[width=0.8\linewidth]{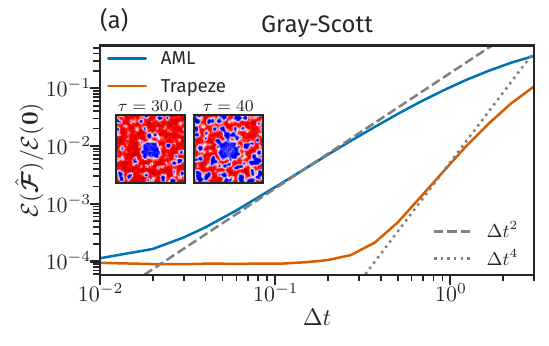} 
    \caption{Numerical benchmark comparing the relative error of SPDE inference for the Gray-Scott model using the Approximate Maximum Likelihood (AML) (\cref{eq:alpha_euler_spde_vector}) and Trapeze (\cref{eq:alpha_Tr_spde_estimator_vector}) estimators versus the sampling interval $\Delta t$ for the base $\mathcal{B}^*$. The plot demonstrates the improved error scaling for the Trapeze method (approximately $\Delta t^4$ dependence, potentially limited by the variance term at small $\Delta t$) compared to the Approximate Maximum Likelihood (AML) method (approximately $\Delta t^2$ dependence) for larger sampling intervals.} %
    \label{fig:spde_delta_t_benchmark_vector}
\end{figure}

For larger $\Delta t$, the Euler approximation \cref{eq:spde_discretized_euler} becomes less accurate. Using the trapeze rule to approximate the integral of the drift term, \( \int_t^{t+\Delta t} \bm{F}(\bm{u}_{t'}) dt' \approx \frac{\Delta t}{2} (\bm{F}(\bm{u}_t) + \bm{F}(\bm{u}_{t+\Delta t})) \), leads to the approximation:
\begin{equation}
 \frac{\Delta\bm{u}_t}{\Delta t} \approx \frac{\bm{F}(\bm{u}_{t}) + \bm{F}(\bm{u}_{t+\Delta t})}{2} + \text{noise term} 
\end{equation}
The more rigorous formulation that leads to this approximation relies on higher-order terms in the stochastic Taylor expansion (see \cref{appendix:mse_dt_calculation}).
Substituting the basis expansion \cref{eq:chap4_basis_expansion_vector_only} and projecting onto the basis functions (analogous to \cref{sec:chap3_large_dt} and \cref{eq:alpha_Tr_estimator} for scalar SDEs), we define the Trapeze estimator \( \hat{\bm{\alpha}}^{Tr} \). This involves modifying the Gram matrix using:
\begin{align}
 (\bm{G}^{Tr})_{kj} &= \left\langle \bm{b}_k\left(\bm{u}_t\right)^{\top} \left( \frac{\bm{b}_j(\bm{u}_t) + \bm{b}_j(\bm{u}_{t+\Delta t})}{2} \right) \right\rangle_t, \label{eq:G_Tr_spde_def_vector}\\
 \hat{\bm{\alpha}}^{Tr} &= (\bm{G}^{Tr})^{-1} \bm{Y}, \label{eq:alpha_Tr_spde_estimator_vector}
\end{align}
where \( \bm{Y} \) is the same input vector as in \cref{eq:Y_spde_def_vector}. This estimator is expected to exhibit improved error scaling with respect to the sampling interval (e.g., bias scaling as \( \mathcal{O}(\Delta t^2) \), leading to overall error \( \mathcal{O}(\Delta t^4) \)) compared to the baseline Euler method (\cref{eq:alpha_euler_spde_vector}), as demonstrated for the Gray-Scott SPDE in \cref{fig:spde_delta_t_benchmark_vector}.

\subsection{Robustness Against Measurement Noise ($\sigma$)}
\label{subsec:chap4_noise_vector}

Consider observations \( \bm{v}_t = \bm{u}_t + \bm{\eta}_t \), where \( \bm{\eta}_t \sim \mathcal{N}(\bm{0}, \sigma^2 \bm{I}) \) is additive Gaussian white measurement noise, assumed independent across time steps $t$ and spatial locations $i$, and also independent of the dynamical noise \( \bm{W}_t \). 

The core problem identified for SDEs persists: the term \( \langle \bm{b}_k(\bm{v}_t)^{\top} (\Delta\bm{v}_t / \Delta t) \rangle_t \) used to compute \( Y_k \) (\cref{eq:Y_spde_def_vector}) contains a bias term involving \( \langle \bm{b}_k(\bm{v}_t)^{\top} (\Delta\bm{\eta}_t / \Delta t) \rangle_t \). This bias arises from the correlation between \( \bm{b}_k(\bm{v}_t) \) (which depends on \( \bm{\eta}_t \)) and \( \Delta\bm{\eta}_t = \bm{\eta}_{t+\Delta t} - \bm{\eta}_t \). This bias scales as \( \mathcal{O}(\sigma^2 / \Delta t) \) and can dominate for small \( \Delta t \) or large \( \sigma \).

We adapt the mitigation strategies developed for SDEs to the vector formulation:

\textbf{1. Shift Estimator:} To eliminate the problematic correlation, evaluate the basis function \( \bm{b}_k \) involved in the correlation at a lagged time. Specifically, modify the definitions of $Y_k$ and $G_{kj}$:
\begin{align}
 Y_k^{\text{shift}} &= \left\langle \bm{b}_k(\bm{v}_{t-\Delta t})^{\top} \left(\frac{\Delta\bm{v}_t}{\Delta t}\right) \right\rangle_t, \label{eq:Y_lag_spde_def_vector}\\
 (G^{\text{shift}})_{kj} &= \left\langle \bm{b}_k(\bm{v}_{t-\Delta t})^{\top} \bm{b}_j(\bm{v}_t) \right\rangle_t, \label{eq:G_lag_spde_def_vector}\\
 \hat{\bm{\alpha}}^{\text{shift}} &= (G^{\text{shift}})^{-1} \bm{Y}^{\text{shift}}. \label{eq:alpha_shift_spde_estimator_vector}
\end{align}
Using \( \bm{b}_k(\bm{v}_{t-\Delta t}) \) in $Y_k$ breaks the correlation with $\Delta \bm{\eta}_t$. The definition for $G_{kj}$ uses a similar lag structure for consistency. This "Shift" estimator removes the leading-order \( \mathcal{O}(\sigma^2 / \Delta t) \) bias.

\textbf{2. Stratonovich-Corrected Estimator:} Analogous to the SDE case (Section~\ref{sec:chap3_measurement_noise}), a Stratonovich-type correction could potentially be derived for the SPDE system. This would involve adding correction terms related to the divergence of the basis functions, accounting for the noise structure. Specifically, it requires computing terms involving gradients of the vector basis functions \( \bm{b}_j(\bm{u}) \) with respect to the full $N$-dimensional state vector \( \bm{u} \). For local basis functions (e.g., depending only on $u_i$ and its neighbors) and simple noise structures (like the spatially uncorrelated noise assumed here), this might be computationally feasible, though intensive due to the high dimension $N$.
However, for more generic cases, such as non-local basis functions or spatially correlated dynamical noise ($D$ being a matrix or operator depending on $\bm{u}$), calculating these high-dimensional gradients becomes computationally prohibitive. Furthermore, as observed in Section~\ref{sec:chap3_comparison}, the Stratonovich estimator often yielded results similar to the simpler Shift estimator for the SDE benchmarks. Given the significant potential computational burden for potentially limited performance gain, we focus on the Shift estimator in the SPDE context.

\begin{figure}[htbp]
    \centering
    \includegraphics[width=0.8\linewidth]{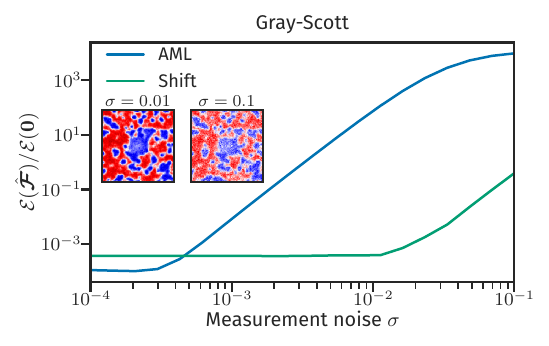} 
    \caption{Numerical benchmark for the Gray-Scott model comparing the relative inference error for the Approximate Maximum Likelihood (AML) (\cref{eq:alpha_euler_spde_vector}) and the Shift (\cref{eq:alpha_shift_spde_estimator_vector}) estimators as a function of measurement noise standard deviation $\sigma$. The Shift estimator demonstrates significant robustness against measurement noise compared to the standard approach, whose error increases substantially with $\sigma$.} 
    \label{fig:spde_noise_benchmark_vector}
\end{figure}

As demonstrated in \cref{fig:spde_noise_benchmark_vector}, the Shift estimator shows significantly improved robustness to measurement noise compared to the Approximate Maximum Likelihood (AML) approach (\cref{eq:alpha_euler_spde_vector}). However, the figure also reveals that the standard method exhibits slightly lower error for very small noise levels ($\sigma \to 0$). This occurs because the time lag introduced in the Shift estimator (\cref{eq:Y_lag_spde_def_vector,eq:G_lag_spde_def_vector}) inherently introduces a small discretization error related to approximating time derivatives and integrals, scaling with the sampling interval $\Delta t$. While this $\Delta t$-related error is present regardless of measurement noise, it becomes the dominant error source for the Shift estimator when $\sigma$ is very small. For the AML estimator, the $\mathcal{O}(\sigma^2/\Delta t)$ bias dominates quickly as $\sigma$ increases, making the Shift estimator preferable in noisy scenarios despite its small intrinsic lag-induced error.

\section{Conclusion}
\label{sec:chap4_conclusion}

This chapter successfully extended the inference methodologies developed for Stochastic Differential Equations (SDEs) in previous chapters to the significantly more complex domain of Stochastic Partial Differential Equations (SPDEs).

The primary strategy employed was spatial discretization, transforming the infinite-dimensional SPDE (\cref{eq:spde_general_vector}) into a high-dimensional system of coupled SDEs (\cref{eq:sde_system_discretized_fd}). This allowed us to adapt the basis function regression framework to estimate the discretized drift vector $\bm{F}(\bm{u})$. The baseline Approximate Maximum Likelihood (AML) estimator (\cref{eq:alpha_euler_spde_vector}) was introduced, leveraging vector dot products to average spatial information. Its performance was numerically validated using the stochastic Gray-Scott model (\cref{sec:gray_scott_spde}), demonstrating the expected convergence behaviour with increasing trajectory length $\tau$, consistent with theoretical predictions (\cref{fig:spde_euler_gray_scott}).

Acknowledging the practical challenges inherent in real-world data, we extended the techniques for handling large sampling intervals ($\Delta t$) and measurement noise ($\sigma$) to the SPDE context. The Trapeze estimator (\cref{eq:alpha_Tr_spde_estimator_vector}) demonstrated improved robustness to larger $\Delta t$, exhibiting better error scaling compared to the baseline method (\cref{fig:spde_delta_t_benchmark_vector}). Similarly, the Shift estimator (\cref{eq:alpha_shift_spde_estimator_vector}) proved effective in mitigating the bias introduced by additive measurement noise, showing significant error reduction compared to the standard approach in noisy regimes (\cref{fig:spde_noise_benchmark_vector}).

While this chapter successfully adapted SDE inference tools to handle temporal discretization and measurement noise issues in the context of discretized SPDEs, several challenges specific to spatially extended systems remain important considerations for future work. These include the computational burden associated with very high dimensionality ($N$), the inference of potentially complex spatial structures in the dynamical noise, the treatment of unknown or non-periodic boundary conditions, and assessing the adequacy of spatial resolution ($\Delta x$). The methods presented here provide a robust foundation for parameter estimation in discretized SPDEs, particularly when the underlying structure is assumed known or can be well-represented by a chosen basis set.

\newpage

\begin{tcolorbox}[
    enhanced, sharp corners, boxrule=0.5pt, colframe=black!75!white,
    colback=white, coltitle=black, fonttitle=\bfseries,
    title=Chapter Takeaways,
    attach boxed title to top left={yshift=-0.1in, xshift=0.15in},
    boxed title style={colback=white, sharp corners, boxrule=0pt,
        frame code={\draw[black!75!white, line width=0.5pt] ([yshift=-1pt]frame.south west) -- ([yshift=-1pt]frame.south east);}
    },
    boxsep=5pt, left=5pt, right=5pt, top=12pt, bottom=5pt,
    breakable 
    ]
    \textbf{Context:} Extended SDE inference to SPDEs:
    \[ \partial_t u = \mathcal{F}[u] + \sqrt{2D}\xi \]
    via spatial discretization of the equation, which leads to a high-dimensional SDE: 
    \[ d\bm{u} = \bm{F}(\bm{u}) dt + \sqrt{2D/\Delta V} d\bm{W}_t. \]
    Suppose that:
    \[ \bm{F}(\bm{u}) = \sum_j \alpha_j \bm{b}_{j}(\bm{u}). \]

    \medskip

    \begin{itemize}
        \item \textbf{SPDE Inference Framework:}  colon
            \begin{itemize}
                \item Baseline Approximate Maximum Likelihood (AML) Estimator (\cref{eq:alpha_euler_spde_vector}): Estimate \( \bm{\alpha} \) via
                \[ \hat{\bm{\alpha}}  = \bm{G}^{-1}\bm{Y}  \]
                with \( G_{kj} = \langle \bm{b}_k(\bm{u}_t)^{\top} \bm{b}_j(\bm{u}_t) \rangle_t \) and  \( Y_k = \langle \bm{b}_k(\bm{u}_t)^{\top} (\Delta\bm{u}_t / \Delta t) \rangle_t \).
                \item Expected NMSE: \( \mathbb{E}[\mathcal{E}(\hat{\bm{F}})] \approx n_{\mathcal{B}}/(2\tau) \) (\cref{eq:expected_mse_spde}). 
            \end{itemize}
        \item \textbf{Challenge: Large Sampling Intervals ($\Delta t$):}
            \begin{itemize}
                \item \textbf{Solution: Trapeze Estimator} (\cref{eq:alpha_Tr_spde_estimator_vector}): 
                \[ \hat{\bm{\alpha}}^{Tr}  = (\bm{G}^{Tr})^{-1} \bm{Y} \]
                with $ (\bm{G}^{Tr})_{kj} = \left\langle \bm{b}_k\left(\bm{u}_t\right)^{\top} \left( \frac{\bm{b}_j(\bm{u}_t) + \bm{b}_j(\bm{u}_{t+\Delta t})}{2} \right) \right\rangle_t$.
            \end{itemize}
        \item \textbf{Challenge: Measurement Noise ($\sigma$):} (Observed field \( \bm{v}_t = \bm{u}_t + \bm{\eta}_t \))
            \begin{itemize}
                \item \textbf{Solution: Shift Estimator} (\cref{eq:alpha_shift_spde_estimator_vector}): Uses lagged basis functions in calculations: 
                \[ \hat{\bm{\alpha}}^{\text{shift}} = (G^{\text{shift}})^{-1} \bm{Y}^{\text{shift}} \]
                with $Y_k^{\text{shift}} = \left\langle \bm{b}_k(\bm{v}_{t-\Delta t})^{\top} \left(\frac{\Delta\bm{v}_t}{\Delta t}\right) \right\rangle_t $ and $(\bm{G}^{\text{shift}})_{kj} = \left\langle \bm{b}_k(\bm{v}_{t-\Delta t})^{\top} \bm{b}_j(\bm{v}_t) \right\rangle_t$.
            \end{itemize}
    \end{itemize}
\end{tcolorbox}

%% file: tex_body/chap3.tex
\part{Model Selection in Stochastic Systems}
\label{part:Model_selection}

\chapter{Introduction to Model Selection in Stochastic Systems}

\epigraph{%
   ``With four parameters I can fit an elephant, and with five I can make him wiggle his trunk.''%
}{\textsc{Enrico Fermi}\autocite{dysonMeetingEnricoFermi2004}}

\chaptertoc{}
\label{chap:intro_model_selection}

Before delving into the specifics of model selection, I'd like to share a brief account of my PhD journey, which took a less direct route than initially envisioned. It began with the question: "How can we learn a Stochastic Partial Differential Equation (SPDE) from data?" However, after working on this question for about a year, my focus took a decisive turn towards a related but distinct problem: "How can we effectively compare and select between different Stochastic Differential Equation (SDE) or SPDE models?"

There were two main motivations for this shift. Firstly, I saw the challenges faced by Yirui Zhang, a postdoc working nearby, who was trying to compare different SDE models learned from cell nucleus trajectories \cite{amiriInferringGeometricalDynamics2024}. It seemed to me that a theoretically grounded framework for this task was missing. This resonated with a growing suspicion I had: it's remarkably easy to introduce superfluous parameters into a model.

My reasons for suspecting this trace back, perhaps unexpectedly, to my long nights in the YouTube era. While I mostly browsed for funny videos, I also followed two French science communicators who significantly shaped my interests during preparatory classes and after: the theoretical physics-oriented channel "Science Étonnante" and the mathematics-focused channel "El Ji."

\begin{figure}
    \centering
    \includegraphics[width=0.7\linewidth]{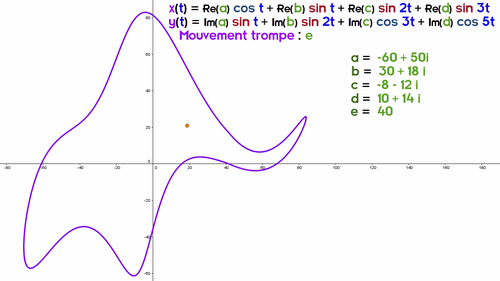}
    \caption{The ``Fermi/von Neumann elephant'' drawn using five parameters and Fourier series. Screenshot from the YouTube video ``Deux (deux ?) minutes pour l'éléphant de Fermi \& Neumann'' by El Ji, illustrating the concept of fitting complex shapes mathematically with a few parameters. \url{https://youtu.be/uazPP0ny3XQ?si=VMn0tiN3eUni5hwT&t=792}.} 
    \label{fig:Fermi_Neumann_elephant}
\end{figure}

So, what's the connection to model selection? When I was about 20, I watched an "El Ji" video titled "Deux (deux ?) minutes pour l'éléphant de Fermi \& Neumann." This video delved into the mathematical challenge sparked by the famous quote often attributed to Enrico Fermi regarding John von Neumann: "With four parameters I can fit an elephant, and with five I can make him wiggle his trunk” \cite{dysonMeetingEnricoFermi2004}. This isn't just an abstract thought experiment. As the video discussed, Mayer, Khairy, and Howard demonstrated in 2010 that you can indeed draw an elephant with just four complex parameters and make its trunk wiggle with a fifth! \cite{mayerDrawingElephantFour2010}. \Cref{fig:Fermi_Neumann_elephant} shows a similar rendering using five parameters, visually capturing the essence of this surprising feat.

While this story is amusing, it planted a crucial seed in my mind: if one can seemingly fit something as complex as an elephant with so few parameters, how much easier might it be to over-parameterize when modeling complex stochastic processes?

\section{Motivation}
\label{sec:intro_motivation}
Building upon the methods for parameter inference of a single Stochastic Differential Equation (SDE) developed in \cref{part:Learning_Langevin_equation} of this thesis, we now address a common and critical challenge in complex systems modeling: the structure of the governing equations themselves is often unknown. Specifically, while SDEs provide a powerful framework, determining the precise functional form of the drift term, $\bm{f}(\bm{x})$, which dictates the system's deterministic dynamics, directly from observational data remains a significant obstacle. 
This part of the thesis focuses on developing and validating a principled methodology for discovering parsimonious and accurate SDE models from time-series data, selecting the most appropriate drift structure from a potentially vast space of candidate functions.

\section{Context: Data-Driven Discovery for Dynamical Systems}
\label{sec:intro_context}
The field of data-driven discovery of governing equations has seen rapid advancements, particularly for deterministic systems described by Ordinary or Partial Differential Equations (ODEs/PDEs) \cite{northReviewDataDrivenDiscovery2023}. Techniques like symbolic regression \cite{schmidtDistillingFreeFormNatural2009,cranmerDiscoveringSymbolicModels2020, cranmerInterpretableMachineLearning2023}, Sparse Identification of Nonlinear Dynamics (SINDy) \cite{bruntonDiscoveringGoverningEquations2016,manganModelSelectionDynamical2017,zhengUnifiedFrameworkSparse2019,championUnifiedSparseOptimization2020,kaptanogluPySINDyComprehensivePython2022,kaptanogluBenchmarkingSparseSystem2023a}, Neural Network methods \cite{raissiPhysicsinformedNeuralNetworks2019,longPDENet20Learning2019,mccabeMultiplePhysicsPretraining2023}, group sparsity \cite{madduLearningPhysicallyConsistent2021, madduStabilitySelectionEnables2022}, or Bayesian methods \cite{northBayesianApproachDataDriven2022} provide tools for inferring equations directly from measurements. 

However, many real-world systems, from molecular processes to ecological dynamics and financial markets, exhibit significant stochasticity that cannot be ignored. Extending data-driven discovery to these systems, typically modeled by SDEs, presents unique challenges. 
While some approaches exist \cite{boninsegnaSparseLearningStochastic2018, gaoAutonomousInferenceComplex2022,callahamNonlinearStochasticModelling2021,huangSparseInferenceActive2022,nabeelDiscoveringStochasticDynamical2025}, they often lack rigorous theoretical grounding or rely on heuristics, particularly regarding the crucial step of model selection to avoid overfitting and ensure interpretability. This work aims to establish a robust statistical framework specifically for model selection in the context of SDE inference.

\section{Problem Formulation}
\label{sec:intro_problem}
We formulate the problem in the same terms as in the previous \cref{part:Learning_Langevin_equation}. 
In the following, to achieve a self-contained part, we recapitulate this formulation. 

We focus on autonomous, first-order SDEs describing Brownian dynamics, a broadly applicable model class:
\begin{equation}
    \dd{\bm{x}_t} = \bm{f}(\bm{x}_t) \dd{t}+ \sqrt{2\bm{D}(\bm{x}_t)} \dd{\bm{W}_t}
    \label{eq:Langevin_chapN}
\end{equation}
Here, $\bm{x}_t \in \mathbb{R}^d$ is the state vector, $\bm{f}(\bm{x})$ is the drift field representing deterministic forces, $\bm{D}(\bm{x})$ is the symmetric positive-definite diffusion matrix characterizing the state-dependent noise intensity, and $\dd{\bm{W}_t}$ is a $d$-dimensional Wiener process. 
We interpret multiplicative noise in the Itô sense. While inferring both $\bm{f}(\bm{x})$ and $\bm{D}(\bm{x})$ is important, we concentrate primarily on discovering the structure of the drift $\bm{f}(\bm{x})$, often the most physically informative component. 

Given an observed time series $\bm{X}_{\tau}=\{\bm{x}_t\}_{t=0,\Delta t, \dots, \tau}$, our goal is to find an inferred drift field $\hat{\bm{f}}(\bm{x})$ that accurately approximates the true $\bm{f}(\bm{x})$. 
We adopt the strategy of representing the drift as a linear combination of basis functions chosen from an initial large library $\mathcal{B}_0=\{\bm{b_i}(\bm{x})\}_{i=1,\dots,n_0}$: 
\begin{equation}
    \hat{\bm{f}}^{\mathcal{B}}(\bm{x}) = \sum_{i=1}^{n_\mathcal{B}} \hat{\alpha}_i^{\mathcal{B}} \bm{b}_i(\bm{x}), \quad \text{where } \mathcal{B} = \{\bm{b_i}(\bm{x})\}_{i=1,\dots,n_{\mathcal{B}}} \subseteq \mathcal{B}_0. 
    \label{eq:basis_chapN}
\end{equation}
The core problem thus becomes twofold:
\begin{enumerate}
    \item For a given basis set $\mathcal{B}$, estimate the coefficients $\hat{\alpha}_i^{\mathcal{B}}$.
    \item Select the "best" subset of basis functions $\mathcal{B}$ from the potentially large initial library $\mathcal{B}_0$ (with $n_0$ basis functions that can be large, e.g., $n_0 \approx 100$), balancing model accuracy with parsimony to avoid overfitting and enhance interpretability. 
\end{enumerate}
We previously addressed the first point in \cref{part:Learning_Langevin_equation} of this thesis. 
The second point, the principled selection of the basis $\mathcal{B}$, is the central focus of this part.

Furthermore, a central assumption underpins the framework developed in this part: we posit the existence of a "true" basis set, $\mathcal{B}^* \subset \mathcal{B}_0$, composed of $n_{\mathcal{B}^*}$ functions, such that the exact drift dynamics can be represented as:
\begin{equation}
    \bm{f}(\bm{x}) = \sum_{i=1}^{n_{\mathcal{B}^*}} \alpha^{\mathcal{B}^*}_i \bm{b}_i(\bm{x}), \quad \text{where } \bm{b_i}(\bm{x}) \in \mathcal{B}^* 
    \label{eq:basis_decompo_exact}
\end{equation}
This "exact decomposition" assumption is, admittedly, an idealization. For most real-world applications, it is likely violated, echoing George Box's famous maxim: ``All models are wrong'' \cite{boxScienceStatistics1976}. However, adopting this assumption provides a necessary theoretical starting point. From my perspective, tackling complex systems often benefits from beginning with a simplified, tractable framework---even if known to be imperfect---rather than immediately confronting the full, potentially intractable, complexity of reality.
This pragmatic stance must be balanced, of course. There is a limit to useful simplification, a boundary famously articulated by Albert Einstein's admonition: ``Everything should be made as simple as possible, but not simpler.'' Our goal, therefore, is to leverage the assumption in \cref{eq:basis_decompo_exact} to build effective model selection tools, while remaining mindful of its inherent limitations when applied to real data.

\section{Outline of \cref{part:Model_selection}}
\label{sec:intro_outline}
The subsequent chapters will develop our approach to SDE model selection:
\begin{itemize}
    \item \textbf{\Cref{chap:likelihood_aic}:} Details the likelihood-based framework for inferring coefficients and introduces standard model selection methods like AIC, BIC, CV, and SINDy, highlighting their limitations for sparse discovery from large libraries. 
    \item \textbf{\Cref{chap:pastis}:} Introduces our novel method, Parsimonious Stochastic Inference (PASTIS), which leverages extreme value theory to derive a statistically principled information criterion specifically designed for sparse selection.
    \item \textbf{\Cref{chap:validation}:} Presents comprehensive validation of PASTIS on various synthetic SDE benchmarks, comparing its performance against existing methods.
    \item \textbf{\Cref{chap:pastis_spde}:} Shows the applicability of PASTIS to SPDEs.
    \item \textbf{\Cref{chap:robustness_sde}:} Discusses the robustness of PASTIS to common data imperfections like finite sampling rates and measurement noise, presenting modifications to handle these issues.
    \item \textbf{\Cref{chap:part2_discussion_conclusion}:} Summarizes the contributions, discusses the advantages and limitations of PASTIS, and suggests directions for future research.
\end{itemize}

%% file: tex_body/chap4.tex
\chapter{Review of Standard Model Selection Methods for SDE Discovery}
\label{chap:likelihood_aic}
\chaptertoc{}

This chapter lays the groundwork for model selection by first summarizing how coefficients of an SDE model are inferred for a \textit{given} set of basis functions using a log-likelihood-based approach. This method was discussed in depth in \cref{part:Learning_Langevin_equation}. 
We then introduce the concept of overfitting in this context and discuss the Akaike Information Criterion (AIC) as a standard method for model comparison, ultimately highlighting its inadequacy for the task of sparse discovery from large function libraries. Then, we introduce the Bayesian Information Criterion (BIC) as another standard method for model comparison, which partially addresses the AIC's limitations. 
Finally, we review common computational approaches like Lasso regularization and the SINDy framework, evaluating their suitability for this task.

\section{Summary for Learning Stochastic Differential Equations} 
\label{sec:likelihood_sfi}
Assuming a specific basis $\mathcal{B}=\{\bm{b_i}(\bm{x})\}_{i=1..n_\mathcal{B}}$ has been chosen, we need to estimate the corresponding coefficients $\hat{\alpha}_i^\mathcal{B}$ in the drift expansion (\cref{eq:basis_chapN}). We follow the approach developed in \cref{part:Learning_Langevin_equation} of this manuscript, which is closely related to \cite{frishmanLearningForceFields2020}, based on maximizing an approximate log-likelihood function for the observed trajectory $\bm{X}_{\tau}$. 

The approximate log-likelihood of observing the trajectory $\bm{X}_{\tau}$ given the inferred drift field $\hat{\bm{f}}^{\mathcal{B}}$ is:
\begin{align}
    \ell(\hat{\bm{f}}^{\mathcal{B}}\mid\bm{X}_{\tau}) = -\frac{\tau}{4} \avg{\left(\frac{\Delta \bm{x}_t}{\Delta t} -  \hat{\bm{f}}^{\mathcal{B}}_t\right)\cdot\bm{\bar{D}}^{-1}\cdot \left(\frac{\Delta \bm{x}_t}{\Delta t} -  \hat{\bm{f}}^{\mathcal{B}}_t \right)}
    \label{eq:likelihood_chapN1}
\end{align}
where $\Delta \bm{x}_t \coloneqq \bm{x}_{t+\Delta t} - \bm{x}_t$, $ \hat{\bm{f}}^{\mathcal{B}}_t \coloneqq \hat{\bm{f}}^{\mathcal{B}}(\bm{x}_t)$, $\avg{\cdot} \coloneqq \frac{1}{\tau} \sum_t (\cdot) \Delta t$ denotes the trajectory average over the total time $\tau$. $\bm{\bar{D}} \coloneqq \frac{1}{2\Delta t} \avg{\Delta \bm{x_t} \otimes \Delta \bm{x_t}}$ serves as an empirical estimate of the (average) diffusion matrix (see \cref{sec:chap_2b_Diffusion_maximum_likelihood_estimator}). In the limit of small time steps ($\Delta t \to 0$) and additive noise ($\bm{D}(\bm{x})=\bm{D}$ constant), this expression corresponds to the true log-likelihood (up to constants that depend on $\bm{\bar{D}}$, see \cref{sec:chap_2b_Diffusion_maximum_likelihood_estimator}). 
While challenges exist for state-dependent diffusion, this formulation remains a practical starting point.

Substituting the basis expansion $\hat{\bm{f}}^{\mathcal{B}} \coloneqq \sum_{i=1}^{n_\mathcal{B}} \hat{\alpha}_i^{\mathcal{B}} \bm{b}_i(\bm{x})$ into \cref{eq:likelihood_chapN1} and maximizing with respect to the coefficients $\hat{\alpha}_i^{\mathcal{B}}$ yields:
\begin{equation}
\hat{\alpha}_i^\mathcal{B} \coloneqq \sum_{j=1}^{n_\mathcal{B}} ({G_\mathcal{B}^{-1}})_{ij} \avg{\frac{\Delta \bm{x}_t}{\Delta t}\cdot\bm{\bar{D}}^{-1}\cdot\bm{b_j}(\bm{x}_t)}
    \label{eq:SFI-Ito_chapN1}
\end{equation}
where $(G_{\mathcal{B}})_{ij} \coloneqq \avg{\bm{b}_i(\bm{x}_t)\cdot\bm{\bar{D}}^{-1}\cdot\bm{b_j}(\bm{x}_t)}$ is the Gram matrix associated with the basis $\mathcal{B}$ under the metric induced by $\bm{\bar{D}}^{-1}$ (detailed demonstration around \cref{eq:chap_2b_alpha_approx_mle}).

Once the coefficients $\hat{\alpha}_i^\mathcal{B}$ are determined, we could try to quantify the "goodness of fit" provided by the basis $\mathcal{B}$ using the log-likelihood evaluated with the reconstructed drift:
\begin{equation}
     \hat{\ell}(\mathcal{B}) \coloneqq \ell(\hat{\bm{f}}^{\mathcal{B}}|\bm{X}_{\tau})
    \label{eq:I_chapN1}
\end{equation}
Next, we will see that looking only at the log-likelihood will lead to selecting a model that contains irrelevant parameters. 

\section{The Challenge of Overfitting in Model Comparison}
\label{sec:likelihood_overfitting}
A naive approach to selecting the best basis $\mathcal{B}$ from the library $\mathcal{B}_0$ would be to simply choose the one maximizing the log-likelihood $\hat{\ell}(\mathcal{B})$. However, this is flawed because the same data $\bm{X}_{\tau}$ is used both to estimate the parameters $\hat{\alpha}_i^\mathcal{B}$ and to evaluate the log-likelihood $ \hat{\ell}(\mathcal{B})$. This reuse of data introduces a systematic bias that favors more complex models (larger $n_\mathcal{B}$), even if they merely fit the noise in the data rather than the underlying structure.

To understand this bias more formally, we relate the log-likelihood $ \hat{\ell}(\mathcal{B})$ to the normalized mean squared error between the true drift $\bm{f}$ and the inferred drift $\hat{\bm{f}}^\mathcal{B}$, defined as $\mathcal{E}(\hat{\bm{f}}^\mathcal{B}) = \frac{1}{4}\avg{(\bm{f} - \hat{\bm{f}}^\mathcal{B} )\cdot\bar{\bm{D}}^{-1}\cdot(\bm{f} - \hat{\bm{f}}^\mathcal{B} )}$. As derived in \cref{apdx:E_from_log_likelihood}, the expected difference in log-likelihood between two models based on bases $\mathcal{B}$ and $\mathcal{C}$ relates to the difference in their expected errors:
\begin{equation}
 \mathbb{E}[\hat{\ell}(\mathcal{C}) - \hat{\ell}(\mathcal{B})] \approx \tau \left(\E{\mathcal{E}(\hat{\bm{f}}^{\mathcal{B}})} - \E{\mathcal{E}(\hat{\bm{f}}^{\mathcal{C}})}\right) + n_{\mathcal{C}} - n_{\mathcal{B}} 
 \label{eq:error_chapN1}
\end{equation}
where $\mathbb{E}[\cdot]$ denotes expectation over trajectory realizations, and $n_{\mathcal{B}}$, $n_{\mathcal{C}}$ are the number of basis functions (parameters) in each model. The crucial terms $+n_{\mathcal{C}}$ and $-n_{\mathcal{B}}$ represent the overfitting bias: each parameter added contributes, on average, one unit to the estimated log-likelihood $\hat{\ell}$, regardless of whether it actually reduces the true error $\mathcal{E}$. Comparing models based solely on $\hat{\ell}$ will therefore systematically favor models with more parameters.

\begin{figure}[htbp]
    \centering
    \includegraphics[width=\linewidth]{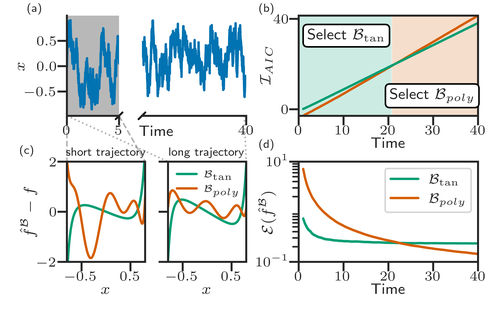} 
    \caption{\textbf{AIC performance in selecting drift models for a 1D system.} (a) Example trajectory from the system with drift $f(x)=-x/(1-x^2)^2$, diffusion $D=0.4$, sampled at $\Delta t=0.001$ (integration step $dt=0.0001$). (b) Average AIC (100 runs) vs. trajectory duration $\tau$ for a 1-parameter model ($\mathcal{B}_{\tan}=\{\tan(x)\}$) and a 9-parameter polynomial model ($\mathcal{B}_{\mathrm{poly}} = \{x^k\}_{k=0..8}$). (c) Drift error $f(x) - \hat{f}(x)$ vs. position $x$ for both models, comparing inferences from short (left) and long (right) trajectories. (d) Average Mean Squared Error ($\mathcal{E}$) of drift reconstruction vs. trajectory duration $\tau$ (100 runs), correlating with the AIC trends in (b).}
    \label{fig:AIC_exemple_chapN1}
\end{figure}

\section{AIC: Akaike's Information Criterion}
\label{sec:likelihood_aic_criterion}
To counteract the overfitting bias identified in \cref{eq:error_chapN1}, standard statistical practice suggests penalizing the log-likelihood based on the number of parameters. The Akaike Information Criterion (AIC) \cite{akaikeNewLookStatistical1974} provides one such correction. It can be defined as:
\begin{equation}
\mathcal{I}_{\mathrm{AIC}}(\mathcal{B}) = \hat{\ell}(\mathcal{B}) - n_\mathcal{B}.
\label{eq:aic_def}
\end{equation}
Based on \cref{eq:error_chapN1}, maximizing $\mathcal{I}_{\mathrm{AIC}}$ is, on average, equivalent to minimizing the true inference error $\mathcal{E}$. Note that, in most textbooks, AIC is defined as $-2\hat{\ell}(\mathcal{B}) + 2n_\mathcal{B}$. Here, for simplicity, we choose to omit the prefactor -2. 
Thus, in this manuscript, we will talk about \textbf{maximizing} the information criterion, whereas in most textbooks the goal is to minimize the information criterion. 

Since its introduction by Akaike in 1974, AIC has become a routine tool across disciplines—from phylogenetic model choice \cite{posadaJModelTestPhylogeneticModel2008,darribaModelTestNGNewScalable2020} and ecological inference \cite{johnsonModelSelectionEcology2004} to cosmological model comparison (coupled with BIC, which will be presented later) \cite{liddleInformationCriteriaAstrophysical2007} and cognitive-model evaluation \cite{wagenmakersAICModelSelection2004}.

AIC works well when comparing a small, pre-defined set of candidate models. \Cref{fig:AIC_exemple_chapN1} illustrates this with a one-dimensional toy model. To illustrate this point, we simulate a one-dimensional trajectory of a particle following an SDE with a non-linear drift $f(x)=\frac{-x}{(1-x)^2}$ and an additive noise $\sqrt{2D}\xi_t$ with $D=0.4$. 
To approximate this drift, we use two different bases: a simple, single-parameter basis ($\mathcal{B}_{\tan} = \{\tan\}$) and a more complex, multi-parameter polynomial basis ($\mathcal{B}_{poly} = \{x^k\}_{k=0..8}$).

As shown, for short trajectories (low data), the simpler model has a higher $\mathcal{I}_{\mathrm{AIC}}$ because the complex model overfits. For long trajectories (high data), the complex model provides a better description and achieves a higher $\mathcal{I}_{\mathrm{AIC}}$. Crucially, the crossover point in $\mathcal{I}_{\mathrm{AIC}}$ aligns with the crossover point in the true error $\mathcal{E}$ (\cref{fig:AIC_exemple_chapN1}b, d). Thus, selecting the model with the highest $\mathcal{I}_{\mathrm{AIC}}$ effectively minimizes the inference error when comparing these two specific models.

\subsection{Link between AIC and evaluation of the log-likelihood on a testing set}
\label{sec:link_AIC_testing_set}

Previously, we related the normalized mean squared error to the log-likelihood $\hat{\ell}(\mathcal{B})$, which was evaluated using the \textit{same} data $\bm{X}_{\tau}$ employed for learning the drift $\hat{\bm{f}}^{\mathcal{B}}$ (see \cref{eq:error_chapN1}). Now, let's consider the scenario where we use separate datasets: a training set $\bm{X}_{1,\tau}$ to estimate the drift $\hat{\bm{f}}^{\mathcal{B}}_{\bm{X}_{1,\tau}}$, and an independent testing set $\bm{X}_{2,\tau}$ to evaluate its performance via the log-likelihood $\ell(\hat{\bm{f}}^{\mathcal{B}}_{{\bm{X}_{1,\tau}}}|\bm{X}_{2,\tau})$. The expected value of this out-of-sample log-likelihood can be related to the expected normalized mean squared error:
\begin{align}
    \mathbb{E}[\ell(\hat{\bm{f}}^{\mathcal{B}}_{{\bm{X}_{1,\tau}}}|\bm{X}_{2,\tau})]
    &= -\frac{\tau}{4} \mathbb{E}\left[\avg{\left(\frac{\Delta \bm{x}_t}{\Delta t} -  \hat{\bm{f}}^{\mathcal{B}}_{\bm{X}_{1,\tau}}\right)\cdot\bm{\bar{D}}^{-1}\cdot \left(\frac{\Delta \bm{x}_t}{\Delta t} - \hat{\bm{f}}^{\mathcal{B}}_{\bm{X}_{1,\tau}}\right)}_{\bm{X}_{2,\tau}}\right] \nonumber\\
    &\approx -\frac{\tau}{4} \mathbb{E}\left[\avg{\left(\bm{f} - \hat{\bm{f}}^{\mathcal{B}}_{\bm{X}_{1,\tau}}\right)\cdot\bm{D}^{-1}\cdot \left(\bm{f} - \hat{\bm{f}}^{\mathcal{B}}_{\bm{X}_{1,\tau}}\right)}_{\bm{X}_{2,\tau}}\right] - \frac{d \tau}{2 \Delta t} \quad (\text{using } \bm{\bar{D}} \approx \bm{D}) \nonumber \\ 
    &= -\tau \, \mathbb{E}[\mathcal{E}(\hat{\bm{f}}^{\mathcal{B}}_{\bm{X}_{1,\tau}})] - \frac{d \tau}{2 \Delta t}
    \label{eq:link_aic_error}
\end{align}
where $d$ is the dimension of $\bm{x_t}$ and we used that the expectation of cross-terms involving noise $\dd{\bm{W}_t}$ and the estimated drift $\hat{\bm{f}}^{\mathcal{B}}_{\bm{X}_{1,\tau}}$ (trained independently) is zero when averaged over the testing set $\bm{X}_{2,\tau}$. 
We also assumed that the expected error $\mathbb{E}[\mathcal{E}(\cdot)]$ is statistically similar whether averaged over $\bm{X}_{1,\tau}$ or $\bm{X}_{2,\tau}$. The term $\frac{d \tau}{2 \Delta t}$ represents the contribution from the noise process squared, which is constant across different drift models $\hat{\bm{f}}^{\mathcal{B}}$.

\Cref{eq:link_aic_error} establishes that computing the log-likelihood on unseen (testing) data leads to the expected prediction error $\mathbb{E}[\mathcal{E}]$ (up to a constant term which has no impact when comparing AIC values among different models). 
This provides a deeper understanding of the Akaike Information Criterion (AIC): it acts as an estimator of this expected out-of-sample log-likelihood performance, thereby serving as a proxy for a model's generalization ability, correcting for the overfitting bias seen when using only the training data (as discussed around \cref{eq:error_chapN1}).

\begin{figure}[htbp]
    \centering
    \includegraphics[width=\linewidth]{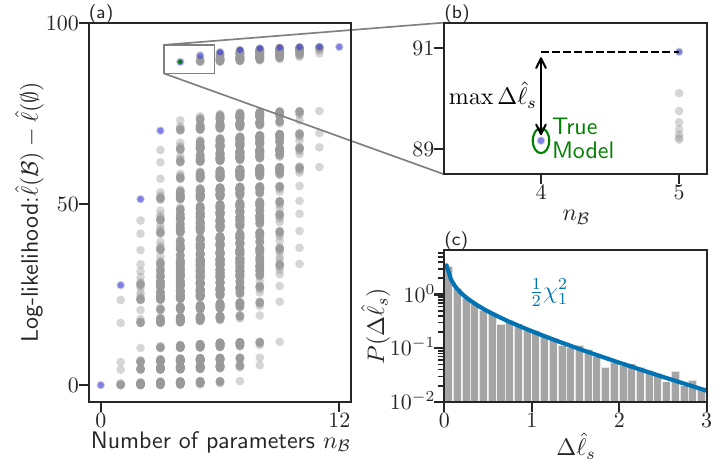}
    \caption{\textbf{Log-likelihood difference statistics for sparse models.}
    (a) Log-likelihood centered around the null model $\mathcal{B} = \emptyset$, $\hat{\ell}(\mathcal{B}) - \hat{l}(\emptyset)$, \emph{versus} the number of parameters $n_\mathcal{B}$ for sub-models $\mathcal{B}$ of a 3-dimensional Ornstein-Uhlenbeck process. The full model space potentially includes terms like $\{x_1 \bm{e_1}, \dots, x_3 \bm{e_3}\}$ (where $\bm{e_i}$ are basis vectors), and the true underlying model has $n^*=4$ non-zero parameters. Blue dots show the average $\hat{\ell}$ for the highest-likelihood model of size $n_\mathcal{B}$ across simulations, while gray dots represent lower-ranked models of the same size. 
    (b) Close-up of panel (a) focusing on the region near the true model, visualizing the expected maximum log-likelihood difference $\mathbb{E}[\Delta\hat{l}^{*}]$ against $n_\mathcal{B}$. The location of the true model $\mathcal{B}^*$ (labeled 'True Model') is indicated.
    (c) Probability distribution $P(\Delta\hat{l}_{s})$ for the log-likelihood difference $\Delta\hat{\ell}_s$ when testing models containing the true parameters plus one superfluous parameter. The exact distribution from simulations (histogram) is shown alongside the theoretical asymptotic $\frac{1}{2}\chi_{1}^{2}$ distribution (solid line).
    }
    \label{fig:Pastis_concept_abc_chapN1}
\end{figure}

\subsection{Limitations of AIC: Inconsistency in Sparse Selection}
\label{sec:likelihood_aic_limits}

While AIC provides an unbiased way to estimate the difference of error $\mathcal{E}$ between two models (\cref{eq:error_chapN1,sec:link_AIC_testing_set}), it exhibits a crucial limitation: \textit{AIC is not a consistent model selection criterion.} 
In statistical terms, consistency means that as the amount of data increases indefinitely, the criterion should select the true underlying model with probability approaching one. AIC does not satisfy this property when faced with overly complex candidate models \cite{lebarbierIntroductionAuCritere2006}. It tends to select models that contain more parameters than necessary, thus including superfluous parameters. 

To understand this inconsistency, let's analyze a fundamental scenario. Assume the true data-generating process is perfectly described by a model using a specific set of basis functions $\mathcal{B}^* \subset \mathcal{B}_0$. This $\mathcal{B}^*$ represents the most parsimonious correct model within our initial library $\mathcal{B}_0$. Now, consider comparing this true model $\mathcal{B}^*$ against an alternative model $\mathcal{B}^* \cup \{s\}$, which includes all the correct basis functions plus exactly one superfluous function $s \in \mathcal{B}_0 \setminus \mathcal{B}^*$.

According to the AIC criterion (\cref{eq:aic_def}), we would prefer the overly complex model $\mathcal{B}^* \cup \{s\}$ if its AIC value is higher than $\mathcal{I}_{\mathrm{AIC}}(\mathcal{B}^*)$:
\begin{equation*}
    \mathcal{I}_{\mathrm{AIC}}(\mathcal{B}^* \cup \{s\}) > \mathcal{I}_{\mathrm{AIC}}(\mathcal{B}^*)
\end{equation*}
Substituting the definition $\mathcal{I}_{\mathrm{AIC}}(\mathcal{B}) = \hat{\ell}(\mathcal{B}) - n_\mathcal{B}$, and noting that $n_{\mathcal{B}^* \cup \{s\}} = n_{\mathcal{B}^*} + 1$, this inequality becomes: 
\begin{align*}
    \Delta \hat{\ell}_s \coloneqq\hat{\ell}(\mathcal{B}^* \cup \{s\}) - \hat{\ell}(\mathcal{B}^*) &> 1
\end{align*}
Thus, AIC incorrectly selects the model containing the superfluous parameter $s$ whenever the increase in the maximized log-likelihood difference $\Delta \hat{\ell}_s$ exceeds the penalty difference, which is exactly 1. 
A graphical illustration of $\Delta \hat{\ell}_s$ is provided in \cref{fig:Pastis_concept_abc_chapN1}(b).

What is the probability of this occurring? The distribution of the log-likelihood difference $\Delta \hat{\ell}_s$, under the null hypothesis that the simpler model $\mathcal{B}^*$ is the true one, can be determined using Wilks' theorem \cite{wilksLargeSampleDistributionLikelihood1938}. This fundamental theorem states that, for nested models where the smaller model is true, twice the difference in their maximized log-likelihood values ($2 \Delta \hat{\ell}_s$) asymptotically follows a chi-squared ($\chi^2$) distribution. The degrees of freedom for this distribution equal the difference in the number of estimated parameters between the two models (which is 1 in our case). A detailed demonstration of Wilks' theorem, which I found to be more pedagogical than comparable explanations available online, is provided in \cref{annexe:WILKS}. Applying the theorem yields:
\begin{equation}
     2 \Delta \hat{\ell}_s \sim \chi_1^2 \quad \implies \quad \Delta \hat{\ell}_s \sim \frac{1}{2}\chi_1^2
    \label{eq:Wilks_chapN1}
\end{equation}
where $\chi_1^2$ denotes the chi-squared distribution with 1 degree of freedom. This theoretical result is empirically verified in \cref{fig:Pastis_concept_abc_chapN1}(c).

Using the $\frac{1}{2}\chi_1^2$ distribution, we can calculate the probability that AIC makes the wrong choice in this specific comparison:
\begin{equation}
 P\left(\text{AIC selects } \mathcal{B}^* \cup \{s\}\right) = P(\Delta \hat{\ell}_s > 1) = P(\chi_1^2 > 2) \approx 0.157
 \label{eq:aic_fail_prob}
\end{equation}
This result is critical: even when comparing the true parsimonious model to a model with just \emph{one} irrelevant parameter, there is a substantial, non-vanishing probability (approximately 16\%) that AIC will favor the larger, over-complete model. This occurs because the superfluous parameter can fit random noise in the data, leading to a spurious increase in likelihood that overcomes the fixed penalty of 1. This reasoning can be extended to any models with $k$ superfluous parameters by computing $P(\chi_k^2 > 2k)$. 
This demonstrates AIC's inconsistency – it does not guarantee convergence to the true model as data increases.

This inconsistency becomes severely problematic in the context of \emph{sparse model selection}. In sparse selection, the goal is not merely to compare two pre-defined models, but to identify the optimal, most parsimonious model $\mathcal{B}^*$ from a potentially very large initial library $\mathcal{B}_0$ containing $n_0$ candidate basis functions. This task involves implicitly comparing the true model $\mathcal{B}^*$ against numerous alternatives containing one or more superfluous terms from $\mathcal{B}_0 \setminus \mathcal{B}^*$.

When $n_0$ is large, there are many opportunities for irrelevant basis functions to randomly produce $\Delta \hat{\ell}_s > 1$. The probability calculated in \cref{eq:aic_fail_prob} applies, approximately, independently to each superfluous candidate term. Consequently, it becomes highly probable that \emph{at least one} of these irrelevant terms will satisfy the condition for inclusion by AIC, leading to the selection of an overly complex model.

Therefore, AIC is generally considered unsuitable for the task of identifying the true sparse structure from large candidate libraries, as it systematically tends to include irrelevant terms. This fundamental limitation motivates the need for alternative criteria with stronger penalties for model complexity, such as the Bayesian Information Criterion (BIC) discussed next. The practical failure of AIC in high-dimensional model comparison settings will be further illustrated through numerical examples in \cref{chap:validation}. 

\section{Cross-Validation: CV}
\label{sec:cross_validation}
Before defining BIC, let us examine another standard method for model comparison called Cross-Validation (CV). 

The overfitting problem highlighted in \cref{sec:likelihood_overfitting} arises because the model's performance is assessed on the same data used for its parameter estimation. A more robust approach to evaluate a model's predictive power is to test it on data it has not seen during training. Cross-Validation (CV) is a widely used set of techniques that formalizes this idea by systematically partitioning the data into training and testing (or validation) sets.

The general principle of CV involves:
\begin{enumerate}
    \item Splitting the available dataset into multiple, typically non-overlapping, segments or "folds".
    \item Iteratively using one fold as the test set and the remaining folds as the training set.
    \item For each iteration (fold), the model parameters (e.g., $\hat{\alpha}_i^\mathcal{B}$) are estimated using the training set.
    \item The performance of this fitted model is then evaluated on the test set, often using the log-likelihood in our context.
    \item The performance metrics from each fold are averaged to provide an overall assessment of the model's generalization capability.
\end{enumerate}
A common variant is $K$-fold Cross-Validation, where the data is divided into $K$ folds.

Let $\bm{X}_{\tau, \text{train}}^{(k)}$ be the training data for the $k$-th fold with $k  \in \llbracket 1, K \rrbracket$, and $\bm{X}_{\tau, \text{test}}^{(k)}$ be the corresponding test data. 
We first estimate the drift $\hat{\bm{f}}^{\mathcal{B}}_{(k)}$ using $\bm{X}_{\tau, \text{train}}^{(k)}$. Then, we evaluate its performance on the unseen test data $\bm{X}_{\tau, \text{test}}^{(k)}$ by computing the log-likelihood $\ell(\hat{\bm{f}}^{\mathcal{B}}_{(k)} | \bm{X}_{\tau, \text{test}}^{(k)})$. The CV score for model $\mathcal{B}$ is then the average over the $K$ folds:
\begin{equation}
    \ell_{\text{CV}}(\mathcal{B}) = \frac{1}{K} \sum_{k=1}^{K} \ell(\hat{\bm{f}}^{\mathcal{B}}_{(k)} | \bm{X}_{\tau, \text{test}}^{(k)})
    \label{eq:cv_score}
\end{equation}
This $\ell_{\text{CV}}(\mathcal{B})$ serves as an estimate of how well the model, when trained on a dataset of a certain size, is expected to perform on new, independent data. By selecting the model $\mathcal{B}$ that maximizes $\ell_{\text{CV}}(\mathcal{B})$, we aim to choose the model with the best generalization ability.

The key idea is that CV directly estimates the expected log-likelihood on a testing set. This provides a more reliable measure of model quality than the in-sample log-likelihood $\hat{\ell}(\mathcal{B})$, as it mitigates the optimistic bias that leads to overfitting. This objective is fundamentally the same as that of AIC, which, as demonstrated in \cref{sec:link_AIC_testing_set}, attempts to achieve this by correcting the in-sample log-likelihood (effectively by subtracting the bias $n_\mathcal{B}$). Therefore, CV will consequently face the same fundamental issue as AIC: it may not consistently select the true minimal model, particularly when faced with a large number of candidate functions.

\section{BIC: Bayesian Information Criterion}
\label{sec:likelihood_bic}

Four years after Akaike published his Akaike Information Criterion \cite{akaikeNewLookStatistical1974}, Schwarz introduced an alternative criterion based on Bayesian arguments \cite{schwarzEstimatingDimensionModel1978}, now widely known as the Bayesian Information Criterion (BIC). BIC provides a framework for model selection that has found broad application across diverse scientific domains. Examples include: guiding substitution-model selection in evolutionary biology and phylogenetics\,\cite{posadaJModelTestPhylogeneticModel2008,darribaModelTestNGNewScalable2020}; studying the genetic structure of natural populations\,\cite{jombartDiscriminantAnalysisPrincipal2010}; ranking competing brain-connectivity architectures in cognitive neuroscience\,\cite{pennyComparingDynamicCausal2012}; determining the number and shape of clusters in model-based machine learning algorithms\,\cite{fraleyEnhancedModelBasedClustering2003}; model selection in time-series econometrics and business forecasting\,\cite{hyndmanAutomaticTimeSeries2008}; and, combined with AIC, weighing alternative dark-energy scenarios in cosmological model comparisons\,\cite{liddleInformationCriteriaAstrophysical2007}.

The derivation of BIC stems from a Bayesian perspective on model comparison. As discussed in \cref{sec:mle_bayes_connection}, for a given model $\mathcal{B}$ with parameters $\bm{\alpha}$, Bayesian inference relies on the prior distribution $\Pi(\bm{\alpha} | \mathcal{B})$ over the parameters. A key quantity in Bayesian model selection is the model evidence (also known as the marginal likelihood), which quantifies the probability of observing the data $\bm{X}_\tau$ given the model $\mathcal{B}$, averaged over all possible parameter values according to their prior distribution:
\begin{equation}
   P(\bm{X}_{\tau}\mid \mathcal{B}) = \int P(\bm{X}_{\tau} \mid \mathcal{B}, \bm{\alpha}) \, \Pi(\bm{\alpha} | \mathcal{B}) \, \dd{\bm{\alpha}}
\end{equation}
The model evidence inherently penalizes model complexity; models with more parameters typically require the prior probability mass $\Pi(\bm{\alpha} | \mathcal{B})$ to be spread more thinly, reducing the average likelihood $P(\bm{X}_{\tau} \mid \mathcal{B}, \bm{\alpha})$ unless the extra parameters significantly improve the fit.

Evaluating this integral directly can be challenging. However, in the limit of large data ($\tau \to \infty$) and under certain regularity conditions, the logarithm of the model evidence can be approximated using a saddle-point method (specifically, the Laplace approximation). This approximation yields:
\begin{equation}
   \ln{P(\bm{X}_{\tau}\mid \mathcal{B})} \approx \hat{\ell}(\mathcal{B}) - \frac{n_{\mathcal{B}}}{2} \ln(\tau)
   \label{eq:bic_approximation}
\end{equation}
where $\hat{\ell}(\mathcal{B})$ is the maximized log-likelihood for model $\mathcal{B}$ (as defined in \cref{eq:I_chapN1}), $n_{\mathcal{B}}$ is the number of free parameters in model $\mathcal{B}$, and $\tau$ represents the amount of data (total trajectory time in our case). A detailed derivation of this approximation is provided in \cref{apdx:BIC}.

Based on this asymptotic result, the Bayesian Information Criterion ($\mathcal{I}_{\text{BIC}}$) is defined as:
\begin{equation}
   \mathcal{I}_{\text{BIC}}(\mathcal{B}) = \hat{\ell}(\mathcal{B}) - \frac{n_{\mathcal{B}}}{2} \ln(\tau)
   \label{eq:bic_def}
\end{equation}
In practice, the model $\mathcal{B}$ that maximizes $\mathcal{I}_{\text{BIC}}$ is selected. As for AIC, BIC is often defined in the literature with a prefactor $-2$. Here, we stick to our previous notation.

Within a fully Bayesian framework, one might aim to calculate the posterior probability of each model given the data, $P(\mathcal{B} | \bm{X}_{\tau})$. Using Bayes' theorem, and assuming a prior probability $P(\mathcal{B})$ over the models, the posterior is:
\begin{equation}
   P(\mathcal{B} |  \bm{X}_{\tau}) = \frac{P(\bm{X}_{\tau} | \mathcal{B}) P(\mathcal{B})}{P(\bm{X}_{\tau})} \propto P(\bm{X}_{\tau} | \mathcal{B}) P(\mathcal{B})
\end{equation}
Using the approximation from \cref{eq:bic_approximation}, we have $P(\bm{X}_{\tau} | \mathcal{B}) \approx \exp(\mathcal{I}_{\text{BIC}}(\mathcal{B}))$. Therefore:
\begin{equation}
   P(\mathcal{B} |  \bm{X}_{\tau}) \propto e^{\mathcal{I}_{\text{BIC}}(\mathcal{B})} P(\mathcal{B})
\end{equation}
If one assumes a uniform prior over the candidate models ($P(\mathcal{B}) = \text{constant}$), then selecting the model that maximizes $\mathcal{I}_{\text{BIC}}$ is equivalent to selecting the model with the highest approximate posterior probability.

A key feature of BIC is its penalty term, $\frac{n_\mathcal{B}}{2}\ln(\tau)$, which increases logarithmically with the amount of data $\tau$. This contrasts sharply with AIC's constant penalty ($n_\mathcal{B}$). For instance, whenever $\frac{1}{2}\ln(\tau) > 1$ (i.e., $\tau > e^2 \approx 7.4$), the BIC penalty per parameter is stronger than AIC's, and this relative penalty continues to grow as $\tau$ increases. This property makes BIC a consistent model selection criterion: given enough data, if the true data-generating model is among the candidates, BIC will select it with probability approaching one \cite{lebarbierIntroductionAuCritere2006}. This addresses the inconsistency issue highlighted for AIC in \cref{sec:likelihood_aic_limits}.

Despite its consistency, BIC is not without limitations. As we will explore empirically in \cref{chap:validation}, the convergence to the true model can be slow. Furthermore, the standard BIC penalty does not explicitly depend on the total number of candidate basis functions ($n_0$) in the initial library. This can be a drawback in high-dimensional sparse selection problems, where controlling the rate of false discoveries among a large number of potential parameters is crucial. These considerations motivate the development of methods specifically designed for sparse discovery, as discussed later (\cref{chap:pastis}).

\section{Lasso Regularization for Sparse Discovery}
\label{sec:lasso}

The information criteria discussed previously (AIC and BIC) and the CV method provide methods for model comparison based on statistical/Bayesian arguments. 
However, selecting the optimal model according to these criteria often requires evaluating a potentially vast number of candidate models ($\mathcal{B} \subset \mathcal{B}_0$). When the initial library $\mathcal{B}_0$ is large (large $n_0$), exhaustively searching through all $2^{n_0}$ possible sub-models becomes computationally infeasible. This combinatorial challenge arises implicitly because AIC and BIC use a penalty based on the number of non-zero parameters ($n_{\mathcal{B}}$), which corresponds to an $L_0$ pseudo-norm ($|\bm{\alpha}|_0 = n_{\mathcal{B}}$). Optimizing objectives involving the non-convex $L_0$ term directly is computationally hard.

In contrast, the Lasso (Least Absolute Shrinkage and Selection Operator) \cite{tibshiraniRegressionShrinkageSelection1996} offers a computationally tractable alternative that performs variable selection within a continuous optimization framework. While its theoretical guarantees for model selection differ from those of criteria like BIC, Lasso provides a practical method for inducing sparsity.

Instead of first selecting a subset of basis functions $\mathcal{B} \subset \mathcal{B}_0$ and then estimating coefficients, Lasso performs coefficient estimation and variable selection simultaneously. It starts with the objective of minimizing the negative approximate log-likelihood (equivalent to maximizing the likelihood, \cref{eq:likelihood_chapN1}). Ignoring constant factors ($\tau/4$), this corresponds to finding the coefficients $\bm{\alpha} = (\alpha_1, \dots, \alpha_{n_0})$ for the \textbf{full} library $\mathcal{B}_0$ that minimize the weighted least-squares error:
\begin{equation}
    \mathcal{L}(\bm{\alpha}) = \avg{\left(\frac{\Delta \bm{x}_t}{\Delta t} - \sum_{k=1}^{n_0} \alpha_k \bm{b}_k(\bm{x}_t)\right)\cdot\bm{\bar{D}}^{-1}\cdot \left(\frac{\Delta \bm{x}_t}{\Delta t} - \sum_{k=1}^{n_0} \alpha_k \bm{b}_k(\bm{x}_t) \right)}
    \label{eq:neg_log_likelihood_objective}
\end{equation}
where the sum runs over all $n_0$ basis functions in the initial library $\mathcal{B}_0$.

Lasso modifies this minimization problem by adding a penalty proportional to the $L_1$-norm of the coefficient vector $\bm{\alpha}$:
\begin{equation}
    \hat{\bm{\alpha}}_{\text{Lasso}} = \arg\min_{\bm{\alpha}} \left[ \mathcal{L}(\bm{\alpha}) + \lambda \sum_{k=1}^{n_0} |\alpha_k| \right]
    \label{eq:lasso_objective}
\end{equation}
Here, $\lambda \ge 0$ is the regularization parameter controlling the penalty strength. The crucial property of the $L_1$ penalty ($\sum |\alpha_k|$) is that, unlike the $L_2$ penalty ($\sum \alpha_k^2$) used in Ridge Regression, it performs automatic variable selection: for sufficiently large $\lambda$, it forces many coefficients $\alpha_k$ to be exactly zero. This yields a sparse model.

By varying $\lambda$, one controls the sparsity level of the resulting coefficient vector $\hat{\bm{\alpha}}_{\text{Lasso}}$. A larger $\lambda$ imposes a stronger penalty, shrinking more coefficients to zero. However, the Lasso method itself does not dictate the choice of $\lambda$. Its optimal value, balancing model fit and sparsity, is typically determined using external procedures such as cross-validation, or by employing information criteria like AIC or BIC adapted for the Lasso context \cite{zouDegreesFreedomLasso2007,usaiLASSOCrossvalidationGenomic2009,bhattacharyaLASSOpenalizedBICMixture2014,ninomiyaAICLassoGeneralized2016}.

From a Bayesian perspective, the Lasso estimate corresponds to the Maximum A Posteriori (MAP) estimate for $\bm{\alpha}$ under independent Laplace (double-exponential) prior distributions for each coefficient: $\Pi(\alpha_k) \propto \exp(-\frac{|\alpha_k|}{\sigma})$. The regularization parameter $\lambda$ is inversely related to the scale parameter $\sigma$ of the Laplace priors.

Information criteria like AIC and BIC provide methods for evaluating or selecting among candidate models. In contrast, penalized regression methods like Lasso aim to perform model selection simultaneously with parameter estimation, offering a different approach to achieving sparsity.

Regarding its application to discovering SDEs, benchmark tests conducted for this work across four different systems indicated limitations. Even when evaluating a wide range of $\lambda$ values, Lasso consistently failed to recover the exact underlying sparse model $\mathcal{B}^*$ in our simulations (see details in \cref{apdx:hyperparams_errors}, particularly \cref{fig:appendix_hyperparameters_total}). Furthermore, the notable success of alternative sparsity-promoting techniques like SINDy \cite{bruntonDiscoveringGoverningEquations2016} (to be introduced later) in the related field of ODE discovery might suggest that standard Lasso faces challenges in identifying dynamical systems accurately from data. Thus, we will not include Lasso in the main benchmark presented in the next chapter.

\section{SINDy: Sparse Identification of Nonlinear Dynamics}
\label{sec:sindy}

Building on the idea of leveraging sparsity for model discovery, the Sparse Identification of Nonlinear Dynamics (SINDy) framework was introduced by Brunton, Proctor, and Kutz \cite{bruntonDiscoveringGoverningEquations2016}. SINDy is specifically designed to identify governing equations directly from time-series data, under the assumption that the underlying dynamics, even if complex, can often be described by only a few dominant terms – embodying the principle of parsimony.

Originally developed for Ordinary Differential Equations (ODEs) of the form $\dot{\bm{x}} = \bm{f}(\bm{x})$, the SINDy algorithm typically proceeds as follows:

1.  \textbf{Data Acquisition and Derivative Estimation:} Collect time-series data $\bm{x}(t)$ for the state variables of interest. Numerically estimate the time derivatives $\dot{\bm{x}}(t)$ from the data (e.g., using finite differences or other differentiation techniques). Store the state variables and derivatives over time, often forming matrices $\mathbf{X} = [\bm{x}(t_1), \bm{x}(t_2), \dots]^{\top}$ and $\dot{\mathbf{X}} = [\dot{\bm{x}}(t_1), \dot{\bm{x}}(t_2), \dots]^{\top}$.

2.  \textbf{Candidate Function Library Construction:} Define a library $\mathcal{B}_0 = \{b_k(\bm{x})\}_{k=1..n_0}$ of candidate functions that might appear in the governing equations. This library can include polynomials (e.g., $1, x_1, x_2, x_1^2, x_1 x_2, \dots$), trigonometric functions, or other domain-specific functions. Evaluate these library functions on the measured data $\mathbf{X}$ to form a large matrix $\mathbf{\Theta}(\mathbf{X})$, where each column $\mathbf{\Theta}_k$ corresponds to the evaluation of a basis function $b_k$ at all time points.

3.  \textbf{Linear Regression Formulation:} Assume that the dynamics can be represented as a linear combination of the library functions:
    \begin{equation}
        \dot{\mathbf{X}} \approx \mathbf{\Theta}(\mathbf{X}) \mathbf{\Xi}
        \label{eq:sindy_linear_system}
    \end{equation}
    Here, $\mathbf{\Xi}$ is a matrix of coefficients, where each column $\bm{\xi}_j$ contains the coefficients multiplying the library functions $b_k$ to reconstruct the derivative of the $j$-th state variable, $\dot{x}_j$. That is, $\dot{x}_j(t) \approx \sum_{k=1}^{n_0} \xi_{kj} b_k(\bm{x}(t))$. Note that this standard formulation uses scalar basis functions $b_k(\bm{x})$; our framework, in contrast, employs vectorial basis functions $\bm{b}_k(\bm{x})$ (\cref{eq:neg_log_likelihood_objective}). The goal is to find a sparse coefficient matrix $\mathbf{\Xi}$.

4.  \textbf{Sparse Regression:} Solve the linear system (\cref{eq:sindy_linear_system}) for $\mathbf{\Xi}$ using a sparse regression technique. The original SINDy paper proposed the Sequentially Thresholded Least Squares (STLSQ) algorithm. STLSQ works iteratively:
    \begin{itemize}
        \item[(a)] Estimate coefficients $\mathbf{\Xi}$ using standard least-squares on \cref{eq:sindy_linear_system}.
        \item[(b)] Identify and set to zero all coefficients $\xi_{kj}$ whose magnitude is below a predefined threshold $\lambda_{\text{SINDy}}$.
        \item[(c)] Refit the least-squares problem using only the remaining (non-zero) candidate functions (columns of $\mathbf{\Theta}$).
        \item[(d)] Repeat steps (b) and (c) until the set of active coefficients converges.
    \end{itemize}
The final non-zero elements in $\hat{\mathbf{\Xi}}$ determine the structure of the identified dynamical system.

A key challenge within this framework is the selection of the threshold $\lambda_{\text{SINDy}}$, which must typically be chosen by the user. Subsequently, methods using AIC or BIC were proposed to automate the selection of $\lambda_{\text{SINDy}}$, although applying likelihood-based criteria directly to ODE data requires careful justification \cite{manganModelSelectionDynamical2017}.

SINDy has proven effective for discovering governing equations in various scientific and engineering domains, particularly for ODEs and PDEs. Its application to SDEs is an active area of research and provides another valuable benchmark (as discussed in this section, \cref{sec:sindy}) for assessing methods aimed at discovering sparse stochastic dynamics from data.

\section{Extending SINDy to SDEs: The SSR-CV Approach}
\label{sec:sindy_sde_ssr_cv}

While the original SINDy framework focused on deterministic systems, its core principle of combining function libraries with sparse regression is adaptable to Stochastic Differential Equations (SDEs). One notable attempt to extend SINDy for SDE discovery was proposed by Boninsegna et al. \cite{boninsegnaSparseLearningStochastic2018}. Their approach aims to identify a sparse representation for the SDE drift term, starting from an initial fit potentially based on minimizing an objective like $\mathcal{L}(\bm{\alpha})$ (\cref{eq:neg_log_likelihood_objective}) using the full library $\mathcal{B}_0$.

To achieve sparsity, Boninsegna et al. introduced a method combining Stepwise Sparse Regression (SSR) with $K$-fold Cross-Validation (CV): 

1.  \textbf{Initial Full Fit:} First, determine the coefficients $\bm{\alpha}$ using a standard linear regression that includes all $n_0$ candidate basis functions from the library $\mathcal{B}_0$.

2.  \textbf{Stepwise Sparse Regression (SSR):} This algorithm proceeds iteratively to prune the model:
    \begin{itemize}
        \item[(a)] Identify the coefficient with the smallest absolute magnitude among the currently active ones.
        \item[(b)] Set this coefficient permanently to zero, effectively removing the corresponding basis function from the model.
        \item[(c)] Re-estimate the remaining non-zero coefficients using regression with the reduced set of basis functions.
        \item[(d)] Repeat steps (a)-(c), generating a sequence of models with decreasing complexity (fewer non-zero coefficients).
    \end{itemize}

3.  \textbf{Cross-Validation (CV) for Model Selection:} The crucial challenge is selecting the optimal level of sparsity from the sequence of models generated by SSR. To address this, the authors employ $K$-fold CV.
For each model produced by SSR (characterized by its number of non-zero coefficients), a CV score is computed. This score typically reflects the model's predictive performance on held-out data segments. The CV score is plotted against the model size (number of active terms). Ideally, the score decreases as irrelevant terms are removed, reaches a minimum or plateau, and then increases sharply as essential terms are pruned (underfitting). The authors propose selecting the model size $n_{\mathcal{B}}$ ($\tilde{n}$ in the paper) corresponding to the point just before this "sharp increase," aiming for a balance between sparsity and predictive power.

Despite its logical structure, a central limitation of this SSR-CV approach is the ambiguity of its stopping criterion. The definition of the "sharp increase" in the CV score, used to select the final model size $n_{\mathcal{B}}$, lacks a precise, quantitative specification in the paper. What constitutes a "sharp" increase is subjective and potentially problem-dependent, making the selection process difficult to automate reliably and potentially introducing user bias. This ambiguity hinders reproducibility.

Furthermore, the unavailability of the original implementation code prevented direct benchmarking of this specific SSR-CV method against the approaches developed or evaluated in this thesis. Nevertheless, the work highlights the challenges and possibilities in adapting sparse regression techniques for SDE discovery.

\section{Conclusion}
\label{sec:likelihood_conclusion}

This chapter provided essential groundwork for model selection in the context of inferring Stochastic Differential Equation (SDE) drift terms from trajectory data. We began by recapping the likelihood-based framework (\cref{sec:likelihood_sfi}) used to estimate the coefficients of a drift expansion for a given set of basis functions $\mathcal{B}$. This established the log-likelihood $\hat{\ell}(\mathcal{B})$ as a fundamental measure of model fit.

However, we quickly identified the inherent challenge of overfitting (\cref{sec:likelihood_overfitting}): naively maximizing $\hat{\ell}(\mathcal{B})$ systematically favors overly complex models. To counteract this, we examined standard model selection techniques. The Akaike Information Criterion (AIC) (\cref{sec:likelihood_aic_criterion}) was introduced as a method designed to minimize prediction error, effectively providing an estimate of the likelihood performance on an independent testing set (\cref{sec:link_AIC_testing_set}). Despite its utility in comparing a small set of models, we demonstrated AIC's critical flaw for sparse discovery: its inconsistency (\cref{sec:likelihood_aic_limits}). Due to its small penalty per parameter, AIC retains a significant, non-vanishing probability of selecting models with superfluous terms, a problem exacerbated when searching among large libraries $\mathcal{B}_0$. 

The Bayesian Information Criterion (BIC) (\cref{sec:likelihood_bic}) was presented as a consistent alternative, employing a penalty that increases with data size ($\ln \tau$). While theoretically sound in the large data limit, BIC's convergence can be slow, and its penalty does not explicitly depend on the size $n_0$ of the initial candidate library, which is a key factor in controlling false discoveries in high-dimensional searches.

We then explored computational approaches tailored for sparsity. Lasso regularization (\cref{sec:lasso}) offers a practical method for simultaneous coefficient estimation and variable selection via an $L_1$ penalty, but its performance relies on tuning the regularization parameter $\lambda$, and it exhibited limitations in accurately recovering SDE models in our context. The SINDy framework (\cref{sec:sindy}), specifically designed for discovering dynamical systems, often employs heuristics like Sequential Thresholded Least Squares (STLSQ). While conceptually appealing and successful in many ODE/PDE applications, a previous adaptation to SDEs (e.g., via SSR-CV) introduces new challenges related to hyperparameter selection (thresholds) and algorithmic choices (ambiguous stopping criteria).

A common thread emerges from evaluating these standard approaches: while AIC/BIC provide statistical grounding and Lasso/SINDy offer computational tools for sparsity, they do not fully resolve the statistical challenge of multiple comparisons inherent in selecting a single parsimonious model from a large library $\mathcal{B}_0$. AIC is inconsistent; BIC lacks dependence on $n_0$; Lasso and SINDy variants often rely on hyperparameter tuning.

This gap motivates the need for an information criterion specifically designed to handle sparse selection from large libraries by rigorously accounting for the multiple comparisons problem. The following chapter, \cref{chap:pastis}, introduces precisely such a method: PASTIS (Parsimonious Stochastic Inference).
Leveraging insights from extreme value theory, PASTIS derives a penalty term that explicitly depends on the library size $n_0$, aiming to provide robust control over the probability of selecting superfluous terms.

%% file: tex_body/chap5.tex
\chapter{PASTIS: Parsimonious Selection via Extreme Value Statistics}
\label{chap:pastis}
\chaptertoc{}

\Cref{chap:likelihood_aic} established that while the Akaike Information Criterion (AIC) offers an unbiased estimate of predictive accuracy when comparing a few models, but it tends to overfit when trying to select a sparse model from a large pool of potential terms. 

This overfitting arises from the inherent multiple comparisons problem. This chapter introduces Parsimonious Stochastic Inference (PASTIS), an information criterion designed to mitigate this issue by leveraging principles from extreme value theory.

\section{Motivation: Handling Multiple Comparisons}
\label{sec:pastis_motivation}

A common model selection strategy involves maximizing an information criterion (e.g., AIC, BIC), which is typically based on the log-likelihood $\hat{\ell}$. Hence, when the information criterion is maximized, several different models are compared. Let's simplify this problem by focusing on comparing the true model (with basis $\mathcal{B}^*$) against alternative models that include single superfluous terms. When the pool of potential superfluous terms ($s \in \mathcal{B}_0 \setminus \mathcal{B}^*$) is large (size $N = n_0 - n^*$), the selection process must account for multiple comparisons. Consequently, it is insufficient to know only the distribution of a single log-likelihood gain $\Delta\hat\ell_s$ (as suggested by \cref{eq:Wilks_chapN1}); we must understand the distribution of the \emph{maximum} gain observed across all potential single superfluous terms, denoted by: 
\begin{equation}
\max_{s\in\mathcal B_0\setminus\mathcal B^*} \Delta\hat\ell_s,
\quad \text{where} \quad
\Delta\hat\ell_s = \hat\ell(\mathcal B^* \cup \{s\}) - \hat\ell(\mathcal B^*),
\end{equation}
where $\mathcal B^*$ denotes the true basis set and $s$ is a single superfluous basis function from the candidate set $\mathcal B_0$ but not in $\mathcal B^*$.

\section{Distribution of the Maximum Log-Likelihood Gain}
\label{sec:pastis_gumbel}

According to Wilks’ theorem (cf. \cref{eq:Wilks_chapN1}, \cref{fig:Pastis_concept_abc_chapN1} and the demonstration in \cref{annexe:WILKS}), each individual log-likelihood gain $\Delta\hat\ell_s$ for a single superfluous parameter converges in distribution to: 
\begin{equation}
    \Delta\hat\ell_s \sim \tfrac{1}{2}\chi^2_1
\end{equation}
as the sample size increases. When we study the maximum of the log-likelihood gain, $\max_{s\in\mathcal B_0\setminus\mathcal B^*} \Delta\hat\ell_s$, we are considering $N = n_0 - n^*$ superfluous terms. 
If we assume these gains $\Delta\hat\ell_s$ are approximately independent and identically distributed (i.i.d.)—an assumption reasonable if the basis functions are nearly orthogonal with respect to the data distribution—then finding the distribution of their maximum becomes a problem addressed by extreme value theory (EVT).

\begin{figure}[htbp] 
    \centering
    \includegraphics[width=\textwidth]{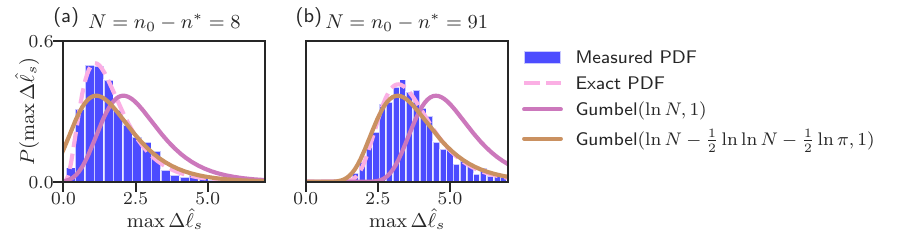} 
    \caption{Distribution of the maximum log-likelihood gain, $\max_{s\in\mathcal B_0\setminus\mathcal B^*} \Delta\hat\ell_s$, resulting from adding a single superfluous term. Panels compare the probability density functions (PDFs) measured from simulations (blue histograms) with various theoretical predictions for (a) a 3-dimensional Ornstein-Uhlenbeck process simulation with $N = n_0 - n^* = 8$ superfluous terms and (b) a 10-dimensional Ornstein-Uhlenbeck process simulation with $N = n_0 - n^* = 91$ terms. Theoretical curves include the exact PDF derived via differentiation of \cref{eq:exact_cumulative_pastis} (dashed pink line), the first-order Gumbel approximation $\text{Gumbel}(\mu=\ln N, \beta=1)$ corresponding to \cref{eq:approx_cumulative_pastis} (solid purple line), and a refined Gumbel approximation with location parameter $\mu_2 \approx \ln N - \frac{1}{2}\ln\ln N - \frac{1}{2}\ln\pi$ derived in \cref{myfootnote_erf_deriv} (solid ochre line).} 
    \label{fig:max_delta_l_dist}
\end{figure}

We previously saw that $\Delta\hat\ell_s \sim \tfrac{1}{2}\chi^2_1$. Thus, the Cumulative Density Function (CDF) of $\Delta\hat\ell_s$ is $P(\Delta\hat\ell_s \le z) = \erf(\sqrt{z})$ for $z \ge 0$. Under the i.i.d. assumption for the $\Delta\hat\ell_s$ terms, the exact CDF of their maximum is the product of the individual CDFs:
\begin{equation}
P\biggl[\max_{s\in\mathcal B_0\setminus\mathcal B^*} \Delta\hat\ell_s < z\biggr]
= \prod_{s\in\mathcal B_0\setminus\mathcal B^*} P\bigl[\Delta\hat\ell_s < z\bigr]
= \bigl(\erf(\sqrt{z})\bigr)^{N},
\label{eq:exact_cumulative_pastis}
\end{equation}
where $N = n_0 - n^* = |\mathcal{B}_0| - |\mathcal{B}^*|$ is the number of models with one superfluous term. The corresponding Probability Density Function (PDF), obtained by differentiation, is depicted by the dashed curve in \cref{fig:max_delta_l_dist}. Despite the underlying independence assumption, this theoretical PDF approximates the measured distribution very well. However, the exact formula \cref{eq:exact_cumulative_pastis} offers limited intuition regarding the underlying behavior, particularly the scaling with $N$. Therefore, we seek a simpler analytical approximation that provides the foundation for deriving the PASTIS criterion.

For large values of $z$, this CDF can be approximated using the asymptotic behavior of the error function\footnote{\label{myfootnote_erf} For large~$z$, recall that
\[
\erfc(\sqrt z)=1-\erf(\sqrt z)\sim \frac{e^{-z}}{\sqrt{\pi z}}
\quad(z\to\infty).
\]
Hence
\[
\erf(\sqrt z)
=1-\erfc(\sqrt z)
\approx1-\frac{e^{-z}}{\sqrt{\pi z}}
\approx\exp\!\Bigl(-\frac{e^{-z}}{\sqrt{\pi z}}\Bigr),
\]
where we used $1-\varepsilon\approx e^{-\varepsilon}$ for small~$\varepsilon$. Dropping the slowly-varying prefactor $(\pi z)^{-1/2}$ inside the exponential yields the simple “Gumbel-style” approximation 
\[
\erf(\sqrt z)\approx e^{-e^{-z}}.
\]
}:
\begin{equation}
P\biggl[\max_{s\in\mathcal B_0\setminus\mathcal B^*} \Delta\hat\ell_s < z\biggr]
\approx \bigl(e^{-e^{-z}}\bigr)^{N}
= \exp\bigl(-N e^{-z}\bigr)
= \exp\bigl(-e^{-z + \ln N}\bigr).
\label{eq:approx_cumulative_pastis}
\end{equation}
This is precisely the CDF of a Gumbel distribution ($\text{Gumbel}(\mu,\beta)$) with location parameter $\mu_1 = \ln N$ and scale parameter $\beta = 1$. This derivation carries a crucial implication: the expected maximum log-likelihood gain from adding a single potentially superfluous term, $\mathbb{E}\bigl[\max_{s\in\mathcal B_0\setminus\mathcal B^*} \Delta\hat\ell_s\bigr]$, does not approach a constant value but instead grows logarithmically with the number of candidate superfluous terms ($N$). Specifically, this first-order Gumbel approximation suggests $\mathbb{E}\bigl[\max_{s\in\mathcal B_0\setminus\mathcal B^*} \Delta\hat\ell_s\bigr] \approx \ln N + \eulergamma$, where $\eulergamma \approx 0.5772$ is the Euler–Mascheroni constant.

However, as shown by the purple curve in \cref{fig:max_delta_l_dist}, this first-order Gumbel approximation (with location $\mu_1 = \ln N$) does not perfectly match the measured PDF. This discrepancy arises from the crude approximation $\erf(\sqrt{z}) \approx e^{-e^{-z}}$ used to arrive at \cref{eq:approx_cumulative_pastis}. Using the more precise intermediate approximation $\erf(\sqrt{z}) \approx \exp(-\frac{e^{-z}}{\sqrt{\pi z}})$ (derived in \cref{myfootnote_erf}) leads to a refined Gumbel approximation. The location parameter $\mu_2$ for this refined distribution is derived as\footnote{\label{myfootnote_erf_deriv} We define $N=n_0 - n^*$. We look for the location $\mu_2$ of the Gumbel approximation:
\begin{equation}
 P\biggl[\max_{s\in\mathcal B_0\setminus\mathcal B^*} \Delta\hat\ell_s < z\biggr]
\approx (e^{-e^{-z}/\sqrt{\pi z}})^N = e^{-e^{-z + \mu_2}} 
\label{eq:G_gumbel_def_footnote} 
\end{equation}
By definition of $\mu_2$ from the previous equation, we have by taking $z=\mu_2$ and the log:
\begin{equation}
N\,\frac{e^{-\mu_2}}{\sqrt{\pi\,\mu_2}}\;=\;1 \implies \ln N \;-\;\mu_2 \;-\;\frac12\bigl(\ln\pi + \ln \mu_2\bigr)
\;=\;0.
\label{eq:log-condition}
\end{equation}
We now make the ansatz $\mu_2 \;=\;\ln N + u_N$ where \(u_N=o(\ln N)\). Plugging into \cref{eq:log-condition}:
\[
\ln N
-\bigl(\ln N + u_N\bigr)
-\frac12\Bigl[\ln\pi + \ln\!\bigl(\ln N + u_N\bigr)\Bigr]
=0
\;\Longrightarrow\;
-\,u_N
-\frac12\ln\pi
-\frac12\ln\!\bigl(\ln N + u_N\bigr)
=0.
\]
Since \(u_N=o(\ln N)\), expand
\(\ln(\ln N + u_N)=\ln\ln N + o(1)\).  Thus $ u_N= -\tfrac12\ln\pi - \tfrac12\ln\ln N + o(1)$. Hence
\[
\boxed{
\mu_2
=\ln N
- \tfrac12\ln\ln N
- \tfrac12\ln\pi
+ o(1).
}
\]
}
\[ \mu_2 \approx \ln N - \tfrac{1}{2}\ln(\ln N) - \tfrac{1}{2} \ln(\pi). \]
This refined Gumbel approximation, with location parameter $\mu_2 \approx \ln N - \tfrac{1}{2}\ln(\ln N) - \tfrac{1}{2} \ln(\pi)$, yields much better agreement with the measured PDF, as illustrated by the ochre curve in \cref{fig:max_delta_l_dist}. This demonstrates the validity of the EVT approach and the importance of higher-order correction terms for precise quantitative description. However, the most crucial insight for designing a robust model selection criterion comes from the \textbf{dominant} term in the location parameter: $\ln N$. This term reveals that the expected maximum log-likelihood gain achieved purely by chance when testing $N$ superfluous candidates does not saturate but grows logarithmically with $N$. This finding explains why fixed-penalty criteria like AIC tend to overfit in large libraries and mandates a penalty that scales with the library size. Recognizing that the $\ln N$ term captures this essential scaling behavior, we will use the insight from this simpler, first-order Gumbel approximation (location $\mu_1 = \ln N$) as the theoretical foundation for constructing the PASTIS penalty in the next section.

\section{The PASTIS Criterion}
\label{sec:pastis_criterion}

To avoid selecting models with superfluous terms, the penalty applied to each parameter in an information criterion should counteract the tendency to select terms just by chance due to multiple comparisons. As derived in \cref{sec:pastis_gumbel}, the maximum log-likelihood gain from a single superfluous term, $\max_{s\in\mathcal B_0\setminus\mathcal B^*} \Delta\hat\ell_s$, has a characteristic scale determined by its expected value, $\mathbb{E}\bigl[\max_{s\in\mathcal B_0\setminus\mathcal B^*} \Delta\hat\ell_s\bigr] \approx \ln(n_0 - n^*) + \gamma$. Therefore, the penalty per parameter should exceed this scale. How can we incorporate this insight into a practical information criterion?

As seen previously (e.g., with AIC, BIC), an information criterion $\mathcal{I}$ is often defined as the maximized log-likelihood $\hat{\ell}(\mathcal{B})$ minus a penalty term that depends on the number of parameters $n_\mathcal{B} = |\mathcal{B}|$ in the model basis $\mathcal{B}$. 
Let's consider a criterion of the form:
\[ \mathcal{I}(\mathcal{B}) = \hat{\ell}(\mathcal{B}) - n_\mathcal{B} \lambda, \]
where $\lambda$ is the penalty per parameter. The preferred model is the one that maximizes $\mathcal{I}$ among the candidates.

Let's analyze the comparison between the true model $\mathcal{B}^*$ and all models $\mathcal{B}_s = \mathcal{B}^* \cup \{s\}$ formed by adding a single superfluous parameter $s \in \mathcal{B}_0 \setminus \mathcal{B}^*$. 

Let $p$ be the probability of incorrectly choosing any of these single-superfluous-term models $\mathcal{B}_s$ instead of the true model $\mathcal{B}^*$ when using the criterion $\mathcal{I}$. An incorrect selection occurs if $\max_{s\in\mathcal B_0\setminus\mathcal B^*} \mathcal{I}(\mathcal{B}_s) > \mathcal{I}(\mathcal{B}^*)$. Thus, we can write the probability of correctly selecting the true model $\mathcal{B}^*$ as:

\begin{align}
   1 - p &= P\biggl[\max_{s\in\mathcal B_0\setminus\mathcal B^*} \bigl(\mathcal{I}(\mathcal{B}_s) - \mathcal{I}(\mathcal{B}^*)\bigr) \le 0\biggr] \nonumber \\
         &= P\biggl[\max_{s\in\mathcal B_0\setminus\mathcal B^*} \bigl( \Delta\hat\ell_s - \lambda (n_{\mathcal{B}_s} - n_{\mathcal{B}^*}) \bigr) \le 0\biggr] \nonumber \\
         &= P\biggl[\max_{s\in\mathcal B_0\setminus\mathcal B^*} \Delta\hat\ell_s \le \lambda \biggr], \label{eq:initial_equation_pastis}
\end{align}
since $n_{\mathcal{B}_s} - n_{\mathcal{B}^*}= |\mathcal{B}_s| - |\mathcal{B}^*| = 1$.

Now, we can use the Gumbel approximation for the CDF of the maximum log-likelihood gain, \cref{eq:approx_cumulative_pastis}, in \cref{eq:initial_equation_pastis}. Substituting the approximation into the probability expression gives:
\begin{equation}
1 - p \approx \exp\bigl(-e^{-\lambda + \ln(n_0-n^*)}\bigr).
\label{eq:derivation_pastis_gumbel_approximation}
\end{equation}
Solving this equation for $\lambda$ allows us to determine the penalty required to achieve the probability $p$ of selecting a model with one superfluous parameter, leading to $\lambda \approx \ln{\frac{n_0}{p}}$\footnote{Let $N = n_0 - n^*$. From \cref{eq:initial_equation_pastis} and the approximation \cref{eq:approx_cumulative_pastis}, we have:
\[ 1 - p \approx P\biggl[\max_{s\in\mathcal B_0\setminus\mathcal B^*} \Delta\hat\ell_s \le \lambda \biggr] \approx \exp\bigl(-e^{-\lambda + \ln N}\bigr) \]
Taking the natural logarithm of both sides and considering $p\ll 1$:
\[ \ln(1 - p) = -e^{-\lambda + \ln N} \implies p \approx e^{-\lambda + \ln N}\]
Taking the natural logarithm again:
\[ \ln{p} = -\lambda + \ln N \implies \lambda = \ln{\frac{N}{p}} \approx \ln{\frac{n_0}{p}} \quad \text{for } n_0 \gg n^*.\] 
QED. }. Hence, we define the Parsimonious Stochastic Inference (PASTIS) criterion as: 
\begin{equation}
\mathcal{I}_{\text{PASTIS}}(\mathcal{B}) = \hat{\ell}(\mathcal{B}) - n_\mathcal{B} \ln \frac{n_0}{p}
\label{eq:PASTIS_chapN2}
\end{equation}
where $\hat{\ell}(\mathcal{B})$ is the maximized log-likelihood for the model basis $\mathcal{B}$, $n_\mathcal{B} = |\mathcal{B}|$ is the number of parameters (basis functions) in $\mathcal{B}$, $n_0 = |\mathcal{B}_0|$ is the total number of candidate basis functions in the initial library, and $p$ is a user-chosen small parameter that corresponds approximately to the probability of selecting a model with superfluous parameters. The model chosen by PASTIS corresponds to the one maximizing $\mathcal{I}_{\text{PASTIS}}$.

The structure of the penalty term, $n_\mathcal{B} \ln(n_0/p)$, combines two key aspects. The factor $n_\mathcal{B}$ imposes a cost proportional to the model complexity (number of terms). The term $\ln(n_0/p)$ serves as the penalty per parameter, incorporating both the size of the search space ($n_0$) and a user-defined significance level ($p$). This penalty magnitude is motivated by the extreme value theory analysis (\cref{sec:pastis_gumbel}), which characterizes the maximum spurious log-likelihood gain expected when comparing many candidate terms.

\subsection*{Extending the argument for models with several superfluous terms}

The primary derivation determined the penalty $\lambda = \ln(n_0/p)$ by considering the risk of adding a single superfluous term ($k=1$). We now aim to show, following a similar line of reasoning albeit with stronger assumptions, that the PASTIS criterion $\mathcal{I}_{\text{PASTIS}}(\mathcal{B}) = \hat{\ell}(\mathcal{B}) - n_\mathcal{B} \ln (n_0/p)$ remains appropriate when considering models with $k = n_\mathcal{B} - n^* > 1$ superfluous terms.

Consider candidate models $\mathcal{B}$ containing $k$ superfluous terms added to the true basis $\mathcal{B}^*$. Under Wilks' theorem, the log-likelihood gain $\Delta\hat{\ell} = \hat{\ell}(\mathcal{B}) - \hat{\ell}(\mathcal{B}^*)$ for such a model asymptotically follows a $\tfrac{1}{2}\chi^2_k$ distribution.

Now, let's consider the number of potential models with $k$ superfluous terms, which is $\binom{n_0 - n^*}{k}$. Assuming, for the sake of extending the argument, that the corresponding log-likelihood gains $\Delta\hat{\ell}$ behave sufficiently independently, we can apply the same reasoning as before%
\footnote{A key difference from the previous reasoning is the presence of a $\chi^2_k$ random variable instead of a $\chi^2_1$ variable.
However, the tails of the $\chi^2_k$ and $\chi^2_1$ distributions behave sufficiently similarly for this analysis.
Thus, the core reasoning remains valid up to the first order in the location parameter.
To see this, consider $x_i \sim \chi^2_k$:
\begin{equation*}
\operatorname{P}[\max(x_1,\dots,x_N) \leq z] = \operatorname{P}[x_1\leq z]^N \approx e^{-N(1 - \operatorname{P}[x_1\leq z])}
\end{equation*}
Furthermore, we have the following tail approximation:
\begin{equation*}
1 - \operatorname{P}[x_1\leq z] \approx \frac{z^{\frac{k}{2}-1}e^{-z/2}}{c}
\end{equation*}
where $c$ is a constant. This approximation is obtained from the  asymptotic series of the upper incomplete gamma function \cite{IncompleteGammaFunction2025}. Thus,
\begin{equation*}
\operatorname{P}[\max(x_1,\dots,x_N) \leq z] \approx e^{e^{-(z + 2\ln N - (k-2)\ln z + 2\ln c)/2}}
\end{equation*}
Hence, by neglecting the term $(\frac{k}{2}-1)\ln z$ (which varies more slowly than $z$), we observe that the leading term of the Gumbel distribution's location parameter remains $2\ln N$. This result leads to $\mu_1= \ln{N}$ as the location parameter when we are dealing with $\frac{1}{2} \chi_k^2$ instead of $\frac{1}{2} \chi_1^2$ as in \cref{eq:approx_cumulative_pastis}. 
}. The core idea was that the penalty should counteract the effect of maximizing over the number of models considered, relative to a target error probability $p$.

Let $\Lambda_k$ be the threshold that the gain $\Delta\hat{\ell}$ must exceed to justify accepting $k$ terms over the true model. Following the previous logic (penalty $\approx \ln(\text{Number of models}/p)$), we expect:
\[
\Lambda_k \approx \ln\frac{\binom{n_0 - n^*}{k}}{p}
\]
Using the approximation $\ln \binom{n_0 - n^*}{k} \approx k \ln(n_0 - n^*) \approx k \ln n_0$ (for large $n_0-n^*$ relative to $k$), the required threshold becomes approximately:
\[
\Lambda_k \approx k \ln n_0 - \ln p
\]

We now check the threshold implemented by the PASTIS criterion. Comparing model $\mathcal{B}$ (with $k$ superfluous terms) to the true model $\mathcal{B}^*$, the criterion selects $\mathcal{B}$ if $\mathcal{I}_{\text{PASTIS}}(\mathcal{B}) > \mathcal{I}_{\text{PASTIS}}(\mathcal{B}^*)$. This condition is equivalent to requiring the log-likelihood gain $\Delta\hat{\ell}$ to exceed the threshold $T_{\text{PASTIS}}$:
\[
\Delta\hat{\ell} > T_{\text{PASTIS}} \quad \text{where} \quad T_{\text{PASTIS}} = (n_\mathcal{B} - n^*) \ln\frac{n_0}{p} = k \ln\frac{n_0}{p}.
\]
Comparing the derived threshold $\Lambda_k \approx k \ln n_0 - \ln p$ with the PASTIS threshold $T_{\text{PASTIS}} = k \ln n_0 - k \ln p$, we see they share the same dominant term ($k \ln n_0$) and differ by $(k-1)\ln p$. For $k=1$, they match exactly. For $k>1$, the PASTIS threshold is larger (more stringent) than the heuristic $\Lambda_k$. Thus, the probability of selecting a model with $k>1$ superfluous parameters is expected to be lower than the probability ($\approx p$) of selecting a model with one superfluous parameter. Therefore, $\mathcal{I}_{\text{PASTIS}}$ strongly discourages the selection of models with many superfluous parameters. 

\section{$p$ is a statistical significance threshold}
\label{sec:p_statistical_significance_threshold} 

The parameter $p$ acts as a \textbf{statistical significance threshold}. As suggested by the derivation in \cref{sec:pastis_criterion} (\cref{eq:initial_equation_pastis} and the subsequent approximation leading to the Gumbel distribution), $p$ approximately represents the target probability of incorrectly selecting a model containing at least one superfluous term over the true parsimonious model $\mathcal{B}^*$, assuming $\mathcal{B}^*$ is contained within the library $\mathcal{B}_0$. More formally, based on the simple Gumbel approximation, we expect:
\begin{equation}
P\left[ \max_{s \in \mathcal{B}_0 \setminus \mathcal{B}^*} \mathcal{I}_{\text{PASTIS}}(\mathcal{B}^* \cup \{s\}) > \mathcal{I}_{\text{PASTIS}}(\mathcal{B}^*) \right] \approx p.
\label{eq:prob_overfit_approx_p}
\end{equation}

By choosing a small $p$ (e.g., $p=0.001$, used in subsequent benchmarks unless otherwise stated), we enforce a strong penalty against adding parameters. This demands significant evidence (a large increase in log-likelihood $\hat{\ell}(\mathcal{B})$) before accepting a new term into the model, thereby controlling the error rate associated with the multiple hypothesis tests inherent in sparse selection procedures. 
The penalty term $n_\mathcal{B} \ln(n_0/p)$ thus explicitly balances model fidelity (via $\hat{\ell}(\mathcal{B})$) against model complexity ($n_\mathcal{B}$), search space size ($n_0$), and the desired level of statistical stringency ($p$).

To quantify the performance of an information criterion in identifying the true model, we define the "Exact Match" accuracy. For a selected model basis $\mathcal{B}$ compared to the true basis $\mathcal{B}^*$, this accuracy is given by:
\begin{equation}
    \text{Exact Match}(\mathcal{B}) = \delta_{\mathcal{B}, \mathcal{B}^*}
    \label{eq:exact_match_def}
\end{equation}
where $\delta_{\mathcal{B}, \mathcal{B}^*}$ equals 1 if $\mathcal{B}=\mathcal{B}^*$, and 0 otherwise. 

When averaged over multiple independent simulations or data realizations (denoted by $\mathbb{E}[\cdot]$), the Exact Match accuracy provides an empirical estimate of the probability of correctly identifying the true model:
\begin{equation}
\mathbb{E}[ \text{Exact Match} ] \approx P[\mathcal{B} = \mathcal{B}^*].
\label{eq:avg_exact_match}
\end{equation}
Consequently, the quantity $1 - \mathbb{E}[\text{Exact Match}]$ estimates the overall probability of selecting an incorrect model. In scenarios where the primary selection error involves adding a single superfluous term to the true model (particularly when sufficient data is available), this empirical error rate $R_p$ is expected to approximate the theoretical error probability $p$ from \cref{eq:prob_overfit_approx_p}:
\begin{equation}
 R_p = 1 - \mathbb{E}[\text{Exact Match}(\tau \to \infty)]  = P\left[ \max_{s \in \mathcal{B}_0 \setminus \mathcal{B}^*} \mathcal{I}_{\text{PASTIS}}(\mathcal{B}^* \cup \{s\}) > \mathcal{I}_{\text{PASTIS}}(\mathcal{B}^*) \right] \approx p.
 \label{eq:empirical_error_approx_p}
\end{equation}

\begin{figure}[htbp]
    \centering
    \includegraphics[width=\linewidth]{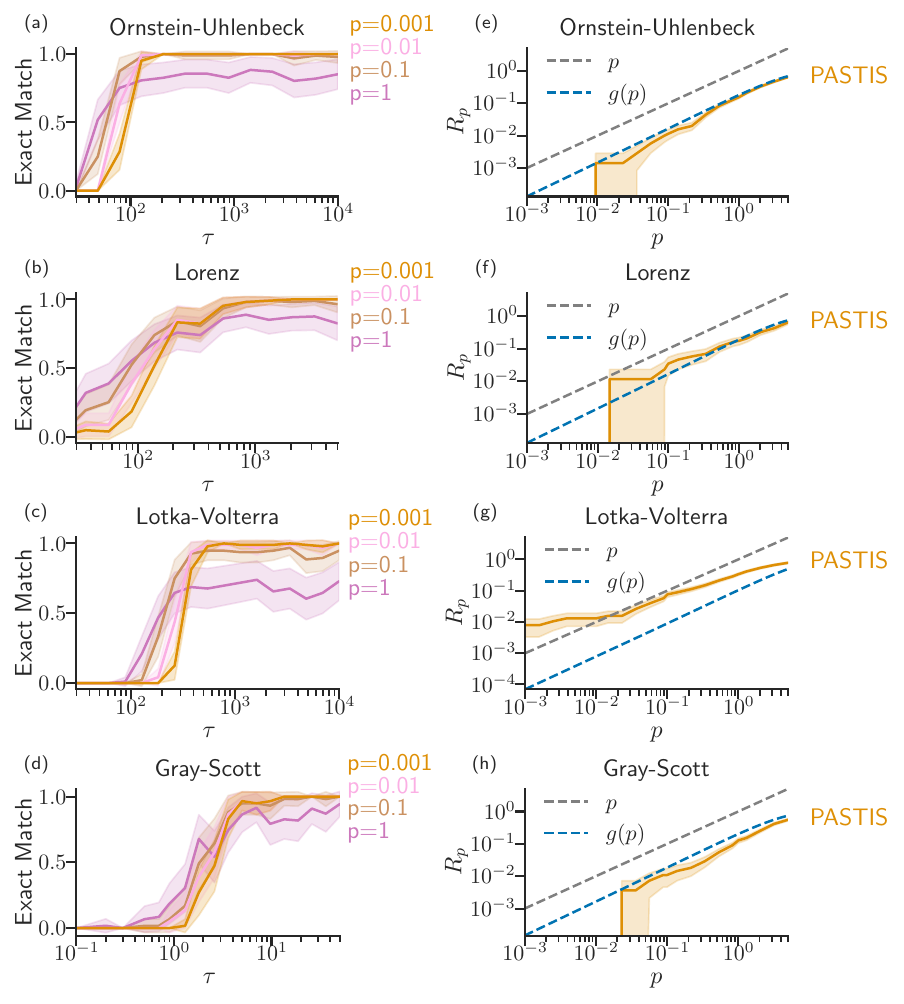} 
    \caption[Influence of p on PASTIS performance across systems]{Influence of the PASTIS control parameter $p$ on model selection performance across various dynamical systems.
    \textbf{(a-d)} Exact match accuracy (\cref{eq:exact_match_def}, averaged over 100 simulations) versus trajectory time $\tau$ for different values of $p$ (indicated by color) applied to simulated data from Ornstein-Uhlenbeck (a), Lorenz (b), Lotka-Volterra (c), and Gray-Scott (d) models. Smaller values of $p$ generally lead to higher asymptotic accuracy but require more data (larger $\tau$) to converge. Shaded areas likely represent variability across simulation runs.
    \textbf{(e-h)} Asymptotic probability of selecting an incorrect model ($R_p = 1 - \mathbb{E}[\text{Exact Match}(\tau \to \infty)]$, averaged over 100 simulations) versus the control parameter $p$, evaluated for long trajectories. Each panel displays simulation results for a specific system—(e) Ornstein-Uhlenbeck, (f) Lorenz, (g) Lotka-Volterra, (h) Gray-Scott—compared against the theoretical prediction $g(p)$ (dashed line) defined in \cref{eq:g_p_definition}.}
    \label{fig:p_influence_OU_chapN2}
\end{figure}

\Cref{fig:p_influence_OU_chapN2} provides empirical validation for these concepts across several dynamical systems that will be presented in detail in the next chapter, and were presented in \cref{sec:chap3_comparison}. 
Panels (a-d) plot the average Exact Match accuracy against the amount of data ($\tau$) for different $p$ values. These plots clearly illustrate the inherent trade-off: smaller $p$ values (corresponding to stricter penalties) achieve higher asymptotic accuracy but require substantially more data to converge compared to larger $p$ values. This confirms that demanding stronger evidence via a smaller $p$ improves the final model accuracy in the large data limit, effectively controlling overfitting.

Panels (e-h) further investigate the relationship between the control parameter $p$ and the empirically measured asymptotic probability of selecting an incorrect model ($R_p$). The simple theory leading to \cref{eq:prob_overfit_approx_p} suggests $R_p\approx p$. However, a more refined theoretical prediction can be obtained by using the exact CDF (\cref{eq:exact_cumulative_pastis}) instead of the simple Gumbel approximation when evaluating the probability in \cref{eq:prob_overfit_approx_p} with the PASTIS penalty $\lambda = \ln(n_0/p)$. This leads to a predicted error rate $g(p)$:
\begin{equation}
   R_p = P\left[ \max_{s \in \mathcal{B}_0 \setminus \mathcal{B}^*} \Delta\hat\ell_s > \ln\frac{n_0}{p} \right] = 1 - \left(\erf\left(\sqrt{\ln{\frac{n_0}{p}}}\right)\right)^{n_0 - n^*} =  g(p).
    \label{eq:g_p_definition}
\end{equation}
\Cref{fig:p_influence_OU_chapN2}(e-h) confirms that for systems satisfying the additive noise assumptions (OU, Lorenz, Gray-Scott - panels e, f, h), the empirically observed error rate $R_p$ closely follows this more precise theoretical prediction $g(p)$ (blue dashed line). While $g(p)$ differs slightly from $p$, both serve as effective controls for the error rate. This alignment validates the effectiveness of the PASTIS criterion, controlled by $p$, and underscores the accuracy of the underlying theoretical framework.

For the Lotka-Volterra system (panels c, g), which involves multiplicative noise rather than the additive noise assumed in the theoretical derivation of the $\Delta\hat\ell_s$ distribution, PASTIS still demonstrates strong performance in identifying the true model (panel c). However, the precise probabilistic interpretation of $p$ through $g(p)$ is less direct. Despite this, \cref{fig:p_influence_OU_chapN2}(g) shows that the empirical error rate $R_p$ still aligns reasonably well with $p$ and $g(p)$. This suggests that $p$ remains a useful parameter for controlling model selection stringency, even when the strict assumptions of the theoretical error rate calculation are not met.

In defining the PASTIS criterion with the penalty term $n_\mathcal{B} \ln(n_0/p)$, we prioritized a balance between theoretical grounding, practical utility, and interpretability. We recognize that the refined theoretical error rate associated with this specific penalty, under ideal assumptions (like additive noise), is given by $g(p)$ defined in \cref{eq:g_p_definition}, which is approximately, but not exactly, equal to the nominal rate $p$.

One could theoretically define an alternative criterion by inverting the function $g(p)$ to find a penalty term, let's call it $\lambda'$, such that the resulting theoretical error rate perfectly matches the target $p$. This would require solving $1 - (\erf(\sqrt{\lambda'}))^{n_0 - n^*} = p$ for $\lambda'$, likely leading to a significantly more complex expression involving the inverse erf function.

However, such a formulation would sacrifice the simple and intuitive structure of the proposed PASTIS penalty. The term $\ln(n_0/p)$ provides a clear, logarithmic relationship between the penalty per parameter, the search space size ($n_0$), and the desired significance level ($p$), directly motivated by the leading-order extreme value approximation. As empirically validated in \cref{fig:p_influence_OU_chapN2}, the simple penalty $n_\mathcal{B}\ln(n_0/p)$ leads to an actual error rate that closely follows the refined prediction $g(p)$ and remains near the target $p$. We believe this elegant and relatively simple form, which demonstrates robust performance, is more likely to be understood and adopted by practitioners than a mathematically exact but potentially obscure alternative.

\section{Comparison with Extended Bayesian Information Criterion (EBIC)}
\label{sec:comparison_ebic}

The challenge of model selection in high-dimensional settings, where the number of potential candidate terms ($n_0$) can be large relative to the sample size (denoted by $\tau$ in our context), has motivated the development of criteria beyond AIC and standard BIC. One notable example is the Extended Bayesian Information Criterion (EBIC), proposed by Chen and Chen \cite{chenExtendedBayesianInformation2008}. Like PASTIS, EBIC aims to mitigate the tendency of standard criteria to select overly complex models when the search space is vast. EBIC utilizes the maximized likelihood within a penalty framework derived from Bayesian principles to score and select models.

\subsection*{Motivation and Formulation of EBIC}
The primary motivation for EBIC is the observation that the standard BIC, while consistent when the number of potential candidates $n_0$ is fixed, becomes too liberal (prone to selecting superfluous terms) when $n_0$ is large or grows with the sample size $\tau$. EBIC addresses this by incorporating the size of the model space into its penalty term, derived from a Bayesian perspective that assigns specific prior probabilities to different models.

To align with the maximization framework used for PASTIS (i.e., $\hat{\ell} - \text{penalty}$), the EBIC criterion for a model $\mathcal{B}$ with $n_{\mathcal{B}}$ parameters can be expressed as:
\begin{equation}
\text{EBIC}_{\gamma}(\mathcal{B}) = \hat{\ell}(\mathcal{B}) - \frac{n_{\mathcal{B}}}{2} \ln \tau - \gamma \ln \binom{n_{0}}{n_{\mathcal{B}}}
\label{eq:ebic_formula_maximise}
\end{equation}
where $\hat{\ell}(\mathcal{B})$ is the maximized log-likelihood, $n_{\mathcal{B}}$ is the number of parameters in model $\mathcal{B}$, $\tau$ is the trajectory duration, $n_0$ is the total number of candidate terms in the initial library, $\binom{n_{0}}{n_{\mathcal{B}}}$ represents the number of possible models with exactly $n_{\mathcal{B}}$ parameters, and $\gamma \in [0, 1]$ is a tuning parameter controlling the strength of the additional penalty.

The crucial addition compared to BIC (which corresponds to $\gamma=0$) is the term $\gamma \ln \binom{n_{0}}{n_{\mathcal{B}}}$. This term explicitly penalizes models based on the size of the combinatorial space they belong to. A larger $\gamma$ imposes a stronger penalty related to this model space complexity, making the criterion more conservative and favoring sparser models.

For large $n_0$ and relatively small $n_{\mathcal{B}}$ ($n_{\mathcal{B}} \ll n_0$), the term $\ln \binom{n_{0}}{n_{\mathcal{B}}} \approx n_{\mathcal{B}} \ln n_0$. Thus, the additional penalty term behaves approximately as $\gamma n_{\mathcal{B}} \ln n_0$. This highlights a connection to PASTIS, where the penalty is $n_{\mathcal{B}} \ln(n_0/p)$, as both involve a penalty proportional to $n_{\mathcal{B}} \ln n_0$.

\subsection*{Key Similarities and Differences}

While both PASTIS (\cref{eq:PASTIS_chapN2}) and EBIC (\cref{eq:ebic_formula_maximise}) aim for parsimony in large model spaces by adding penalties dependent on the library size $n_0$, their derivations and interpretations differ:

\begin{itemize}
    \item \textbf{Foundation:} EBIC modifies the standard Bayesian derivation by assigning a prior probability $p(\mathcal{B})$ to model $\mathcal{B}$ that is inversely related to the size of the space of models with $n_{\mathcal{B}}$ parameters, specifically $p(\mathcal{B}) \propto [\binom{n_0}{n_{\mathcal{B}}}]^{-\gamma}$. PASTIS uses a frequentist approach grounded in Extreme Value Theory to control the risk of selecting spurious terms based on their maximum expected likelihood gain.

    \item \textbf{Parameter Interpretation:} The EBIC parameter $\gamma$ tunes the strength of the prior penalization based on model space size. While values like $\gamma=1$ provide strong control, and consistency results can guide its choice, $\gamma$ lacks a direct probabilistic interpretation in terms of error control. In contrast, the parameter $p$ in PASTIS is designed to approximate the probability of selecting a model with at least one superfluous term, providing a more direct link to statistical significance thresholds, as discussed in \Cref{sec:p_statistical_significance_threshold}.
\end{itemize}

In essence, EBIC offers a Bayesian extension to handle large model spaces by adjusting priors, while PASTIS provides an alternative rooted in controlling error rates via extreme value statistics, offering an interpretable control parameter $p$.

\section{Implementation: Model Space Search}
\label{sec:pastis_search}
Selecting the model $\mathcal{B} \subseteq \mathcal{B}_0$ that maximizes an information criterion (AIC, BIC, PASTIS) or the log-likelihood computed with CV involves searching a space of $2^{n_0}$ possibilities, which is computationally infeasible for large $n_0$ (NP-complete). 

We employ a practical greedy search strategy inspired by the Single Best Replacement (SBR) developed in \cite{soussenBernoulliGaussianDeconvolution2011}:
\begin{enumerate}
    \item \textbf{Initialization:} Start multiple parallel searches from different initial models: the null model ($\mathcal{B}=\emptyset$), the full library ($\mathcal{B}=\mathcal{B}_0$), and $n_0$ random subsets of $\mathcal{B}_0$.
    \item \textbf{Iteration:} In each step, randomly propose either adding a single function from $\mathcal{B}_0 \setminus \mathcal{B}$ to the current basis $\mathcal{B}$, or removing a single function from $\mathcal{B}$.
    \item \textbf{Acceptance:} Calculate the information criterion for the proposed model. If it is higher than the criterion for the current model, accept the change. Otherwise, keep the current model.
    \item \textbf{Termination:} Repeat the iteration until no single addition or removal improves the criterion for a certain number of steps (convergence).
\end{enumerate}
The final selected model is the one with the highest information criterion found across all parallel searches. While not guaranteed to find the global optimum, this stochastic hill-climbing approach is computationally efficient and performs well in practice for finding sparse models, rapidly converging when a sparse true model exists and is favored by the criterion.

\section{Conclusion}
\label{sec:pastis_conclusion}

This chapter introduced the Parsimonious Stochastic Inference (PASTIS) criterion, a novel approach designed to address the challenge of model selection within large candidate libraries, a common scenario in system identification where standard criteria like AIC and BIC often falter due to the multiple comparisons problem. We demonstrated how leveraging Extreme Value Theory (EVT) allows for characterizing the distribution of the maximum spurious log-likelihood gain expected when considering many potential model terms. This analysis formed the basis for the PASTIS penalty term, $n_\mathcal{B} \ln(n_0/p)$, which explicitly incorporates the size of the initial library ($n_0$) and a user-defined significance level ($p$) that approximately controls the probability of selecting superfluous terms.

We explored the theoretical underpinnings of PASTIS, including the derivation based on the Gumbel approximation and the interpretation of $p$ as a statistical threshold, supported by empirical simulations showing its effectiveness in controlling the error rate. Furthermore, a comparison with the Extended Bayesian Information Criterion (EBIC) highlighted the distinct theoretical foundations—frequentist EVT for PASTIS versus Bayesian priors for EBIC—and the useful interpretation of the PASTIS parameter compared to the EBIC parameter. 
Finally, we outlined a practical greedy search algorithm for implementing AIC, BIC, PASTIS, and CV to navigate the vast model space efficiently. 

Having established the theoretical motivation and formulation of PASTIS, the crucial next step is to rigorously evaluate its practical performance and compare it against established methods. The following chapter (\Cref{chap:validation}) undertakes this task through comprehensive benchmarking on synthetic data generated from known stochastic dynamical systems, assessing the ability of PASTIS to accurately and reliably recover the true underlying equations of motion.

\newpage

\begin{tcolorbox}[
    enhanced, 
    sharp corners, 
    boxrule=0.5pt, 
    colframe=black!75!white, 
    colback=white, 
    coltitle=black, 
    fonttitle=\bfseries, 
    title=Key Takeaways: \Cref{chap:pastis} -- The PASTIS Criterion, 
    attach boxed title to top left={yshift=-0.1in, xshift=0.15in},
    boxed title style={ 
        colback=white, 
        sharp corners, 
        boxrule=0pt, 
        frame code={
            \draw[black!75!white, line width=0.5pt]
                ([yshift=-1pt]frame.south west) -- ([yshift=-1pt]frame.south east);
        }
    },
    boxsep=5pt, 
    left=5pt,
    right=5pt,
    top=12pt, 
    bottom=5pt
    ]
    \begin{itemize}
        \item \textbf{Problem Addressed:} Standard criteria like AIC overfit when selecting sparse models from large candidate libraries ($n_0$ terms) due to the multiple comparisons problem. BIC does not explicitly account for library size $n_0$.
        \item \textbf{Core Idea:} Analyze the distribution of the \emph{maximum} log-likelihood gain ($\max \Delta\hat\ell_s$) obtained when adding single \emph{superfluous} terms to the true model.
        \item \textbf{Extreme Value Theory (EVT):} Under the assumption that individual gains $\Delta\hat\ell_s$ (approx. $\frac{1}{2}\chi^2_1$) are i.i.d., the distribution of their maximum can be approximated using EVT.
        \item \textbf{Gumbel Distribution:} The distribution of $\max \Delta\hat\ell_s$ is approximated by a Gumbel distribution, implying the expected maximum spurious gain grows logarithmically with the number of superfluous terms ($n_0-n^*$), dominated by a $\sim \ln n_0$ scaling.
        \item \textbf{PASTIS Criterion:} A new criterion is proposed to counteract this effect:
            \[
            \mathcal{I}_{\text{PASTIS}}(\mathcal{B}) = \hat{\ell}(\mathcal{B}) - n_\mathcal{B} \ln \frac{n_0}{p}
            \]
        \item \textbf{Penalty Term:} The penalty term $n_\mathcal{B} \ln(n_0/p)$ incorporates: 
            \begin{itemize} 
                \item Scaling with model complexity ($n_\mathcal{B}$).
                \item Penalty per parameter accounting for library size ($n_0$) and significance level ($p$). 
            \end{itemize}
        \item \textbf{Parameter $p$ Interpretation:} $p$ acts as a statistical significance threshold, approximately controlling the probability of selecting a model with at least one superfluous term over the true model ($\approx p$ via simple Gumbel, more accurately given by $g(p)$ \cref{eq:g_p_definition}). Choosing small $p$ enforces parsimony. A typical choice used in benchmarks is $p=0.001$.
        \item \textbf{Implementation:} Maximizing $\mathcal{I}_{\text{PASTIS}}$ requires searching the model space; a practical greedy search algorithm is employed.
    \end{itemize}
\end{tcolorbox}

%% file: tex_body/chap6.tex
\chapter{Validation and Benchmarking}
\label{chap:validation}
\chaptertoc{}

Having introduced the PASTIS information criterion (\cref{eq:PASTIS_chapN2}) in \cref{chap:pastis}, this chapter presents a comprehensive validation of its performance. We use synthetic data generated from known Stochastic Differential Equations (SDEs), allowing us to assess the ability of PASTIS to correctly identify the underlying equations and compare its effectiveness against established model selection methods.

\section{Methodology}
\label{sec:validation_methodology}

\subsection{Benchmark Systems}
\label{subsec:benchmark_systems}
\begin{figure}[htbp]
    \centering
    \includegraphics[width=\linewidth]{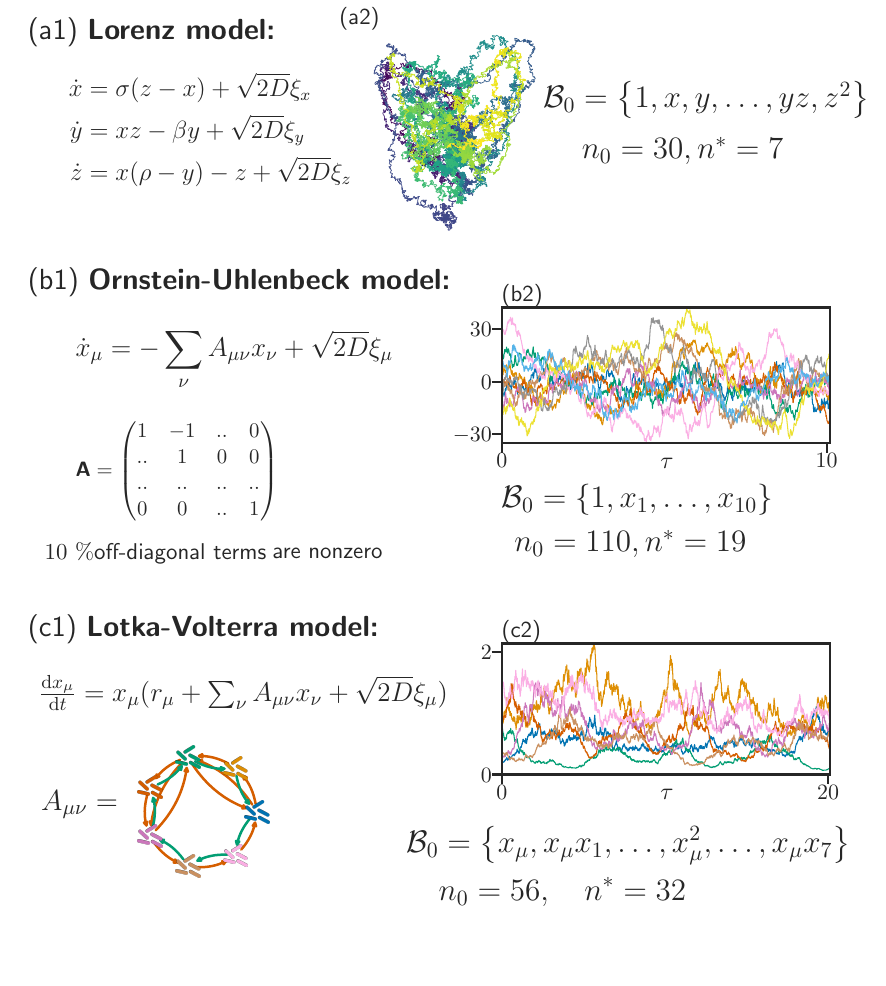} 
    \caption[Benchmark dynamical systems for model selection]{Overview of the three benchmark dynamical systems used for testing model selection methods. Each system is presented with its governing equations, the structure of the candidate function library $\mathcal{B}_0$, the total number of candidate terms ($n_0$), the number of true terms ($n^*$), and a visualization of sample dynamics.
    \textbf{(a1, a2)} The out-of-equilibrium Lorenz system.
    \textbf{(b1, b2)} A high-dimensional linear Ornstein-Uhlenbeck process.
    \textbf{(c1, c2)} The Lotka-Volterra predator-prey model exhibiting population dynamics.
    These systems represent a diverse set of challenges for sparse model identification, varying in dimensionality ($n^*$ ranging from 7 to 32), library size ($n_0$ ranging from 30 to 110), linearity, and type of dynamics (out-of-equilibrium, stochastic linear, population dynamics).}
    \label{fig:benchmark_systems}
\end{figure}

We evaluate PASTIS on three distinct dynamical systems exhibiting different characteristics, summarized in \cref{fig:benchmark_systems}. These systems allow us to test the method across varying dimensions, nonlinearities, noise types, and library sizes.

\begin{enumerate}
    \item \textbf{Stochastic Lorenz System} (\cref{fig:benchmark_systems}(a)): A low-dimensional ($d=3$) system modeled with additive noise $\sqrt{2D}\xi$ in each component. 
    The true model contains $n^*=7$ terms describing the dynamics. The candidate library $\mathcal{B}_0$ used for selection comprises $n_0=30$ polynomial terms up to second order in the state variables $x, y, z$ (e.g., $1, x, y, \dots, yz, z^2$).

    \item \textbf{High-Dimensional Ornstein-Uhlenbeck (OU) Process} (\cref{fig:benchmark_systems}(b)): A $d=10$-dimensional linear stochastic system $\dot{\bm{x}} = -\bm{A}\bm{x} + \sqrt{2D}\boldsymbol{\xi}$ with additive noise. 
    The interaction matrix $\bm{A}$ is sparse (specifically, constructed with 10\% off-diagonal non-zero terms and all diagonal terms equal to 1), meaning the true dynamics involve only $n^*=19$ linear terms across all equations. 
    The candidate library $\mathcal{B}_0$ includes all possible linear terms $\{1, x_1, \dots, x_{10}\}$ for each component equation $\dot{x}_\mu$, leading to a library size of $n_0=110$ potential terms.

    \item \textbf{Generalized Lotka-Volterra (LV) System} (\cref{fig:benchmark_systems}(c)): A model of $d=7$ interacting species with dynamics $\dot{x}_\mu = x_\mu (r_\mu + \sum_\nu A_{\mu\nu} x_\nu + \sqrt{2D}\xi_\mu)$, exhibiting multiplicative noise. The interaction matrix $A_{\mu\nu}$ representing the ecological network is sparse, which leads to a true model of $n^*=32$ terms across all equations. The candidate library $\mathcal{B}_0$ for identifying the dynamics of each species $\dot{x}_\mu$ includes linear ($x_\nu$) and quadratic ($x_\nu x_\kappa$) interaction terms (e.g., $x_\mu x_1, \dots, x_\mu x_7, \dots, x_\mu^2$), resulting in $n_0=56$ candidate terms.
\end{enumerate}
For each system, synthetic trajectories are generated using numerical integration schemes. Specific parameters used in the simulations and further details on the library construction can be found in \cref{apdx:simulations}.

\subsection{Performance Metrics}
\label{subsec:perf_metrics}
We assess model selection performance using two primary metrics:
\begin{enumerate}
    \item \textbf{Exact Match Accuracy:} The fraction of independent simulation runs for which the selected basis $\mathcal{B}$ exactly matches the true basis $\mathcal{B}^*$ that generated the data (defined in \cref{eq:exact_match_def}). This is a stringent measure of successful identification, averaged over multiple runs.
    \item \textbf{Prediction Error:} The mean squared error $\mathcal{E}(\hat{\bm{f}}^\mathcal{B})$ of the inferred drift field $\hat{\bm{f}}^\mathcal{B}$ from the selected model $\mathcal{B}$, evaluated on an independent, long test trajectory. It is normalized by a measure of the true drift variance: $\text{Prediction Error} = \mathcal{E}(\hat{\bm{f}}^\mathcal{B}) / \avg{\bm{f}\cdot(4\bm{\bar{D}})^{-1}\cdot\bm{f}}$, where $\bm{\bar{D}}$ is the approximate diffusion matrix. 
    This metric measures the predictive quality of the selected model, even if it is not an exact structural match.
\end{enumerate}

\subsection{Compared Algorithms}
\label{subsec:compared_algos}
We compare PASTIS (using $p=0.001$) against several standard or state-of-the-art methods:
\begin{itemize}
    \item \textbf{AIC:} Akaike Information Criterion (\cref{eq:aic_def}).
    \item \textbf{BIC:} Bayesian Information Criterion (\cref{eq:bic_def}).
    \item \textbf{CV:} $K$-fold Cross-Validation (here, $K=7$) selecting the model minimizing prediction error on held-out folds (see \cref{sec:cross_validation}). 
    \item \textbf{SINDy:} Sparse Identification of Nonlinear Dynamics \cite{bruntonDiscoveringGoverningEquations2016}, using the Sequential Thresholding Least Squares (STLSQ) algorithm from the PySINDy library \cite{kaptanogluPySINDyComprehensivePython2022} with a threshold of 0.5. While primarily designed for ODEs, we apply it for comparison; considerations regarding hyperparameter choices like the threshold are discussed in \cref{apdx:hyperparams_errors}.
\end{itemize}
For AIC, BIC, CV, and PASTIS, the same underlying likelihood framework (\cref{sec:likelihood_sfi}) and greedy search algorithm (\cref{sec:pastis_search}) are employed, ensuring that performance differences primarily reflect the selection criterion itself.

\section{Performance Evaluation}
\label{sec:validation_results}

\begin{figure}[htbp] 
    \centering
    \includegraphics[width=\textwidth]{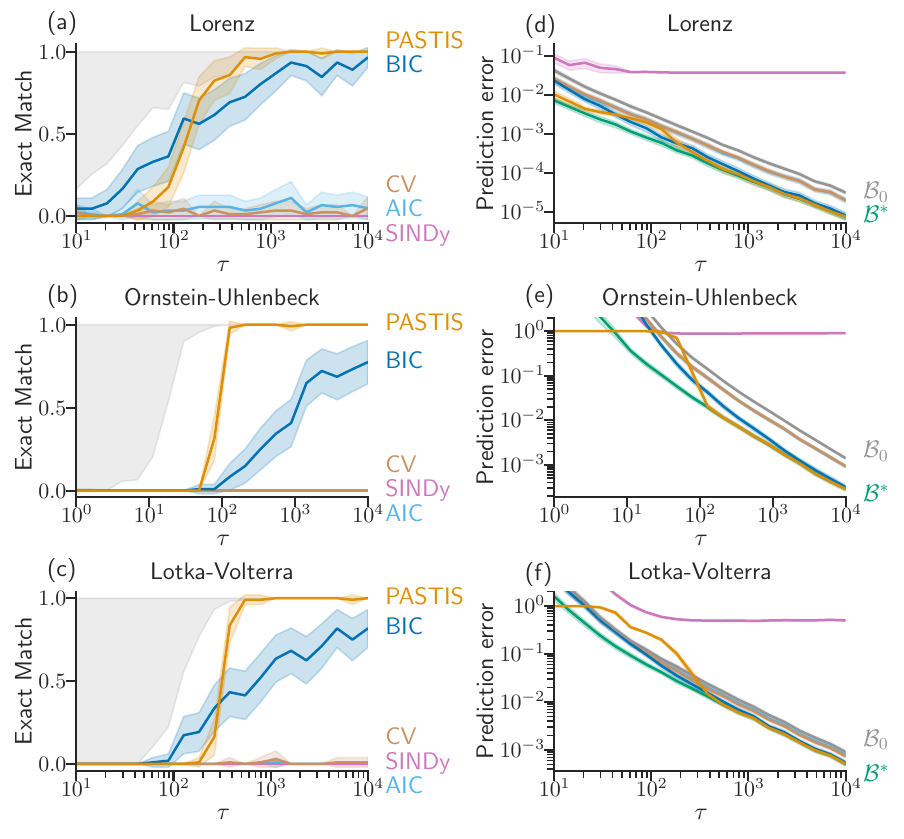} 
    \caption[Benchmark comparison of model selection methods]{Benchmark comparison of PASTIS against standard model selection methods (AIC, BIC, Cross-Validation (CV), and SINDy) across the three dynamical systems presented in \cref{fig:benchmark_systems}.
    \textbf{(Left column, a-c)} Exact Match accuracy (\cref{eq:exact_match_def}, averaged over 100 simulations) versus trajectory time $\tau$ for the Lorenz (a), Ornstein-Uhlenbeck (b), and Lotka-Volterra (c) systems. Different colors represent different selection criteria as indicated by the legend in (c) and (f). The gray area indicates cases where the true model $\mathcal{B}^*$ does not maximize $\hat{\ell}$ among models with $n^*$ parameters. 
    \textbf{(Right column, d-f)} Corresponding Prediction Error (defined in \cref{subsec:perf_metrics}) of the inferred model versus trajectory time $\tau$ for the Lorenz (d), Ornstein-Uhlenbeck (e), and Lotka-Volterra (f) systems. Lower values indicate better predictive performance. All curves are averages over 100 simulations, with shaded areas representing 3 standard errors.} 
    \label{fig:benchmark_comparison}
\end{figure}

\Cref{fig:benchmark_comparison} summarizes the key results across the three benchmark systems. A discussion of potential sources contributing to errors or performance differences observed in these benchmarks can be found in \cref{apdx:hyperparams_errors}.

\subsection{Exact Match Accuracy}
\label{subsec:results_accuracy}
\begin{figure}[htbp] 
    \centering
    \includegraphics[width=\textwidth]{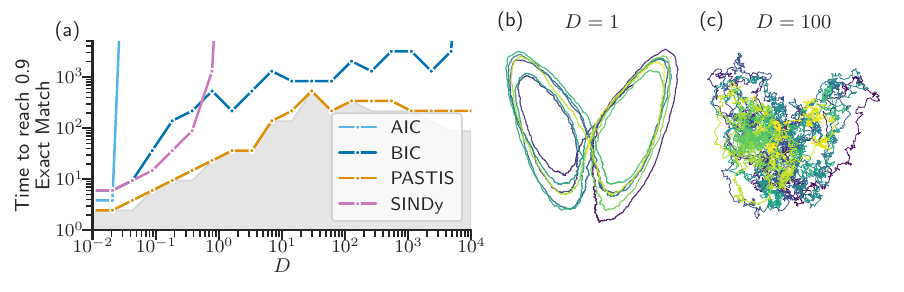}
    \caption[Efficiency vs. Noise for Lorenz System]{Efficiency, noise robustness, and system dynamics for the stochastic Lorenz system (\Cref{fig:benchmark_systems}(a)). \textbf{(a)}~Average trajectory time $\tau$ (log scale) needed to reach 0.9 Exact Match accuracy (\Cref{eq:exact_match_def}) versus noise amplitude $D$ (log scale), comparing PASTIS, AIC, BIC, and SINDy (legend inset). The gray area indicates cases where the true model $\mathcal{B}^*$ maximize  $\hat{\ell}$ value among models with $n^*$ parameters for 90 \% of cases. \textbf{(b, c)}~Sample trajectories illustrating the Lorenz dynamics for noise levels $D=1$ (b) and $D=100$ (c) for a trajectory duration $\tau=50$, corresponding to points within the range explored in panel (a).} 
\label{fig:lorenz_efficiency_vs_noise}
\end{figure}

Across all systems (\cref{fig:benchmark_comparison}, panels a-c), PASTIS generally demonstrates superior or competitive performance in identifying the exact underlying model structure, especially with sufficient data.
\begin{itemize}
    \item \textbf{PASTIS:} With an increasing amount of data ($\tau$), the exact match accuracy converges towards a high value (typically consistent with $1-g(p)$ or $1-p$), confirming its ability to control false inclusions according to the chosen significance level $p$. 
    It reliably recovers the correct sparse structure in diverse settings.
    
    \item \textbf{AIC and CV:} These methods often exhibit lower exact match accuracy, particularly for larger $\tau$. They tend to select models with superfluous terms, confirming the theoretical tendency of AIC and CV to overfit when comparing multiple models (discussed in \cref{sec:likelihood_aic_limits,sec:cross_validation}). 
    
    \item \textbf{BIC:} BIC, like PASTIS, shows asymptotic consistency properties, with accuracy generally increasing with $\tau$. However, its convergence is slower than PASTIS, often requiring more data to reach comparable accuracy levels. This difference stems from BIC's penalty ($\log \tau$) not explicitly accounting for the library size $n_0$. Nevertheless, for Lorenz and Lotka-Volterra systems, in the intermediate data-duration regime BIC tends to have a higher chance to select the true model. 
    
    \item \textbf{SINDy:} SINDy's performance is highly dependent on the noise level. In these examples (\cref{fig:benchmark_comparison}), it may perform well in some cases with very low dynamical noise (\cref{fig:lorenz_efficiency_vs_noise} and \cref{apdx:hyperparams_errors_sindy}) but is very sensitive to the noise inherent in the stochastic dynamics. 
    It always fails to identify the correct terms despite careful threshold tuning (\cref{apdx:hyperparams_errors_sindy}).
\end{itemize}

To further investigate the robustness and efficiency of these methods, particularly concerning the impact of noise, \Cref{fig:lorenz_efficiency_vs_noise} presents a complementary analysis for the stochastic Lorenz system. Instead of plotting accuracy versus time for a fixed noise level, this figure shows the average trajectory time $\tau$ required by each method to reach a high level of confidence (0.9 Exact Match accuracy) as the noise amplitude $D$ is varied. This metric highlights how efficiently each algorithm converges to the correct model under different noise conditions. As seen in \Cref{fig:lorenz_efficiency_vs_noise}, PASTIS consistently requires relatively short trajectories across a wide range of noise amplitudes, indicating good robustness. BIC also demonstrates robustness, although it generally requires more data than PASTIS. Conversely, the performance of AIC and SINDy degrades significantly with increasing noise. This reinforces the observation that criteria like PASTIS and BIC, which employ stronger penalties, are better suited for reliable sparse identification in stochastic systems, especially when noise levels are non-negligible.

The gray shaded areas in the accuracy plots of \cref{fig:benchmark_comparison,fig:lorenz_efficiency_vs_noise} show scenarios where the true model $\mathcal{B}^*$ does not achieve the maximum log-likelihood value ($\hat{\ell}$) among all possible models having the correct number of parameters ($n^*$). 
In these specific scenarios, identifying the true model becomes intrinsically impossible for any method that relies strictly on comparing likelihood values within the class of $n^*$-parameter models. Consequently, the upper boundary suggested by this gray region represents a practical performance limit for such likelihood-based identification strategies. The observation that PASTIS's accuracy closely approaches this empirical bound indicates its near-optimal performance in leveraging likelihood information, while the small remaining gap might suggest room for minor refinements. 

\subsection{Prediction Error}
\label{subsec:results_prediction}
Examining the prediction error (\cref{fig:benchmark_comparison}, panels d-f) provides insight into the quality of the selected models, even if they don't perfectly match the true structure.
\begin{itemize}
    \item \textbf{PASTIS:} In the long trajectory regime (large $\tau$), PASTIS consistently identifies the true model $\mathcal{B}^*$ (as highlighted by the Exact Match accuracy in \cref{fig:benchmark_comparison}(a-c)). 
    Consequently, the models selected by PASTIS yield low prediction error, often achieving the best predictive performance among the tested methods in this regime (\cref{fig:benchmark_comparison}(d-f)). 
    For very short trajectories, the strong penalty inherent in PASTIS requires substantial evidence to include terms. With insufficient data, this leads to the selection of the null model. Notably, this error from the null model may still be lower than the prediction error that would arise from fitting the true model structure $\mathcal{B}^*$ with parameters estimated from such severely limited data, due to high parameter variance in the latter case. During the transition in intermediate data regimes, PASTIS may select a model that slightly underfits relative to methods with weaker penalties like BIC, potentially resulting in a transiently higher prediction error before it converges to the optimal sparse model with more data.
    \item \textbf{AIC and CV:} The models selected by AIC and CV lead to lower prediction error than the full model $B_0$ by removing a few superfluous terms (usually selecting $2 n^*$ basis functions for CV \cref{fig:appendix_hyperparameters_total}). However, these selected models do not achieve the prediction error attainable with the true model $\mathcal{B}^*$.
    \item \textbf{BIC:} The Bayesian Information Criterion employs a penalty term equal to $\frac{n_{\mathcal{B}}}{2}\ln{\tau}$. 
    For small amounts of data (small $\tau$), this penalty is relatively weak (even negative for $\tau<1$). Consequently, BIC may initially favor overly complex models, as the penalty might be insufficient to counteract moderate increases in likelihood gained from including superfluous terms. This behavior stems partly from the BIC derivation relying on asymptotic approximations that may not hold for small $\tau$. While the selected model isn't necessarily the full library $\mathcal{B}_0$, this initial tendency towards complexity can lead to suboptimal prediction performance in the low-data regime (\cref{fig:benchmark_comparison}(d-f)). In intermediate data regimes, where PASTIS has not yet converged to the true sparse model $\mathcal{B}^*$ due to its stronger penalty, BIC selects a slightly over-complete model that can temporarily achieve lower prediction error than the sparser model favored by PASTIS (visible in \cref{fig:benchmark_comparison}(d-f)). 
    However, BIC is known to be statistically consistent; its penalty strengthens sufficiently with increasing $\tau$, ensuring it will eventually select the true model asymptotically, given enough data. 
    Our simulations reflect this property (\cref{fig:benchmark_comparison}(a-c)), but also highlight that BIC's convergence to the exact model can be considerably slower than PASTIS, often requiring much longer trajectories ($\tau$) to achieve high Exact Match accuracy.
    \item \textbf{SINDy:} The performance of the SINDy algorithm, particularly using the Sequential Thresholding Least Squares (STLSQ) method employed here, is known to be sensitive to its threshold hyperparameter and the noise level in the data. Despite our attention to selecting this threshold hyperparameter, in all of the scenarios tested (\cref{fig:benchmark_comparison}(a-c)), SINDy struggles to identify the true model structure, often yielding low Exact Match accuracy across different data regimes ($\tau$). Consequently, the resulting models frequently exhibit poor predictive capabilities, as reflected by a substantially higher prediction error compared to methods like PASTIS or BIC in most of the benchmark cases (\cref{fig:benchmark_comparison}(d-f)). 
\end{itemize}
Overall, PASTIS demonstrates a strong ability to select models that are both structurally accurate (high Exact Match) and predictively skillful (low Prediction Error).

\section{Conclusion}
\label{sec:validation_conclusion}

This chapter presented a comprehensive evaluation of the PASTIS information criterion, validating its performance against established model selection techniques using synthetic data from diverse benchmark systems: the Lorenz model, the high-dimensional linear Ornstein-Uhlenbeck process, and the nonlinear Lotka-Volterra system with multiplicative noise. Performance was assessed based on the stringent Exact Match accuracy metric, measuring the ability to identify the precise underlying model structure, and the Prediction Error, quantifying the predictive skill of the selected models.

The comparative benchmarks consistently highlighted the advantages of the PASTIS approach. Across the tested systems, PASTIS demonstrated a superior capability for accurately identifying the true sparse model, generally converging much faster than the asymptotically consistent BIC and decisively outperforming AIC and CV, which tended to select overly complex models. Furthermore, the models selected by PASTIS consistently exhibited low prediction errors, often matching or exceeding the predictive performance of models chosen by other criteria, underscoring the link between structural accuracy and predictive reliability. While SINDy can be effective for deterministic systems, its performance degraded significantly in these stochastic settings, failing to reliably identify the correct models despite careful tuning.

In summary, the validation demonstrates that PASTIS, grounded in extreme value theory, provides a robust, statistically principled, and highly effective method for sparse system identification. It offers a compelling combination of rapid convergence to the true model structure and controllable error rates, making it a valuable tool for discovering parsimonious models of stochastic dynamical systems from data.

%% file: tex_body/chap7.tex
\chapter{Sparse Model Selection for SPDEs using PASTIS}
\label{chap:pastis_spde}
\chaptertoc{}

\section{Introduction}
\label{sec:chap5_intro}

\Cref{chap:spde_inference} established a framework for inferring the coefficients of Stochastic Partial Differential Equations (SPDEs) by discretizing them into high-dimensional systems of Stochastic Differential Equations (SDEs). We developed robust estimators for the coefficients \( \bm{\alpha} \) of a given set of vector basis functions \( \{\bm{b}_j(\bm{u})\} \) representing the drift term \( \bm{{F}}(\bm{u}) = \sum_j \alpha_j \bm{b}_j(\bm{u}) \).

This chapter addresses the challenge of model selection by extending the Parsimonious Stochastic Inference (PASTIS) criterion, introduced in \cref{chap:pastis} for SDEs, to the problem of sparse model selection for SPDEs. Our goal is to leverage the statistical foundation of PASTIS, particularly its ability to handle large candidate libraries derived from extreme value theory, to identify the minimal, physically relevant set of terms governing the spatio-temporal dynamics described by an SPDE, directly from noisy, discretely sampled data.

\section{Applying PASTIS to Discretized SPDE Systems}
\label{sec:chap5_pastis_adapt}

The core idea is to apply the PASTIS model selection framework directly to the high-dimensional SDE system obtained after spatial discretization, as formulated in \cref{chap:spde_inference}. Recall the discretized system (e.g., its Euler-Maruyama form): 
\begin{equation}
    \Delta\bm{u}_t = \bm{{F}}(\bm{u}_t) \Delta t + \sqrt{\frac{2D}{\Delta V}} \Delta \bm{W}_t , 
    \label{eq:chap5_sde_system_recall}
\end{equation}
where \( \bm{u}(t) \) is the vector of field values on a spatial grid with $N$ points, and $\Delta V = (\Delta x)^d$ is the volume of a grid cell. 
We assume the drift \( \bm{{F}}(\bm{u}) \) belongs to the span of a large candidate library \( \mathcal{B}_0 = \{\bm{b}_1, \dots, \bm{b}_{n_0}\} \) of vector basis functions, such that the true drift corresponds to a sparse subset \( \mathcal{B}^* \subset \mathcal{B}_0 \):
\begin{equation}
 \bm{{F}}(\bm{u}) = \sum_{\bm{b}_j \in \mathcal{B}^*} \alpha_j^* \bm{b}_j(\bm{u}).
\end{equation}
The task is to identify this sparse subset \( \mathcal{B}^* \) from data. From a basis $\mathcal{B}$, we denote the reconstructed drift by $\bm{{\hat{F}}}^{\mathcal{B}}(\bm{u}_t)$: 
\begin{equation}
    \bm{{\hat{F}}}^{\mathcal{B}}(\bm{u}_t) =  \sum_{\bm{b}_j \in \mathcal{B}} \hat{\alpha}_j^{\mathcal{B}} \bm{b}_j(\bm{u}_t). 
\end{equation}
where $ \hat{\bm{\alpha}}^{\mathcal{B}} = (\bm{G}_{\mathcal{B}})^{-1} \avg{ \bm{b}_k(\bm{u}_t)^\top \left(\frac{\Delta\bm{u}_t}{\Delta t}\right)}_t$ with $(G_{\mathcal{B}})_{kj} = \avg{ \bm{b}_k(\bm{u}_t)^\top \bm{b}_j(\bm{u}_t) }_t$, as derived in \cref{chap:spde_inference} (see \cref{eq:alpha_euler_spde_vector}). 

We apply the PASTIS criterion (\cref{eq:PASTIS_chapN2}) to select the optimal basis \( \mathcal{B} \subseteq \mathcal{B}_0 \):
\begin{equation}
    \mathcal{I}_{\text{PASTIS}}(\mathcal{B}) = \hat{\ell}(\mathcal{B}) - n_\mathcal{B} \log \frac{n_0}{p},
    \label{eq:chap5_pastis_recall}
\end{equation}
where:
\begin{itemize}
    \item \( \hat{\ell}(\mathcal{B}) \) is the maximized log-likelihood calculated for the model defined by the selected basis functions in \( \mathcal{B} \). This likelihood is computed using the framework from \cref{chap:spde_inference}. The time average \( \langle\dots\rangle_t = \frac{1}{N_t} \sum_{k=1}^{N_t} (\cdot) \) averages the scalar dot products over the $N_t$ available time steps in the observed trajectory. Hence, the estimated likelihood $\hat{\ell}(\mathcal{B})$ is: 
    \begin{equation}
            \hat{\ell}(\mathcal{B}) = -\frac{\tau}{4} \avg{\left(\frac{\Delta \bm{u}_t}{\Delta t} -  \bm{{\hat{F}}}^{\mathcal{B}}(\bm{u}_t)\right)\cdot \bar{D}_{\text{eff}}^{-1}\cdot \left(\frac{\Delta \bm{u}_t}{\Delta t} -  \bm{{\hat{F}}}^{\mathcal{B}}(\bm{u}_t) \right)}_t
    \end{equation}
    with $\bar{D}_{\text{eff}} = \frac{1}{2\Delta t N} \avg{(\Delta \bm{u}_t)^\top (\Delta \bm{u}_t)}_t$ being the effective noise intensity on the grid. 
    \item \( n_\mathcal{B} = |\mathcal{B}| \) is the number of basis functions selected in the model \( \mathcal{B} \).
    \item \( n_0 = |\mathcal{B}_0| \) is the total number of candidate vector basis functions considered. This term is crucial, as SPDE libraries often contain many terms (polynomials, derivatives of various orders), making \( n_0 \) large.
    \item \( p \) is the user-defined significance parameter controlling the trade-off between model complexity and goodness-of-fit, interpreted as an approximate upper bound on the probability of including a superfluous term (see \cref{chap:pastis}).
\end{itemize}
The model \( \mathcal{B} \) that maximizes \( \mathcal{I}_{\text{PASTIS}} \) is chosen as the best representation of the underlying SPDE dynamics. The selection process involves a search strategy conducted with the greedy forward-backward search algorithm discussed in \cref{sec:pastis_search}, applied to the set of vector basis functions \( \mathcal{B}_0 \). 

\section{Benchmark Case Study: Gray-Scott System}
\label{sec:chap5_benchmark_gs}

\begin{figure}[htbp]
    \centering
    \includegraphics[width=0.8\textwidth]{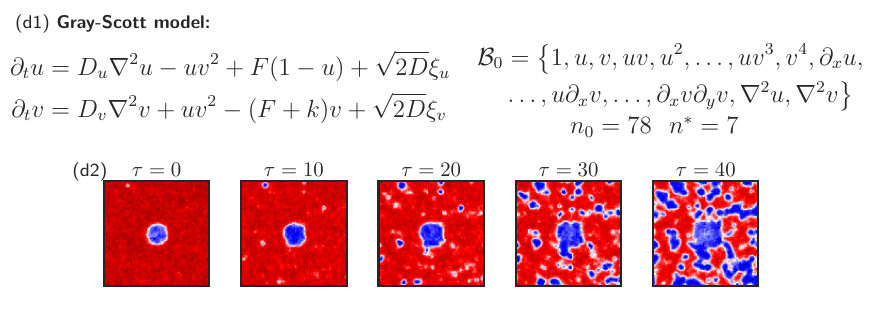} 
    \caption[Gray-Scott benchmark system details]{Details of the Gray-Scott benchmark system. \textbf{(d1)} Governing SPDEs, candidate library $\mathcal{B}_0$ with $n_0=78$ terms, and true model size with $n^*=7$ terms. \textbf{(d2)} Sample spatio-temporal pattern formation for the vector field $u$ over several trajectory lengths. See \cref{apdx:simulations} for simulation details.} 
    \label{fig:gray_scott_details}
\end{figure}

To evaluate the effectiveness of applying PASTIS to SPDEs, we focus on the Gray-Scott reaction-diffusion system, whose details are presented in \cref{fig:gray_scott_details}.

This system involves two coupled fields ($u, v$) exhibiting pattern formation. The inference task is particularly demanding due to our choice of a large candidate library ($n_0=78$) required to represent potential reaction terms (polynomials up to order 4) and spatial coupling (various first and second-order derivatives), compared to the truly sparse underlying dynamics ($n^*=7$ terms). 

\begin{figure}[htbp]
    \centering
    \includegraphics[width=\textwidth]{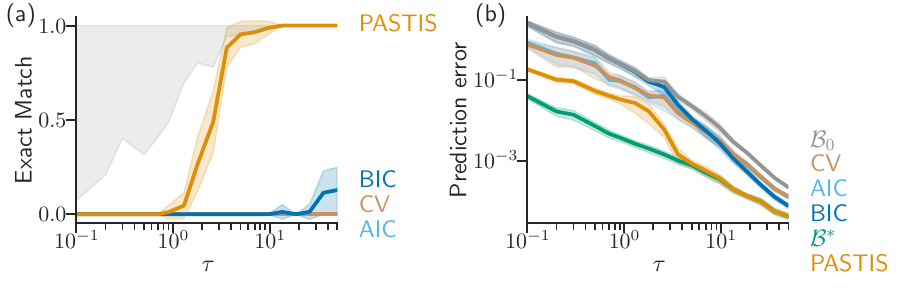} 
    \caption[PASTIS performance on Gray-Scott system]{Benchmark comparison for the Gray-Scott SPDE. \textbf{(a)} Exact Match accuracy vs. trajectory time $\tau$. \textbf{(b)} Prediction Error vs. $\tau$. Methods compared: PASTIS (orange), BIC (blue), AIC (light blue), CV (light brown). Shaded areas indicate standard error over simulations. The gray region in the left panel indicates regimes where the true model does not maximize the likelihood among models with $n^*$ terms.} 
    \label{fig:benchmark_gs_comparison}
\end{figure}

We applied PASTIS and other standard model selection criteria (AIC, BIC, CV) to synthetic data generated from the Gray-Scott system, using the inference framework from \cref{chap:spde_inference}. The performance was evaluated based on Exact Match Accuracy and Prediction Error as a function of trajectory length $\tau$. \Cref{fig:benchmark_gs_comparison} presents the results.

\subsection{Performance Analysis}
\label{subsec:chap5_gs_results}

\textbf{Exact Match Accuracy (\cref{fig:benchmark_gs_comparison}, panel a):} 
The results clearly demonstrate PASTIS's ability to identify the correct sparse structure of the SPDE. As the amount of data ($\tau$) increases, PASTIS (orange curve) rapidly converges to nearly 100\% accuracy, successfully distinguishing the 7 true terms from the 78 candidates. In contrast, AIC and CV (light blue and light brown curves) fail to identify the true model, selecting overly complex models consistent with their known tendency to overfit, especially when $n_0 \gg n^*$. 
BIC (blue curve) shows improvement over AIC/CV but converges significantly slower than PASTIS, failing to reach high accuracy within the simulated time range. PASTIS's accuracy closely approaches the theoretical limit imposed by data likelihood (upper edge of the gray region), indicating near-optimal performance.

\textbf{Prediction Error (\cref{fig:benchmark_gs_comparison}, panel b):} 
The prediction error results mirror the accuracy findings. PASTIS selects models that are not only structurally correct but also predictively accurate, achieving the lowest prediction error for sufficient data. The overfitted models selected by AIC and CV yield significantly higher prediction errors. 
BIC's prediction error improves as it selects sparser models than AIC/CV but remains higher than PASTIS due to its slower convergence to the true structure.

These results strongly indicate that the PASTIS criterion, derived from principles accounting for large search spaces, successfully extends to the high-dimensional SDE systems arising from SPDE discretization. Its penalty term effectively counteracts the multiple comparisons problem inherent in searching the large library ($\mathcal{B}_0$) needed to represent potential SPDE terms, allowing it to reliably recover the true, sparse governing equations.

\section{Discussion and Challenges}
\label{sec:chap5_discussion}

The successful application of PASTIS to the Gray-Scott system demonstrates the potential of this approach for discovering SPDEs from data. By leveraging the framework of \cref{chap:spde_inference}, PASTIS provides a statistically principled way to enforce parsimony.

However, the challenges inherent in SPDE inference, as discussed in \cref{sec:chap4_spde_challenges_vector}, remain pertinent:
\begin{itemize}
    \item \textbf{Basis Function Choice:} While PASTIS selects from a given library $\mathcal{B}_0$, the initial construction of this library (e.g., deciding the maximum polynomial order or types of derivatives) remains critical and problem-dependent.
    \item \textbf{Other Factors:} Issues like complex spatial noise correlations, non-periodic boundary conditions, and the influence of spatial discretization accuracy ($\Delta x$) require further investigation within the PASTIS framework.
\end{itemize}
Despite these challenges, the ability of PASTIS to handle large $n_0$ effectively makes it a promising candidate for SPDE identification where candidate libraries naturally become very large.

\section{Conclusion}
\label{sec:chap5_conclusion}

This chapter demonstrated the successful extension of the PASTIS information criterion to the challenging problem of sparse model selection for Stochastic Partial Differential Equations. By applying PASTIS to the high-dimensional SDE system resulting from spatial discretization (developed in \cref{chap:spde_inference}), we leverage its theoretical foundation in extreme value statistics to effectively manage the large candidate libraries inherent in SPDE inference.

The benchmark results on the Gray-Scott reaction-diffusion system clearly showed PASTIS's superior ability to identify the true, sparse governing SPDE compared to standard methods like AIC, BIC, and CV. It achieved high Exact Match accuracy rapidly and yielded models with excellent predictive performance. This confirms that PASTIS provides a robust and statistically principled framework for discovering parsimonious SPDE models from noisy, discrete spatio-temporal data, offering a significant advantage when searching through complex and extensive libraries of potential physical terms. While computational scaling remains a challenge for very large systems, PASTIS represents a powerful tool for data-driven discovery in complex spatio-temporal dynamics.

\newpage

\begin{tcolorbox}[
    enhanced, 
    sharp corners, 
    boxrule=0.5pt, 
    colframe=black!75!white, 
    colback=white, 
    coltitle=black, 
    fonttitle=\bfseries, 
    title=Key Takeaways: \Cref{chap:pastis} -- Sparse SPDE Selection with PASTIS, 
    attach boxed title to top left={yshift=-0.1in, xshift=0.15in},
    boxed title style={ 
        colback=white, 
        sharp corners, 
        boxrule=0pt, 
        frame code={
            \draw[black!75!white, line width=0.5pt]
                ([yshift=-1pt]frame.south west) -- ([yshift=-1pt]frame.south east);
        }
    },
    boxsep=5pt, 
    left=5pt,
    right=5pt,
    top=12pt, 
    bottom=5pt
    ]
    \begin{itemize}
        \item \textbf{Context:} Extended PASTIS from SDEs (\cref{chap:pastis}) to SPDEs by applying it to the discretized high-dimensional SDE system:
        \[ \Delta\bm{u}_t = \bm{{F}}(\bm{u}_t) \Delta t + \sqrt{\frac{2D}{\Delta V}} \Delta \bm{W}_t \] 
        where $\bm{{F}}(\bm{u}) = \sum_j \alpha_j \bm{b}_{j}(\bm{u})$.
        \item \textbf{PASTIS for SPDEs:} The criterion remains $\mathcal{I}_{\text{PASTIS}}(\mathcal{B}) = \hat{\ell}(\mathcal{B}) - n_\mathcal{B} \ln \frac{n_0}{p}$, where:
            \begin{itemize}
                \item $\hat{\ell}(\mathcal{B})$ is the log-likelihood from the discretized SPDE system.
                \item $n_0$ is the number of candidate vector basis functions for the SPDE terms.
            \end{itemize}
        \item \textbf{Benchmark (Gray-Scott):} PASTIS successfully identified the $n^*=7$ true terms from an $n_0=78$ library.
            \begin{itemize}
                \item Achieved high Exact Match accuracy, outperforming AIC, CV, and converging faster than BIC.
                \item Resulted in low Prediction Error, reflecting accurate model structure.
            \end{itemize}
        \item \textbf{Significance:} PASTIS effectively handles the large libraries typical of SPDE inference, controlling for multiple comparisons to find parsimonious, physically relevant models.
    \end{itemize}
\end{tcolorbox}

%% file: tex_body/chap8.tex
\chapter{Robustness to Data Imperfections}
\label{chap:robustness_sde} 
\chaptertoc{}

Real-world experimental data often deviate from the idealized conditions assumed in the basic likelihood formulation (e.g., Eq.~\eqref{eq:likelihood_chapN1}). Two common challenges are large sampling time intervals ($\Delta t$) between measurements, and the presence of measurement noise ($\sigma$) superimposed on the true system state. Standard inference methods can yield biased or inaccurate results when applied naively to such data. This chapter discusses modifications to the inference framework and the PASTIS methodology to enhance robustness against these data imperfections, building upon the robust coefficient estimators introduced in Chapter~\ref{chap:advanced_inference}. More precisely, we will leverage the robust parameter estimators introduced in \Cref{chap:advanced_inference}: the trapeze rule adaptation for large sampling intervals ($\Delta t$) and the shift estimator for measurement noise ($\sigma$). 
These estimators will be used in conjunction with modified likelihood formulations, which we will derive in this chapter, to build a robust inference framework capable of handling these data imperfections.

\section{Challenges of Non-Ideal Data for Model Selection}
\label{sec:robust_challenges}

The basic SDE coefficient estimators (e.g., Eq.~\eqref{eq:SFI-Ito_chapN1}) and the associated likelihood functions (e.g., Eq.~\eqref{eq:likelihood_chapN1}) typically rely on approximations valid in the limit of small sampling intervals, $\Delta t \to 0$. When applying model selection criteria like PASTIS, both the parameter estimation step and the likelihood calculation itself must be robust to data imperfections. As discussed in Chapter~\ref{chap:advanced_inference}, the main issues are:
\begin{itemize}
    \item \textbf{Large $\Delta t$:} When the time between samples ($\Delta t$) is significant relative to the system's characteristic timescales, the finite difference $\Delta \bm{x}_t / \Delta t$ poorly approximates the instantaneous velocity $\dd{\bm{x}}/\dd{t}$.
    This introduces discretization errors (bias) in the estimation of both the drift $\bm{f}$ and potentially the diffusion $\bm{D}$, affecting subsequent model selection.
    \item \textbf{Measurement Noise:} If observations are corrupted by noise, $\bm{y}_t = \bm{x}_t + \bm{\eta}_t$ with $\bm{\eta}_t \sim \mathcal{N}(\bm{0}, \bm{\sigma}^2)$, standard estimators based directly on $\bm{y}_t$ suffer from biases often scaling as $\mathcal{O}(\sigma^2 / \Delta t)$. These biases can dominate the true signal, especially for high noise levels or small sampling intervals, leading to incorrect inference and model selection.
\end{itemize}
Addressing these sources of error is crucial for reliably applying SDE/SPDE inference and model selection techniques to experimental or real-world observational data.

\section{Robustness to Large Sampling Intervals (PASTIS-$\Delta t$)}
\label{sec:robust_deltat}

To improve robustness against large sampling intervals $\Delta t$, we need to adapt the likelihood calculation used within PASTIS. Recalling the derivation of the likelihood (Chapter~\ref{subsection:likelihood_derivation}), a key step often involves approximating the increment $\Delta \bm{x}_t$ using the Euler-Maruyama scheme: $\Delta \bm{x}_t \approx \bm{f}(\bm{x}_t) \Delta t + \sqrt{2 \bm{D}}\Delta \bm{W}_t$. This approximation breaks down for large $\Delta t$.

A more accurate approach starts from the integral form and uses a higher-order approximation for the drift integral, inspired by the trapeze rule: 
\begin{equation}
    \Delta \bm{x}_t \approx \frac{1}{2} (\bm{f}(\bm{x}_{t+\Delta t}) + \bm{f}(\bm{x}_{t}))\Delta t + \int_t^{t+\Delta t} \sqrt{2\bm{D}(\bm{x}_{t'})}\dd{\bm{W}_{t'}} 
    \label{eq:trapez_approx_dx}
\end{equation}
The residual term, $\Delta \bm{x}_t - \frac{\bm{f}(\bm{x}_{t+\Delta t}) + \bm{f}(\bm{x}_{t})}{2} \Delta t$, approximates the noise integral $\int_t^{t+\Delta t} \sqrt{2\bm{D}(\bm{x}_{t'})}\dd{\bm{W}_{t'}}$, which is Gaussian with zero mean and covariance related to $\bm{D}$ over the interval $\Delta t$. This suggests constructing a likelihood based on this residual.

Furthermore, the standard diffusion estimator $\bar{\bm{D}} = \left\langle \frac{\Delta \bm{x}_t \Delta \bm{x}_t^\top}{2 \Delta t} \right\rangle$ can be biased by drift effects when $\Delta t$ is large.
Indeed, this estimator satisfies the approximation
\[
    \bar{\bm{D}} \approx \left\langle \bm{D} \right\rangle + \left\langle \bm{f}\bm{f}^\top \right\rangle \frac{\Delta t}{2},
\]
which shows that it approximates $\left\langle \bm{D} \right\rangle$ with a bias term of order $\mathcal{O}(\Delta t)$ dependent on the drift.
A more robust estimator, designed to cancel this leading-order drift bias by using three points ($\bm{x}_{t+\Delta t}, \bm{x}_t, \bm{x}_{t-\Delta t}$), is given by:
\begin{equation}
    \bm{\hat{D}}_{\Delta t}(\bm{x}_t) = \frac{1}{4 \Delta t}(\Delta\bm{x}_{t} - \Delta \bm{x}_{t-\Delta t}) (\Delta\bm{x}_{t} - \Delta \bm{x}_{t-\Delta t})^\top.
    \label{eq:3pt_diffusion_estimator}
\end{equation}
A similar version was introduced in \cite{ragwitzIndispensableFiniteTime2001}. Combining these ideas leads to a modified log-likelihood function, $\ell_{\Delta t}$, designed to be more accurate for larger $\Delta t$: 
\begin{multline}
  \ell_{\Delta t}(\bm{f}\mid\bm{X}_{\tau}) =
  - \tau \avg{\left(\frac{\Delta \bm{x}_t}{\Delta t} - \frac{\bm{f}(\bm{x}_{t+\Delta t}) + \bm{f}(\bm{x}_{t})}{2}  \right)\cdot (4\bar{\bm{D}}_{\Delta t})^{-1} \cdot \left(\frac{\Delta \bm{x}_t}{\Delta t} - \frac{\bm{f}(\bm{x}_{t+\Delta t}) + \bm{f}(\bm{x}_{t})}{2}  \right)} \\
  -\sum_{\alpha,\beta,\gamma} \frac{\tau}{2} \avg{\hat{D}_{\Delta t,\gamma \beta }(\bm{x}_t)\pdv{f_\alpha(\bm{x}_t)}{x_\beta}(\bar{\bm{D}}_{\Delta t})^{-1}_{\gamma\alpha}} 
  \label{eq:log_likelihood_stato_conceptual} 
\end{multline}
where $\bar{\bm{D}}_{\Delta t} = \avg{ \bm{\hat{D}}_{\Delta t}(\bm{x}_t)}$ denotes the time-averaged robust diffusion estimate. 
The correction term $\sum_{\alpha,\beta,\gamma} \frac{\tau}{2} \avg{\hat{D}_{\Delta t,\gamma \beta }(\bm{x}_t)\pdv{f_\alpha(\bm{x}_t)}{x_\beta}(\bar{\bm{D}}_{\Delta t})^{-1}_{\gamma\alpha}}$ accounts for interactions between dynamical noise and $\bm{f}(\bm{x}_{t+\Delta t})$, ensuring consistency in the small $\Delta t$ limit such that $\lim_{\Delta t \to 0} \ell_{\Delta t}(\bm{f}\mid\bm{X}_{\tau}) = \ell(\bm{f}\mid\bm{X}_{\tau})$. It can also be seen as a Stratonovich transformation of the approximate log-likelihood $\ell$. 

For model selection, we use this robust likelihood $\ell_{\Delta t}$ within the PASTIS criterion. The coefficients $\hat{\bm{\alpha}}$ for a given basis $\mathcal{B}$ used to evaluate $\bm{f} = \sum \hat{\alpha}_j \bm{b}_j$ inside $\ell_{\Delta t}$ should also be estimated robustly. To achieve that, we use the trapeze estimator developed in \cref{chap:advanced_inference} (\cref{sec:chap3_large_dt}), where we replace $\bar{\bm{D}}$ by $\bar{\bm{D}}_{\Delta t}$ for consistency: 
    \begin{equation}
        \bm{\alpha}^{Tr}_i = \sum_j (\bm{G}_{\mathcal{B}}^{\Delta t})^{-1}_{i,j} \avg{\bm{b_j}(\bm{x}_t)^\top \cdot \bar{\bm{D}}_{\Delta t}^{-1} \cdot \left(\frac{\Delta \bm{x}_t}{\Delta t}\right) }, 
        \label{eq:SFI-Trape_chapN5_robustD}
    \end{equation}
where the modified Gram matrix is $\left({G_\mathcal{B}^{\Delta t}}\right)_{ij} = \avg{\bm{b_j}(\bm{x}_t)^\top \cdot \bar{\bm{D}}_{\Delta t}^{-1} \cdot \frac{(\bm{b_i}(\bm{x}_{t+\Delta t}) + \bm{b_i}(\bm{x}_t))}{2}}$. 
Note that this estimator $\hat{\bm{\alpha}}^{Tr}$ is not obtained by minimizing the robust likelihood $\ell_{\Delta t}$.
However, as demonstrated in \cref{sec:chap3_large_dt}, the trapeze estimator provides a reliable and practical improvement for handling larger time steps $\Delta t$. 
Therefore, we employ this previously studied estimator here for its robustness.

We refer to the PASTIS criterion applied using this $\Delta t$-robust likelihood $\ell_{\Delta t}$ \cref{eq:log_likelihood_stato_conceptual} and trapeze coefficient estimation framework \cref{eq:SFI-Trape_chapN5_robustD} as \textbf{PASTIS-$\bm{\Delta t}$}: 
\begin{equation}
   \mathcal{I}_{\text{PASTIS}-\Delta t} = \hat{\ell}_{\Delta t}(\mathcal{B}) - n_{\mathcal{B}} \ln{\frac{n_0}{p}} 
\end{equation}

\begin{figure}[htbp]
    \centering
    \includegraphics[width=\textwidth]{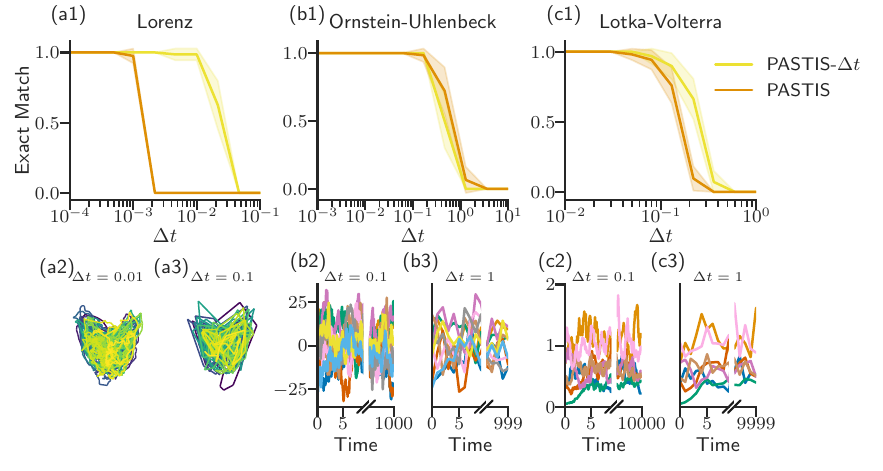} 
    \caption[PASTIS robustness to sampling interval $\Delta t$]{Comparison of PASTIS model selection robustness against increasing sampling intervals ($\Delta t$) using different underlying likelihood estimators.
    \textbf{(Top row, a1-c1)} Exact Match accuracy versus sampling interval $\Delta t$ (log scale) for the Lorenz (a1), Ornstein-Uhlenbeck (b1), and Lotka-Volterra (c1) systems. Performance is compared between standard PASTIS (orange), utilizing the Approximate Maximum Likelihood (AML), and 'PASTIS-$\Delta t$' (yellow), employing the likelihood framework adapted for larger $\Delta t$ (e.g., based on Eq.~\eqref{eq:log_likelihood_stato_conceptual} and Eq.~\eqref{eq:SFI-Trape_chapN5_robustD}).
    \textbf{(Bottom row, a2-c3)} Sample trajectories or phase portraits illustrating the dynamics of each system captured at different sampling intervals ($\Delta t = 0.01, 0.1$ or $1$). The different sampling intervals are specifically chosen to show the transition regime for the PASTIS-$\Delta t$ method.}
    \label{fig:pastis_delta_t_robustness}
\end{figure}

Figure~\ref{fig:pastis_delta_t_robustness} empirically validates the effectiveness of this $\Delta t$-robust approach, revealing system-dependent improvements. The top panels (a1-c1) directly compare the Exact Match accuracy of standard PASTIS (orange), which relies on the Approximate Maximum Likelihood (AML), against PASTIS-$\Delta t$ (yellow), which uses the modified likelihood framework. The degree of improvement varies significantly across the systems tested:
\begin{itemize}
    \item For the \textbf{Lorenz system (a1)}, the improvement is dramatic. Standard PASTIS fails rapidly as the sampling interval $\Delta t$ increases, while PASTIS-$\Delta t$ maintains high accuracy over a substantially wider range of $\Delta t$.
    \item For the \textbf{Lotka-Volterra system (c1)}, characterized by multiplicative noise, PASTIS-$\Delta t$ demonstrates a slight improvement in robustness, extending the range of reliable model identification compared to the standard version.

    \item For the linear \textbf{Ornstein-Uhlenbeck process (b1)}, PASTIS-$\Delta t$ shows limited improvement over standard PASTIS. Both methods exhibit relatively similar performance degradation as $\Delta t$ increases, which contrasts with the significant gains PASTIS-$\Delta t$ provided for the nonlinear systems studied (such as the Lorenz system). This observation suggests that the specific discretization challenges targeted by PASTIS-$\Delta t$ are less critical for this linear benchmark. \textbf{Although} the trapeze estimator $\hat{\bm{\alpha}}^{Tr}$ is designed for greater robustness in coefficient estimation (\cref{sec:chap3_large_dt}), the overall model selection accuracy also depends heavily on the likelihood estimation $\ell_{\Delta t}$. 
    
    It is possible that for the linear OU process at large $\Delta t$, challenges related to accurately estimating the diffusion $\bar{\bm{D}}_{\Delta t}$ or the specific form of the correction term within $\ell_{\Delta t}$ (which depends on the constant gradient of $\bm{f}$) become the limiting factors. This would diminish the relative advantage gained from the improved drift handling in PASTIS-$\Delta t$.
\end{itemize}

The bottom panels (a2-c3) provide visual context by showing system trajectories sampled at different $\Delta t$, illustrating the increasing challenge for inference as the sampling becomes sparser. These results confirm that incorporating estimators and likelihood formulations designed to handle larger time steps can be crucial for applying PASTIS successfully, particularly for nonlinear systems sensitive to discretization errors, although the magnitude of the benefit depends on the specific dynamics.

\section{Correction for Measurement Noise (PASTIS-$\sigma$)}
\label{sec:robust_sigma}

To handle additive measurement noise in the observations, $\bm{y}_t = \bm{x}_t + \bm{\eta}_t$ where $\bm{\eta}_t \sim \mathcal{N}(\bm{0}, \bm{\sigma}^2)$ (assuming $\bm{\sigma}^2$ is the covariance matrix), standard likelihood calculations and coefficient estimators become biased, particularly for small $\Delta t$. 
To create a robust version of PASTIS, termed \textbf{PASTIS-$\sigma$}, we adapt the underlying components:

\begin{enumerate}

    \item \textbf{Robust Diffusion Estimator:} Measurement noise also biases standard diffusion estimators. We employ a corrected estimator $\bm{\hat{D}_{\sigma}}$ designed to be less sensitive to measurement noise, such as the one proposed by Vestergaard et al. \cite{vestergaardOptimalEstimationDiffusion2014}:
    \begin{equation}
\bm{\hat{D}_{\sigma}}(\bm{y}_t) = \frac{1}{2 \Delta t} \Delta\bm{y}_{t} \Delta\bm{y}_{t}^\top + \frac{1}{\Delta t}\avg{\Delta \bm{y}_{t+\Delta t} \Delta \bm{y}_t^\top}
    \label{eq:robust_diffusion_estimator_sigma} 
    \end{equation}
    
    \item \textbf{Shift Likelihood Formulation:} We modify the likelihood calculation to mitigate the primary bias term ($\mathcal{O}(\bm{\sigma}^2 / \Delta t)$) caused by the correlation between the observed state and its increment due to noise. This is achieved by evaluating the drift term contribution using lagged observations within the likelihood expression:
    \begin{multline}
    \ell^{\text{shift}}(\bm{f}\mid\bm{Y}) \approx -\tau \left\langle \left(\frac{\Delta \bm{y}_{t}}{\Delta t} - \sum_{i=1}^{n_{\mathcal{B}}} \alpha_i \bm{b}_i(\bm{y}_{t-\Delta t})\right)^\top \left(4\bm{\bar{D}_{\sigma}}\right)^{-1} \right. \\ \left. \left(\frac{\Delta \bm{y}_{t}}{\Delta t} - \sum_{k=1}^{n_{\mathcal{B}}} \alpha_k \bm{b}_k(\bm{y}_{t-\Delta t})\right) \right\rangle 
    \label{eq:shift_likelihood_sigma}
    \end{multline}
    with $\bm{\bar{D}_{\sigma}}=\avg{\bm{\hat{D}_{\sigma}}(\bm{y}_t)}$. 
    Evaluating the basis functions $\bm{b}_k$ at the lagged time $t-\Delta t$ breaks the problematic correlation with the noise $\bm{\eta}_t$ affecting $\Delta \bm{y}_t$. Note that this shifting introduces a small error that scales with $\Delta t$, meaning it works best when measurement noise is the dominant error source compared to discretization error.

    \item \textbf{Shift Coefficient Estimator:} We require that the coefficients $\bm{\alpha}$ used within the shift likelihood \eqref{eq:shift_likelihood_sigma} are also estimated robustly.
    We use the Shift coefficient estimator introduced previously (\cref{chap:advanced_inference}):
    \begin{equation}
        \hat{\alpha}^{\text{shift}}_i = \sum_j (\bm{G}_{\mathcal{B}}^{\text{shift}})^{-1}_{ij} \avg{\bm{b_j}(\bm{y}_{t-\Delta t})^\top \cdot \bm{\bar{D}_{\sigma}}^{-1} \cdot \left(\frac{\Delta \bm{y}_{t}}{\Delta t}\right)}, 
        \label{eq:SFI-Shift_chapN5}
    \end{equation}
    where the shift Gram matrix is $\left({G_\mathcal{B}^{\text{shift}}}\right)_{ij} = \avg{\bm{b_i}(\bm{y}_{t-\Delta t})^{\top} \cdot \bm{\bar{D}_{\sigma}}^{-1} \cdot \bm{b_j}(\bm{y}_{t})}$. 
\end{enumerate}
PASTIS-$\sigma$ then proceeds by applying the standard PASTIS criterion (Eq.~\eqref{eq:PASTIS_chapN2}) but using $\ell^{\text{shift}}$ calculated with $\hat{\bm{\alpha}}^{\text{shift}}$ and $\bm{\bar{D}_{\sigma}}$): 
\begin{equation}
   \mathcal{I}_{\text{PASTIS}-\sigma} = \hat{\ell}^{\text{shift}}(\mathcal{B}) - n_{\mathcal{B}} \ln{\frac{n_0}{p}} 
\end{equation}

\begin{figure}[htbp]
    \centering
    \includegraphics[width=\textwidth]{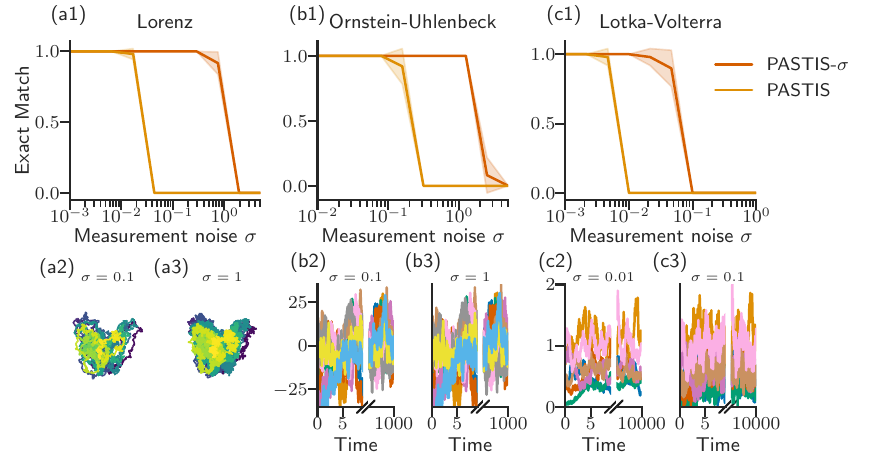}
    \caption[PASTIS robustness to measurement noise $\sigma$]{Comparison of PASTIS model selection robustness against increasing measurement noise standard deviation ($\sigma$).
    \textbf{(Top row, a1-c1)} Exact Match accuracy versus measurement noise $\sigma$ (log scale) for the Lorenz (a1), Ornstein-Uhlenbeck (b1), and Lotka-Volterra (c1) systems. Performance is compared between standard PASTIS (orange), using estimators sensitive to measurement noise, and 'PASTIS-$\sigma$' (red), employing the Shift likelihood and coefficient estimators combined with a robust diffusion estimate (Eqs.~\eqref{eq:shift_likelihood_sigma}-\eqref{eq:SFI-Shift_chapN5}).
    \textbf{(Bottom row, a2-c3)} Sample trajectories or phase portraits illustrating system dynamics observed with different levels of measurement noise ($\sigma = 0.01, 0.1$ or $1$). The levels of measurement noise are specifically chosen to show the transition regime for the PASTIS-$\sigma$ method.}
    \label{fig:pastis_sigma_robustness}
\end{figure}

Figure~\ref{fig:pastis_sigma_robustness} demonstrates the practical effectiveness of the PASTIS-$\sigma$ framework. The top panels (a1-c1) compare the Exact Match accuracy of standard PASTIS (orange) against PASTIS-$\sigma$ (red) as the measurement noise level $\sigma$ increases. For all three benchmark systems (Lorenz, OU, Lotka-Volterra), the standard PASTIS approach fails rapidly, with accuracy dropping to zero even for moderate noise levels. This highlights the severe impact of measurement noise bias on standard likelihood-based selection. In stark contrast, PASTIS-$\sigma$ maintains nearly perfect accuracy across a significantly wider range of noise intensities, showcasing the success of the Shift estimation framework in mitigating noise-induced bias. The bottom panels (a2-c3) illustrate the appearance of the data under different noise levels. These results confirm that PASTIS-$\sigma$ provides a crucial enhancement for applying sparse model discovery reliably to noisy experimental or observational data.

Note that simultaneously correcting for large sampling intervals $\Delta t$ and high measurement noise $\sigma$ remains a significant challenge.

\section{Conclusion}
\label{sec:robust_conclusion}

This chapter addressed the critical issue of applying sparse model selection techniques, specifically PASTIS, to data affected by real-world imperfections: large sampling intervals ($\Delta t$) and measurement noise ($\sigma$). We identified the biases these imperfections introduce into standard likelihood calculations and coefficient estimations, which can severely compromise model selection accuracy.

To overcome these limitations, we introduced two enhanced versions of the PASTIS framework:
\begin{itemize}
    \item \textbf{PASTIS-$\Delta t$}, which incorporates higher-order time integration approximations (inspired by the trapeze rule) into the likelihood calculation, along with robust diffusion estimation, mitigating discretization bias associated with large $\Delta t$. 
    \item \textbf{PASTIS-$\sigma$}, which employs a Shift estimator framework for both coefficients and likelihood evaluation, coupled with robust diffusion estimation, to effectively counteract the bias induced by measurement noise.
\end{itemize}

Numerical benchmarks demonstrated the practical value of these adaptations. PASTIS-$\Delta t$ showed substantially improved performance over standard PASTIS when sampling intervals were large, particularly for nonlinear systems like the Lorenz model, although the degree of improvement was system-dependent. PASTIS-$\sigma$ provided dramatic gains in robustness, maintaining high model identification accuracy across a wide range of measurement noise levels where the standard approach failed entirely.

These results underscore the necessity of accounting for data imperfections when inferring models from experimental or observational time series. While simultaneously addressing both large $\Delta t$ and significant $\sigma$ remains an ongoing challenge, the PASTIS-$\Delta t$ and PASTIS-$\sigma$ methods presented here provide crucial extensions, enabling more reliable and accurate sparse discovery of stochastic dynamics from non-ideal data commonly encountered in scientific and engineering applications.

%% file: tex_body/chap9.tex
\chapter{Discussion on PASTIS}
\label{chap:part2_discussion_conclusion}

The preceding chapters in \cref{part:Model_selection} (specifically, \cref{chap:likelihood_aic} through \cref{chap:robustness_sde}) detailed the development, validation, extension, and refinement of the Parsimonious Stochastic Inference (PASTIS) framework for sparse model selection of stochastic dynamics. This discussion covers the broader context, strengths, limitations, and implications arising from this body of work.

The core motivation, outlined in \cref{chap:likelihood_aic}, was the inadequacy of standard information criteria like AIC and BIC for reliably identifying sparse governing equations from large candidate libraries---a common scenario in data-driven modeling of complex systems. AIC tends to overfit by including superfluous terms (see \cref{sec:likelihood_aic_limits}), while BIC, though consistent, may converge slowly and does not explicitly penalize based on the library size $n_0$ (discussed in \cref{sec:likelihood_bic}). Introduced in \cref{chap:pastis}, PASTIS directly addresses this by leveraging extreme value statistics to formulate a penalty, $n_\mathcal{B} \log(n_0/p)$, that accounts for the multiple comparisons inherent in searching large model spaces. This principled approach provides a statistically grounded method for controlling the inclusion of non-essential terms, tunable via the parameter $p$, which can be interpreted as approximately the probability of including a superfluous term (detailed in \cref{sec:pastis_criterion}).

Validation across diverse benchmark systems (Lorenz, OU, Lotka-Volterra), presented in \cref{chap:validation}, confirmed the practical advantages of PASTIS. As shown in \cref{subsec:results_accuracy}, it consistently outperformed AIC, BIC, CV, and SINDy in accurately identifying the true sparse model structure, especially demonstrating faster convergence than BIC and avoiding the overfitting pitfalls of AIC and CV. Furthermore, the models selected by PASTIS yielded excellent predictive performance (\cref{subsec:results_prediction}).

The successful extension of PASTIS to SPDEs, demonstrated in \cref{chap:pastis_spde} using the Gray-Scott system, highlighted the framework's scalability to high-dimensional problems arising from spatial discretization.

Recognizing the practical limitations of real-world data (\cref{chap:robustness_sde}), the development of PASTIS-$\Delta t$ (\cref{sec:robust_deltat}) and PASTIS-$\sigma$ (\cref{sec:robust_sigma}) significantly enhanced the framework's robustness. By incorporating robust coefficient estimators and modified likelihood formulations, these extensions effectively mitigate biases arising from large sampling intervals ($\Delta t$) and measurement noise ($\sigma$), respectively. The benchmarks presented in \cref{fig:pastis_delta_t_robustness,fig:pastis_sigma_robustness} showed substantial improvements, particularly the dramatic gains of PASTIS-$\sigma$ against noise, making the approach much more applicable to experimental data.

Despite the demonstrated strengths of the PASTIS framework, several limitations warrant discussion and point towards important directions for future research, expanding on \cref{sec:chap5_discussion}.

\textbf{Computational Scalability and Optimality:} A primary challenge lies in the computational cost of model selection. Searching the full model space involves evaluating $2^{n_0}$ candidates, which is computationally intractable for large $n_0$. The greedy search algorithm detailed in \cref{sec:pastis_search} provides a pragmatic solution for moderate $n_0$ (e.g., $n_0< 100$), but it is susceptible to converging to local optima and does not guarantee identification of the globally best model. Future work could explore alternative search strategies, although guaranteeing global optimality remains inherently difficult for this NP-hard problem. 

\textbf{Dependence on Basis Library:} The effectiveness of PASTIS is sensitive to the initial choice of the candidate basis library $\mathcal{B}_0$. This selection remains a user-driven process, often relying heavily on domain expertise. Furthermore, the PASTIS framework implies that larger libraries (increasing $n_0$) demand more data for reliable model identification, as reflected in the penalty term's dependence on $n_0$. Consequently, constructing an appropriately sized and relevant initial library $\mathcal{B}_0$ becomes particularly critical when data is scarce, often necessitating the careful application of prior knowledge.

\textbf{Model Structure Assumptions:} The developments presented in this thesis primarily addressed systems where the drift term is linear in the parameters, i.e., $\bm{f}(\bm{x})=\sum_i \alpha_i \bm{b}_i(\bm{x})$. An important extension would be to investigate how PASTIS can be adapted for models with non-linear parameter dependencies, for instance, of the form $\bm{f}(\bm{x})=\sum_i \bm{b}_i(\bm{x}, \alpha_i)$. This would broaden the class of systems analyzable by the PASTIS framework.

\textbf{Handling Partial Observability:} A common scenario in practice is that only a subset of system variables is measured. The presence of hidden states poses significant challenges for system identification and model selection. Integrating the PASTIS methodology with techniques designed for partially observed systems is a crucial step towards applying the framework to more complex, real-world experimental data.

\textbf{Combined Robustness Challenges:} While robustness enhancements were developed, they primarily addressed large sampling intervals ($\Delta t$) and measurement noise ($\sigma$) in isolation (\cref{chap:robustness_sde}). Developing estimators and selection criteria that are simultaneously robust to both effects, especially when dealing with state-dependent diffusion, remains a significant challenge for future work (as noted in \cref{sec:robust_conclusion}).

\textbf{Underlying Theoretical Assumptions:} The theoretical derivation of PASTIS itself relies on certain assumptions whose validity might not hold universally in practice. Key assumptions include the approximate independence of log-likelihood gains for superfluous terms and the reliance on asymptotic limits (\cref{sec:pastis_criterion}). Potential deviations from these ideal conditions might explain observations such as the slight discrepancy noted for the Lotka-Volterra system with multiplicative noise (\cref{fig:p_influence_OU_chapN2}). Furthermore, the extension to SPDEs (\cref{chap:pastis_spde}) implicitly assumes factors like adequate spatial resolution and potentially simplified boundary conditions or noise structures; the impact of these assumptions merits further investigation (\cref{sec:chap5_conclusion}).

Nonetheless, the PASTIS framework, including its robust variants, represents a significant step forward. It provides a theoretically motivated, validated, and practical methodology specifically tailored for uncovering sparse representations of stochastic dynamics (SDEs) from data, navigating the challenges posed by large model spaces and imperfect measurements. Importantly, the core statistical foundation of PASTIS, which employs extreme value theory to derive a penalty accounting for the vast size of the model search space, is not inherently restricted to SDEs and can be applied to any likelihood maximization problems. 
This suggests potential for generalizing the PASTIS principle to other scientific domains confronted with similar challenges of selecting parsimonious models from large candidate libraries via likelihood-based inference. Its demonstrated ability to control false discoveries statistically makes it a valuable tool for robust scientific inference across disciplines where stochasticity and model complexity are prevalent.

%% file: tex_body/conc.tex
\chapter{General Conclusion and Perspectives}
\label{chap:global_conclusion}

This thesis addressed the challenge of discovering mathematical models of stochastic dynamical systems directly from observational time-series data. Recognizing that such systems, often described by Stochastic Differential Equations (SDEs) or Stochastic Partial Differential Equations (SPDEs), are ubiquitous across scientific and engineering disciplines, the goal was to develop and validate a robust, data-driven workflow encompassing both parameter inference and model structure selection, particularly when faced with realistic data limitations.

\Cref{part:Introdution_langevin} laid the necessary groundwork by first introducing the theoretical framework of SDEs, rooted in the Langevin equation and requiring the tools of stochastic calculus (\cref{chap:langevin_intro,chap:sde_toolbox,chap:inference_methods}).

Then in \Cref{part:Learning_Langevin_equation}, we established foundational methods for inferring the drift and diffusion components of SDEs from discrete data, focusing on practical approximations for maximum likelihood estimators (\cref{chap:inference_langevin}). Acknowledging that real-world data rarely conform to idealized assumptions, a significant focus of \Cref{part:Learning_Langevin_equation} was on enhancing the robustness of these inference techniques (\cref{chap:advanced_inference}). We developed specific estimators, namely the trapeze estimator and the Shift estimator, to effectively mitigate the biases introduced by large sampling intervals ($\Delta t$) and additive measurement noise ($\sigma$), respectively. 

This pursuit of robustness extended to the inference of SPDEs (\cref{chap:spde_inference}), where we adapted these techniques to handle the high-dimensional systems arising from spatial discretization, while acknowledging the associated computational and conceptual challenges. The key outcome of \Cref{part:Learning_Langevin_equation} was thus a suite of validated inference methods capable of providing reliable parameter estimates even when data are sparse in time or corrupted by noise.

Building upon robust parameter estimation, \Cref{part:Model_selection} tackled the crucial, subsequent step of model selection: identifying the most parsimonious and accurate model structure from a potentially vast library of candidate terms. We first examined standard methods like AIC, BIC, and CV, highlighting their limitations, particularly AIC's and CV's propensity to select models with superfluous terms (\cref{chap:likelihood_aic}). 

To overcome this, we proposed the Parsimonious Stochastic Inference (PASTIS) criterion, a novel approach grounded in extreme value statistics designed specifically to handle the multiple comparisons problem inherent in searching large model spaces (\cref{chap:pastis}). The PASTIS penalty term explicitly incorporates the library size ($n_0$) and a user-tunable significance parameter ($p$), offering statistical control over the inclusion of superfluous parameters. 
Extensive validation (\cref{chap:validation}) demonstrated PASTIS's superior performance compared to established criteria in accurately recovering the true sparse structure of various SDE systems and yielding models with high predictive accuracy. The framework's utility was further confirmed by its successful application to SPDE model selection (\cref{chap:pastis_spde}). 

Finally, mirroring the efforts in \Cref{part:Learning_Langevin_equation}, we enhanced PASTIS itself to be robust against data imperfections, developing PASTIS-$\Delta t$ and PASTIS-$\sigma$ variants that maintain high selection accuracy even with large sampling intervals or significant measurement noise (\cref{chap:robustness_sde}).

In synthesis, this thesis presents an integrated and robust methodology for data-driven discovery of stochastic dynamical models. By first establishing reliable techniques for parameter inference under non-ideal conditions (\Cref{part:Learning_Langevin_equation}) and then developing a statistically principled criterion for selecting the most appropriate sparse model structure (\Cref{part:Model_selection}), the work provides a comprehensive workflow applicable to challenging real-world problems. The PASTIS framework, including its robust extensions, stands as the central contribution, offering a powerful tool that balances model fidelity with parsimony while explicitly accounting for the statistical challenges of large search spaces and imperfect data.

The methods developed herein have broad applicability in fields relying on SDE and SPDE models, including physics, quantitative biology, neuroscience, finance, and climate science. By enabling more accurate and reliable model identification from experimental or observational data, this work can contribute to a deeper understanding of the underlying mechanisms governing complex stochastic phenomena.

Furthermore, the core principle of parsimonious model selection embodied by PASTIS extends beyond the domain of stochastic dynamics. For instance, in computational biology, Direct Coupling Analysis (DCA) infers protein contacts or interaction partners from Multiple Sequence Alignments (MSAs) using Potts models \cite{morcosDirectcouplingAnalysisResidue2011,bitbolInferringInteractionPartners2016,rossetAdabmDCA20Flexible2025}. The PASTIS framework could potentially refine these models by identifying and setting to zero irrelevant coupling parameters ($J_{ij}$), potentially improving the analysis. 

Another possible avenue lies in statistical physics and machine learning, specifically in the context of spiked Wigner model inference, where identifying the underlying sparse structure is a key challenge that aligns well with the capabilities developed in this thesis \cite{mukherjeeConsistentModelSelection2025,barbierMutualInformationSymmetric2016}.

Looking ahead, several avenues for future research emerge. Methodologically, further enhancements could include refining the model search strategy (e.g., improving greedy search efficiency or exploring alternatives), developing estimators robust to the simultaneous challenges of large sampling intervals ($\Delta t$) and significant measurement noise ($\sigma$), extending the framework to accommodate more complex noise structures, and incorporating techniques to infer dynamics even when some system variables are unobserved (hidden dynamics). A crucial complementary direction involves applying the complete inference and selection workflow to challenging datasets from specific scientific domains to uncover novel dynamical insights and further validate the framework's practical utility.

In conclusion, the PASTIS framework, including its robust variants, represents a significant advancement in the data-driven discovery of stochastic dynamical systems. It provides a theoretically motivated, validated, and practical methodology specifically tailored for uncovering sparse representations of SDEs and discretized SPDEs from observational data. By navigating the challenges posed by large model spaces and imperfect measurements, PASTIS offers a powerful approach to scientific inference. Importantly, its core statistical foundation, which employs extreme value theory to derive a penalty accounting for the vastness of the model search space, is not inherently restricted to stochastic dynamics. This suggests promising potential for generalizing the PASTIS principle to other scientific domains confronted with similar challenges of selecting parsimonious models from large candidate libraries via likelihood-based inference. Its demonstrated ability to control false discoveries statistically makes it a valuable tool for robust scientific inquiry across disciplines where stochasticity and model complexity are prevalent.

\epigraph{%
"je rentre à\,ma\,maison."}{Eric~Cartman, \textit{South Park}}

%% file: tex_behind/annexes.tex
\addpart{ANNEXES}

\input{tex_behind/annexe0}
\input{tex_behind/annexe1}

\input{tex_behind/annexe_Wilks_theorem}
\input{tex_behind/appendix_part_2}
\input{tex_behind/annexe_TP_FP_FN}

%% file: tex_behind/annexe0.tex
\chapter{Neglecting Correlation in Higher-Order Calculations}
\label{appendix:OU_neglecting_correlation}
Here, we motivate our widely used approximation for neglecting the correlation between $G$ and $\int \dd{W_t}$. This approximation is used, for instance, in calculating the mean squared error, taking a form such as:
\begin{equation}
    A = \E{G^{-1}_{i,j}G^{-1}_{i,k}\int_0^\tau b_i(x_t) dW_t \int_0^\tau b_j(x_s) dW_s} \approx \E{G}^{-1}_{i,j} \E{G}^{-1}_{i,k} \E{\int_0^\tau b_i(x_t) b_j(x_t) \dd{t}}
\end{equation}
To motivate this approximation, we treat, as an example, an Ornstein-Uhlenbeck process in one dimension:
\begin{equation}
    \dd{x_t} = a x_t \dd{t} + \dd{W_t}
\end{equation}
where $W_t$ is a Wiener process. The significant advantage of this very simple model is that an exact analytical solution exists:
\begin{equation}
    x_t = \int_0^t e^{-a (t-s)}  \dd{W_s}
\end{equation}
with $x_0 = 0$. If we use a simple basis $\mathcal{B} = \{x\}$, the Gram matrix is a scalar and is equal to
\begin{equation}
        G = \int_0^\tau x_t^2 \dd{t} = \int_0^\tau \left(\int_0^t e^{-a (t-s)}  \dd{W_s}\right)^2 \dd{t}
\end{equation}
We are going to express $G$ as a sum of its average and its fluctuating part. To do so, we need to write:
\begin{align}
    &\dd{\left(x_t^2e^{2at}\right)} = 2\int_0^t e^{a (s+t)} \dd{W_s} \dd{W_t}  + e^{2at} \dd{t}\\
    &\implies x_\tau^2e^{2a\tau} = \frac{e^{2a\tau}-1}{2a} + 2\int_0^{\tau}\int_0^t e^{a(s + t )}  \dd{W_{s}} \dd{W_t} \\
    &\implies x_\tau^2 = \frac{1 - e^{-2a\tau}}{2a} + 2\int_0^{\tau}\int_0^t e^{-a(2\tau - (s + t))}  \dd{W_{s}} \dd{W_t}
\end{align}
where we used $x_0 = 0$ and, for two functions $u,v$: $d(uv) = v\dd{u} + u\dd{v} + \dd{u} \dd{v}$ with the convention that $\dd{W_t}^2 = \dd{t}$. Another possible way to obtain this result is to split the integral $\int_0^\tau \int_0^\tau \dd{W_s} \dd{W_t}$ into two parts: one where $t=s$ and one where $t\neq s$. Thus,
\begin{align}
    G &= \int_0^\tau \frac{1 - e^{-2a t}}{2a} \dd{t} + 2\int_0^\tau  \int_0^{t}\int_0^s e^{-a(2 t - (s' + s))}  \dd{W_{s'}} \dd{W_s}  \dd{t} \\
    &= \frac{\tau}{2a} - \frac{1}{4 a^2}\left(1 - e^{-2a\tau}\right) + 2 \int_0^{\tau}\int_0^s  \int_s^\tau e^{-a(2 t - (s' + s))} \dd{t} \dd{W_{s'}} \dd{W_s}   \\
    &= \underbrace{\frac{\tau}{2a} - \frac{1}{4 a^2}\left(1 - e^{-2a\tau}\right)}_{\E{G}} + \underbrace{\int_0^{\tau}\int_0^s\frac{1}{a} e^{a(s+s')}\left(e^{-2as} - e^{-2a\tau} \right) \dd{W_{s'}} \dd{W_s}}_{I_{(1,1)}[f_1]}
\end{align}
We also obtain that $\E{\left(I_{(1,1)}[f_1]\right)^2} = \frac{\tau}{2 a^3} + \mathcal{O}(1)$. Thus, for large $\tau$, the fluctuations of $I_{(1,1)}[f_1]^2$ become negligible with respect to $\E{G}$. Hence, we write that $G^{-1} \approx \E{G}^{-1} - \E{G}^{-2}I_{(1,1)}[f_1]$. We also use the fact that:
\begin{align}
    \left(\int_0^\tau x_s \dd{W_s}\right)^2 &= G +  2 \int_0^\tau \int_0^s x_s x_{s'} \dd{W_s'}\dd{W_s}\\
    &= \E{G} + I_{(1,1)}[f_1] +  2 \int_0^\tau \int_0^s x_s x_{s'} \dd{W_s'}\dd{W_s}
\end{align}
To go deeper into the calculation, we need to take into account the fluctuations of $x_s x_{s'}$ with $s>s'$:
\begin{align}
    x_s x_{s'} = \frac{1}{2a} \left(e^{-a (s-s')} - e^{-a(s+s')}\right) +  \left(2\int_0^{s'}\int_0^{s''} + \int_{s'}^{s}\int_0^{s'}\right) e^{-a(s + s' - s''-s''')}\dd{W_{s'''}}\dd{W_{s''}}
\end{align}
Thus, we can write:
\begin{equation}
    \left(\int_0^\tau x_s \dd{W_s}\right)^2 = \E{G} + I_{(1,1)}[f_1] +  I_{(1,1)}[f_2] + I_{(1,1,1,1)}[f_3]
\end{equation}
where
\begin{align}
    I_{(1,1)}[f_2] &= \frac{1}{a}\int_0^\tau \int_0^s \left(e^{-a (s-s')} - e^{-a(s+s')}\right) \dd{W_s'}\dd{W_s}\\
    I_{(1,1,1,1)}[f_3] &= 2 \int_0^\tau \int_0^s\left(2\int_0^{s'}\int_0^{s''} + \int_{s'}^{s}\int_0^{s'}\right) e^{-a(s + s' - s''-s''')}\dd{W_{s'''}}\dd{W_{s''}}\dd{W_{s'}}\dd{W_{s}}
\end{align}
Given this simplification, we are able to calculate:
\begin{align}
    A &= \E{\left(\E{G}^{-1} - \E{G}^{-2}I_{(1,1)}[f_1]\right)^2\left(\E{G} + I_{(1,1)}[f_1] +  I_{(1,1)}[f_2] + I_{(1,1,1,1)}[f_3]\right)}\\
    &= \E{G}^{-1} - \E{G}^{-3}\E{I_{(1,1)}[f_1]\left( I_{(1,1)}[f_1] + 2I_{(1,1)}[f_2]\right)}\\
    &= \E{G}^{-1}\left(1 - \E{G}^{-2}\E{I_{(1,1)}[f_1]\left( I_{(1,1)}[f_1] + 2I_{(1,1)}[f_2]\right)}\right)
\end{align}
Furthermore, given the structure of $I_{(1,1)}$, we have:
\begin{align}
    \E{I_{(1,1)}[f_1]^2} &= \int_0^\tau \int_0^s f_1^2 \dd{s'}\dd{s}\\
    &= \frac{\tau}{2 a^3} - \frac{5}{8  a^{4}} + \frac{4  {\left(2  a \tau + 1\right)} e^{-2  a \tau} + e^{-4  a \tau}}{8  a^{4}}
\end{align}
Similarly, we obtain:
\begin{equation}
    \E{I_{(1,1)}[f_1]I_{(1,1)}[f_2]} = \frac{{\left(2 \, a^{2} \tau^{2} + 4 \, a \tau + {\left(2 \, a \tau - 3\right)} e^{\left(2 \, a \tau\right)} + 3\right)} e^{\left(-2 \, a \tau\right)}}{4 \, a^{4}}
\end{equation}
Finally, we obtain:
\begin{equation}
    A = \E{G}^{-1}\left(1 - \E{G}^{-2}\frac{3\tau}{2a^3} + \mathcal{O}(\tau^{-2})\right) = \E{G}^{-1}\left(1 - \mathcal{O}(\tau^{-1})\right)
\end{equation}
Hence, we have shown that our approximation, where $G$ is treated as independent of $\int_0^\tau b_i(x_t) \dd{W_t} \int_0^{\tau} b_i(x_s) \dd{W_s}$, holds up to a factor of $\frac{1}{\tau}$ for the case of an Ornstein-Uhlenbeck process in one dimension. This result could be generalized to a multi-dimensional Ornstein-Uhlenbeck process. But can we generalize it to any other type of process? Certainly not. For instance, consider a pure Brownian process ($\dd{x_t} = \dd{W_t}$); in this case, $x_t = \int_0^{\tau} \dd{W_s}$. Thus, $x_t$ depends equally on all previous realizations of $\dd{W_s}$. If we try to fit this process with a basis $\mathcal{B} = \{x\}$, the Gram matrix will be $G=\frac{\tau}{2} + I_{(1,1)}[2(\tau-s)]$. Its fluctuations are $\E{4 I_{(1,1)}^2} = \frac{\tau^4}{8}$, which are of the same order of magnitude as $\E{G}$. Thus, it would not be possible to write $E[G^{-1}] \approx \E{G}^{-1} + o(\E{G}^{-1})$ and, by extension, to neglect any type of cross-correlation. I think that the key ingredient for neglecting the cross-correlation is that the drift $f$ confines the trajectory to a finite region of space. More precisely, this implies the existence of attractors. Hence, each kick received by the particle from the dynamical noise will be dampened as the trajectory converges to the attractor position. Unfortunately, it is unclear how to derive a general demonstration without linearizing the original SDE to match the Ornstein-Uhlenbeck case.

%% file: tex_behind/annexe1.tex
\chapter{Bias and Mean Squared Error for large $\Delta t$}
\label{appendix:mse_dt_calculation}

This section provides a detailed derivation of the bias and Mean Squared Error (MSE) for estimators of the drift parameters ${\alpha}$. While similar to the estimators discussed in the main text, the versions analyzed here are evaluated under the simplification that $4 {\bar{D}} = {I}$, where ${I}$ is the identity matrix.

We consider the SDE system in one dimension:
$$ \dd{x_t} = \underbrace{f(x_t)}_{\sum_k \alpha_k {b}_k(x_t)} \dd{t}+ \underbrace{g(x_t)}_{\sqrt{2D(x_t)}} \dd{W_t} $$

Under this condition, we evaluate the following estimators:
\begin{itemize}
    \item  The \textbf{Approximate Maximum Likelihood (AML) estimator}, defined using the matrix $G_{ij} = \langle {b}_i(x_t) \cdot {b}_j(x_t) \rangle$:
    \begin{equation}
        \hat{\alpha}_i = ({G}^{-1})_{ij} \left\langle \frac{{x}_{t+\Delta t} - {x}_{t}}{\Delta t} \cdot {b}_j(x_t) \right\rangle
        \label{eq:appendix_em_estimator}
    \end{equation}

    \item  The \textbf{Trapeze estimator}, defined using the matrix $G^{Tr}_{ij}$:
    \begin{equation}
        \hat{\alpha}_i^{(Tr)} = (({G}^{Tr})^{-1})_{ij} \left\langle \frac{{x}_{t+\Delta t} - {x}_{t}}{\Delta t} \cdot {b}_j(x_t) \right\rangle
        \label{eq:appendix_tr_estimator}
    \end{equation}
   with $G^{Tr}_{ij} = \avg{\frac{{b}_j(x_t) + {b}_j({x}_{t+\Delta t})}{2} \cdot {b}_i(x_t)}$,
\end{itemize}
We now proceed to derive the bias and MSE for these two estimators.

\section{For the Approximate Maximum Likelihood (AML) estimator}
\label{appendix:section_euler_maruyama_mse}
We define the following integrals:
\begin{align}
    I_{(0),t} &= \int_t^{t+\Delta t}\dd{s} = \Delta t\\
    I_{(1),t} &= \int_t^{t+\Delta t} \dd{W_t}\\
    I_{(0,1),t} &= \int_t^{t+\Delta t} \int_t^{s_1} \dd{W_{s_1}} \dd{s_2}\\
    I_{(1,0),t} &= \int_t^{t+\Delta t} \int_t^{s_1} \dd{s_1} \dd{W_{s_2}}\\
    I_{(0,0),t} &= \int_t^{t+\Delta t} \int_t^{s_1} \dd{s_1} \dd{s_2} = \Delta t^2 / 2\\
    I_{(1,1),t} &=  \int_t^{t+\Delta t} \int_t^{s_1} \dd{W_{s_1}} \dd{W_{s_2}}\\
    I_{(1,1,1),t} &=  \int_t^{t+\Delta t} \int_t^{s_1} \int_t^{s_2}\dd{W_{s_1}} \dd{W_{s_2}}\dd{W_{s_3}}
\end{align}
 Then, from \cite[p.~182]{kloedenStochasticTaylorExpansions1992}, we can write the Taylor expansion of $x_{t+\Delta t}$ as:
\begin{multline}
    x_{t+\Delta t} = x_{t} + f I_{(0),t} + g I_{(1),t} + L_0 f I_{(0,0),t} + L_0 g I_{(0,1),t}\\
    + L_1 f I_{(1,0),t} + L_1 g I_{(1,1),t} + L_1 L_1 g I_{(1,1,1),t} + \mathcal{O}(\Delta t^3) + \mathcal{O}_\text{fluc}(\Delta t^{2})
    \label{eq:appendix_taylor_x}
\end{multline}
where $f = f(x_t)$, $g = g(x_t)$, $\mathcal{O}(\Delta t^3)$ includes terms with $I_{(0,0,0),t}$, and $O_{\text{fluc}}(\Delta t^2)$ contains terms such as $I_{(1,1,1,1),t}, I_{(0,1,1),t}$. Then, by manipulating the previous equation, we obtain:
\begin{multline}
    \frac{x_{t+\Delta t} - x_{t}}{\Delta t} = f + g \frac{I_{(1),t}}{\Delta t} + L_0 f \frac{\Delta t}{2} + L_0 g \frac{I_{(0,1),t}}{\Delta t}\\
    + L_1 f \frac{I_{(1,0),t}}{\Delta t} + L_1 g \frac{I_{(1,1),t}}{\Delta t} + L_1 L_1 g \frac{I_{(1,1,1),t}}{\Delta t} + \mathcal{O}(\Delta t^2) + \mathcal{O}_\text{fluc}(\Delta t^{1})
    \label{eq:appendix_taylor_x_1}
\end{multline}
Since we assume that $f$ can be decomposed in terms of basis functions such that $f(x) = \sum_i \alpha_i b_i$, we directly see that our Approximate Maximum Likelihood (AML) estimator $\hat{\alpha}_i = G^{-1}_{ij} \avg{\frac{x_{t+\Delta t} - x_{t}}{\Delta t} \cdot b_j}$ can be written as:
\begin{multline}
    \hat{\alpha}_i = \alpha_i + G^{-1}_{ij} \frac{\Delta t}{2} \avg{L_0 f b_j} + \\
    + G^{-1}_{ij}\pqty{ \avg{b_j\pqty{g \frac{I_{(1),t}}{\Delta t}  + L_0 g \frac{I_{(0,1),t}}{\Delta t}
    + L_1 f \frac{I_{(1,0),t}}{\Delta t} + L_1 g \frac{I_{(1,1),t}}{\Delta t} + L_1 L_1 g \frac{I_{(1,1,1),t}}{\Delta t}}}} + \mathcal{O}(\Delta t^2) + \mathcal{O}_\text{fluc}(\Delta t^{1})
    \label{eq:appendix_taylor_x_2}
\end{multline}

From the definition of $I$, all integrals containing stochastic terms $dW_t$ $\left(I_{(1),t}, I_{(1,0),t}, I_{(0,1),t}, I_{(1,1,1),t}\right)$ will average to zero.  Thus, the bias of $\hat{\alpha}_i$ is:
\begin{equation}
    \E{\hat{\alpha}_i - \alpha_i} = \E{G^{-1}_{ij} \frac{\Delta t}{2} \avg{ b_j L_0 f}} + \mathcal{O}(\Delta t^2)
\end{equation}
For the Mean Squared Error (MSE) $\E{(\hat{\alpha}_i- \alpha_i)^2}$, the computation is a bit more involved. To illustrate the calculation, consider the expectation of:
\begin{align}
    \E{\avg{g \frac{I_{(1),t}}{\Delta t} b_j} \avg{g \frac{I_{(1),t}}{\Delta t}b_k} } &=\frac{1}{N^2}\E{\sum_{t\in \{t_0, \cdots, \tau\}} \sum_{t'\in \{t_0, \cdots, \tau\}} g(x_t) \frac{I_{(1),t}}{\Delta t} b_j(x_{t}) g(x_{t'}) \frac{I_{(1),t'}}{\Delta t} b_k(x_{t'})}\\
    &= \frac{1}{\Delta t N^2}\E{\sum_{t\in \{t_0, \cdots, \tau\}} g(x_t)^2 b_j(x_{t})  b_k(x_{t})}\\
    &= \frac{1}{\tau} \E{\avg{g^2 b_j b_k}}
\end{align}
From this example, we see that all higher-order terms in the MSE coming from products such as $I_{(1),t} I_{(1,0),t}$, $I_{(0,1),t}I_{(1),t}$, and $I_{(1,1,1),t}I_{(1),t}$ will be of order $\frac{\Delta t}{\tau}$. By performing some basic algebra and neglecting the correlation between $G_{ij}^{-1}$ and $I_{(1),t}, I_{(1,0),t}, \dots$, we obtain:
\begin{equation}
    \E{(\hat{\alpha}_i- \alpha_i)^2} = \E{G_{i,j}^{-1} G_{ik}^{-1}} \E{\frac{\Delta t^2}{4} \avg{b_j L_0 f} \avg{b_k L_0 f} + \frac{1}{\tau} \pqty{\avg{g^2 b_j b_k} + \frac{\Delta t}{2}\avg{gL_0g + gL_1f}}}
    \label{eq:appendix_mse_alpha_dt}
\end{equation}
Here, the MSE can be decomposed into the squared bias and the variance.
For a long trajectory ($\tau\to \infty$), the variance, being proportional to $\frac{1}{\tau}$, will be negligible with respect to the bias. For a small trajectory, particular attention should be given to both contributions. An idea, not yet explored, is that since all terms in \cref{eq:appendix_mse_alpha_dt} can be estimated, it could be used to determine whether the error primarily stems from the large sampling interval $\Delta t$, the total time of acquiring data, or both.

\section{For the Trapeze-estimator}
\label{appendix:section_trapeze_mse}
In this section, we derive the bias and the Mean Squared Error (MSE) of the Trapeze estimator. The goal for this estimator is to be more robust against large sampling intervals than the Approximate Maximum Likelihood (AML) estimator. To achieve this, we look at the Taylor expansion of the drift $f$ around $x_t$:
\begin{equation}
    f(x_{t+\Delta t}) = f + L_0 f I_{(0),t} + L_1 f I_{(1),t}+ L_0 L_0 f I_{(0,0),t} + \mathcal{O}(\Delta t^3) + \mathcal{O}_\text{fluc}(\Delta t)
\end{equation}
Then, by using $I_{(0),t} = \Delta t$, $I_{(0,0),t} = \frac{1}{2} \Delta t^2$, and $I_{(0),t} I_{(0),t} = 2 I_{(0,0),t}$, and manipulating the previous equation, we obtain:
\begin{align*}
    f &= f(x_{t+\Delta t}) -  L_0 fI_{(0),t}
    - L_1 f I_{(1),t} - L_0 L_0 f I_{(0,0),t} + \mathcal{O}(\Delta t^2) + \mathcal{O}_\text{fluc}(\Delta t)\\
    \implies f \frac{I_{(0),t}}{2} &= f(x_{t+\Delta t})\frac{I_{(0),t}}{2} - L_0 fI_{(0),t}\frac{I_{(0),t}}{2}  - L_1 f I_{(1),t}\frac{I_{(0),t}}{2} - L_0 L_0 f I_{(0,0),t} \frac{I_{(0),t}}{2} + \mathcal{O}(\Delta t^3) + \mathcal{O}_\text{fluc}(\Delta t^2)\\
    \implies f \frac{I_{(0),t}}{2} &= f(x_{t+\Delta t})\frac{I_{(0),t}}{2} - L_0 fI_{(0,0),t}  - L_1 f\frac{I_{(0,1),t} + I_{(1,0),t}}{2} - \frac{3}{2} L_0 L_0 f I_{(0,0,0),t} + \mathcal{O}(\Delta t^3) + \mathcal{O}_\text{fluc}(\Delta t^2)
\end{align*}
where we used $I_{(0),t} I_{(1),t} = I_{(1,0),t} + I_{(0,1),t}$  (\cite[p.~172]{kloedenStochasticTaylorExpansions1992}). By writing $f I_{(0),t} = \frac{1}{2} f I_{(0),t} + \frac{1}{2} f I_{(0),t}$ and injecting the last equation into \cref{eq:appendix_taylor_x}, we obtain:
\begin{multline}
    x_{t+\Delta t} - x_{t} =  \frac{f(x_t) + f(x_{t+\Delta t})}{2}\Delta t  - \frac{1}{2} L_0L_0 f I_{(0,0,0),t}+ g I_{(1),t} + L_1 f \frac{I_{(1,0),t} - I_{(0,1),t}}{2} \\
 +L_0 g I_{(0,1),t}  + L_1 g I_{(1,1),t} + L_1 L_1 g I_{(1,1,1),t} + \mathcal{O}(\Delta t^4) + \mathcal{O}_\text{fluc}(\Delta t^{2})
 \label{eq:appendix_final_taylor}
\end{multline}
By assuming that the drift $f$ can be decomposed in terms of basis functions ($f(x_t) = \sum_i \alpha_i b_i(x_t)$), we can derive the inferred parameters $\hat{\alpha}_i^{Tr}$ (see \cref{eq:alpha_Tr_estimator}). By using \cref{eq:alpha_Tr_estimator} and \cref{eq:appendix_final_taylor}, we directly observe that:
\begin{multline}
    \hat{\alpha}_i^{Tr} = \alpha_i  - \frac{\Delta t^2}{12} (G^{Tr})^{-1}_{ij}  \avg{b_j L_0 L_0 f } +\\+ \sum_j \left(G^{Tr}\right)^{-1}_{ij} \avg{\frac{b_j}{\Delta t} \pqty{g I_{(1),t} + L_1 f \frac{I_{(1,0),t} - I_{(0,1),t}}{2}
 +L_0 g I_{(0,1),t}  + L_1 g I_{(1,1),t} + L_1 L_1 g I_{(1,1,1),t}}}
 \label{eq:appendix_alpha_fluct}
\end{multline}
By neglecting the correlation between $\left(G^{Tr}\right)^{-1}_{ij}$ and the integrals $I_{(0,1),t}, I_{(1,0),t}, I_{(1,1),t}, I_{(1,1,1),t}$ in \cref{eq:appendix_alpha_fluct}, we obtain:
\begin{equation}
\E{\hat{\alpha}_i^{Tr} - \alpha_i} =  -\frac{\Delta t^2}{12} \E{ (G^{Tr})^{-1}_{ij} \avg{b_j L_0 L_0 f }} + \mathcal{O}(\Delta t^3)
\end{equation}
Hence, by using $G^{Tr}$ as the Gram matrix instead of $G$ in the parameter estimation formula (e.g., \cref{eq:alpha_Tr_estimator}), the bias is reduced from $\mathcal{O}(\Delta t)$ to $\mathcal{O}(\Delta t^2)$.

To obtain the Mean Squared Error $\E{\pqty{\hat{\alpha}_i^{Tr} - \alpha_i}^2}$, we need the following expectations:
\begin{align*}
    \E{I_{(1),t}^2} &= \Delta t\\
    \E{I_{(1),t} I_{(0,1),t}} &= \frac{1}{2} \Delta t^2\\
    \E{I_{(1),t} I_{(1,0),t}}  &= \frac{1}{2} \Delta t^2\\
    \E{I_{(1),t} I_{(1,1,1),t}}  &= \E{I_{(1),t} \left(I_{(1),t}^3 - 3 \Delta t I_{(1),t}  \right)} = 3 \Delta t^2 - 3 \Delta t^2 = 0 \\ 
    \E{I_{(1),t}I_{(1,1),t}} &= 0\\
    \E{I_{(1,1),t}I_{(1,1),t}} &= \E{\pqty{\frac{1}{2}(I_{(1),t}^2-\Delta t)}^2} = \E{\frac{1}{4}\pqty{I_{(1),t}^4 - 2 \Delta t I_{(1),t}^2 + \Delta t^2}} = \frac{1}{2} \Delta t^2 
\end{align*}
Hence, from \cref{eq:appendix_alpha_fluct}, by neglecting the correlation between $G^{Tr}$ and the integrals $I_{(0,1),t}, I_{(1,0),t}, I_{(1,1),t}, I_{(1,1,1),t}$, we obtain:
\begin{multline}
    \E{(\hat{\alpha}^{Tr}_i - \alpha_i)^2} = \E{(G^{Tr})_{i,j}^{-1} (G^{Tr})_{ik}^{-1}} \\ \E{\frac{\Delta t^4}{144} \avg{b_j L_0 L_0 f}  \avg{b_k L_0 L_0 f} + \frac{1}{\tau} \pqty{\avg{g^2 b_j b_k} + \frac{\Delta t}{2}\avg{gL_0g}}}
    \label{eq:appendix_mse_alpha_dt_tr}
\end{multline}

Hence, for a long trajectory, the term proportional to $\frac{1}{\tau}$ becomes negligible, and the dominant term is of order $\mathcal{O}(\Delta t^4)$, as confirmed by simulation in the main text. Nevertheless, numerical verification of the prefactor was not performed as part of this research due to time constraints.

%% file: tex_behind/annexe_Wilks_theorem.tex
\chapter{Wilks' Theorem}
\label{annexe:WILKS}

\section{Introduction}

This chapter derives key asymptotic results for Maximum Likelihood Estimation (MLE) based on a data trajectory $X_N = \{x_1, ..., x_N\}$ of size $N$, when the parameter vector is subject to linear constraints. We first establish the large-sample distribution of the constrained MLE. Subsequently, we derive Wilks' theorem, which corresponds to the asymptotic distribution of the Likelihood Ratio Test (LRT) statistic used for testing such constraints. These results are fundamental for inference in constrained parameter spaces, particularly when analyzing time series or other sequential data. We will primarily utilize the total Fisher information matrix derived from the full data trajectory $X_N$.

\section{Asymptotic Distribution of the Constrained MLE}

\subsection{Model Setup and Reparameterization}

Let $\ell_N(\boldsymbol{\alpha}) := \ell(\boldsymbol{\alpha} \mid X_N)$ be the log-likelihood function based on the data trajectory $X_N$, where $\boldsymbol{\alpha} \in \R^d$ is the parameter vector. We consider the case where $\boldsymbol{\alpha}$ is constrained to lie in a linear subspace defined by a full-rank matrix $\bm{H} \in \R^{d \times r}$, with $r < d$. The constraint is expressed as:
\[
\boldsymbol{\alpha} = \bm{H} \boldsymbol{\epsilon},
\]
for some reduced parameter vector $\boldsymbol{\epsilon} \in \R^r$.

Under this constraint, the log-likelihood becomes a function of $\boldsymbol{\epsilon}$:
\[
\tilde{\ell}_N(\boldsymbol{\epsilon}) := \ell_N(\bm{H} \boldsymbol{\epsilon}).
\]
Let $\hat{\boldsymbol{\epsilon}}$ denote the MLE of $\boldsymbol{\epsilon}$, obtained by maximizing $\tilde{\ell}_N(\boldsymbol{\epsilon})$. The corresponding constrained MLE for $\boldsymbol{\alpha}$ is then $\hat{\boldsymbol{\alpha}}_{c} = \bm{H} \hat{\boldsymbol{\epsilon}}$. We denote the true parameter values as $\boldsymbol{\alpha}_0$ and $\boldsymbol{\epsilon}_0$, satisfying $\boldsymbol{\alpha}_0 = \bm{H} \boldsymbol{\epsilon}_0$.

\subsection{Asymptotic Distribution of the Reduced Parameter Estimator}

Under standard regularity conditions for MLEs (applied to the sequence of log-likelihoods $\ell_N$) and assuming model identifiability, the MLE $\hat{\boldsymbol{\epsilon}}$ is consistent ($\hat{\boldsymbol{\epsilon}} \to \boldsymbol{\epsilon}_0$ as $N \to \infty$) and asymptotically normal.

Let $\mathcal{I}_{\boldsymbol{\alpha}}(\boldsymbol{\alpha})$ denote the total Fisher information matrix for the parameter $\boldsymbol{\alpha}$ based on the sample $X_N$, and $\mathcal{I}_{\boldsymbol{\alpha}_0} = \mathcal{I}_{\boldsymbol{\alpha}}(\boldsymbol{\alpha}_0)$ be this matrix evaluated at the true parameter value $\boldsymbol{\alpha}_0$: 
\[
\mathcal{I}_{\boldsymbol{\alpha}}(\boldsymbol{\alpha}) = -\mathbb{E}\left[ \frac{\partial^2 \ell_N(\boldsymbol{\alpha})}{\partial \boldsymbol{\alpha} \partial \boldsymbol{\alpha}^\top} \right].
\]
Similarly, let $\mathcal{I}_{\boldsymbol{\epsilon}}(\boldsymbol{\epsilon})$ be the total Fisher information matrix for the reduced parameter $\boldsymbol{\epsilon}$, and $\mathcal{I}_{\boldsymbol{\epsilon}_0} = \mathcal{I}_{\boldsymbol{\epsilon}}(\boldsymbol{\epsilon}_0)$:
\[
\mathcal{I}_{\boldsymbol{\epsilon}}(\boldsymbol{\epsilon}) = -\mathbb{E}\left[ \frac{\partial^2 \tilde{\ell}_N(\boldsymbol{\epsilon})}{\partial \boldsymbol{\epsilon} \partial \boldsymbol{\epsilon}^\top} \right].
\]
Using the chain rule ($\frac{\partial^2 \tilde{\ell}_N}{\partial \boldsymbol{\epsilon} \partial \boldsymbol{\epsilon}^\top} = \bm{H}^\top \frac{\partial^2 \ell_N}{\partial \boldsymbol{\alpha} \partial \boldsymbol{\alpha}^\top} \bm{H}$), we find the relationship between the information matrices evaluated at the true values:
\[
\mathcal{I}_{\boldsymbol{\epsilon}_0} = \bm{H}^\top \mathcal{I}_{\boldsymbol{\alpha}_0} \bm{H}.
\]
The asymptotic distribution of $\hat{\boldsymbol{\epsilon}}$ can be expressed using this total information. For large $N$, the distribution of the estimator around the true value is approximately Gaussian:
\[
(\hat{\boldsymbol{\epsilon}} - \boldsymbol{\epsilon}_0) \sim  \N\left(0, \mathcal{I}_{\boldsymbol{\epsilon}_0}^{-1}\right).
\]
More formally, $\sqrt{N}(\hat{\boldsymbol{\epsilon}} - \boldsymbol{\epsilon}_0) \xrightarrow{d} \N(0, \lim_{N\to\infty} [N \mathcal{I}_{\boldsymbol{\epsilon}_0}^{-1}])$ (assuming $\mathcal{I}_{\boldsymbol{\epsilon}_0}$ is $O(N)$ and the limit of $N\mathcal{I}_{\boldsymbol{\epsilon}_0}^{-1}$ exists, or alternatively, if $\mathcal{I}_{\boldsymbol{\epsilon}_0}^*$ is the average Fisher information, then $\sqrt{N}(\hat{\boldsymbol{\epsilon}} - \boldsymbol{\epsilon}_0) \xrightarrow{d} \N(0, (\mathcal{I}_{\boldsymbol{\epsilon}_0}^*)^{-1})$). However, working with the total information $\mathcal{I}_{\boldsymbol{\epsilon}_0}$ directly (as defined) simplifies many expressions for the non-scaled estimator.

\subsection{Asymptotic Distribution of the Constrained Estimator}

The constrained estimator $\hat{\boldsymbol{\alpha}}_{c}$ is a linear transformation of $\hat{\boldsymbol{\epsilon}}$:
\[
\hat{\boldsymbol{\alpha}}_{c} = \bm{H} \hat{\boldsymbol{\epsilon}}.
\]
Using the large-sample approximate distribution for $\hat{\boldsymbol{\epsilon}}$, we find the approximate distribution of $\hat{\boldsymbol{\alpha}}_{c}$:
\[
(\hat{\boldsymbol{\alpha}}_{c} - \boldsymbol{\alpha}_0) = \bm{H} (\hat{\boldsymbol{\epsilon}} - \boldsymbol{\epsilon}_0) \sim  \N\left(0, \bm{H} \mathcal{I}_{\boldsymbol{\epsilon}_0}^{-1} \bm{H}^\top \right).
\]
Substituting the expression for $\mathcal{I}_{\boldsymbol{\epsilon}_0}$, we obtain that the distribution of the constrained estimator around the true value is:
\[
\boxed{
(\hat{\boldsymbol{\alpha}}_{c} - \boldsymbol{\alpha}_0) \sim  \N\left(0, \bm{H} (\bm{H}^\top \mathcal{I}_{\boldsymbol{\alpha}_0} \bm{H})^{-1} \bm{H}^\top \right).
}
\]

\section{Wilks' Theorem: Asymptotic Distribution of the Likelihood Ratio Statistic}

We now derive the asymptotic distribution of the LRT statistic for testing the null hypothesis $H_0: \boldsymbol{\alpha} = \bm{H}\boldsymbol{\epsilon}$ for some $\boldsymbol{\epsilon} \in \R^r$ against the alternative hypothesis $H_A: \boldsymbol{\alpha} \in \R^d$.

\subsection{Estimators and Log-likelihood Expansion}

Let $\hat{\boldsymbol{\alpha}}_{u}$ be the unconstrained MLE maximizing $\ell_N(\boldsymbol{\alpha})$ over $\boldsymbol{\alpha} \in \R^d$. Let $\hat{\boldsymbol{\alpha}}_{c} = \bm{H}\hat{\boldsymbol{\epsilon}}$ be the constrained MLE, as defined previously.

We use a second-order Taylor expansion of $\ell_N(\boldsymbol{\alpha})$ around the true parameter $\boldsymbol{\alpha}_0$:
\[
\ell_N(\boldsymbol{\alpha}) \approx \ell_N(\boldsymbol{\alpha}_0) + (\boldsymbol{\alpha}-\boldsymbol{\alpha}_0)^\top \nabla \ell_N(\boldsymbol{\alpha}_0) + \frac{1}{2} (\boldsymbol{\alpha}-\boldsymbol{\alpha}_0)^\top \nabla^2 \ell_N(\boldsymbol{\alpha}_0) (\boldsymbol{\alpha}-\boldsymbol{\alpha}_0).
\]
Under standard regularity conditions for large $N$:
\begin{itemize}
    \item The Hessian approximates the negative total Fisher information: $\nabla^2 \ell_N(\boldsymbol{\alpha}_0) \approx -\mathcal{I}_{\boldsymbol{\alpha}_0}$.
    \item The score vector at the true parameter is asymptotically normal: $\nabla \ell_N(\boldsymbol{\alpha}_0) \sim  \N(0, \mathcal{I}_{\boldsymbol{\alpha}_0})$.
    \item The estimators relate to the score:
     \begin{align*}
     \hat{\boldsymbol{\alpha}}_{u} - \boldsymbol{\alpha}_0 &\approx \mathcal{I}_{\boldsymbol{\alpha}_0}^{-1} \nabla \ell_N(\boldsymbol{\alpha}_0) \\
     \hat{\boldsymbol{\alpha}}_{c} - \boldsymbol{\alpha}_0 &\approx \bm{H}(\bm{H}^\top \mathcal{I}_{\boldsymbol{\alpha}_0} \bm{H})^{-1} \bm{H}^\top \nabla \ell_N(\boldsymbol{\alpha}_0)
     \end{align*}
\end{itemize}
Substituting these approximations back into the Taylor expansion for $\ell_N(\boldsymbol{\alpha})$ around $\boldsymbol{\alpha}_0$, and using $\nabla^2 \ell_N(\boldsymbol{\alpha}_0) \approx -\mathcal{I}_{\boldsymbol{\alpha}_0}$, we get:
\begin{align*}
\ell_N(\hat{\boldsymbol{\alpha}}_u) &\approx \ell_N(\boldsymbol{\alpha}_0) + (\hat{\boldsymbol{\alpha}}_u - \boldsymbol{\alpha}_0)^\top \nabla \ell_N(\boldsymbol{\alpha}_0) - \frac{1}{2} (\hat{\boldsymbol{\alpha}}_u - \boldsymbol{\alpha}_0)^\top \mathcal{I}_{\boldsymbol{\alpha}_0} (\hat{\boldsymbol{\alpha}}_u - \boldsymbol{\alpha}_0) \\
&\approx \ell_N(\boldsymbol{\alpha}_0) + \nabla \ell_N(\boldsymbol{\alpha}_0)^\top \mathcal{I}_{\boldsymbol{\alpha}_0}^{-1} \nabla \ell_N(\boldsymbol{\alpha}_0) - \frac{1}{2} \nabla \ell_N(\boldsymbol{\alpha}_0)^\top \mathcal{I}_{\boldsymbol{\alpha}_0}^{-1} \mathcal{I}_{\boldsymbol{\alpha}_0} \mathcal{I}_{\boldsymbol{\alpha}_0}^{-1} \nabla \ell_N(\boldsymbol{\alpha}_0) \\
&\approx \ell_N(\boldsymbol{\alpha}_0) + \frac{1}{2} \nabla \ell_N(\boldsymbol{\alpha}_0)^\top \mathcal{I}_{\boldsymbol{\alpha}_0}^{-1} \nabla \ell_N(\boldsymbol{\alpha}_0).
\end{align*}
Similarly, for the constrained estimator:
\[
\ell_N(\hat{\boldsymbol{\alpha}}_c) \approx \ell_N(\boldsymbol{\alpha}_0) + \frac{1}{2} \nabla \ell_N(\boldsymbol{\alpha}_0)^\top \bm{H}(\bm{H}^\top \mathcal{I}_{\boldsymbol{\alpha}_0} \bm{H})^{-1} \bm{H}^\top \nabla \ell_N(\boldsymbol{\alpha}_0).
\]

\subsection{Difference in Log-Likelihoods}

The difference between the maximized log-likelihoods under the unconstrained and constrained models is approximately:
\begin{align*}
\ell_N(\hat{\boldsymbol{\alpha}}_u) - \ell_N(\hat{\boldsymbol{\alpha}}_c) &\approx \frac{1}{2} \nabla \ell_N(\boldsymbol{\alpha}_0)^\top \left[ \mathcal{I}_{\boldsymbol{\alpha}_0}^{-1} - \bm{H}(\bm{H}^\top \mathcal{I}_{\boldsymbol{\alpha}_0} \bm{H})^{-1} \bm{H}^\top \right] \nabla \ell_N(\boldsymbol{\alpha}_0).
\end{align*}

\subsection{Likelihood Ratio Statistic and its Distribution}

The Likelihood Ratio (LR) statistic is defined as:
\[
\lambda = 2 \left[ \ell_N(\hat{\boldsymbol{\alpha}}_u) - \ell_N(\hat{\boldsymbol{\alpha}}_c) \right].
\]
Substituting the approximation for the difference:
\[
\lambda \approx \nabla \ell_N(\boldsymbol{\alpha}_0)^\top \left[ \mathcal{I}_{\boldsymbol{\alpha}_0}^{-1} - \bm{H}(\bm{H}^\top \mathcal{I}_{\boldsymbol{\alpha}_0} \bm{H})^{-1} \bm{H}^\top \right] \nabla \ell_N(\boldsymbol{\alpha}_0).
\]
Let $\mathcal{I}_{\boldsymbol{\alpha}_0}^{1/2}$ be a matrix square root of $\mathcal{I}_{\boldsymbol{\alpha}_0}$. We know that $\bm{Z} = \mathcal{I}_{\boldsymbol{\alpha}_0}^{-1/2} \nabla \ell_N(\boldsymbol{\alpha}_0)$ is asymptotically $\N(0, \bm{I}_d)$, where $\bm{I}_d$ is the $d \times d$ identity matrix. We can rewrite $\nabla \ell_N(\boldsymbol{\alpha}_0) = \mathcal{I}_{\boldsymbol{\alpha}_0}^{1/2} \bm{Z}$. Substituting this into the expression for $\lambda$:
\begin{align*}
\lambda &\approx \bm{Z}^\top \mathcal{I}_{\boldsymbol{\alpha}_0}^{1/2} \left[ \mathcal{I}_{\boldsymbol{\alpha}_0}^{-1} - \bm{H}(\bm{H}^\top \mathcal{I}_{\boldsymbol{\alpha}_0} \bm{H})^{-1} \bm{H}^\top \right] \mathcal{I}_{\boldsymbol{\alpha}_0}^{1/2} \bm{Z} \\
&\approx \bm{Z}^\top \left[ \bm{I}_d - \mathcal{I}_{\boldsymbol{\alpha}_0}^{1/2} \bm{H}(\bm{H}^\top \mathcal{I}_{\boldsymbol{\alpha}_0} \bm{H})^{-1} \bm{H}^\top \mathcal{I}_{\boldsymbol{\alpha}_0}^{1/2} \right] \bm{Z}.
\end{align*}
Let $\bm{P} = \mathcal{I}_{\boldsymbol{\alpha}_0}^{1/2} \bm{H}(\bm{H}^\top \mathcal{I}_{\boldsymbol{\alpha}_0} \bm{H})^{-1} \bm{H}^\top \mathcal{I}_{\boldsymbol{\alpha}_0}^{1/2}$. This matrix $\bm{P}$ is a projection matrix onto the subspace spanned by the columns of $\mathcal{I}_{\boldsymbol{\alpha}_0}^{1/2} \bm{H}$. Thus, we have $\text{rank}(P)=\text{rank}(\mathcal{I}_{\boldsymbol{\alpha}_0}^{1/2} \bm{H})=\text{rank}(\bm{H})=r$, because $\mathcal{I}_{\boldsymbol{\alpha}_0}$ is full rank. Then, the expression simplifies to:
\[
\lambda \approx \bm{Z}^\top (\bm{I}_d - \bm{P}) \bm{Z}.
\]
The matrix $\bm{I}_d - \bm{P}$ is also a projection matrix, projecting onto the orthogonal complement of the subspace spanned by the columns of $\mathcal{I}_{\boldsymbol{\alpha}_0}^{1/2} \bm{H}$. Its rank is $d - \text{rank}(\bm{P}) = d - r$.

By Cochran's theorem, a quadratic form $\bm{Z}^\top \bm{A} \bm{Z}$ where $\bm{Z} \sim \N(0, \bm{I})$ and $\bm{A}$ is an idempotent matrix (projection) follows a chi-squared distribution with degrees of freedom equal to the rank of $\bm{A}$.
Therefore,
\[
\boxed{
\lambda = 2 \left[ \ell_N(\hat{\boldsymbol{\alpha}}_u) - \ell_N(\hat{\boldsymbol{\alpha}}_c) \right] \xrightarrow{d} \chi^2_{d-r}.
}
\]
where $d-r$ can be seen as the number of independent linear constraints imposed by the null hypothesis.

This demonstration was inspired by the methods presented in lecture notes from Charles J. Geyer available at \url{https://www.stat.umn.edu/geyer/8112/notes/tests.pdf}.

\section{Summary}

We have derived two key asymptotic results for MLE under the linear constraint $\boldsymbol{\alpha} = \bm{H}\boldsymbol{\epsilon}$:
\begin{enumerate}
    \item The constrained MLE $\hat{\boldsymbol{\alpha}}_c$ is asymptotically normal with mean $\boldsymbol{\alpha}_0$ and covariance matrix $\bm{H}(\bm{H}^\top \mathcal{I}_{\boldsymbol{\alpha}_0}\bm{H})^{-1} \bm{H}^\top$. (This assumes $\mathcal{I}_{\boldsymbol{\alpha}_0}$ is the total Fisher information, consistent with the boxed formula earlier. The original had a $1/n$ factor here which seemed inconsistent.)
    \item The Likelihood Ratio Test statistic $\lambda = 2[\ell_N(\hat{\boldsymbol{\alpha}}_u) - \ell_N(\hat{\boldsymbol{\alpha}}_c)]$ for testing the validity of the constraint asymptotically follows a chi-squared distribution with $d-r$ degrees of freedom, where $d-r$ is the number of independent linear restrictions imposed by the constraint. This is also known as Wilks' theorem.
\end{enumerate}
These results allow for constructing confidence intervals and performing hypothesis tests regarding the linear constraints.

%% file: tex_behind/appendix_part_2.tex
\chapter{Mathematical Derivations and Simulation Details for Part III}
\label{apdx:part2_details}

This appendix provides supplementary mathematical details for the derivations presented in \cref{part:Model_selection} and lists the parameters used in the benchmark simulations.

\section{Estimating the Error $\mathcal{E}$ from the Log-Likelihood}
\label{apdx:E_from_log_likelihood}
We aim to prove Eq.~\eqref{eq:error_chapN1}, which connects the inference error $\mathcal{E}(\bm{\hat{f}}^{\mathcal{B}})$ to the estimated log-likelihood $\ell(\bm{\hat{f}}^{\mathcal{B}}\mid\bm{X}_{\tau})$. For simplicity, we assume the estimated diffusion $\bm{\bar{D}}$ equals the true (potentially averaged) diffusion matrix $\bm{D}$. We also assume $\Delta t$ is small enough to justify the Euler-Maruyama approximation $\Delta \bm{x}_t \approx \bm{f}(\bm{x}_t)\Delta t + \Delta \bm{\Xi}_t$, where $\Delta \bm{\Xi_t} = \int_t^{t+\Delta t} \sqrt{2\bm{D}(\bm{x}_s)} \bm{\xi}(s) \dd{s}$.

Expanding the negative log-likelihood (Eq.~\eqref{eq:likelihood_chapN1}):
\begin{align*}
-\frac{4}{\tau} \ell\left(\bm{X}_{\tau} | \hat{\bm{f}}^\mathcal{B}\right) &= \avg{\left(\frac{\Delta \bm{x}_t}{\Delta t} - \hat{\bm{f}}^\mathcal{B}_t\right)\cdot\mathbf{D}^{-1}\cdot \left(\frac{\Delta \bm{x}_t}{\Delta t} - \hat{\bm{f}}^\mathcal{B}_t \right)} \\
&\approx \avg{\left( (\bm{f}_t - \hat{\bm{f}}^\mathcal{B}_t) + \frac{\Delta \bm{\Xi}_t}{\Delta t} \right)\cdot\mathbf{D}^{-1}\cdot \left( (\bm{f}_t - \hat{\bm{f}}^\mathcal{B}_t) + \frac{\Delta \bm{\Xi}_t}{\Delta t} \right)} \\
&= \underbrace{\avg{(\bm{f}_t - \hat{\bm{f}}^\mathcal{B}_t)\cdot\mathbf{D}^{-1}\cdot (\bm{f}_t - \hat{\bm{f}}^\mathcal{B}_t)}}_{4\mathcal{E}(\hat{\bm{f}}^\mathcal{B})} + 2 \avg{(\bm{f}_t - \hat{\bm{f}}^\mathcal{B}_t)\cdot\mathbf{D}^{-1}\cdot \frac{\Delta \bm{\Xi}_t}{\Delta t}} \\
&\quad + \underbrace{\avg{\frac{\Delta \bm{\Xi}_t}{\Delta t}\cdot\mathbf{D}^{-1}\cdot \frac{\Delta \bm{\Xi}_t}{\Delta t}}}_{C'}
\end{align*}
The term $C'$ involves only the noise increments and is independent of the model $\mathcal{B}$, thus irrelevant for model comparison. Let $C = \tau C'/4$. We focus on the cross term. Using the Itô isometry property $\E{\Delta \Xi_{t, \alpha} \Delta \Xi_{t, \beta}} = 2 D_{\alpha \beta}(\bm{x}_t) \Delta t + \mathcal{O}(\Delta t^2)$ and the fact that $\Delta \bm{\Xi}_t$ is independent of past states (including $\bm{f}_t$), we have $\E{\avg{\bm{f}_t \cdot\mathbf{D}^{-1}\cdot \frac{\Delta \bm{\Xi}_t}{\Delta t}}} = 0$.

However, the term involving the inferred force $\hat{\bm{f}}^\mathcal{B}_t$ requires care because $\hat{\bm{f}}^\mathcal{B}_t$ depends on the entire trajectory, including future noise increments, through the coefficient calculation Eq.~\eqref{eq:SFI-Ito_chapN1}. Let $\hat{\bm{f}}^\mathcal{B}_t = \sum_i \hat{\alpha}_i^\mathcal{B} \bm{b}_i(\bm{x}_t)$.
\begin{align*}
\E{ -2 \avg{\hat{\bm{f}}^\mathcal{B}_t \cdot\mathbf{D}^{-1}\cdot \frac{\Delta \bm{\Xi}_t}{\Delta t}}} &= \E{ -2 \sum_i \hat{\alpha}_i^\mathcal{B} \avg{\bm{b}_i(\bm{x}_t) \cdot\mathbf{D}^{-1}\cdot \frac{\Delta \bm{\Xi}_t}{\Delta t}} } \\
&= \E{ -2 \sum_{i,j} \left( G_\mathcal{B}^{-1} \right)_{ij} \avg{\frac{\Delta \bm{x}_s}{\Delta s}\cdot\mathbf{D}^{-1}\cdot\bm{b_j}(\bm{x}_s)} \avg{\bm{b}_i(\bm{x}_t) \cdot\mathbf{D}^{-1}\cdot \frac{\Delta \bm{\Xi}_t}{\Delta t}} }
\end{align*}
Substituting $\Delta \bm{x}_s \approx \bm{f}_s \Delta s + \Delta \bm{\Xi}_s$ and keeping the leading terms related to noise correlations (neglecting correlations between $G_\mathcal{B}^{-1}$ and noise, which yield higher-order terms):
The dominant contribution comes from $s=t$:
\begin{align*}
&\approx \E{ -2 \sum_{i,j} \left( G_\mathcal{B}^{-1} \right)_{ij} \avg{\frac{\Delta \bm{\Xi}_t}{\Delta t}\cdot\mathbf{D}^{-1}\cdot\bm{b_j}(\bm{x}_t)} \avg{\bm{b}_i(\bm{x}_t) \cdot\mathbf{D}^{-1}\cdot \frac{\Delta \bm{\Xi}_t}{\Delta t}}} \\
&\approx \E{ - \frac{2}{\tau} \sum_{i,j} \left( G_\mathcal{B}^{-1} \right)_{ij} \sum_t \left( \frac{\Delta \bm{\Xi}_t}{\Delta t}\cdot\mathbf{D}^{-1}\cdot\bm{b_j}(\bm{x}_t) \right) \left( \bm{b}_i(\bm{x}_t) \cdot\mathbf{D}^{-1}\cdot \frac{\Delta \bm{\Xi}_t}{\Delta t} \right) \Delta t} \\
&\approx \E{ - \frac{4}{\tau \Delta t} \sum_{i,j} \left( G_\mathcal{B}^{-1} \right)_{ij} \sum_t \bm{b}_i(\bm{x}_t) \cdot\mathbf{D}^{-1}\cdot \bm{b_j}(\bm{x}_t) \Delta t} \quad \text{(Using Itô isometry properties)} \\
&= \E{ - \frac{4}{\tau} \sum_{i,j} \left( G_\mathcal{B}^{-1} \right)_{ij} (G_\mathcal{B})_{ji}}= - \frac{4 n_\mathcal{B}}{\tau}.
\end{align*}
Thus, taking the expectation of the negative log-likelihood expansion:
\begin{equation*}
\E{-\frac{4}{\tau} \ell\left(\bm{X}_{\tau} | \hat{\bm{f}}^\mathcal{B}\right)} \approx 4 \E{\mathcal{E}(\hat{\bm{f}}^\mathcal{B})} - \frac{4 n_\mathcal{B}}{\tau} + C'
\end{equation*}
Multiplying by $-\tau/4$:
\begin{equation}
\E{\ell\left(\bm{X}_{\tau} | \hat{\bm{f}}^\mathcal{B}\right)} \approx -\tau \E{\mathcal{E}(\hat{\bm{f}}^\mathcal{B})} + n_\mathcal{B} - C 
\label{eq:expected_logL_apdx}
\end{equation}
Taking the difference between two models $\mathcal{B}$ and $\mathcal{C}$ cancels the constant $C$ and yields Eq.~\eqref{eq:error_chapN1}:
\begin{equation*}
\E{\ell(\mathcal{C}) - \ell(\mathcal{B})} \approx \tau\E{\mathcal{E}(\hat{\bm{f}}^{\mathcal{B}}) -  \mathcal{E}(\hat{\bm{f}}^{\mathcal{C}})} + n_{\mathcal{C}} - n_{\mathcal{B}}.
\end{equation*}

\section{Derivation of the Bayesian Information Criterion (BIC)}
\label{apdx:BIC}
We derive the BIC formula $\mathcal{I}_{\mathrm{BIC}}(\mathcal{B}) = \mathcal{I}(\mathcal{B}) - \frac{n_\mathcal{B}}{2}\log(\tau)$ presented in \cref{eq:bic_def}. BIC approximates the log marginal likelihood (or Bayesian evidence) for a model $\mathcal{B}$. The marginal likelihood is obtained by integrating the likelihood over the prior distribution of the parameters $\bm{\alpha}^\mathcal{B} = \{\alpha_1^\mathcal{B}, \dots, \alpha_{n_\mathcal{B}}^\mathcal{B}\}$: 
\begin{equation}
    P(\bm{X}_{\tau}|\mathcal{B}) = \int P(\bm{X}_{\tau}|\bm{\alpha}^\mathcal{B}, \mathcal{B}) \Pi_\mathcal{B}(\bm{\alpha}^\mathcal{B}) \dd{\bm{\alpha}^\mathcal{B}}
    \label{eq:posterior_apdx}
\end{equation}
where $P(\bm{X}_{\tau}|\bm{\alpha}^\mathcal{B}, \mathcal{B}) = \frac{1}{Z} e^{\ell(\bm{X}_{\tau}|\bm{\alpha}^\mathcal{B})}$ is the likelihood ($Z$ is a normalization constant independent of $\bm{\alpha}^\mathcal{B}$) and $\Pi_\mathcal{B}(\bm{\alpha}^\mathcal{B})$ is the prior. 

We use Laplace's approximation for the integral. We Taylor-expand the log-likelihood $\ell(\bm{X}_{\tau}|\bm{\alpha}^\mathcal{B})$ around its maximum, which occurs at the inferred parameters $\hat{\bm{\alpha}}^\mathcal{B}$:
\begin{equation*}
\ell(\bm{X}_{\tau}|\bm{\alpha}^\mathcal{B}) \approx \ell(\bm{X}_{\tau}|\hat{\bm{\alpha}}^\mathcal{B}) + (\bm{\alpha}^\mathcal{B}-\hat{\bm{\alpha}}^\mathcal{B})^T \nabla \ell + \frac{1}{2} (\bm{\alpha}^\mathcal{B}-\hat{\bm{\alpha}}^\mathcal{B})^T \nabla^2 \ell (\bm{\alpha}^\mathcal{B}-\hat{\bm{\alpha}}^\mathcal{B})
\end{equation*}
At the maximum, $\nabla \ell = 0$. The Hessian matrix $\nabla^2 \ell$ can be calculated from Eq.~\eqref{eq:likelihood_chapN1} (assuming parameters are $F_k^\mathcal{B} = \alpha_k^\mathcal{B}$):
$\frac{\partial^2 \ell}{\partial \alpha_i^\mathcal{B} \partial \alpha_j^\mathcal{B}} = -\frac{\tau}{2} \avg{\bm{b}_i \cdot \bar{\mathbf{D}}^{-1} \cdot \bm{b}_j} = -\frac{\tau}{2} (G_\mathcal{B})_{ij}$. 
So, $\nabla^2 \ell = -\frac{\tau}{2} \bm{G}_\mathcal{B}$.
\begin{equation*}
\ell(\bm{X}_{\tau}|\bm{\alpha}^\mathcal{B}) \approx \ell(\bm{X}_{\tau}|\hat{\bm{\alpha}}^\mathcal{B}) - \frac{\tau}{4} (\bm{\alpha}^\mathcal{B}-\hat{\bm{\alpha}}^\mathcal{B})^T \bm{G}_\mathcal{B} (\bm{\alpha}^\mathcal{B}-\hat{\bm{\alpha}}^\mathcal{B})
\end{equation*}
Substituting this into the integral and assuming the prior $\Pi_\mathcal{B}(\bm{\alpha}^\mathcal{B})$ is approximately constant around $\hat{\bm{\alpha}}^\mathcal{B}$:
\begin{align*}
P(\bm{X}_{\tau}|\mathcal{B}) &\approx \frac{1}{Z} e^{\ell(\bm{X}_{\tau}|\hat{\bm{\alpha}}^\mathcal{B})} \Pi_\mathcal{B}(\hat{\bm{\alpha}}^\mathcal{B}) \int e^{- \frac{\tau}{4} (\bm{\alpha}^\mathcal{B}-\hat{\bm{\alpha}}^\mathcal{B})^T \bm{G}_\mathcal{B} (\bm{\alpha}^\mathcal{B}-\hat{\bm{\alpha}}^\mathcal{B})} \dd{\bm{\alpha}^\mathcal{B}} \\
&= \frac{1}{Z} e^{\ell(\bm{X}_{\tau}|\hat{\bm{\alpha}}^\mathcal{B})} \Pi_\mathcal{B}(\hat{\bm{\alpha}}^\mathcal{B}) \sqrt{\frac{(2\pi)^{n_\mathcal{B}}}{\det(\frac{\tau}{2} \bm{G}_\mathcal{B})}} \\
&= \frac{e^{\ell(\bm{X}_{\tau}|\hat{\bm{\alpha}}^\mathcal{B})}}{Z} \Pi_\mathcal{B}(\hat{\bm{\alpha}}^\mathcal{B}) \left(\frac{4\pi}{\tau}\right)^{\frac{n_\mathcal{B}}{2}} (\det\bm{G}_\mathcal{B})^{-\frac{1}{2}}
\end{align*}
Taking the logarithm:
\begin{equation*}
\log P(\bm{X}_{\tau}|\mathcal{B}) \approx \ell(\bm{X}_{\tau}|\hat{\bm{\alpha}}^\mathcal{B}) + \log \Pi_\mathcal{B}(\hat{\bm{\alpha}}^\mathcal{B}) - \frac{n_\mathcal{B}}{2} \log \tau + \frac{n_\mathcal{B}}{2}\log(4\pi) - \frac{1}{2}\log(\det\bm{G}_\mathcal{B}) - \log Z
\end{equation*}
For large $\tau$, the dominant term involving $n_\mathcal{B}$ is $-\frac{n_\mathcal{B}}{2} \log \tau$. The terms involving the prior and the determinant of the Gram matrix are typically $\mathcal{O}(1)$ or grow slower than $\log \tau$. Neglecting constant terms and terms that do not scale with $n_\mathcal{B} \log \tau$, we obtain the BIC approximation:
\begin{equation*}
\log P(\bm{X}_{\tau}|\mathcal{B}) \approx \ell(\bm{X}_{\tau}|\hat{\bm{\alpha}}^\mathcal{B}) - \frac{n_\mathcal{B}}{2} \log \tau
\end{equation*}
Since $\mathcal{I}(\mathcal{B}) = \ell(\bm{X}_{\tau}|\hat{\bm{\alpha}}^\mathcal{B}) - \ell(\bm{X}_{\tau}|0)$, maximizing BIC is equivalent to maximizing $\mathcal{I}_{\mathrm{BIC}}(\mathcal{B}) = \mathcal{I}(\mathcal{B}) - \frac{n_\mathcal{B}}{2}\log(\tau)$, ignoring the model-independent $\ell(\bm{X}_{\tau}|0)$. The use of total time $\tau$ instead of the number of data points $N=\tau/\Delta t$ arises naturally from the SDE likelihood formulation and is consistent with \cite{frishmanLearningForceFields2020}.


\section{Simulations Details and Parameters}
\label{apdx:simulations}
All simulations use the Euler–Maruyama method with simulation time step $\dd{t}$, sampling interval $\Delta t$ (with $\Delta t = 10 \dd{t}$), and total time $\tau$. For the prediction error \cref{subsec:perf_metrics}, the independent test trajectories have duration $\tau_{\mathcal{E}}$. In all cases, initial conditions are taken after a thermalization period $\tau_{\text{therm.}}=10$. Results are averaged over 100 simulations.

\begin{itemize}
    \item \textbf{Lorenz System} (\cref{fig:p_influence_OU_chapN2,fig:benchmark_comparison,fig:pastis_delta_t_robustness,fig:pastis_sigma_robustness}): The system parameters are $\sigma = 10, \rho=28, \beta=7/3$ \cref{fig:benchmark_systems}. The diffusion constant is $D=100$, the sampling interval $\Delta t = 2 \times 10^{-4}$, and $\tau_{\mathcal{E}}=20$. Specifically for \cref{fig:pastis_sigma_robustness,fig:pastis_delta_t_robustness}, we use $\tau=4 \times 10^{3}$. The total library $\mathcal{B}_0$ contains monomials up to order 2 in $x,y,z$ for each component ($n_0 = 3 \times 10 = 30$). The true model $\mathcal{B}^*$ has $n^*=7$ basis functions.

    \item \textbf{Ornstein-Uhlenbeck} (\cref{fig:p_influence_OU_chapN2,fig:benchmark_comparison,fig:pastis_delta_t_robustness,fig:pastis_sigma_robustness}): $\dv{\bm{x}}{t} = -\bm{A}\bm{x} + \sqrt{2\bm{D}}\bm{\xi}(t)$. Dimension $d=10$ with $\bm{x} \in \mathbb{R}^d$. Matrix $\bm{A}$ has diagonal elements equal to 1 ($A_{ii}=1$), and 10\% of off-diagonal elements are randomly set to $\pm 1$, with other elements being 0. Thus, the true model has $n^* = d + 0.1 d(d-1) \approx 19$ parameters. The simulation details are $\bm{D} = 100 \bm{I}$, $\dd{t}=0.001$, $\Delta t = 0.01$, $\tau_{\mathcal{E}}=10^3$. The total library $\mathcal{B}_0$: $\{\bm{e_j}\}_{j=1\dots d} \cup \{x_k \bm{e}_j\}_{j,k=1..d}$ has $n_0 = d + d^2 = 110$ functions. 

    \item \textbf{Lotka-Volterra} (\cref{fig:p_influence_OU_chapN2,fig:benchmark_comparison,fig:pastis_delta_t_robustness,fig:pastis_sigma_robustness}): $\dv{x_i}{t} = x_i (r_i + \sum_j A_{ij} x_j) + \sqrt{2D x_i^2} \xi_i(t)$ with dimension $d=10$, $r_i=1$ and $D=0.05$. The interaction matrix has diagonal terms equal to $-1$ ($A_{ii}=-1$), and the off-diagonal $A_{ij}$ are sparse with $\pm 1$ values as in \cref{fig:benchmark_systems} and \cref{sec:lotka_volterra_benchmark}. For the simulations, we use $\dd{t}=0.001$, $\Delta t = 0.01$, $\tau_{\mathcal{E}}=100$. The total basis is presented in \cref{fig:benchmark_systems} and \cref{sec:lotka_volterra_benchmark}. 

    \item \textbf{Gray-Scott} (\cref{fig:gray_scott_details,fig:benchmark_gs_comparison}): The Gray-Scott model is composed of 2 SPDEs: $\pdv{u}{t} = D_u \nabla^2 u - uv^2 + F(1-u) + \sqrt{2D}\xi_u$ and $\pdv{v}{t} = D_v \nabla^2 v + uv^2 - (F+k)v + \sqrt{2D}\xi_v$. The parameters are $D_u=0.2097, D_v=0.105, F=0.029, k=0.057$, and $D=0.001$. For the simulations, we used a square lattice $100 \times 100$ ($L_x=L_y=100$) with periodic boundaries, $\Delta x = \Delta y = 1$, with simulation time step $\dd{t}=0.001$, and sampling interval of the data $\Delta t = 0.01$. For the prediction error, we use independent trajectories of duration $\tau_{\mathcal{E}}=10$. The total library is presented in \cref{fig:benchmark_systems}. 
\end{itemize}

%% file: tex_behind/annexe_TP_FP_FN.tex
\section{Benchmarking and Hyperparameter Tuning for Comparison Methods}
\label{apdx:hyperparams_errors}

\begin{figure*}[!htbp]
    \centering
    \includegraphics[width=\textwidth]{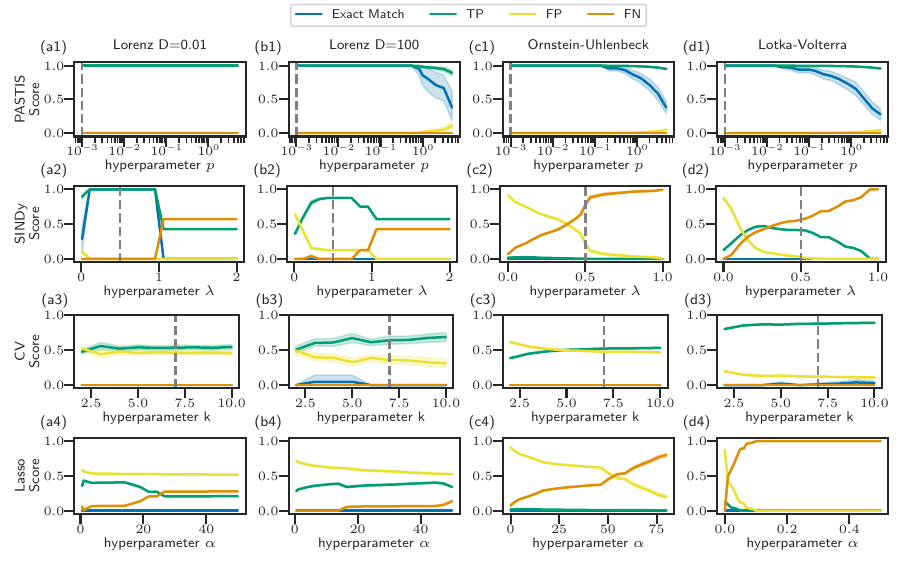}
    \caption{Comparison of model selection performance versus hyperparameters for PASTIS, SINDy, k-fold CV, and LASSO across four benchmark systems (columns: Lorenz D=0.01, Lorenz D=100, Ornstein-Uhlenbeck, Lotka-Volterra). Each row corresponds to a method, plotting performance metrics against its primary hyperparameter: \(p\) for PASTIS (row 1, panels a1-d1), STLSQ threshold \(\lambda\) for SINDy (row 2, panels a2-d2), number of folds \(k\) for k-fold CV (row 3, panels a3-d3), and regularization strength \(\alpha\) for LASSO (row 4, panels a4-d4). Performance metrics are Exact Match (blue), True Positives (TP, green), False Positives (FP, yellow), and False Negatives (FN, orange). The vertical dashed lines indicate default or chosen parameter values used in main text comparisons where applicable (e.g., \(p=0.001\) for PASTIS, \(\lambda=0.5\) for SINDy). Note that the x-axis scale and range vary between rows depending on the relevant hyperparameter.}
    \label{fig:appendix_hyperparameters_total}
\end{figure*}

To ensure a fair comparison between PASTIS and existing methods (SINDy, k-fold CV, LASSO), we performed careful hyperparameter tuning for these comparison methods. \cref{fig:appendix_hyperparameters_total} consolidates the results, showing performance metrics versus the primary hyperparameter for each method across the four benchmark systems. The metrics shown are Exact Match (blue), True Positives (TP, green), False Positives (FP, yellow), and False Negatives (FN, orange), defined as:
\begin{equation}
    \text{TP} = \frac{\lvert \mathcal{B} \cap \mathcal{B}^* \rvert}{\lvert \mathcal{B} \cup \mathcal{B}^* \rvert}, \qquad
    \text{FP} = \frac{\lvert \mathcal{B} - \mathcal{B}^* \rvert}{\lvert \mathcal{B} \cup \mathcal{B}^* \rvert}, \qquad
    \text{FN} = \frac{\lvert \mathcal{B}^* - \mathcal{B} \rvert}{\lvert \mathcal{B} \cup \mathcal{B}^* \rvert},
    \label{eq:appendix_tp_fp_fn}
\end{equation}
where \(\mathcal{B}\) is the model selected by the algorithm and \(\mathcal{B}^*\) is the true model.

\subsection{SINDy (STLSQ)}
\label{apdx:hyperparams_errors_sindy}
The performance of SINDy depends significantly on its STLSQ threshold hyperparameter \(\lambda\), as shown in the second row of \cref{fig:appendix_hyperparameters_total} (panels a2-d2). The optimal \(\lambda\) varies with the system and noise level. For low noise (Lorenz D=0.01, panel a2), a sharp threshold exists where the exact match rate is high. However, for noisier systems (panels b2-d2), this performance window disappears. Even near-optimal \(\lambda\) values (e.g., \(\lambda \approx 0.5-1.0\) for Lorenz D=100, \(\lambda \approx 0.1-0.5\) for OU and LV), SINDy often selects models with substantial numbers of both false positives (FP, yellow lines) and false negatives (FN, orange lines). This highlights its sensitivity and difficulty in achieving exact sparse recovery in challenging stochastic settings. The value \(\lambda=0.5\) used in the main text represents a reasonable choice across the noisy systems tested.

\subsection{k-Fold Cross-Validation (CV)}

The third row of \cref{fig:appendix_hyperparameters_total} (panels a3-d3) shows the performance of k-fold CV versus the number of folds \(k\). The results are largely insensitive to \(k\) in the range 2-10. More importantly, k-fold CV consistently fails to achieve a high exact match rate for any \(k\) across all systems. It primarily suffers from a high rate of false positives (FP, yellow lines), indicating that, like AIC, it does not sufficiently penalize model complexity when selecting from a large library in this SDE context.

\subsection{LASSO}

The fourth row of \cref{fig:appendix_hyperparameters_total} (panels a4-d4) shows the performance of LASSO versus its regularization parameter \(\alpha\). Similar to SINDy, performance is sensitive to \(\alpha\). However, across the tested range, LASSO failed to achieve a high exact match rate for any of the stochastic systems. It typically exhibits a trade-off where reducing false positives (by increasing \(\alpha\)) simultaneously increases false negatives, failing to identify the correct sparse structure. Its overall performance was significantly worse than PASTIS (row 1) and often worse than SINDy.

\subsection{Conclusion}

Based on these comprehensive hyperparameter analyses presented in \cref{fig:appendix_hyperparameters_total}, the comparisons in the main text (\cref{fig:p_influence_OU_chapN2,fig:benchmark_comparison,fig:pastis_delta_t_robustness,fig:pastis_sigma_robustness}) are well-justified, using representative or near-optimal parameters for the benchmark methods. The figure clearly demonstrates the superior robustness and accuracy of PASTIS in achieving exact model recovery across different systems compared to SINDy, k-fold CV, and LASSO, especially when considering the challenge of minimizing both false positives and false negatives simultaneously.